\documentclass{article}

\usepackage{microtype}
\usepackage{graphicx}
\usepackage{subfigure}
\usepackage{booktabs} 

\usepackage{hyperref}

\usepackage[accepted]{icml2021}

\usepackage{titletoc}

\usepackage[utf8]{inputenc} \usepackage[T1]{fontenc}    \usepackage{hyperref}       \usepackage{url}            \usepackage{booktabs}       \usepackage{amsfonts}       \usepackage{nicefrac}       \usepackage{microtype}

\usepackage{xargs}

\usepackage{booktabs}   \usepackage{tabularx}		\usepackage{makecell}

\usepackage{graphicx} 	\usepackage{xcolor} 		\usepackage{pifont}			\usepackage{caption}    \usepackage{tikz}			  \usepackage{float}      \usepackage{pgfplots}
\pgfplotsset{compat=1.15}	\usetikzlibrary{shapes,
                pgfplots.groupplots,
                shadings,
                calc,
                arrows,
                backgrounds,
                colorbrewer,
                shadows.blur,
                math}

\usetikzlibrary{external}
\usepackage{environ}    \newif\ifshowtikz       \showtikztrue

\ifshowtikz
\else
\RenewEnviron{tikzpicture}[1][]{}
\fi

\definecolor{oo_rot}{RGB}{204,0,0}
\definecolor{oo_blau}{RGB}{51,153,255}
\definecolor{oo_gruen}{RGB}{0,153,0}
\definecolor{oo_gelb}{RGB}{153,153,102}
\definecolor{light_red}{RGB}{235,135,117}

\usepackage{amsfonts}    \usepackage{amsmath} 		 \usepackage{amssymb}
\usepackage{mathtools} 	 \usepackage{nicefrac}

\usepackage{xspace} 		\usepackage{microtype} 	\usepackage{url} 			  \usepackage[]{hyperref}
\hypersetup{
    colorlinks=true,
    linkcolor=firebrick,
    citecolor=firebrick,
    urlcolor=firebrick
}
\usepackage[]{cleveref}

\usepackage{blindtext}

\usepackage[
colorinlistoftodos,prependcaption,textsize=tiny, shadow
]{todonotes} 						

\newcommandx{\unsure}[1]{\todo[linecolor=red,backgroundcolor=red!25,bordercolor=red]{#1}}				

\newcommandx{\change}[1]{\todo[linecolor=blue,backgroundcolor=blue!25,bordercolor=blue]{#1}}			

\newcommandx{\note}[1]{\todo[linecolor=orange,backgroundcolor=orange!25,bordercolor=orange]{#1}}		

\newcommandx{\improvement}[1]{\todo[linecolor=violet,backgroundcolor=violet!25,bordercolor=violet]{#1}}

\newcommand*{\eg}{e.g.\@\xspace}
\newcommand*{\ie}{i.e.\@\xspace}

\newcommand{\explainmath}[1]{\ensuremath{&&\text{#1}}}

\definecolor{TUred}{RGB}{141,45,57}
\definecolor{TUdark}{RGB}{55,65,74}
\definecolor{TUgold}{RGB}{174,159,109}
\definecolor{TUgray}{RGB}{175,179,183}
\definecolor{ERC_ora}{RGB}{233,93,15}
\definecolor{firebrick}{RGB}{197,57,50}

\newcommand{\goldenRatio}{1.61803398875}
\newcommand{\goldenRatioInv}{0.61803398875}

\newcommand{\viviturl}{\url{https://github.com/jbzrE7bp/vivit}}

\newcommand{\vivit}{\mbox{\normalfont\textsc{ViViT}}\xspace}
\newcommand{\bfvivit}{\mbox{\normalfont\textbf{\textsc{ViViT}}\xspace}}
\newcommand{\vivittitle}{\textsc{ViViT}\xspace} \newcommand{\cockpit}{\mbox{\normalfont\textsc{Cockpit}}\xspace}
\newcommand{\deepobs}{\mbox{\normalfont\textsc{DeepOBS}}\xspace}
\newcommand{\pyhessian}{\mbox{\normalfont\textsc{PyHessian}}\xspace}
\newcommand{\backpack}{\mbox{\normalfont\textsc{BackPACK}}\xspace}
 \newcommand{\kfac}{\mbox{\normalfont\textsc{K-FAC}}\xspace}

\newcommand{\pytorch}{\mbox{\normalfont\textsc{PyTorch}}\xspace}

\newcommand{\sgd}{\textsc{SGD}\xspace}
\newcommand{\ggn}{\textsc{GGN}\xspace}
\newcommand{\adam}{\textsc{Adam}\xspace}

\newcommand{\cifarten}{\textsc{CIFAR-10}\xspace}
\newcommand{\cifarhun}{\textsc{CIFAR-100}\xspace}

\newcommand{\fmnist}{\textsc{Fashion-MNIST}\xspace}
\newcommand{\imagenet}{\textsc{ImageNet}\xspace}

\newcommand{\allcnnc}{\textsc{All-CNN-C}\xspace}
\newcommand{\threecthreed}{\textsc{3c3d}\xspace}
\newcommand{\twoctwod}{\textsc{2c2d}\xspace}

\newcommand{\resnetthirtytwo}{\textsc{ResNet-32}\xspace}

\newcommand{\mc}{\textsc{MC}\xspace} 

\usepackage{amsmath,amsfonts,bm}

\def\eqref#1{equation~\ref{#1}}

\def\floor#1{\lfloor #1 \rfloor}
\def\1{\bm{1}}

\def\vtheta{{\bm{\theta}}}
\def\va{{\bm{a}}}
\def\vb{{\bm{b}}}

\def\ve{{\bm{e}}}

\def\vg{{\bm{g}}}

\def\vs{{\bm{s}}}

\def\vv{{\bm{v}}}

\def\vx{{\bm{x}}}
\def\vy{{\bm{y}}}
\def\vz{{\bm{z}}}

\def\mG{{\bm{G}}}
\def\mH{{\bm{H}}}
\def\mI{{\bm{I}}}

\def\mP{{\bm{P}}}

\def\mR{{\bm{R}}}

\def\mU{{\bm{U}}}
\def\mV{{\bm{V}}}
\def\mW{{\bm{W}}}

\DeclareMathAlphabet{\mathsfit}{\encodingdefault}{\sfdefault}{m}{sl}
\SetMathAlphabet{\mathsfit}{bold}{\encodingdefault}{\sfdefault}{bx}{n}

\def\gL{{\mathcal{L}}}

\def\gO{{\mathcal{O}}}

\def\sR{{\mathbb{R}}}
\def\sS{{\mathbb{S}}}

\def\sX{{\mathbb{X}}}
\def\sY{{\mathbb{Y}}}

\newcommand{\E}{\mathbb{E}}

\newcommand{\R}{\mathbb{R}}

\DeclareMathOperator{\Tr}{Tr}
\DeclareMathOperator{\rank}{rank}
\DeclareMathOperator{\diag}{diag}
\DeclareMathOperator{\vecspan}{span}

\newcommand{\jac}{\mathrm{J}}
\newcommand{\mGtilde}{\mathbf{\tilde{G}}}
\newcommand{\vetilde}{\mathbf{\tilde{e}}}
\newcommand{\vstilde}{\mathbf{\tilde{s}}}

\icmltitlerunning{\bfvivit{}: Curvature access through the generalized Gauss-Newton's low-rank structure}

\newcommand{\vivitpapertitle}{\vivittitle: Curvature access through the \\
  generalized Gauss-Newton's low-rank structure}

\begin{document}

\twocolumn[
\icmltitle{\vivitpapertitle}

\icmlsetsymbol{equal}{*}

\begin{icmlauthorlist}
\icmlauthor{Felix Dangel}{equal,uni_tue}
\icmlauthor{Lukas Tatzel}{equal,uni_tue}
\icmlauthor{Philipp Hennig}{uni_tue_mpi}
\end{icmlauthorlist}

\icmlaffiliation{uni_tue}{University of T{\"u}bingen, Maria-von-Linden-Stra{\ss}e 6, T{\"u}bingen, Germany}
\icmlaffiliation{uni_tue_mpi}{University of T{\"u}bingen \& MPI for Intelligent Systems, T{\"u}bingen, Germany}

\icmlcorrespondingauthor{Felix Dangel}{fdangel@tue.mpg.de}
\icmlcorrespondingauthor{Lukas Tatzel}{lukas.tatzel@uni-tuebingen.de}

\icmlkeywords{Machine Learning, ICML}

\vskip 0.3in
]

\printAffiliationsAndNotice{\icmlEqualContribution} 

\begin{abstract}
Curvature in form of the Hessian or its generalized Gauss-Newton (\ggn)
  approximation is valuable for algorithms that rely on a local model for
  the loss to train, compress, or explain deep networks.
Existing methods based on implicit multiplication via automatic
  differentiation or Kronecker-factored block diagonal approximations do not
  consider noise in the mini-batch.
We present \vivit, a curvature model that leverages the \ggn's low-rank
  structure without further approximations. It allows for efficient
  computation of eigenvalues, eigenvectors, as well as
  per-sample first- and second-order directional derivatives.
The representation is computed in parallel with gradients in one backward
  pass and offers a fine-grained cost-accuracy trade-off, which allows it to
  scale.
We demonstrate this by conducting performance benchmarks and substantiate \vivit's usefulness by studying the impact of noise on the \ggn{}'s structural properties during neural network training.
\end{abstract}

\section{Introduction \& Motivation}
\label{sec:introduction}

The large number of trainable parameters in deep neural networks imposes computational constraints on the information that can be made available to optimization algorithms.
Standard machine learning libraries \citep{abadi2015tensorflow,
  paszke2019pytorch} mainly provide access to first-order information in the form of \emph{average} mini-batch gradients.
This is a limitation that complicates the development of novel methods that may
outperform the state-of-the-art: They must use the same objects to remain easy to
implement and use, and to rely on the highly optimized code of those libraries.
There is evidence that this has led to stagnation in the performance of
first-order optimizers \citep{schmidt2021descending}. Here, we thus study how to
provide efficient access to richer information, namely higher-order derivatives and their distribution across the mini-batch.

Recent advances in automatic differentiation \citep{bradbury2020jax,
  dangel2020backpack} have made such information more readily accessible through
vectorization of algebraic structure in the differentiated loss. We leverage
and extend this functionality to efficiently access curvature in form of the
Hessian's generalized Gauss-Newton (\ggn) approximation. It offers practical
advantages over the Hessian and is established for training
\citep{martens2010deep, martens2015optimizing}, compressing
\citep{singh2020woodfisher}, or adding uncertainty to \citep{ritter2018scalable,
  ritter2018online, kristiadi2020being} neural networks.
It is also
linked theoretically to the natural gradient method \citep{amari2000natural} via the Fisher information matrix \citep[Section 9.2]{martens2020new}.

Traditional ways to access curvature fall into two categories. Firstly, repeated
automatic differentiation allows for matrix-free exact multiplication with the
Hessian \citep{pearlmutter1994fast} and \ggn \citep{schraudolph2002fast}.
Iterative linear and eigensolvers can leverage such functionality to
compute Newton steps \citep{martens2010deep, zhang2017blockdiagonal,
  gargiani2020promise} and spectral properties \citep{sagun2017eigenvalues,
  sagun2018empirical, adams2018estimating, ghorbani2019investigation,
  papyan2019spectrum, yao2019pyhessian, granziol2021deep} on arbitrary
architectures thanks to the generality of automatic differentiation. However,
repeated matrix-vector products
are potentially detrimental to performance.

Secondly, \kfac (Kronecker-factored approximate curvature)
\citep{martens2015optimizing, grosse2016kroneckerfactored, botev2017practical,
  martens2018kroneckerfactored} constructs an explicit light-weight
representation of the \ggn based on its algebraic Kronecker structure.
The computations are streamlined via gradient backpropagation and the resulting
matrices are cheap to store and invert. This allows \kfac to scale: It
has been used successfully with large mini-batches~\citep{osawa2019large}.
One reason for this efficiency is that \kfac only
approximates the \ggn's block diagonal, neglecting interactions across layers.
Such terms could be useful, however, for  applications like uncertainty
quantification with Laplace approximations \citep{ritter2018scalable,
  ritter2018online, kristiadi2020being, daxberger2021laplace} that currently rely on \kfac. Moreover,
due to its specific design for optimization, the Kronecker representation does
not become more accurate with more data. It remains a
simplification, exact only under assumptions unlikely to be met in
practice \citep{martens2015optimizing}. This might be a downside for applications
that depend on a precise curvature proxy.

\begin{figure}[!tb]
  \centering
  \begin{flushleft}
    (a)
  \end{flushleft}
  \vspace{-1\baselineskip}
  \begin{minipage}[t]{0.33\linewidth}
    \centering
\pgfkeys{/pgfplots/spectrumdefault/.style={
    width=1.38\linewidth,
    height=48mm,
    every axis plot/.append style={line width = 1.2pt},
    every axis plot post/.append style={
      mark size=1, mark options={opacity=0.3}
    },
    tick pos = left,
    xmajorticks = true,
    ymajorticks = true,
    ylabel near ticks,
    xlabel near ticks,
    xtick align = inside,
    ytick align = inside,
    legend cell align = left,
    legend columns = 1,
    legend pos = north west,
    legend style = {
      fill opacity = 0.9,
      text opacity = 1,
      font = \small,
    },
    xticklabel style = {font = \small, inner xsep = 0ex},
    xlabel style = {font = \small},
    axis line style = {black},
    yticklabel style = {font = \small, inner ysep = 0ex},
    ylabel style = {font = \small, inner ysep = 0ex},
    title style = {font = \small, inner ysep = 0ex, yshift = -0.75ex},
    grid = major,
    grid style = {dashed}
  }
}
\pgfkeys{/pgfplots/spectrumdefaultleft/.style={
    spectrumdefault,
    title=\empty,
    ymax=3e-4,
    xlabel=\phantom{eigenvalues},
  }}
\pgfkeys{/pgfplots/spectrumdefaultcenter/.style={
    spectrumdefault,
    ylabel=\empty,
    yticklabels=\empty,
    title=\empty,
    ymax=3e-4,
  }}
\pgfkeys{/pgfplots/spectrumdefaultright/.style={
    spectrumdefault,
    ylabel=\empty,
    yticklabels=\empty,
    title=\empty,
    ymax=3e-4,
    xlabel=\phantom{eigenvalues},
  }}
 \pgfkeys{/pgfplots/zmystyle/.style={spectrumdefaultleft,
        title={mb, exact}}}
    \begin{tikzpicture}

\definecolor{color0}{rgb}{0.4,0.4,0.4}

\begin{axis}[
axis line style={white!80!black},
log basis x={10},
tick pos=left,
title={full\_batch\_exact, one\_group, N=128, D=895210},
xlabel={eigenvalues},
xmin=0.0001, xmax=35.8794860839844,
xmode=log,
ylabel={density},
ymin=1.11705508533742e-06, ymax=1,
ymode=log,
zmystyle
]
\draw[draw=white,fill=color0] (axis cs:0.0001,1.11705508533742e-06) rectangle (axis cs:0.000155434142340701,0.99871426815899);
\draw[draw=white,fill=color0] (axis cs:0.000155434142340701,1.11705508533742e-06) rectangle (axis cs:0.000241597726051892,1.11705508533742e-06);
\draw[draw=white,fill=color0] (axis cs:0.000241597726051892,1.11705508533742e-06) rectangle (axis cs:0.000375525353403394,1.11705508533742e-06);
\draw[draw=white,fill=color0] (axis cs:0.000375525353403394,1.11705508533742e-06) rectangle (axis cs:0.00058369461233445,5.58528041795857e-06);
\draw[draw=white,fill=color0] (axis cs:0.00058369461233445,1.11705508533742e-06) rectangle (axis cs:0.000907260714570931,2.68093507478535e-05);
\draw[draw=white,fill=color0] (axis cs:0.000907260714570931,1.11705508533742e-06) rectangle (axis cs:0.00141019291048744,4.80334210778594e-05);
\draw[draw=white,fill=color0] (axis cs:0.00141019291048744,1.11705508533742e-06) rectangle (axis cs:0.00219192125576551,7.26086604072757e-05);
\draw[draw=white,fill=color0] (axis cs:0.00219192125576551,1.11705508533742e-06) rectangle (axis cs:0.00340699400468264,9.04815617377603e-05);
\draw[draw=white,fill=color0] (axis cs:0.00340699400468264,1.11705508533742e-06) rectangle (axis cs:0.00529563191077756,0.000105003294068835);
\draw[draw=white,fill=color0] (axis cs:0.00529563191077756,1.11705508533742e-06) rectangle (axis cs:0.00823122004203757,0.000123993251732419);
\draw[draw=white,fill=color0] (axis cs:0.00823122004203757,1.11705508533742e-06) rectangle (axis cs:0.0127941262765169,0.000138514984063382);
\draw[draw=white,fill=color0] (axis cs:0.0127941262765169,1.11705508533742e-06) rectangle (axis cs:0.0198864404478903,0.000137397927730282);
\draw[draw=white,fill=color0] (axis cs:0.0198864404478903,1.11705508533742e-06) rectangle (axis cs:0.0309103181522725,0.000130695589731351);
\draw[draw=white,fill=color0] (axis cs:0.0309103181522725,1.11705508533742e-06) rectangle (axis cs:0.0480451879147667,0.000115056801067177);
\draw[draw=white,fill=color0] (axis cs:0.0480451879147667,1.11705508533742e-06) rectangle (axis cs:0.0746786257712956,9.27156744040709e-05);
\draw[draw=white,fill=color0] (axis cs:0.0746786257712956,1.11705508533742e-06) rectangle (axis cs:0.116076081479435,7.26086604072757e-05);
\draw[draw=white,fill=color0] (axis cs:0.116076081479435,1.11705508533742e-06) rectangle (axis cs:0.180421861710252,5.5852815410002e-05);
\draw[draw=white,fill=color0] (axis cs:0.180421861710252,1.11705508533742e-06) rectangle (axis cs:0.280437173344456,3.79799140794064e-05);
\draw[draw=white,fill=color0] (axis cs:0.280437173344456,1.11705508533742e-06) rectangle (axis cs:0.435895115192459,2.23411254153434e-05);
\draw[draw=white,fill=color0] (axis cs:0.435895115192459,1.11705508533742e-06) rectangle (axis cs:0.677529833804408,1.45217310832009e-05);
\draw[draw=white,fill=color0] (axis cs:0.677529833804408,1.11705508533742e-06) rectangle (axis cs:1.05311268627626,5.58528041795857e-06);
\draw[draw=white,fill=color0] (axis cs:1.05311268627626,1.11705508533742e-06) rectangle (axis cs:1.63689667179461,2.2341114184372e-06);
\draw[draw=white,fill=color0] (axis cs:1.63689667179461,1.11705508533742e-06) rectangle (axis cs:2.54429630280743,1.11705508533742e-06);
\draw[draw=white,fill=color0] (axis cs:2.54429630280743,1.11705508533742e-06) rectangle (axis cs:3.95470513687488,2.23411141854822e-06);
\draw[draw=white,fill=color0] (axis cs:3.95470513687488,1.11705508533742e-06) rectangle (axis cs:6.1469620116051,3.351167751648e-06);
\draw[draw=white,fill=color0] (axis cs:6.14696201160511,1.11705508533742e-06) rectangle (axis cs:9.55447768274709,4.46822408474777e-06);
\draw[draw=white,fill=color0] (axis cs:9.55447768274709,1.11705508533742e-06) rectangle (axis cs:14.8509204413116,4.46822408485879e-06);
\draw[draw=white,fill=color0] (axis cs:14.8509204413116,1.11705508533742e-06) rectangle (axis cs:23.0834008176524,1.11705508533742e-06);
\draw[draw=white,fill=color0] (axis cs:23.0834008176524,1.11705508533742e-06) rectangle (axis cs:35.8794860839844,1.11705508533742e-06);
\end{axis}

\end{tikzpicture}
   \end{minipage}
  \hspace{1.9ex}
  \begin{minipage}[t]{0.33\linewidth}
    \centering
\pgfkeys{/pgfplots/spectrumdefault/.style={
    width=1.38\linewidth,
    height=48mm,
    every axis plot/.append style={line width = 1.2pt},
    every axis plot post/.append style={
      mark size=1, mark options={opacity=0.3}
    },
    tick pos = left,
    xmajorticks = true,
    ymajorticks = true,
    ylabel near ticks,
    xlabel near ticks,
    xtick align = inside,
    ytick align = inside,
    legend cell align = left,
    legend columns = 1,
    legend pos = north west,
    legend style = {
      fill opacity = 0.9,
      text opacity = 1,
      font = \small,
    },
    xticklabel style = {font = \small, inner xsep = 0ex},
    xlabel style = {font = \small},
    axis line style = {black},
    yticklabel style = {font = \small, inner ysep = 0ex},
    ylabel style = {font = \small, inner ysep = 0ex},
    title style = {font = \small, inner ysep = 0ex, yshift = -0.75ex},
    grid = major,
    grid style = {dashed}
  }
}
\pgfkeys{/pgfplots/spectrumdefaultleft/.style={
    spectrumdefault,
    title=\empty,
    ymax=3e-4,
    xlabel=\phantom{eigenvalues},
  }}
\pgfkeys{/pgfplots/spectrumdefaultcenter/.style={
    spectrumdefault,
    ylabel=\empty,
    yticklabels=\empty,
    title=\empty,
    ymax=3e-4,
  }}
\pgfkeys{/pgfplots/spectrumdefaultright/.style={
    spectrumdefault,
    ylabel=\empty,
    yticklabels=\empty,
    title=\empty,
    ymax=3e-4,
    xlabel=\phantom{eigenvalues},
  }}
 \pgfkeys{/pgfplots/zmystyle/.style={spectrumdefaultcenter,
        title={sub, exact}}}
    \begin{tikzpicture}

\definecolor{color0}{rgb}{0.4,0.4,0.4}

\begin{axis}[
axis line style={white!80!black},
log basis x={10},
tick pos=left,
title={frac\_batch\_exact, one\_group, N=128, D=895210},
xlabel={eigenvalues},
xmin=0.0001, xmax=35.8794860839844,
xmode=log,
ylabel={density},
ymin=1.11705508533742e-06, ymax=1,
ymode=log,
zmystyle
]
\draw[draw=white,fill=color0] (axis cs:0.0001,1.11705508533742e-06) rectangle (axis cs:0.000155434142340701,0.999840260942811);
\draw[draw=white,fill=color0] (axis cs:0.000155434142340701,1.11705508533742e-06) rectangle (axis cs:0.000241597726051892,1.11705508533742e-06);
\draw[draw=white,fill=color0] (axis cs:0.000241597726051892,1.11705508533742e-06) rectangle (axis cs:0.000375525353403394,1.11705508533742e-06);
\draw[draw=white,fill=color0] (axis cs:0.000375525353403394,1.11705508533742e-06) rectangle (axis cs:0.00058369461233445,1.11705508533742e-06);
\draw[draw=white,fill=color0] (axis cs:0.00058369461233445,1.11705508533742e-06) rectangle (axis cs:0.000907260714570931,1.11705508533742e-06);
\draw[draw=white,fill=color0] (axis cs:0.000907260714570931,1.11705508533742e-06) rectangle (axis cs:0.00141019291048744,1.11705508533742e-06);
\draw[draw=white,fill=color0] (axis cs:0.00141019291048744,1.11705508533742e-06) rectangle (axis cs:0.00219192125576551,1.11705508533742e-06);
\draw[draw=white,fill=color0] (axis cs:0.00219192125576551,1.11705508533742e-06) rectangle (axis cs:0.00340699400468264,1.11705508533742e-06);
\draw[draw=white,fill=color0] (axis cs:0.00340699400468264,1.11705508533742e-06) rectangle (axis cs:0.00529563191077756,1.11705508533742e-06);
\draw[draw=white,fill=color0] (axis cs:0.00529563191077756,1.11705508533742e-06) rectangle (axis cs:0.00823122004203757,1.11705508533742e-06);
\draw[draw=white,fill=color0] (axis cs:0.00823122004203757,1.11705508533742e-06) rectangle (axis cs:0.0127941262765169,1.11705508533742e-06);
\draw[draw=white,fill=color0] (axis cs:0.0127941262765169,1.11705508533742e-06) rectangle (axis cs:0.0198864404478903,5.58528041795857e-06);
\draw[draw=white,fill=color0] (axis cs:0.0198864404478903,1.11705508533742e-06) rectangle (axis cs:0.0309103181522725,6.70233675105834e-06);
\draw[draw=white,fill=color0] (axis cs:0.0309103181522725,1.11705508533742e-06) rectangle (axis cs:0.0480451879147667,1.00535057505797e-05);
\draw[draw=white,fill=color0] (axis cs:0.0480451879147667,1.11705508533742e-06) rectangle (axis cs:0.0746786257712956,1.11705620837905e-05);
\draw[draw=white,fill=color0] (axis cs:0.0746786257712956,1.11705508533742e-06) rectangle (axis cs:0.116076081479435,1.22876184168903e-05);
\draw[draw=white,fill=color0] (axis cs:0.116076081479435,1.11705508533742e-06) rectangle (axis cs:0.180421861710252,1.78729000826112e-05);
\draw[draw=white,fill=color0] (axis cs:0.180421861710252,1.11705508533742e-06) rectangle (axis cs:0.280437173344456,2.45752380816539e-05);
\draw[draw=white,fill=color0] (axis cs:0.280437173344456,1.11705508533742e-06) rectangle (axis cs:0.435895115192459,2.23411254152324e-05);
\draw[draw=white,fill=color0] (axis cs:0.435895115192459,1.11705508533742e-06) rectangle (axis cs:0.677529833804408,2.23411254153434e-05);
\draw[draw=white,fill=color0] (axis cs:0.677529833804408,1.11705508533742e-06) rectangle (axis cs:1.05311268627626,1.56387874163006e-05);
\draw[draw=white,fill=color0] (axis cs:1.05311268627626,1.11705508533742e-06) rectangle (axis cs:1.63689667179461,1.00535057505797e-05);
\draw[draw=white,fill=color0] (axis cs:1.63689667179461,1.11705508533742e-06) rectangle (axis cs:2.54429630280743,5.58528041795857e-06);
\draw[draw=white,fill=color0] (axis cs:2.54429630280743,1.11705508533742e-06) rectangle (axis cs:3.95470513687488,1.11705508533742e-06);
\draw[draw=white,fill=color0] (axis cs:3.95470513687488,1.11705508533742e-06) rectangle (axis cs:6.1469620116051,3.351167751648e-06);
\draw[draw=white,fill=color0] (axis cs:6.14696201160511,1.11705508533742e-06) rectangle (axis cs:9.55447768274709,4.46822408485879e-06);
\draw[draw=white,fill=color0] (axis cs:9.55447768274709,1.11705508533742e-06) rectangle (axis cs:14.8509204413116,3.351167751648e-06);
\draw[draw=white,fill=color0] (axis cs:14.8509204413116,1.11705508533742e-06) rectangle (axis cs:23.0834008176524,3.351167751648e-06);
\draw[draw=white,fill=color0] (axis cs:23.0834008176524,1.11705508533742e-06) rectangle (axis cs:35.8794860839844,1.11705508533742e-06);
\end{axis}

\end{tikzpicture}
   \end{minipage}
  \hspace{-3.5ex}
  \begin{minipage}[t]{0.33\linewidth}
    \centering
\pgfkeys{/pgfplots/spectrumdefault/.style={
    width=1.38\linewidth,
    height=48mm,
    every axis plot/.append style={line width = 1.2pt},
    every axis plot post/.append style={
      mark size=1, mark options={opacity=0.3}
    },
    tick pos = left,
    xmajorticks = true,
    ymajorticks = true,
    ylabel near ticks,
    xlabel near ticks,
    xtick align = inside,
    ytick align = inside,
    legend cell align = left,
    legend columns = 1,
    legend pos = north west,
    legend style = {
      fill opacity = 0.9,
      text opacity = 1,
      font = \small,
    },
    xticklabel style = {font = \small, inner xsep = 0ex},
    xlabel style = {font = \small},
    axis line style = {black},
    yticklabel style = {font = \small, inner ysep = 0ex},
    ylabel style = {font = \small, inner ysep = 0ex},
    title style = {font = \small, inner ysep = 0ex, yshift = -0.75ex},
    grid = major,
    grid style = {dashed}
  }
}
\pgfkeys{/pgfplots/spectrumdefaultleft/.style={
    spectrumdefault,
    title=\empty,
    ymax=3e-4,
    xlabel=\phantom{eigenvalues},
  }}
\pgfkeys{/pgfplots/spectrumdefaultcenter/.style={
    spectrumdefault,
    ylabel=\empty,
    yticklabels=\empty,
    title=\empty,
    ymax=3e-4,
  }}
\pgfkeys{/pgfplots/spectrumdefaultright/.style={
    spectrumdefault,
    ylabel=\empty,
    yticklabels=\empty,
    title=\empty,
    ymax=3e-4,
    xlabel=\phantom{eigenvalues},
  }}
 \pgfkeys{/pgfplots/zmystyle/.style={spectrumdefaultright,
        title={mb, mc}}}
    \begin{tikzpicture}

\definecolor{color0}{rgb}{0.4,0.4,0.4}

\begin{axis}[
axis line style={white!80!black},
log basis x={10},
tick pos=left,
title={full\_batch\_mc, one\_group, N=128, D=895210},
xlabel={eigenvalues},
xmin=0.0001, xmax=35.8794860839844,
xmode=log,
ylabel={density},
ymin=1.11705508533742e-06, ymax=1,
ymode=log,
zmystyle
]
\draw[draw=white,fill=color0] (axis cs:0.0001,1.11705508533742e-06) rectangle (axis cs:0.000155434142340701,0.999858133844141);
\draw[draw=white,fill=color0] (axis cs:0.000155434142340701,1.11705508533742e-06) rectangle (axis cs:0.000241597726051892,1.11705508533742e-06);
\draw[draw=white,fill=color0] (axis cs:0.000241597726051892,1.11705508533742e-06) rectangle (axis cs:0.000375525353403394,1.11705508533742e-06);
\draw[draw=white,fill=color0] (axis cs:0.000375525353403394,1.11705508533742e-06) rectangle (axis cs:0.00058369461233445,1.11705508533742e-06);
\draw[draw=white,fill=color0] (axis cs:0.00058369461233445,1.11705508533742e-06) rectangle (axis cs:0.000907260714570931,1.11705508533742e-06);
\draw[draw=white,fill=color0] (axis cs:0.000907260714570931,1.11705508533742e-06) rectangle (axis cs:0.00141019291048744,1.11705508533742e-06);
\draw[draw=white,fill=color0] (axis cs:0.00141019291048744,1.11705508533742e-06) rectangle (axis cs:0.00219192125576551,1.11705508533742e-06);
\draw[draw=white,fill=color0] (axis cs:0.00219192125576551,1.11705508533742e-06) rectangle (axis cs:0.00340699400468264,1.11705508533742e-06);
\draw[draw=white,fill=color0] (axis cs:0.00340699400468264,1.11705508533742e-06) rectangle (axis cs:0.00529563191077756,1.11705508533742e-06);
\draw[draw=white,fill=color0] (axis cs:0.00529563191077756,1.11705508533742e-06) rectangle (axis cs:0.00823122004203757,1.11705508533742e-06);
\draw[draw=white,fill=color0] (axis cs:0.00823122004203757,1.11705508533742e-06) rectangle (axis cs:0.0127941262765169,1.11705508533742e-06);
\draw[draw=white,fill=color0] (axis cs:0.0127941262765169,1.11705508533742e-06) rectangle (axis cs:0.0198864404478903,1.11705508533742e-06);
\draw[draw=white,fill=color0] (axis cs:0.0198864404478903,1.11705508533742e-06) rectangle (axis cs:0.0309103181522725,4.46822408474777e-06);
\draw[draw=white,fill=color0] (axis cs:0.0309103181522725,1.11705508533742e-06) rectangle (axis cs:0.0480451879147667,6.70233675116937e-06);
\draw[draw=white,fill=color0] (axis cs:0.0480451879147667,1.11705508533742e-06) rectangle (axis cs:0.0746786257712956,1.34046747499901e-05);
\draw[draw=white,fill=color0] (axis cs:0.0746786257712956,1.11705508533742e-06) rectangle (axis cs:0.116076081479435,1.22876184168903e-05);
\draw[draw=white,fill=color0] (axis cs:0.116076081479435,1.11705508533742e-06) rectangle (axis cs:0.180421861710252,1.45217310832009e-05);
\draw[draw=white,fill=color0] (axis cs:0.180421861710252,1.11705508533742e-06) rectangle (axis cs:0.280437173344456,1.8989956415822e-05);
\draw[draw=white,fill=color0] (axis cs:0.280437173344456,1.11705508533742e-06) rectangle (axis cs:0.435895115192459,2.01070127490328e-05);
\draw[draw=white,fill=color0] (axis cs:0.435895115192459,1.11705508533742e-06) rectangle (axis cs:0.677529833804408,2.34581817484432e-05);
\draw[draw=white,fill=color0] (axis cs:0.677529833804408,1.11705508533742e-06) rectangle (axis cs:1.05311268627626,1.56387874163006e-05);
\draw[draw=white,fill=color0] (axis cs:1.05311268627626,1.11705508533742e-06) rectangle (axis cs:1.63689667179461,1.22876184168903e-05);
\draw[draw=white,fill=color0] (axis cs:1.63689667179461,1.11705508533742e-06) rectangle (axis cs:2.54429630280743,3.351167751648e-06);
\draw[draw=white,fill=color0] (axis cs:2.54429630280743,1.11705508533742e-06) rectangle (axis cs:3.95470513687488,1.11705508533742e-06);
\draw[draw=white,fill=color0] (axis cs:3.95470513687488,1.11705508533742e-06) rectangle (axis cs:6.1469620116051,3.351167751648e-06);
\draw[draw=white,fill=color0] (axis cs:6.14696201160511,1.11705508533742e-06) rectangle (axis cs:9.55447768274709,5.58528041795857e-06);
\draw[draw=white,fill=color0] (axis cs:9.55447768274709,1.11705508533742e-06) rectangle (axis cs:14.8509204413116,2.23411141854822e-06);
\draw[draw=white,fill=color0] (axis cs:14.8509204413116,1.11705508533742e-06) rectangle (axis cs:23.0834008176524,3.351167751648e-06);
\draw[draw=white,fill=color0] (axis cs:23.0834008176524,1.11705508533742e-06) rectangle (axis cs:35.8794860839844,1.11705508533742e-06);
\end{axis}

\end{tikzpicture}
   \end{minipage}

  \vspace{-2ex}

  \begin{flushleft}
    (b)
  \end{flushleft}
  \vspace{-0.5\baselineskip}
  \begin{tikzpicture}
\node[inner sep=0pt] (pdf_plot) at (0,0){\includegraphics[width=\the\columnwidth]{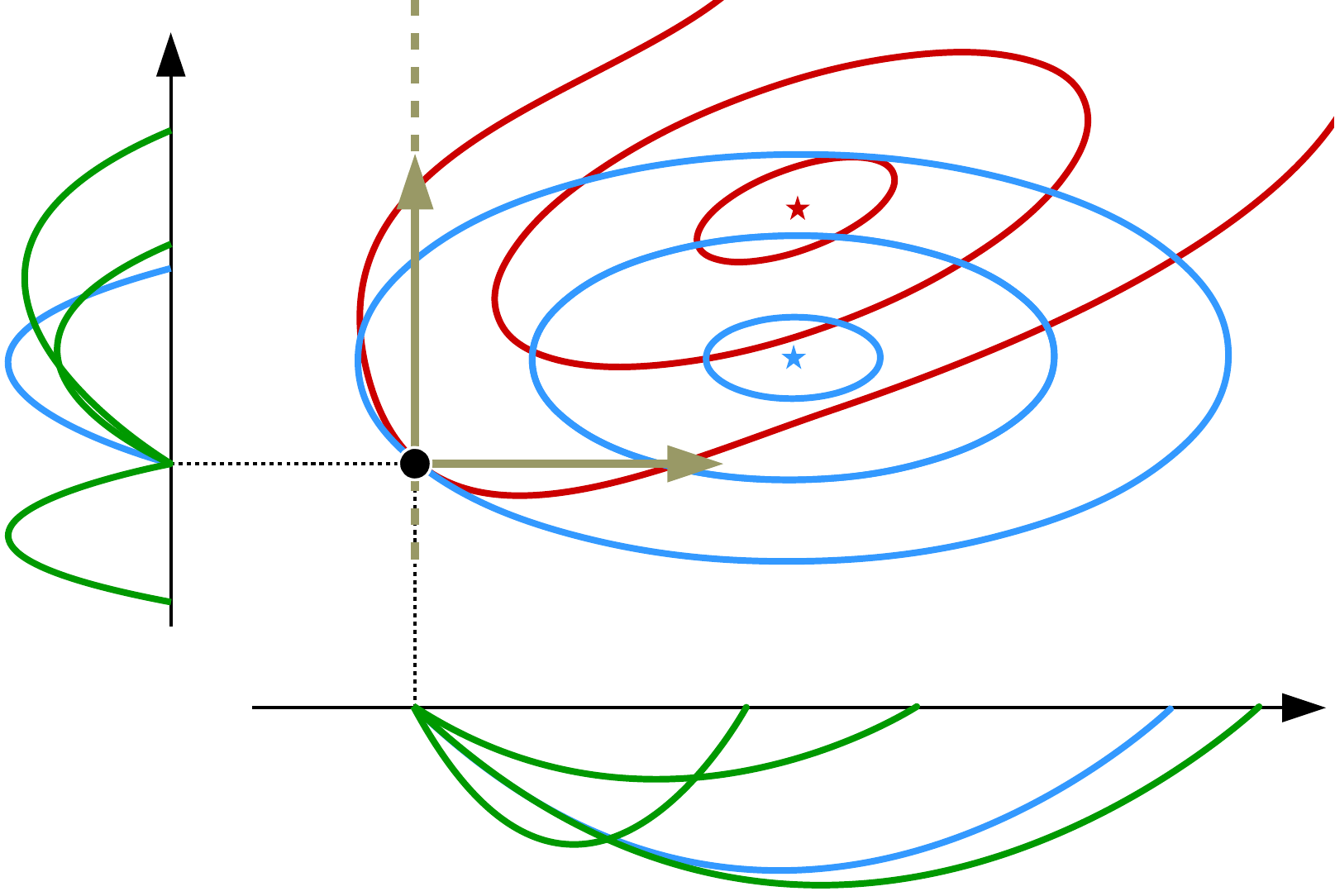}};

\coordinate (origin) at (pdf_plot.south west);

\node[anchor=south west, inner sep=0pt] at ($ (origin) + (22mm, 23mm) $)
{$\vtheta_t$};

\node[anchor=south west, inner sep=0pt] at ($ (origin) + (42mm, 23mm) $)
{$\textcolor{oo_gelb}{\ve_k}$};

\node[anchor=south west, inner sep=0pt] at ($ (origin) + (69mm, 47mm) $)
{$\textcolor{oo_rot}{\gL}$};

\node[anchor=south west, inner sep=0pt] at ($ (origin) + (69mm, 28mm) $)
{$\textcolor{oo_blau}{q}$};

\node[anchor=south west, inner sep=0pt] at ($ (origin) + (23mm, 50mm) $)
{$\textcolor{oo_gelb}{\mathcal{E}}$};

\node[anchor=south west, inner sep=0pt, rotate=21] at ($ (origin) + (55mm, 3mm) $)
{$\textcolor{oo_blau}{\gamma_k} ,\textcolor{oo_blau}{\lambda_k}$};

\node[anchor=south west, inner sep=0pt, rotate=30] at ($ (origin) + (66mm, -1mm) $)
{$\textcolor{oo_gruen}{\gamma_{nk}} ,\textcolor{oo_gruen}{\lambda_{nk}}$};
\end{tikzpicture}
   \vspace{-0.5\baselineskip}

  \vspace{-1ex}

  \caption{
  \textbf{Overview of \vivittitle's quantities:}
\textbf{(a)} \ggn eigenvalue distribution of \deepobs' \threecthreed architecture on
\cifarten \cite{schneider2019deepobs}
for settings with different costs on a mini-batch of size $N = 128$.
From left to right: Exact \ggn,
exact GGN on a mini-batch fraction,
\mc approximation of the \ggn.
\textbf{(b)} Pictorial illustration: Loss function $\textcolor{oo_rot}{\gL}$
from \Cref{eq:objective-function}, quadratic model $\textcolor{oo_blau}{q}$
around $\vtheta_t \in \mathbb{R}^2$ from \Cref{eq:quadratic_model} (both
represented by their contour lines). The low-rank structure provides efficient
access to the \ggn{}'s eigenvectors $\{\textcolor{oo_gelb}{\ve_k}\}$, along
which $\textcolor{oo_blau}{q}$ decouples into one-dimensional parabolas
characterized by the directional derivatives $\textcolor{oo_blau}{\gamma_k},
\textcolor{oo_blau}{\lambda_k}$ and per-sample contributions
$\textcolor{oo_gruen}{\gamma_{nk}}, \textcolor{oo_gruen}{\lambda_{nk}}$
(\Cref{eq:gammas-lambdas}). $\textcolor{oo_gelb}{\mathcal{E}}$ is the \ggn{}'s
top-$1$ eigenspace.}
  \label{fig:visual_abstract}
\end{figure}

Here, we propose \vivit (inspired by $\mV\mV^\top$ in
\Cref{eq:ggn-factorization}), a curvature model that leverages the
\ggn's low-rank structure. Like \kfac, its representation is computed in
parallel with gradients. But it allows a cost-accuracy trade-off, ranging from
the \emph{exact} \ggn to an approximation that has the cost of a single gradient
computation. Our contributions are as follows:
\begin{itemize}
\item We highlight the \ggn's low-rank structure, and with it a structural
limit for the inherent curvature information contained in a mini-batch.

\item We present how to compute various \ggn properties efficiently by exploiting this
  structure (\Cref{fig:visual_abstract}): The exact eigenvalues, eigenvectors,
  and per-sample directional derivatives. In contrast to other methods, these
  quantities allow modeling curvature noise.

\item We introduce approximations that allow a flexible
  trade-off between computational cost and accuracy. We also provide a fully-featured
  efficient implementation in \pytorch \citep{paszke2019pytorch} on top of the
  \backpack \citep{dangel2020backpack} package.\footnotemark

\footnotetext{Code at \viviturl{}.}

\item We empirically demonstrate scalability and efficiency of leveraging the
  \ggn's low-rank structure through benchmarks on different deep neural network
  architectures. Finally, we use \vivit's quantities to study the \ggn, and
  how it is affected by noise, during training.
\end{itemize}

The main focus of this work is demonstrating that many interesting curvature
properties, including uncertainty, can be computed efficiently. Practical
applications of this curvature uncertainty are discussed in
\Cref{sec:use_cases}.

\section{Notation \& Method}
\label{sec:method}
Consider a model $f: \Theta \times \sX \rightarrow \sY$ and a dataset
$\{(\vx_n, \vy_n) \in \sX \times \sY\}_{n=1}^N$. For simplicity we use $N$ for
both the mini-batch and training set size. The network, parameterized by
$\vtheta \in \Theta$, maps a sample $\vx_n$ to a prediction $\hat{\vy}_n$.
Predictions are scored by a convex loss function $\ell : \sY \times \sY
\rightarrow \R$ (\eg cross-entropy or square loss), which compares to
the ground truth $\vy_n$. The training objective $\mathcal{L}: \Theta
\rightarrow \R$ is the empirical risk
\begin{equation}
  \label{eq:objective-function}
  \textstyle
  \gL(\vtheta) = \frac{1}{N} \sum_{n=1}^N \ell(f(\vtheta, \vx_n), \vy_n)\,.
\end{equation}
We use $\ell_n(\vtheta) = \ell(f(\vtheta, \vx_n), \vy_n)$ and $f_n(\vtheta) =
f(\vtheta, \vx_n)$ for per-sample losses and predictions. For gradients, we
write $\vg_n(\vtheta) = \nabla_{\vtheta}\ell_n(\vtheta)$ and $\vg(\vtheta) =
\nabla_{\vtheta} \gL(\vtheta)$, suppressing $\vtheta$ if unambiguous. We also
set $\Theta = \sR^D$ and $\sY = \sR^C$ with $D,C$ the model parameter and
prediction space dimension, respectively. For classification, $C$ is the number of classes.

\paragraph{Hessian \& \ggn:}
Two-fold chain rule application to the split $\ell \circ f$ decomposes the
Hessian of \Cref{eq:objective-function} into two parts $\nabla_{\vtheta}^2
\gL(\vtheta) = \mG(\vtheta) + \mR(\vtheta) \in \sR^{D\times D}$; the positive
semi-definite \ggn
\begin{equation}
  \textstyle
  \label{eq:ggn}
  \mG =
  \frac{1}{N}
  \sum_{n=1}^N
\left(
    \jac_{\vtheta} f_n\right)^\top
\big(
    \nabla_{f_n}^2\ell_n\big)
\left(
    \jac_{\vtheta} f_n\right)
=
  \frac{1}{N}
  \sum_{n=1}^N
  \mG_n\end{equation}
and a residual $\mR = \nicefrac{1}{N}\sum_{n=1}^N\sum_{c=1}^C (\nabla_{\vtheta}^2 [f_n]_c) \left[\nabla_{f_n}\ell_n\right]_c$. Here, we use the Jacobian $\jac_{\va} \vb$ that contains partial
derivatives of $\vb$ with respect to $\va$, $[\jac_{\va} \vb]_{ij} = \partial
[\vb]_i / \partial [\va]_j$. As the residual may alter the Hessian's
definiteness -- undesirable in many applications -- we focus on the
\ggn.
Section \ref{subsec:approx_quality} provides empirical evidence that the curvature's top eigenspace is largely unaffected by this simplification.

\paragraph{Low-rank structure:} By basic inequalities, \Cref{eq:ggn}
has $\rank(\mG) \le NC$.\footnote{We assume the overparameterized deep learning setting ($NC < D$)
  and suppress the trivial rank bound $D$.} To make this
explicit, we factorize the positive semi-definite Hessian
$\nabla_{f_n}^2\ell_n= \sum_{c=1}^C \vs_{nc} \vs_{nc}^\top$, where $\vs_{nc} \in \sR^C$ and denote its
backpropagated version by
$\vv_{nc} = (\jac_{\vtheta} f_n)^\top \vs_{nc} \in \sR^D$. Absorbing sums into matrix multiplications, we
arrive at the \ggn's outer product representation that lies at the heart of the
\vivit concept,
\begin{equation}
  \textstyle
  \label{eq:ggn-factorization}
  \mG
  =
  \frac{1}{N}
  \sum_{n=1}^{N}
  \sum_{c=1}^{C}
  \vv_{nc} \vv_{nc}^\top
  =
  \mV \mV^\top
\end{equation}
with $\mV = \frac{1}{\sqrt{N}}
(\vv_{11}, \vv_{12}, \dots{}, \vv_{NC})
\in \sR^{D\times NC}\,$.
$\mV$ allows for
\emph{exact} computations with the explicit \ggn matrix, at linear rather than
quadratic memory cost in $D$. We first formulate the extraction of relevant \ggn
properties from this factorization, before addressing how to further approximate
$\mV$ to reduce memory and computation costs.

\subsection{Computing the full \ggn eigenspectrum}
\label{sec:computing-full-ggn-eigenspectrum}

Each \ggn eigenvalue $\lambda \in \sR$ is a root of the characteristic
polynomial $\det(\mG - \lambda \mI_D)$ with identity matrix $\mI_D \in
\sR^{D\times D}$. Leveraging the factorization of 
\Cref{eq:ggn-factorization} and the matrix determinant lemma, the
$D$-dimensional eigenproblem reduces to that of the much smaller Gram matrix
$\mGtilde = \mV^\top \mV \in \sR^{NC \times NC}$ which contains pairwise scalar
products of $\vv_{nc}$ (see \Cref{sec:relation-ggn-gram-eigenvalues}),
\begin{align}
  \det(\mG - \lambda \mI_D) = 0
  \quad
  \Leftrightarrow
  \quad
  \det(\mGtilde - \lambda \mI_{NC}) = 0\,.
  \label{eq:ggn-eigenvalues}
\end{align}
With at least $D-NC$ trivial solutions,
the \ggn curvature is zero along most directions in parameter space. Nontrivial
solutions that give rise to curved directions are fully-contained in the Gram
matrix, and hence \textit{much} cheaper to compute.

Despite various Hessian spectral studies which rely on iterative eigensolvers
and implicit matrix multiplication \citep{sagun2017eigenvalues,
  sagun2018empirical, adams2018estimating, ghorbani2019investigation,
  papyan2019spectrum, yao2019pyhessian, granziol2021deep}, we are not aware of
works that efficiently extract the \textit{exact} \ggn spectrum from its Gram
matrix. In contrast to those techniques, this matrix can be computed in parallel
with gradients in a single backward pass, which results in less sequential
overhead.
We demonstrate in \Cref{subsec:scalability} that exploiting the low-rank structure for computing the leading eigenpairs is superior to a power iteration based on matrix-free multiplication in terms of runtime.

Eigenvalues themselves can help identify reasonable hyperparameters, like
learning rates \citep{lecun1993automatic}. But we can also reconstruct the
associated eigenvectors. These are directions along which curvature information
is contained in the mini-batch. Let $\tilde{\sS}_+ = \{(\lambda_k,
\vetilde_k)\:|\: \lambda_k \neq 0, \mGtilde \vetilde_k = \lambda_k \vetilde_k
\}_{k=1}^K$ denote the nontrivial Gram spectrum\footnote{In the following, we
  assume ordered eigenvalues, \ie $\lambda_{1} \ge \lambda_{2} \ge \ldots \ge
  \lambda_{K}$, for convenience.} with orthonormal eigenvectors $\vetilde_j^\top
\vetilde_k = \delta_{jk}$ ($\delta$ represents the Kronecker delta and $K =
\mathrm{rank}(\mG)$). Then, the transformed vectors $\ve_k =
\nicefrac{1}{\sqrt{\lambda_k}} \mV \vetilde_k$ $(k=1, ..., K)$ are orthonormal
eigenvectors of $\mG$ associated to eigenvalues $\lambda_k$ (see
\Cref{sec:relation-ggn-gram-eigenvectors}), i.e. for all $(\lambda_k,
\vetilde_k) \in \tilde{\sS}_+$
\begin{equation}
  \label{eq:ggn-eigenvectors}
  \mGtilde \vetilde_k = \lambda_k \vetilde_k
  \; \implies \;
  \mG \ve_k = \lambda_k \ve_k \,.
\end{equation}
The eigenspectrum also provides access to the \ggn's pseudo-inverse based on $\mV$
and $\tilde{\sS}_+$, required by \eg second-order methods.\footnote{
  \Cref{sec:implicit-multiplication-inverse} describes implicit multiplication
  with $\mG^{-1}$.}

\subsection{Computing directional derivatives}
\label{sec:comp-direct-deriv}

Various algorithms rely on a local quadratic
approximation of the loss landscape. For instance,
optimization methods adapt their parameters by stepping into the minimum of the local proxy.
Curvature, in the form of the Hessian or \ggn, allows to build a quadratic model given by the Taylor expansion.
Let $q$ denote the
quadratic model for the loss around position $\vtheta_t \in \Theta$ that
uses curvature represented by the \ggn,
\begin{equation}
  q(\vtheta)
  = \text{const}
  + (\vtheta - \vtheta_t)^\top \vg(\vtheta_t)
  + \frac{1}{2} (\vtheta - \vtheta_t)^\top \mG(\vtheta_t) (\vtheta - \vtheta_t)\,.
  \label{eq:quadratic_model}
\end{equation}
At its base point $\vtheta_t$, the shape of $q$ along an arbitrary normalized
direction $\ve \in \Theta$ (\ie $\lVert \ve \rVert = 1$) is determined by the
local gradient and curvature. Specifically, the projection of
\Cref{eq:quadratic_model} onto $\ve$
gives rise to the (scalar) first-and second-order directional derivatives
\begin{subequations}
  \label{eq:directional-derivatives}
  \begin{alignat}{4}
    \gamma_{\ve}
    &= \ve^\top \nabla_{\vtheta} q(\vtheta_t)
    &&= \ve^\top \vg(\vtheta_t) &&\in \sR\,,
    \\
    \label{eq:directional-derivatives-lambda}
    \lambda_{\ve}
    &= \ve^\top \nabla_{\vtheta}^2 q(\vtheta_t) \, \ve
    &&= \ve^\top \mG(\vtheta_t) \, \ve &&\in \sR\,.
  \end{alignat}
\end{subequations}
As $\mG$'s characteristic directions are its eigenvectors, they form a
natural basis for the quadratic model. Denoting $\gamma_k =
\gamma_{\ve_k}$ and $\lambda_k = \lambda_{\ve_k}$ the directional gradient and
curvature along eigenvector $\ve_k$, we see from
\Cref{eq:directional-derivatives-lambda} that the directional curvature indeed
coincides with the \ggn's eigenvalue.

Analogous to the gradient and \ggn, the directional derivatives $\gamma_k$ and $\lambda_k$ inherit the sum structure of the loss function from \Cref{eq:objective-function}, \ie they decompose into contributions from individual samples.
Let $\gamma_{nk}$ and
$\lambda_{nk}$ denote these first- and second-order derivatives contributions of
sample $\vx_n$ in direction $k$, \ie
\begin{subequations}
  \label{eq:gammas-lambdas}
  \begin{align}
    \label{eq:gammas}
    \gamma_{nk}
    &= \ve_k^\top \vg_n
      = \frac{\vetilde_k^\top \mV^\top \vg_n}{\sqrt{\lambda_k}}\,,
    \\
    \label{eq:lambdas}
    \lambda_{nk}
    &= \ve_k^\top \mG_n \ve_k
= \frac{\lVert \mV_n^\top \mV \vetilde_k \rVert^2}{\lambda_k}\,,
  \end{align}
\end{subequations}
where $\mV_n \in \sR^{D \times C}$ is
a scaled sub-matrix of $\mV$ with fixed sample index. Note that directional derivatives can be evaluated efficiently with the Gram matrix eigenvectors
without explicit access to the associated directions in parameter space.

In \Cref{eq:directional-derivatives}, gradient $\vg$ and curvature $\mG$ are sums over $\vg_n$ and $\mG_n$, respectively, from which follows the relationship between  directional derivatives and per-sample contributions
$\gamma_k = \nicefrac{1}{N}\sum_{n=1}^N\gamma_{nk}$ and
$\lambda_k = \nicefrac{1}{N} \sum_{n=1}^N\lambda_{nk}$. \Cref{fig:visual_abstract}b shows a pictorial view of the quantities provided by \vivit.

Access to per-sample directional gradients $\gamma_{nk}$
and curvatures $\lambda_{nk}$ along $\mG$'s natural directions
is a distinct feature of \vivit.
These quantities provide geometric information about the local loss landscape
\emph{as well as} about the model's directional curvature stochasticity over the mini-batch.

\subsection{Computational complexity}
\label{sec:method-complexity}
So far, we have formulated the computation of the \ggn{}'s eigenvalues
(\Cref{eq:ggn-eigenvalues}), including eigenvectors (\Cref{eq:ggn-eigenvectors}),
and per-sample directional derivatives (\Cref{eq:gammas-lambdas}). Now, we
analyze their computational complexity in more detail to identify critical
performance factors. Those limitations can effectively be addressed with
approximations that allow the costs to be decreased in a fine-grained fashion.
We substantiate our theoretical analysis with empirical performance measurements in \Cref{subsec:scalability}.

\paragraph{Relation to gradient computation:} Machine learning libraries are
optimized to backpropagate signals $\nicefrac{1}{N} \nabla_{f_n} \ell_n$ and
accumulate the result into the mini-batch gradient $\vg = \nicefrac{1}{N}
\sum_{n=1}^N[\jac_{\vtheta} f_n ]^\top \nabla_{f_n} \ell_n$.
Each column $\vv_{nc}$
of $\mV$ also
involves applying the Jacobian, but to a different vector
$\vs_{nc}$ from the loss Hessian's symmetric factorization.
For popular loss
functions, like square and cross-entropy loss, this factorization is
analytically known and available at negligible overhead. Hence, computing $\mV$
basically costs $C$ gradient computations as it involves $NC$ backpropagations,
while the gradient requires $N$. However, the practical overhead is  expected to be smaller: Computations can re-use information from \backpack's vectorized Jacobians and enjoy
additional speedup on parallel processors like GPUs.

\paragraph{Stage-wise discarding $\mV$:}
The columns of $\mV$ correspond to backpropagated vectors.
During backpropagation, sub-matrices of $\mV$, associated to parameters in the
current layer, become available once at a time
and can be discarded immediately after their use.
This allows for memory savings
without any approximations.

One example is the Gram matrix $\mGtilde$ formed by pairwise scalar products of
$\{\vv_{nc}\}_{n=1,c=1}^{N, C}$ in $\gO((NC)^2D)$ operations. The spectral
decomposition $\tilde{\sS}_+$ has additional cost of $\gO((NC)^3)$.
Similarly, the terms for the
directional derivatives in \Cref{eq:gammas-lambdas} can be built up stage-wise:
First-order derivatives $\{\gamma_{nk}\}_{n=1,k=1}^{N,K}$ require the
vectors $\{ \mV^\top \vg_n \in \sR^{NC} \}_{n=1}^N$ that cost $\gO(N^2CD)$
operations. Second-order derivatives are basically for free, as $\{
\mV_n^\top \mV \in \sR^{C\times NC} \}_{n=1}^N$ is available from $\mGtilde$.

\paragraph{\ggn eigenvectors:}
Transforming an eigenvector $\vetilde_k$
of the Gram matrix to the \ggn eigenvector $\ve_k$
through application of $\mV$ (\Cref{eq:ggn-eigenvectors}) costs $\gO(NCD)$
operations. However, repeated application of $\mV$ can be avoided for
sums of the form $\sum_k (\nicefrac{c_k}{\sqrt{\lambda_k}}) \ve_k $ with arbitrary weights
$c_k \in \sR$. The summation can be performed in the Gram space at negligible
overhead, and only the resulting vector $\sum_k c_k \vetilde_k$ needs to be
transformed.
For a practical example -- computing damped Newton steps -- see \Cref{sec:performance-experiments}.

\subsection{Approximations \& Implementation}
\label{sec:approximations}

Although the \ggn's representation by $\mV$ has linear memory cost in $D$, it
requires memory equivalent to $NC$ model copies.\footnotemark~Of course, this is
infeasible for many networks and data sets, \eg \imagenet ($C=1000$). So far,
our formulation was concerned with \emph{exact} computations. We now present
approximations
that
allow $N$, $C$ and $D$ in the above cost analysis to be replaced by smaller
numbers, enabling \vivit to trade-off accuracy and performance.

\footnotetext{Our implementation uses a more memory-efficient approach that
  avoids expanding $\mV$ for linear layers by leveraging structure in their
  Jacobian (see \Cref{sec:optimized-gram-matrix}).}

\paragraph{\mc approximation \& curvature sub-sampling:}
To reduce the scaling in $C$, we can approximate the factorization
$\nabla^2_{f_n}\ell_n(\vtheta) = \sum_{c=1}^C \vs_{nc} \vs_{nc}^\top$ by a smaller
set of vectors.
One principled approach is to draw \mc samples $\{\vstilde_{nm}\}$ such that $\E_m[ \vstilde_{nm}
\vstilde_{nm}^\top] = \nabla^2_{f_n}\ell_n(\vtheta)$ as in \cite{dangel2020backpack}.
This reduces the scaling of
backpropagated vectors from $C$ to the number of \mc samples $M$ 
($M=1$ in the
following if not specified otherwise).
A common
independent approximation to reduce the scaling in $N$ is computing
curvature on a mini-batch subset \citep{byrd2011use, zhang2017blockdiagonal}.

\paragraph{Parameter groups (block-diagonal approximation):} Some applications,
\eg computing Newton steps, require $\mV$ to be kept in memory for performing
the transformation from Gram space into the parameter space. Still, we can reduce costs by using the \ggn's diagonal blocks $\{\mG^{(i)}\}_{i=1}^L$ of each
layer, rather than the full matrix $\mG$. Such
blocks are available during backpropagation and can thus be used and discarded step by step. In addition to the
previously described approximations for reducing the costs in $N$ and $C$, this
technique tackles scaling in $D$.

\paragraph{Implementation details:}
\backpack's functionality allows us to efficiently
compute individual gradients and $\mV$ in a single backward pass, using either
an exact or \mc-factorization of the loss Hessian. To reduce memory consumption,
we extend its implementation with a protocol to support mini-batch sub-sampling
and parameter groups. By hooks into the package's extensions, we can discard
buffers as soon as possible during backpropagation, effectively implementing all
discussed approximations and optimizations.

In \Cref{sec:experiments}, we specifically address how the above approximations affect runtime and memory requirements, and study their impact on structural properties of the \ggn.

\section{Experiments}
\label{sec:experiments}

For the practical use of the \vivit{} concept, it is essential that (i) the computations are efficient and (ii) that we gain an understanding of how sub-sampling noise and the approximations introduced in \Cref{sec:approximations} alter the structural properties of the \ggn{}.
In the following, we therefore empirically investigate \vivit{}'s scalability and approximation properties in the context of deep learning.
The insights from this analysis substantiate \vivit{}'s value as a monitoring tool for deep learning optimization.

\paragraph{Experimental setting:}
Architectures
include three deep convolutional neural networks from \deepobs
\cite{schneider2019deepobs}
(\twoctwod on \fmnist{},
\threecthreed on \cifarten
and
\allcnnc on \cifarhun),
as well as
residual networks from \citet{he2016deep} on \cifarten based on \citet{idelbayev2018proper} -- all are equipped with cross-entropy loss.
Based on the approximations presented in \Cref{sec:approximations}, we
distinguish the following cases:
\begin{itemize}
\item \textbf{mb, exact:} Exact \ggn with all mini-batch samples.
  Backpropagates $NC$ vectors.
\item \textbf{mb, mc:} \mc-approximated \ggn with all mini-batch samples.
  Backpropagates $N M$ vectors with $M$ the number of \mc{}-samples.
\item \textbf{sub, exact:} Exact \ggn on a subset of mini-batch samples
  ($\floor{\nicefrac{N}{8}}$ as in \cite{zhang2017blockdiagonal}). Backpropagates
  $\floor{\nicefrac{N}{8}} C$ vectors.
\item \textbf{sub, mc:} \mc-approximated \ggn on a subset of mini-batch
  samples. Backpropagates $\floor{\nicefrac{N}{8}} M$ vectors with $M$ the number of \mc{}-samples.
\end{itemize}

\subsection{Scalability}
\label{subsec:scalability}

We now complement the theoretical computational complexity analysis from
\Cref{sec:method-complexity} with empirical studies.
Results were generated on a workstation with an Intel Core i7-8700K CPU (32\,GB)
and one NVIDIA GeForce RTX 2080 Ti GPU (11\,GB). We use $M=1$ in the following.

\begin{figure}[tb]
    
    \begin{minipage}[t]{0.49\linewidth}
      \centering

      \begin{flushleft}
        \vspace{1ex}
        (a)
      \end{flushleft}

      \vspace{-1.0\baselineskip}
      \begin{small}
        $N_{\text{crit}}$ (eigenvalues)
      \end{small}
      \vspace{0.15\baselineskip}

      \begin{small}
        \begin{tabular}{lll}
    \toprule
    $_{\text{\tiny{\ggn}}}$$^{\text{\tiny{Data}}}$ & mb & sub \\
    \midrule
    exact & 909
              & 4375 \\
    mc   & 3840
              & 6626 \\
    \bottomrule
\end{tabular}       \end{small}
    \end{minipage}
    \hfill
    \begin{minipage}[t]{0.49\linewidth}
      \centering

      \begin{flushleft}
        \vspace{1ex}
        \phantom{(a)}
      \end{flushleft}
      \vspace{-1.0\baselineskip}

      \begin{small}
        $N_{\text{crit}}$ (top eigenpair)
      \end{small}
      \vspace{0.15\baselineskip}

      \begin{small}
        \begin{tabular}{lll}
    \toprule
    $_{\text{\tiny{\ggn}}}$$^{\text{\tiny{Data}}}$ & mb & sub \\
    \midrule
    exact & 677
              & 3184 \\
    mc   & 3060
              & 6029 \\
    \bottomrule
\end{tabular}       \end{small}
    \end{minipage}

    \begin{flushleft}
      \vspace{1ex}
      (b)
    \end{flushleft}

    \vspace{-1.3\baselineskip}

\pgfkeys{/pgfplots/performancedefault/.style={
    width=1.04\linewidth,
    height=\goldenRatioInv*1.04\linewidth,
    every axis plot/.append style={line width = 1.2pt},
    every axis plot post/.append style={
      mark size=2, mark options={opacity=0.9, solid, line width = 1pt}
    },
    tick pos = left,
    xmajorticks = true,
    ymajorticks = true,
    ylabel near ticks,
    xlabel near ticks,
    xtick align = inside,
    ytick align = inside,
    legend cell align = left,
    legend columns = 3,
legend style = {
      fill opacity = 0.9,
      text opacity = 1,
      font = \small,
      at={(1, 1.025)},
      anchor=south east,
    },
    xticklabel style = {font = \small, inner xsep = 0ex},
    xlabel style = {font = \small},
    axis line style = {black},
    yticklabel style = {font = \small, inner ysep = 0ex},
    ylabel style = {font = \small, inner ysep = 0ex},
    title style = {font = \small, inner ysep = 0ex, yshift = -0.75ex},
    grid = major,
    grid style = {dashed},
    title = {},
  }
}
 \pgfkeys{/pgfplots/zmystyle/.style={performancedefault}}
    \begin{tikzpicture}

\definecolor{color0}{rgb}{0.937254901960784,0.231372549019608,0.172549019607843}
\definecolor{color1}{rgb}{0.274509803921569,0.6,0.564705882352941}
\definecolor{color2}{rgb}{0.870588235294118,0.623529411764706,0.0862745098039216}
\definecolor{color3}{rgb}{0.501960784313725,0.184313725490196,0.6}

\begin{axis}[
axis line style={white!80!black},
legend style={fill opacity=0.8, draw opacity=1, text opacity=1, at={(0.03,0.97)}, anchor=north west, draw=white!80!black},
tick pos=left,
title={cifar10\_3c3d, N=128, cuda, one\_group},
xlabel={top eigenpairs (\(\displaystyle k\))},
xmin=0.55, xmax=10.45,
ylabel={time [s]},
ymin=0.00910443848057122, ymax=1.89239674986061,
ymode=log,
zmystyle
]
\addplot [, color0, dashed, mark=pentagon*, mark size=3, mark options={solid}]
table {1 0.10646684000676
2 0.263332082999113
3 0.41296657500061
4 0.562371585998335
5 0.961040356996818
6 1.08828064100089
7 1.20301098700293
8 1.29626815899974
9 1.36217651501647
10 1.48477111599641
};
\addlegendentry{power iteration}
\addplot [, black, dashed, mark=*, mark size=3, mark options={solid}]
table {1 0.185383651005395
2 0.186310245997447
3 0.186981072998606
4 0.187526319001336
5 0.18765962299949
6 0.18803448399558
7 0.187778580999293
8 0.186061766995408
9 0.189352801004134
10 0.188033979997272
};
\addlegendentry{mb, exact}
\addplot [, color1, dashed, mark=diamond*, mark size=3, mark options={solid}]
table {1 0.0260587410011794
2 0.0255732080040616
3 0.0259397419940797
4 0.0259047899962752
5 0.0255934300002991
6 0.0258599759981735
7 0.0261660430041957
8 0.02605828599917
9 0.0260880819987506
10 0.0260066910050227
};
\addlegendentry{sub, exact}
\addplot [, color2, dashed, mark=square*, mark size=3, mark options={solid}]
table {1 0.0172249570023268
2 0.0170504369962146
3 0.017168151003716
4 0.0170781389970216
5 0.0171006899981876
6 0.0171182029953343
7 0.0170811919961125
8 0.0169726210006047
9 0.017144368001027
10 0.0171162970000296
};
\addlegendentry{mb, mc}
\addplot [, color3, dashed, mark=triangle*, mark size=3, mark options={solid,rotate=180}]
table {1 0.0127098220036714
2 0.0126403089961968
3 0.0126599400027771
4 0.0123849169976893
5 0.011915481001779
6 0.0116139689998818
7 0.0116039499989711
8 0.0117499150001095
9 0.0118982629937818
10 0.0121629770001164
};
\addlegendentry{sub, mc}
\end{axis}

\end{tikzpicture}

  \vspace{-2ex}
  \caption{\textbf{GPU memory and run time performance:} 
  Performance measurements for the 
  \threecthreed architecture ($D = 895,\!210$) 
  on \cifarten ($C=10$).
  \textbf{(a)} Critical batch sizes $N_{\text{crit}}$
    for computing eigenvalues and the top eigenpair. 
    \textbf{(b)} Run time comparison with a power iteration for extracting 
    the $k$ leading eigenpairs using a batch of size $N=128$.
  }
  \label{fig:performance-cifar10-3c3d-cuda_main}
\end{figure}
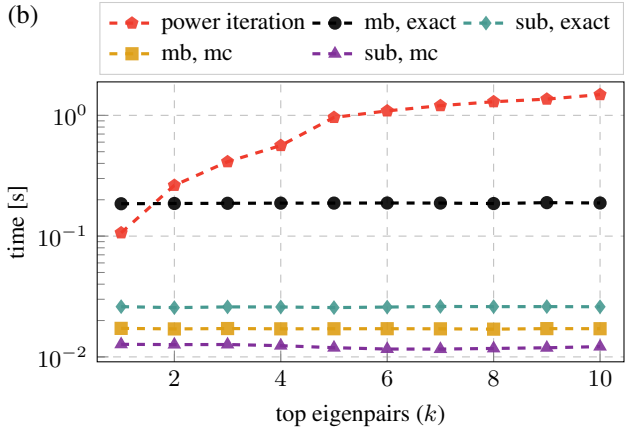

\paragraph{Memory performance:}
We consider two tasks:
\begin{enumerate}
  \item \textbf{Computing eigenvalues:} The nontrivial eigenvalues $\{\lambda_{k}\,|\,
          (\lambda_{k}, \vetilde_{k}) \in \tilde{\sS}_+\}$ are obtained by forming and
        eigen-decomposing the Gram matrix $\mGtilde$, allowing stage-wise discarding
        of $\mV$ (see
        \Cref{sec:computing-full-ggn-eigenspectrum,sec:method-complexity}).
        \label{item:task-eigenvalues}

  \item \textbf{Computing the top eigenpair:} For $(\lambda_{1}, \ve_{1})$, we
        compute the Gram matrix spectrum $\tilde{\sS}_{+}$, extract its top eigenpair
        $(\lambda_{1}, \vetilde_{1})$, and transform it into parameter space by
        \Cref{eq:ggn-eigenvectors}, \ie $(\lambda_{1}, \ve_{1} =
          \nicefrac{1}{\sqrt{\lambda_{1}}} \mV \vetilde_{1} )$. This requires more
        memory than task~\ref{item:task-eigenvalues} as $\mV$ must be stored.
        \label{item:task-eigenvectors}
\end{enumerate}
\vspace{-4mm}
As a comprehensive measure for memory performance, we use the largest batch size
before our system runs out of memory -- we call this the critical batch size
$N_{\text{crit}}$.

\Cref{fig:performance-cifar10-3c3d-cuda_main}a tabularizes the critical batch sizes
on GPU for the \threecthreed architecture on \cifarten. As expected, computing
eigenpairs requires more memory and leads to consistently smaller critical batch
sizes in comparison to computing only eigenvalues. Yet, they all exceed the
traditional batch size used for training ($N=128$, see
\citet{schneider2019deepobs}), even when using the exact \ggn. With \vivit{}'s
approximations, the memory overhead can be reduced to significantly increase the
applicable batch size.

We report similar results for more architectures, a block-diagonal approximation (as in \citet{zhang2017blockdiagonal}),
and on CPU in
\Cref{sec:performance-experiments}, where we also benchmark a third task --
computing damped Newton steps.

\paragraph{Runtime performance:} Here, we consider the task of computing the $k$ leading eigenvectors and eigenvalues of a matrix.
A power iteration that computes eigenpairs iteratively via matrix-vector products serves as a reference. For a fixed value of $k$,
we repeat both approaches $20$ times and report the shortest time.

For the power iteration, we adapt the implementation from the \pyhessian library
\cite{yao2019pyhessian} and replace its Hessian-vector product by a matrix-free
\ggn-vector product \cite{schraudolph2002fast} through \pytorch's automatic
differentiation. We use the same default hyperparameters for the termination
criterion.
Similar to task~\ref{item:task-eigenvalues}, our method obtains the top-$k$
eigenpairs\footnote{In contrast to the power iteration that is restricted to
  dominating eigenpairs, our approach allows choosing arbitrary eigenpairs.}
by computing $\tilde{\sS}_{+}$, extracting its leading eigenpairs
and transforming the
eigenvectors $\vetilde_{1}, \vetilde_{2}, \ldots, \vetilde_{k}$ into parameter space by application of $\mV$ (see \Cref{eq:ggn-eigenvectors}).

\begin{figure}
  \centering
  \includegraphics[scale=1.0]{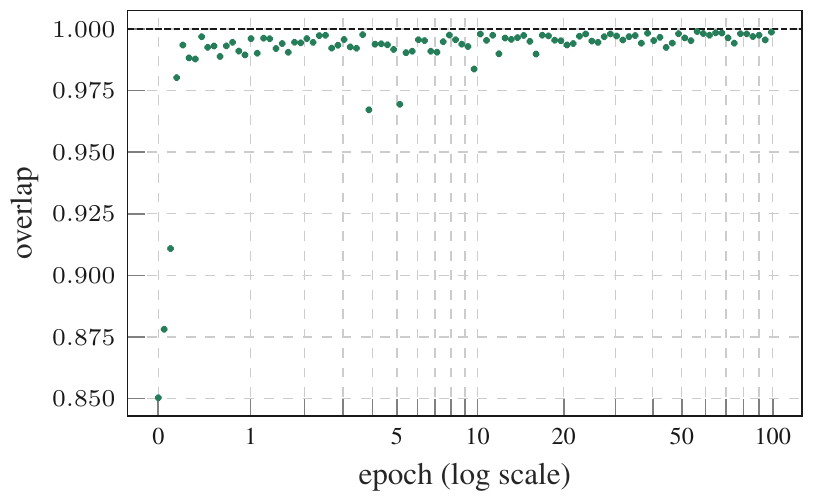}
  \caption{
    \textbf{Full-batch \ggn vs. full-batch Hessian:}
    Overlap between the top-$C$ eigenspaces of the full-batch \ggn and full-batch Hessian during training of the \threecthreed network on \cifarten with \sgd{}.
  }
  \label{fig:approx_GGN_Hessian}
\end{figure}

\Cref{fig:performance-cifar10-3c3d-cuda_main}b shows the GPU runtime for the
\threecthreed architecture on \cifarten, using a mini-batch of size $N=128$.
Without any approximations to the \ggn, our method already outperforms the power
iteration for $k>1$ and increases \textit{much} slower in run time as more leading
eigenpairs are requested. This means that, relative to the transformation of
each eigenvector from the Gram space into the parameter space through $\mV$, the
run time mainly results from computing $\mV,\mGtilde$, and eigendecomposing the
latter. This is consistent with the computational complexity of those operations
in $NC$
(compare \Cref{sec:method-complexity}) and allows for efficient extraction of a
large number of eigenpairs. The run time curves of the approximations confirm
this behavior by featuring the same flat profile. Additionally, they require
significantly less time than the exact mini-batch computation.
Results for more network architectures, a block-diagonal
approximation and on CPU are reported in \Cref{sec:performance-experiments}.

\subsection{Approximation quality}
\label{subsec:approx_quality}

\vivit is based on the Hessian's generalized Gauss-Newton approximation (see \Cref{eq:ggn}).
In practice, the \ggn is only computed on a mini-batch which yields a statistical estimator for the \textit{full-batch} \ggn (\ie the \ggn evaluated on the entire training set).
Additionally, we introduce curvature sub-sampling and an \mc approximation (see Section \ref{sec:approximations}), i.e. further approximations that alter the curvature's structural properties.
In this section, we compare quantities at different stages within this hierarchy of approximations.
We use the test problems from above and train the networks with both \sgd and \adam{} (details in \Cref{sec:training_of_nns}).

\paragraph{\ggn vs. Hessian:}

First, we empirically study the relationship between the \ggn and the Hessian in the deep learning context.
To capture \textit{solely} the effect of neglecting the residual $\mR$ (see \Cref{eq:ggn}), we consider the noise-free case and compute $\mH$ and $\mG$ on the entire training set.

We characterize both curvature matrices by their top-$C$ eigenspace: the space spanned by the eigenvectors to the $C$
largest eigenvalues.
This is a $C$-dimensional subspace of the parameter space $\Theta$, on which the loss function is subject to particularly strong curvature.
The \textit{overlap} between these spaces serves as the comparison metric.
Let $\{ \ve_c^\mU \}_{c=1}^C$ the set of orthonormal eigenvectors to the $C$ largest eigenvalues of some symmetric matrix $\mU$ and $\mathcal{E}^\mU = \vecspan (\ve_1^\mU, ..., \ve_C^\mU)$.
The projection onto this subspace $\mathcal{E}^\mU$ is given by the projection matrix $\mP^\mU = (\ve_1^\mU, ..., \ve_C^\mU) (\ve_1^\mU, ..., \ve_C^\mU)^\top$. As in \citet{gurari2018gradient}, we define the overlap between two top-$C$ eigenspaces $\mathcal{E}^\mU$ and $\mathcal{E}^\mV$ of the matrices $\mU$ and $\mV$ by
\vspace{-2mm}
\begin{equation}
\text{overlap}(\mathcal{E}^\mU, \mathcal{E}^\mV)
= \frac{\Tr{}(\mP^\mU \mP^\mV)}
{\sqrt{\Tr{}(\mP^\mU) \Tr{}(\mP^\mV)}}
\in [0, 1]\, .
\label{eq:overlap_eigenspaces}
\end{equation}
If $\text{overlap}(\mathcal{E}^\mU, \mathcal{E}^\mV) = 0$, then $\mathcal{E}^\mU$ and $\mathcal{E}^\mV$ are orthogonal to each other; if the overlap is $1$, the subspaces are identical.

\Cref{fig:approx_GGN_Hessian} shows the overlap between the full-batch \ggn and Hessian during training of the \threecthreed network on \cifarten with \sgd{}.
Except for a short phase at the beginning of the optimization procedure (note the log scale for the epoch-axis), a strong agreement ($\text{overlap} \geq 0.85$) between the top-$C$ eigenspaces is observed.
We make similar observations with the other test problems (see \Cref{sec:ggn_vs_hessian}), yet to a slightly lesser extent for \cifarhun{}.
Consequently, we identify the \ggn as an interesting object, since it consistently shares relevant structure with the Hessian matrix.

\paragraph{Eigenspace under noise and approximations:}

\begin{figure}[t]
\centering
\includegraphics[scale=1.0]{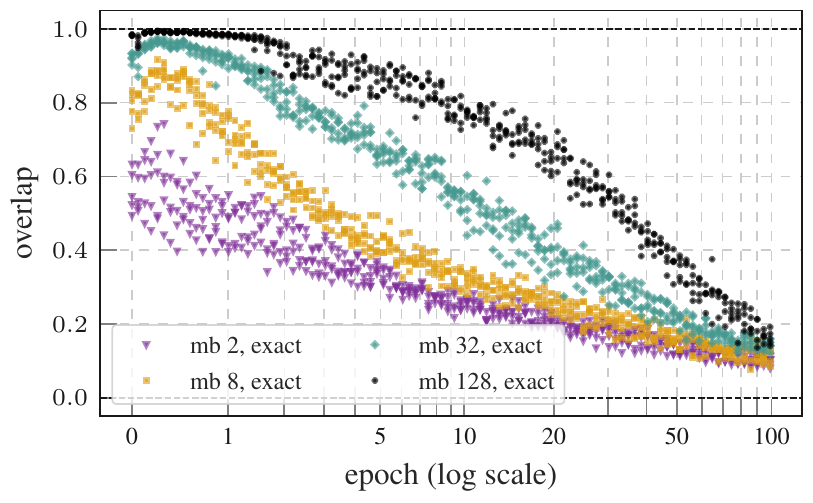}

\caption{
\textbf{Mini-batch \ggn vs. full-batch \ggn{}:}
Overlap between the top-$C$ eigenspaces of the mini-batch \ggn and full-batch \ggn during training of the \threecthreed network on \cifarten with \sgd{}.
For each mini-batch size, $5$ different mini-batches are drawn.
}
\label{fig:approx_eigenspace_bs}
\end{figure}

\vivit uses mini-batching to compute a statistical estimator of the full-batch \ggn{}.
This approximation alters the top-$C$ eigenspace, as shown in
\Cref{fig:approx_eigenspace_bs}:
With decreasing mini-batch size, the approximation carries less and less structure of its full-batch counterpart, as indicated by dropping overlaps.
In addition, at constant batch size, a decrease in approximation quality can be observed over the course of training.
This might be a valuable insight for the design of second-order optimization methods, where this structural decay could lead to performance degradation over the course of the optimization, which has to be compensated for by a growing batch-size (\eg \citet{martens2010deep} reports that the optimal batch size grows during training).

\begin{figure}[t]
\centering
\includegraphics[scale=1.0]{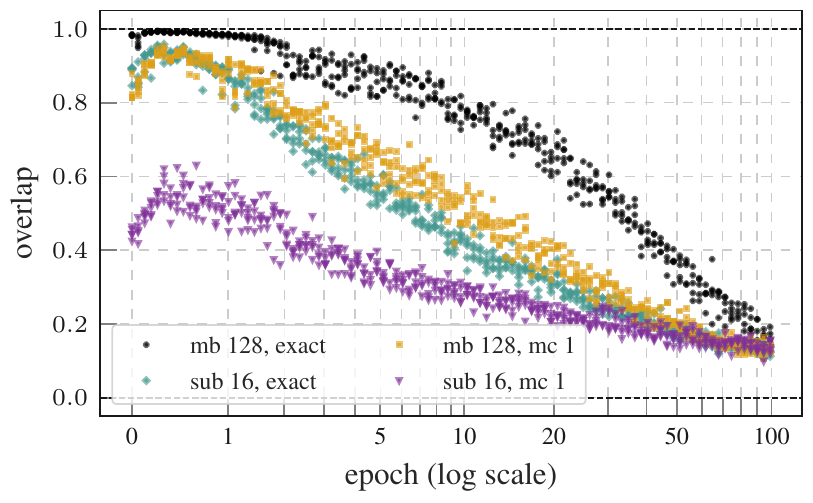}

\caption{
\textbf{Approximations vs. full-batch \ggn{}:}
Overlap between the top-$C$ eigenspaces of the mini-batch \ggn{}, \vivit{}'s approximations and the full-batch \ggn  during training of the \threecthreed network on \cifarten with \sgd{}.
Each approximation is evaluated on $5$ mini-batches.
}
\label{fig:approx_eigenspace_vivit}
\end{figure}

In order to allow for a fine-grained cost-accuracy trade-off, \vivit introduces
\textit{further} approximations to the mini-batch \ggn{} (see \Cref{sec:approximations}).
\Cref{fig:approx_eigenspace_vivit} shows the overlap between these \ggn approximations and the full-batch \ggn{}.\footnote{
A comparison with the mini-batch \ggn as ground truth can be found in \Cref{sec:eigenspace_noise}
}
The order of the approximations is as expected: With increasing computational effort, the approximations improve and, despite the greatly reduced computational effort compared to the exact mini-batch \ggn{}, significant structure of the top-$C$ eigenspace is preserved.
Details and results for the other test problems are reported in \Cref{sec:eigenspace_noise}.

So far, our analysis is based on the top-$C$ eigenspace of the curvature matrices.
We extend this analysis by studying the effect of noise and approximations on the curvature \textit{magnitude} along the top-$C$ directions in \Cref{sec:curvature_noise}.

\subsection{Per-sample directional derivatives}
\label{subsec:directional_derivatives}

A unique feature of \vivit{}'s quantities is that they provide a notion of
\textit{curvature uncertainty} through \textit{per-sample} first- and
second-order directional derivatives
(see \Cref{eq:gammas-lambdas}).
To quantify noise in the directional derivatives, we compute their signal-to-noise ratios (SNRs). For each direction $\ve_k$, the SNR is given by the squared empirical mean divided by the empirical variance of the $N$ mini-batch samples $\{\gamma_{nk}\}_{n=1}^N$ and $\{\lambda_{nk}\}_{n=1}^N$, respectively.

\begin{figure}
\centering
\includegraphics[scale=1.0]{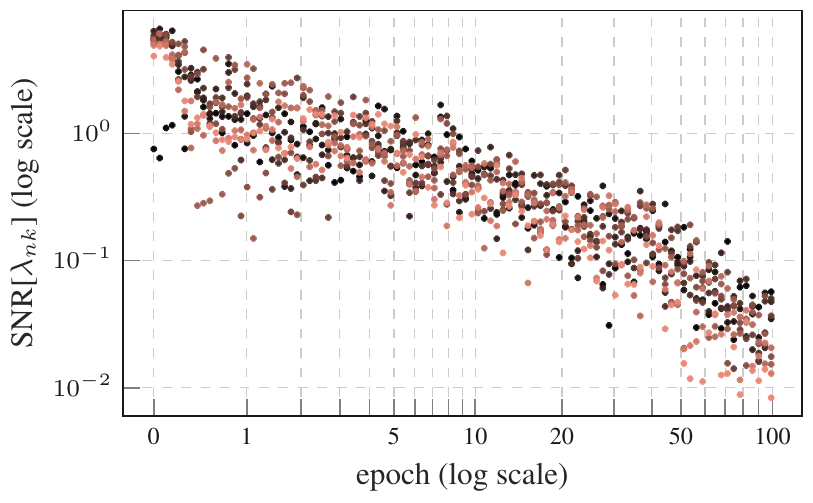}
\caption{
\textbf{Directional curvature SNRs:}
Curvature SNRs along each of the mini-batch \ggn{}'s top-$C$ eigenvectors during training of the \threecthreed network on \cifarten with \sgd{}.
At fixed epoch, the SNR for the most curved direction is shown in
\tikz\draw[white,fill={light_red},line width=0mm] (0,0) circle (.8ex);
and the SNR for the direction with the smallest curvature is shown in
\tikz\draw[white,fill={black}] (0,0) circle (.8ex);.
}
\label{fig:directional_derivatives}
\end{figure}

\Cref{fig:directional_derivatives} shows curvature SNRs
during training the \threecthreed network on \cifarten with \sgd.
The curvature signal along the top-$C$ eigenvectors decreases from $\text{SNR} > 1$ by two orders of magnitude.
In comparison, the directional gradients do not exhibit such a pattern (see \Cref{sec:directional_derivatives}).
Results for the other test cases can be found in \Cref{sec:directional_derivatives}.

In this section, we have given a glimpse of the \textit{very rich} quantities that can be efficiently computed under the \vivit concept. In \Cref{sec:use_cases}, we discuss the practical use of those quantities -- curvature uncertainty in particular.

\section{Related work}
\label{sec:related}

\paragraph{\ggn spectrum \& low-rank structure:} Other works point out the
\ggn's low-rank structure. \citet{botev2017practical} present the rank bound and
propose an alternative to \kfac based on backpropagating a decomposition of the
loss Hessian. \citet{papyan2019measurements} presents the factorization in
\Cref{eq:ggn-factorization} and studies the eigenvalue spectrum's
hierarchy for cross-entropy loss. In this setting, the \ggn further decomposes
into summands, some of which are then analyzed through similar Gram matrices.
These can be obtained as contractions of $\mGtilde$, but our approach goes
beyond them as it does not neglect terms. We are not aware of works that
obtain the exact spectrum \emph{and} leverage a highly-efficient
fully-parallel implementation. This may be because, until recently
\citep{bradbury2020jax, dangel2020backpack}, vectorized Jacobians
required to perform those operations efficiently were not available.

\paragraph{Efficient operations with low-rank matrices in deep learning:}
\citet{chen2021fast} use \Cref{eq:ggn-factorization} for element-wise
evaluation of the \ggn in fully-connected feed-forward neural networks. They
also present a variant based on \mc sampling. This element-wise evaluation is
then used to construct hierarchical matrix approximations of the \ggn. \vivit
instead leverages the global low-rank structure that also enjoys efficient
eigen-decomposition.

Another prominent low-rank matrix in deep learning is the un-centered gradient
covariance (sometimes called empirical Fisher). \citet{singh2020woodfisher}
describe implicit multiplication with its inverse and apply it for
neural network compression, assuming the empirical Fisher as Hessian proxy.
However, this assumption has limitations, specifically for optimization
\citep{kunstner2019limitations}. In principle though, the low-rank structure
also permits the application of our methods from \Cref{sec:method}.

\section{Use cases}
\label{sec:use_cases}

Aiming to provide a well-founded, theoretical and empirical evaluation, we have
here consciously focused on studying the approximation quality of \vivit's
quantities, as well as on demonstrating the efficiency of their
computation. We believe it is interesting in itself that the low-rank structure
provides access to quantities that would previously have been costly. 
Here, we still want to briefly address possible use cases -- their full development and assessment, however, will amount to separate paper(s). They include:
\begin{itemize}
\item \textbf{Monitoring tool:} Our computationally efficient curvature model
  provides geometric \textit{and} stochastic information about the local loss
  landscape and can be used by tools like \cockpit \citep{schneider2021cockpit}
  to debug optimizers or to gain insights into the optimization problem itself
  (as in \Cref{subsec:approx_quality,subsec:directional_derivatives}).

\item \textbf{Second-order optimization:} The quantities provided by \vivit{},
  in particular the first- and second-order directional derivatives, can be used
  to build a stochastic quadratic model of the loss function and perform
  Newton-like parameter updates. In contrast to existing second-order methods,
  \textit{per-sample} quantities contain information about the reliability of
  that quadratic model. This offers a new dimension for improving second-order
  methods through statistics on the mini-batch \textit{distribution} of the
  directional derivatives (e.g. for variance-adapted step sizes), potentially
  increasing the method's performance and stability.
\end{itemize}

\section{Conclusion}
\label{sec:conclusion}

We have presented \vivit, a curvature model based on the low-rank structure of
the Hessian's generalized Gauss-Newton (\ggn) approximation. This structure
allows for efficient extraction of \textit{exact} curvature properties, such as
the \ggn{}'s full eigenvalue spectrum and directional gradients and curvatures
along the associated eigenvectors. \vivit's quantities scale by approximations
that allow for a fine-grained cost-accuracy trade-off. In contrast to
alternatives, these quantities offer a notion of curvature uncertainty across
the mini-batch in the form of directional derivatives.

We empirically demonstrated the efficiency of leveraging the \ggn's low-rank
structure and substantiated its usefulness by studying characteristics of
curvature noise on various deep learning architectures.

The low-rank representation is efficiently computed in parallel with gradients
during a single backward pass. As it mainly relies on vectorized Jacobians, it
is general enough to be integrated into existing machine learning libraries in
the future. For now, we provide an efficient open-source implementation in
\pytorch \cite{paszke2019pytorch} by extending the existing \backpack
\cite{dangel2020backpack} library.

\bibliographystyle{icml2021}

\clearpage
\appendix

\renewcommand{\theequation}{S.\arabic{equation}}

\renewcommand{\thefigure}{S.\arabic{figure}}

\renewcommand{\thetable}{S.\arabic{table}}

\onecolumn

\begin{center}
\icmltitle{\vivitpapertitle \\[3mm] \large Supplementary Material}
\end{center}

\section{Mathematical details}
\label{sec:math-details}
\subsection{Reducing the \ggn eigenvalue problem to the Gram matrix}
\label{sec:relation-ggn-gram-eigenvalues}
For \Cref{eq:ggn-eigenvalues}, consider the left hand side of the \ggn's
characteristic polynomial $\det(\mG - \lambda \mI_D) = 0$. Inserting the \vivit
factorization (\Cref{eq:ggn-factorization}) and using the matrix determinant
lemma yields
\begin{align*}
  \det\!\left(
  - \lambda \mI_D + \mG
  \right)
  &=
    \det\!\left(
    - \lambda \mI_D + \mV\mV^\top
    \right)
    \explainmath{(Low-rank structure (\ref{eq:ggn-factorization}))}
  \\
  &=
    \det\!\left(
    \mI_{NC} + \mV^\top (-\lambda \mI_D)^{-1} \mV
    \right)
    \det\!\left(
    -\lambda \mI_D
    \right)
    \explainmath{(Matrix determinant lemma)}
  \\
  &=
    \det\!\left(
    \mI_{NC} - \frac{1}{\lambda} \mV^\top \mV
    \right)
    (-\lambda)^D
    \explainmath{\empty}
  \\
  &=
    \left(-\frac{1}{\lambda}\right)^{NC}
    \det\!\left(
    \mV^\top \mV - \lambda \mI_{NC}
    \right)
    (-\lambda)^D
    \explainmath{\empty}
  \\
  &=
    (-\lambda)^{D - NC}
    \det\!\left(
    \mGtilde - \lambda \mI_{NC}
    \right)\,.
    \explainmath{(Gram matrix)}
\end{align*}
Setting the above expression to zero reveals that the \ggn's spectrum decomposes
into $D-NC$ zero eigenvalues and the Gram matrix spectrum obtained from
$\det(\mGtilde - \lambda \mI_{NC} )= 0$.

\subsection{Relation between \ggn and Gram matrix eigenvectors}
\label{sec:relation-ggn-gram-eigenvectors}

Assume the nontrivial Gram matrix spectrum $\tilde{\sS}_+ = \{(\lambda_k,
\vetilde_k)\:|\: \lambda_k \neq 0, \mGtilde \vetilde_k = \lambda_k \vetilde_k
\}_{k=1}^K$ with orthonormal eigenvectors $\vetilde_j^\top \vetilde_k =
\delta_{jk}$ ($\delta$ represents the Kronecker delta) and $K =
\mathrm{rank}(\mG)$. We now show that $\ve_k = \nicefrac{1}{\sqrt{\lambda_k}}
\mV \vetilde_k$ are normalized eigenvectors of $\mG$ and inherit orthogonality
from $\vetilde_k$.

To see the first, consider right-multiplication of the \ggn with $\ve_k$, then
expand the low-rank structure,
\begin{align*}
  \mG \ve_k
  &=
    \frac{1}{\sqrt{\lambda_k}}
    \mV \mV^\top  \mV \vetilde_k
    \explainmath{(\Cref{eq:ggn-factorization} and definition of $\ve_k$)}
  \\
  &=
    \frac{1}{\sqrt{\lambda_k}}
    \mV \mGtilde \vetilde_k
    \explainmath{(Gram matrix)}
  \\
  &=
    \lambda_k\frac{1}{\sqrt{\lambda_k}}
    \mV \vetilde_k
    \explainmath{(Eigenvector property of $\vetilde_k$)}
  \\
  &= \lambda_k \ve_k\,.
\end{align*}
Orthonormality of the $\ve_k$ results from the Gram matrix eigenvector
orthonormality,
\begin{align*}
  \ve_j^\top \ve_{k}
  &=
    \left(
    \frac{1}{\sqrt{\lambda_j}} \vetilde_j^\top \mV ^\top
    \right)
    \left(
    \frac{1}{\sqrt{\lambda_k}} \mV \vetilde_k
    \right)
    \explainmath{(Definition of $\ve_j, \ve_k$)}
  \\
  &=
    \frac{1}{\sqrt{\lambda_j\lambda_k}} \vetilde_j^\top \mGtilde \vetilde_k
    \explainmath{(Gram matrix)}
  \\
  &=
    \frac{\lambda_k}{\sqrt{\lambda_j\lambda_k}} \vetilde_j^\top \vetilde_k
    \explainmath{(Eigenvector property of $\vetilde_k$)}
  \\
  &=
    \delta_{jk}\,.
    \explainmath{(Orthonormality)}
\end{align*}

\section{Experimental details}
\label{sec:experimental-details}

Throughout this section, we use the notation introduced in \Cref{sec:experiments} (see \Cref{tab:notation_cases}).
\begin{table}[ht]
  \centering
  \caption{
\textbf{Notation for curvature approximations:}
The notation is introduced in \Cref{sec:experiments}.
This table recapitulates the abbreviations (referring to the approximations introduced in \Cref{sec:approximations}) and provides corresponding explanations. 
}
  \label{tab:notation_cases}
  \vspace{1ex}
  \begin{normalsize}
    \begin{tabular}{ll}
      \toprule
      Abbreviation
      & Explanation \\
      \midrule
      \textbf{mb, exact}
      & \makecell[tl]{Exact \ggn with all mini-batch samples.\\
      Backpropagates $NC$ vectors.}
      \\
      \textbf{mb, mc}
      & \makecell[tl]{ \mc-approximated \ggn with all mini-batch samples.\\
      Backpropagates $N M$ vectors with $M$ the number of \mc{}-samples.}
      \\
      \textbf{sub, exact}
      & \makecell[tl]{Exact \ggn on a subset of mini-batch samples ($\floor{\nicefrac{N}{8}}$ as in \cite{zhang2017blockdiagonal}).\\
      Backpropagates $\floor{\nicefrac{N}{8}} C$ vectors.}
      \\
      \textbf{sub, mc}
      & \makecell[tl]{\mc-approximated \ggn on a subset of mini-batch samples.\\
      Backpropagates $\floor{\nicefrac{N}{8}} M$ vectors with $M$ the number of \mc{}-samples.}
      \\
      \bottomrule
    \end{tabular}
  \end{normalsize}
\end{table}

\paragraph{\ggn spectra (\Cref{fig:visual_abstract}a):} To obtain the spectra of \Cref{fig:visual_abstract}a
we initialize the respective architecture, then draw a mini-batch and evaluate
the \ggn eigenvalues under the described approximations, clipping the Gram
matrix eigenvalues at $10^{-4}$. Figures \ref{fig:spectrum_1} and \ref{fig:spectrum_2} provide the spectra for
all used architectures with both the full \ggn and a per-layer block-diagonal
approximation.

\begin{figure}[p]
  \centering
  
  \begin{small}
    \textbf{\fmnist \twoctwod}
  \end{small}

  \begin{minipage}{0.495\linewidth}
    \centering
    \begin{small}
      \textbf{Full network}
    \end{small}
    \vspace{1ex}

    \begin{minipage}[t]{0.33\linewidth}
      \centering
\pgfkeys{/pgfplots/spectrumdefault/.style={
    width=1.38\linewidth,
    height=\goldenRatio*1.38\linewidth,
    every axis plot/.append style={line width = 1.2pt},
    every axis plot post/.append style={
      mark size=1, mark options={opacity=0.3}
    },
    tick pos = left,
    xmajorticks = true,
    ymajorticks = true,
    ylabel near ticks,
    xlabel near ticks,
    xtick align = inside,
    ytick align = inside,
    legend cell align = left,
    legend columns = 1,
    legend pos = north west,
    legend style = {
      fill opacity = 0.9,
      text opacity = 1,
      font = \small,
    },
    xticklabel style = {font = \small, inner xsep = 0ex},
    xlabel style = {font = \small},
    axis line style = {black},
    yticklabel style = {font = \small, inner ysep = 0ex},
    ylabel style = {font = \small, inner ysep = 0ex},
    title style = {font = \small, inner ysep = 0ex, yshift = -0.75ex},
    grid = major,
    grid style = {dashed}
  }
}
\pgfkeys{/pgfplots/spectrumdefaultleft/.style={
    spectrumdefault,
    title=\empty,
    ymax=3e-4,
    xlabel=\phantom{eigenvalues},
  }}
\pgfkeys{/pgfplots/spectrumdefaultcenter/.style={
    spectrumdefault,
    ylabel=\empty,
    yticklabels=\empty,
    title=\empty,
    ymax=3e-4,
  }}
\pgfkeys{/pgfplots/spectrumdefaultright/.style={
    spectrumdefault,
    ylabel=\empty,
    yticklabels=\empty,
    title=\empty,
    ymax=3e-4,
    xlabel=\phantom{eigenvalues},
  }}
 \pgfkeys{/pgfplots/zmystyle/.style={spectrumdefaultleft,
          title={exact, mb}}}
      \begin{tikzpicture}

\begin{axis}[
axis line style={white!80!black},
log basis x={10},
tick pos=left,
title={full\_batch\_exact, one\_group, N=128, D=3274634},
xlabel={eigenvalues},
xmin=0.0001, xmax=61.7406616210938,
xmode=log,
ylabel={density},
ymin=3.05377545894428e-07, ymax=1,
ymode=log,
zmystyle
]
\draw[draw=white,fill=black] (axis cs:0.0001,3.05377545894428e-07) rectangle (axis cs:0.000158370712903008,0.999648510336027);
\draw[draw=white,fill=black] (axis cs:0.000158370712903008,3.05377545894428e-07) rectangle (axis cs:0.000250812827054069,3.05377545894428e-07);
\draw[draw=white,fill=black] (axis cs:0.000250812827054069,3.05377545894428e-07) rectangle (axis cs:0.000397214062257717,3.05377545894428e-07);
\draw[draw=white,fill=black] (axis cs:0.000397214062257717,3.05377545894428e-07) rectangle (axis cs:0.000629070742148544,3.05377545894428e-07);
\draw[draw=white,fill=black] (axis cs:0.000629070742148544,3.05377545894428e-07) rectangle (axis cs:0.000996263819004891,6.10755185070358e-07);
\draw[draw=white,fill=black] (axis cs:0.000996263819004891,3.05377545894428e-07) rectangle (axis cs:0.00157779011255278,9.16132908121924e-06);
\draw[draw=white,fill=black] (axis cs:0.00157779011255278,3.05377545894428e-07) rectangle (axis cs:0.002498757449363,3.51184284087308e-05);
\draw[draw=white,fill=black] (axis cs:0.002498757449363,3.05377545894428e-07) rectangle (axis cs:0.0039572999862732,5.52733525925658e-05);
\draw[draw=white,fill=black] (axis cs:0.0039572999862732,3.05377545894428e-07) rectangle (axis cs:0.0062672041999715,5.74109960665753e-05);
\draw[draw=white,fill=black] (axis cs:0.0062672041999715,3.05377545894428e-07) rectangle (axis cs:0.0099254159705821,5.13034432836118e-05);
\draw[draw=white,fill=black] (axis cs:0.0099254159705821,3.05377545894428e-07) rectangle (axis cs:0.0157189520311999,4.27528693875739e-05);
\draw[draw=white,fill=black] (axis cs:0.0157189520311999,3.05377545894428e-07) rectangle (axis cs:0.0248942163926931,3.23700296563695e-05);
\draw[draw=white,fill=black] (axis cs:0.0248942163926931,3.05377545894428e-07) rectangle (axis cs:0.0394251479727254,2.25979452037389e-05);
\draw[draw=white,fill=black] (axis cs:0.0394251479727254,3.05377545894428e-07) rectangle (axis cs:0.062437887907471,1.55742595033587e-05);
\draw[draw=white,fill=black] (axis cs:0.0624378879074709,3.05377545894428e-07) rectangle (axis cs:0.0988833282006426,1.12989725552287e-05);
\draw[draw=white,fill=black] (axis cs:0.0988833282006426,3.05377545894428e-07) rectangle (axis cs:0.156602231813579,6.41293032885792e-06);
\draw[draw=white,fill=black] (axis cs:0.156602231813579,3.05377545894428e-07) rectangle (axis cs:0.248012070945185,4.88604213308929e-06);
\draw[draw=white,fill=black] (axis cs:0.248012070945185,3.05377545894428e-07) rectangle (axis cs:0.392778484841403,3.05377629825575e-06);
\draw[draw=white,fill=black] (axis cs:0.392778484841403,3.05377545894428e-07) rectangle (axis cs:0.622046086572963,3.05377629814473e-06);
\draw[draw=white,fill=black] (axis cs:0.622046086572963,3.05377545894428e-07) rectangle (axis cs:0.985138821890862,1.83226574166306e-06);
\draw[draw=white,fill=black] (axis cs:0.985138821890862,3.05377545894428e-07) rectangle (axis cs:1.56017137531285,1.52688810248713e-06);
\draw[draw=white,fill=black] (axis cs:1.56017137531285,3.05377545894428e-07) rectangle (axis cs:2.47085452959162,3.05377545894428e-07);
\draw[draw=white,fill=black] (axis cs:2.47085452959162,3.05377545894428e-07) rectangle (axis cs:3.91310993331051,3.05377545894428e-07);
\draw[draw=white,fill=black] (axis cs:3.91310993331051,3.05377545894428e-07) rectangle (axis cs:6.19722009806228,3.05377545894428e-07);
\draw[draw=white,fill=black] (axis cs:6.19722009806228,3.05377545894428e-07) rectangle (axis cs:9.8145816494697,6.10755185070358e-07);
\draw[draw=white,fill=black] (axis cs:9.8145816494697,3.05377545894428e-07) rectangle (axis cs:15.5434229267129,1.83226574166306e-06);
\draw[draw=white,fill=black] (axis cs:15.5434229267129,3.05377545894428e-07) rectangle (axis cs:24.6162296985648,9.16132824135266e-07);
\draw[draw=white,fill=black] (axis cs:24.6162296985648,3.05377545894428e-07) rectangle (axis cs:38.9848984634591,6.10755185070358e-07);
\draw[draw=white,fill=black] (axis cs:38.9848984634591,3.05377545894428e-07) rectangle (axis cs:61.7406616210938,3.05377545894428e-07);
\end{axis}

\end{tikzpicture}
     \end{minipage}
    \hspace{1.9ex}
    \begin{minipage}[t]{0.33\linewidth}
      \centering
\pgfkeys{/pgfplots/spectrumdefault/.style={
    width=1.38\linewidth,
    height=\goldenRatio*1.38\linewidth,
    every axis plot/.append style={line width = 1.2pt},
    every axis plot post/.append style={
      mark size=1, mark options={opacity=0.3}
    },
    tick pos = left,
    xmajorticks = true,
    ymajorticks = true,
    ylabel near ticks,
    xlabel near ticks,
    xtick align = inside,
    ytick align = inside,
    legend cell align = left,
    legend columns = 1,
    legend pos = north west,
    legend style = {
      fill opacity = 0.9,
      text opacity = 1,
      font = \small,
    },
    xticklabel style = {font = \small, inner xsep = 0ex},
    xlabel style = {font = \small},
    axis line style = {black},
    yticklabel style = {font = \small, inner ysep = 0ex},
    ylabel style = {font = \small, inner ysep = 0ex},
    title style = {font = \small, inner ysep = 0ex, yshift = -0.75ex},
    grid = major,
    grid style = {dashed}
  }
}
\pgfkeys{/pgfplots/spectrumdefaultleft/.style={
    spectrumdefault,
    title=\empty,
    ymax=3e-4,
    xlabel=\phantom{eigenvalues},
  }}
\pgfkeys{/pgfplots/spectrumdefaultcenter/.style={
    spectrumdefault,
    ylabel=\empty,
    yticklabels=\empty,
    title=\empty,
    ymax=3e-4,
  }}
\pgfkeys{/pgfplots/spectrumdefaultright/.style={
    spectrumdefault,
    ylabel=\empty,
    yticklabels=\empty,
    title=\empty,
    ymax=3e-4,
    xlabel=\phantom{eigenvalues},
  }}
 \pgfkeys{/pgfplots/zmystyle/.style={spectrumdefaultcenter,
          title={exact, sub}}}
      \begin{tikzpicture}

\definecolor{color0}{rgb}{0.274509803921569,0.6,0.564705882352941}

\begin{axis}[
axis line style={white!80!black},
log basis x={10},
tick pos=left,
title={frac\_batch\_exact, one\_group, N=128, D=3274634},
xlabel={eigenvalues},
xmin=0.0001, xmax=61.7406616210938,
xmode=log,
ylabel={density},
ymin=3.05377545894428e-07, ymax=1,
ymode=log,
zmystyle
]
\draw[draw=white,fill=color0] (axis cs:0.0001,3.05377545894428e-07) rectangle (axis cs:0.000158370712903008,0.999956330996288);
\draw[draw=white,fill=color0] (axis cs:0.000158370712903008,3.05377545894428e-07) rectangle (axis cs:0.000250812827054069,3.05377545894428e-07);
\draw[draw=white,fill=color0] (axis cs:0.000250812827054069,3.05377545894428e-07) rectangle (axis cs:0.000397214062257717,3.05377545894428e-07);
\draw[draw=white,fill=color0] (axis cs:0.000397214062257717,3.05377545894428e-07) rectangle (axis cs:0.000629070742148544,3.05377545894428e-07);
\draw[draw=white,fill=color0] (axis cs:0.000629070742148544,3.05377545894428e-07) rectangle (axis cs:0.000996263819004891,3.05377545894428e-07);
\draw[draw=white,fill=color0] (axis cs:0.000996263819004891,3.05377545894428e-07) rectangle (axis cs:0.00157779011255278,3.05377545894428e-07);
\draw[draw=white,fill=color0] (axis cs:0.00157779011255278,3.05377545894428e-07) rectangle (axis cs:0.002498757449363,3.05377545894428e-07);
\draw[draw=white,fill=color0] (axis cs:0.002498757449363,3.05377545894428e-07) rectangle (axis cs:0.0039572999862732,3.05377545894428e-07);
\draw[draw=white,fill=color0] (axis cs:0.0039572999862732,3.05377545894428e-07) rectangle (axis cs:0.0062672041999715,3.05377545894428e-07);
\draw[draw=white,fill=color0] (axis cs:0.0062672041999715,3.05377545894428e-07) rectangle (axis cs:0.0099254159705821,3.05377545894428e-07);
\draw[draw=white,fill=color0] (axis cs:0.0099254159705821,3.05377545894428e-07) rectangle (axis cs:0.0157189520311999,3.05377545894428e-07);
\draw[draw=white,fill=color0] (axis cs:0.0157189520311999,3.05377545894428e-07) rectangle (axis cs:0.0248942163926931,9.16132824135266e-07);
\draw[draw=white,fill=color0] (axis cs:0.0248942163926931,3.05377545894428e-07) rectangle (axis cs:0.0394251479727254,3.05377629825575e-06);
\draw[draw=white,fill=color0] (axis cs:0.0394251479727254,3.05377545894428e-07) rectangle (axis cs:0.062437887907471,6.10755268968199e-06);
\draw[draw=white,fill=color0] (axis cs:0.0624378879074709,3.05377545894428e-07) rectangle (axis cs:0.0988833282006426,9.16132908121924e-06);
\draw[draw=white,fill=color0] (axis cs:0.0988833282006426,3.05377545894428e-07) rectangle (axis cs:0.156602231813579,8.24519616380247e-06);
\draw[draw=white,fill=color0] (axis cs:0.156602231813579,3.05377545894428e-07) rectangle (axis cs:0.248012070945185,6.71830796792282e-06);
\draw[draw=white,fill=color0] (axis cs:0.248012070945185,3.05377545894428e-07) rectangle (axis cs:0.392778484841403,3.35915393743168e-06);
\draw[draw=white,fill=color0] (axis cs:0.392778484841403,3.05377545894428e-07) rectangle (axis cs:0.622046086572963,2.74839865907982e-06);
\draw[draw=white,fill=color0] (axis cs:0.622046086572963,3.05377545894428e-07) rectangle (axis cs:0.985138821890862,2.44302101990389e-06);
\draw[draw=white,fill=color0] (axis cs:0.985138821890862,3.05377545894428e-07) rectangle (axis cs:1.56017137531285,1.52688810248713e-06);
\draw[draw=white,fill=color0] (axis cs:1.56017137531285,3.05377545894428e-07) rectangle (axis cs:2.47085452959162,3.05377545894428e-07);
\draw[draw=white,fill=color0] (axis cs:2.47085452959162,3.05377545894428e-07) rectangle (axis cs:3.91310993331051,3.05377545894428e-07);
\draw[draw=white,fill=color0] (axis cs:3.91310993331051,3.05377545894428e-07) rectangle (axis cs:6.19722009806228,3.05377545894428e-07);
\draw[draw=white,fill=color0] (axis cs:6.19722009806228,3.05377545894428e-07) rectangle (axis cs:9.8145816494697,9.16132824246288e-07);
\draw[draw=white,fill=color0] (axis cs:9.8145816494697,3.05377545894428e-07) rectangle (axis cs:15.5434229267129,1.83226574155203e-06);
\draw[draw=white,fill=color0] (axis cs:15.5434229267129,3.05377545894428e-07) rectangle (axis cs:24.6162296985648,6.10755185070358e-07);
\draw[draw=white,fill=color0] (axis cs:24.6162296985648,3.05377545894428e-07) rectangle (axis cs:38.9848984634591,6.10755185070358e-07);
\draw[draw=white,fill=color0] (axis cs:38.9848984634591,3.05377545894428e-07) rectangle (axis cs:61.7406616210938,3.05377545894428e-07);
\end{axis}

\end{tikzpicture}
     \end{minipage}
    \hspace{-3.5ex}
    \begin{minipage}[t]{0.33\linewidth}
      \centering
\pgfkeys{/pgfplots/spectrumdefault/.style={
    width=1.38\linewidth,
    height=\goldenRatio*1.38\linewidth,
    every axis plot/.append style={line width = 1.2pt},
    every axis plot post/.append style={
      mark size=1, mark options={opacity=0.3}
    },
    tick pos = left,
    xmajorticks = true,
    ymajorticks = true,
    ylabel near ticks,
    xlabel near ticks,
    xtick align = inside,
    ytick align = inside,
    legend cell align = left,
    legend columns = 1,
    legend pos = north west,
    legend style = {
      fill opacity = 0.9,
      text opacity = 1,
      font = \small,
    },
    xticklabel style = {font = \small, inner xsep = 0ex},
    xlabel style = {font = \small},
    axis line style = {black},
    yticklabel style = {font = \small, inner ysep = 0ex},
    ylabel style = {font = \small, inner ysep = 0ex},
    title style = {font = \small, inner ysep = 0ex, yshift = -0.75ex},
    grid = major,
    grid style = {dashed}
  }
}
\pgfkeys{/pgfplots/spectrumdefaultleft/.style={
    spectrumdefault,
    title=\empty,
    ymax=3e-4,
    xlabel=\phantom{eigenvalues},
  }}
\pgfkeys{/pgfplots/spectrumdefaultcenter/.style={
    spectrumdefault,
    ylabel=\empty,
    yticklabels=\empty,
    title=\empty,
    ymax=3e-4,
  }}
\pgfkeys{/pgfplots/spectrumdefaultright/.style={
    spectrumdefault,
    ylabel=\empty,
    yticklabels=\empty,
    title=\empty,
    ymax=3e-4,
    xlabel=\phantom{eigenvalues},
  }}
 \pgfkeys{/pgfplots/zmystyle/.style={spectrumdefaultright,
          title={mc, full}}}
      \begin{tikzpicture}

\definecolor{color0}{rgb}{0.870588235294118,0.623529411764706,0.0862745098039216}

\begin{axis}[
axis line style={white!80!black},
log basis x={10},
tick pos=left,
title={full\_batch\_mc, one\_group, N=128, D=3274634},
xlabel={eigenvalues},
xmin=0.0001, xmax=61.7406616210938,
xmode=log,
ylabel={density},
ymin=3.05377545894428e-07, ymax=1,
ymode=log,
zmystyle
]
\draw[draw=white,fill=color0] (axis cs:0.0001,3.05377545894428e-07) rectangle (axis cs:0.000158370712903008,0.999961217038515);
\draw[draw=white,fill=color0] (axis cs:0.000158370712903008,3.05377545894428e-07) rectangle (axis cs:0.000250812827054069,3.05377545894428e-07);
\draw[draw=white,fill=color0] (axis cs:0.000250812827054069,3.05377545894428e-07) rectangle (axis cs:0.000397214062257717,3.05377545894428e-07);
\draw[draw=white,fill=color0] (axis cs:0.000397214062257717,3.05377545894428e-07) rectangle (axis cs:0.000629070742148544,3.05377545894428e-07);
\draw[draw=white,fill=color0] (axis cs:0.000629070742148544,3.05377545894428e-07) rectangle (axis cs:0.000996263819004891,3.05377545894428e-07);
\draw[draw=white,fill=color0] (axis cs:0.000996263819004891,3.05377545894428e-07) rectangle (axis cs:0.00157779011255278,3.05377545894428e-07);
\draw[draw=white,fill=color0] (axis cs:0.00157779011255278,3.05377545894428e-07) rectangle (axis cs:0.002498757449363,3.05377545894428e-07);
\draw[draw=white,fill=color0] (axis cs:0.002498757449363,3.05377545894428e-07) rectangle (axis cs:0.0039572999862732,3.05377545894428e-07);
\draw[draw=white,fill=color0] (axis cs:0.0039572999862732,3.05377545894428e-07) rectangle (axis cs:0.0062672041999715,3.05377545894428e-07);
\draw[draw=white,fill=color0] (axis cs:0.0062672041999715,3.05377545894428e-07) rectangle (axis cs:0.0099254159705821,3.05377545894428e-07);
\draw[draw=white,fill=color0] (axis cs:0.0099254159705821,3.05377545894428e-07) rectangle (axis cs:0.0157189520311999,3.05377545894428e-07);
\draw[draw=white,fill=color0] (axis cs:0.0157189520311999,3.05377545894428e-07) rectangle (axis cs:0.0248942163926931,3.05377545894428e-07);
\draw[draw=white,fill=color0] (axis cs:0.0248942163926931,3.05377545894428e-07) rectangle (axis cs:0.0394251479727254,3.05377545894428e-07);
\draw[draw=white,fill=color0] (axis cs:0.0394251479727254,3.05377545894428e-07) rectangle (axis cs:0.062437887907471,2.44302101990389e-06);
\draw[draw=white,fill=color0] (axis cs:0.0624378879074709,3.05377545894428e-07) rectangle (axis cs:0.0988833282006426,6.10755268968199e-06);
\draw[draw=white,fill=color0] (axis cs:0.0988833282006426,3.05377545894428e-07) rectangle (axis cs:0.156602231813579,9.16132908121924e-06);
\draw[draw=white,fill=color0] (axis cs:0.156602231813579,3.05377545894428e-07) rectangle (axis cs:0.248012070945185,7.32906324627468e-06);
\draw[draw=white,fill=color0] (axis cs:0.248012070945185,3.05377545894428e-07) rectangle (axis cs:0.392778484841403,5.49679741144115e-06);
\draw[draw=white,fill=color0] (axis cs:0.392778484841403,3.05377545894428e-07) rectangle (axis cs:0.622046086572963,3.66453157649659e-06);
\draw[draw=white,fill=color0] (axis cs:0.622046086572963,3.05377545894428e-07) rectangle (axis cs:0.985138821890862,2.74839865907982e-06);
\draw[draw=white,fill=color0] (axis cs:0.985138821890862,3.05377545894428e-07) rectangle (axis cs:1.56017137531285,1.52688810248713e-06);
\draw[draw=white,fill=color0] (axis cs:1.56017137531285,3.05377545894428e-07) rectangle (axis cs:2.47085452959162,6.10755185070358e-07);
\draw[draw=white,fill=color0] (axis cs:2.47085452959162,3.05377545894428e-07) rectangle (axis cs:3.91310993331051,3.05377545894428e-07);
\draw[draw=white,fill=color0] (axis cs:3.91310993331051,3.05377545894428e-07) rectangle (axis cs:6.19722009806228,3.05377545894428e-07);
\draw[draw=white,fill=color0] (axis cs:6.19722009806228,3.05377545894428e-07) rectangle (axis cs:9.8145816494697,1.2215104633112e-06);
\draw[draw=white,fill=color0] (axis cs:9.8145816494697,3.05377545894428e-07) rectangle (axis cs:15.5434229267129,1.22151046342222e-06);
\draw[draw=white,fill=color0] (axis cs:15.5434229267129,3.05377545894428e-07) rectangle (axis cs:24.6162296985648,9.16132824135266e-07);
\draw[draw=white,fill=color0] (axis cs:24.6162296985648,3.05377545894428e-07) rectangle (axis cs:38.9848984634591,6.10755185070358e-07);
\draw[draw=white,fill=color0] (axis cs:38.9848984634591,3.05377545894428e-07) rectangle (axis cs:61.7406616210938,3.05377545894428e-07);
\end{axis}

\end{tikzpicture}
     \end{minipage}
  \end{minipage}
  \hfill
  \begin{minipage}{0.495\linewidth}
    \centering
    \begin{small}
      \textbf{Block-diagonal approximation}
    \end{small}
    \vspace{1ex}

    \begin{minipage}[t]{0.33\linewidth}
      \centering
\pgfkeys{/pgfplots/spectrumdefault/.style={
    width=1.38\linewidth,
    height=\goldenRatio*1.38\linewidth,
    every axis plot/.append style={line width = 1.2pt},
    every axis plot post/.append style={
      mark size=1, mark options={opacity=0.3}
    },
    tick pos = left,
    xmajorticks = true,
    ymajorticks = true,
    ylabel near ticks,
    xlabel near ticks,
    xtick align = inside,
    ytick align = inside,
    legend cell align = left,
    legend columns = 1,
    legend pos = north west,
    legend style = {
      fill opacity = 0.9,
      text opacity = 1,
      font = \small,
    },
    xticklabel style = {font = \small, inner xsep = 0ex},
    xlabel style = {font = \small},
    axis line style = {black},
    yticklabel style = {font = \small, inner ysep = 0ex},
    ylabel style = {font = \small, inner ysep = 0ex},
    title style = {font = \small, inner ysep = 0ex, yshift = -0.75ex},
    grid = major,
    grid style = {dashed}
  }
}
\pgfkeys{/pgfplots/spectrumdefaultleft/.style={
    spectrumdefault,
    title=\empty,
    ymax=3e-4,
    xlabel=\phantom{eigenvalues},
  }}
\pgfkeys{/pgfplots/spectrumdefaultcenter/.style={
    spectrumdefault,
    ylabel=\empty,
    yticklabels=\empty,
    title=\empty,
    ymax=3e-4,
  }}
\pgfkeys{/pgfplots/spectrumdefaultright/.style={
    spectrumdefault,
    ylabel=\empty,
    yticklabels=\empty,
    title=\empty,
    ymax=3e-4,
    xlabel=\phantom{eigenvalues},
  }}
 \pgfkeys{/pgfplots/zmystyle/.style={spectrumdefaultleft, ymax = 5e-4,
          title={exact, mb}}}
      \begin{tikzpicture}

\begin{axis}[
axis line style={white!80!black},
log basis x={10},
tick pos=left,
title={full\_batch\_exact, layerwise\_group, N=128, D=3274634},
xlabel={eigenvalues},
xmin=0.0001, xmax=16.3758640289307,
xmode=log,
ylabel={density},
ymin=3.05377545894428e-07, ymax=1,
ymode=log,
zmystyle
]
\draw[draw=white,fill=black] (axis cs:0.0001,3.05377545894428e-07) rectangle (axis cs:0.000151286489554207,0.99875466998624);
\draw[draw=white,fill=black] (axis cs:0.000151286489554207,3.05377545894428e-07) rectangle (axis cs:0.000228876019216353,4.67227786964169e-05);
\draw[draw=white,fill=black] (axis cs:0.000228876019216353,3.05377545894428e-07) rectangle (axis cs:0.000346258494903833,7.32906333022804e-05);
\draw[draw=white,fill=black] (axis cs:0.000346258494903833,3.05377545894428e-07) rectangle (axis cs:0.000523842321723243,0.000123372566122692);
\draw[draw=white,fill=black] (axis cs:0.000523842321723243,3.05377545894428e-07) rectangle (axis cs:0.000792502659334351,0.000157269484068001);
\draw[draw=white,fill=black] (axis cs:0.000792502659334351,3.05377545894428e-07) rectangle (axis cs:0.00119894945293068,0.000164903925046733);
\draw[draw=white,fill=black] (axis cs:0.00119894945293068,3.05377545894428e-07) rectangle (axis cs:0.0018138485388682,0.000143832867945592);
\draw[draw=white,fill=black] (axis cs:0.0018138485388682,3.05377545894428e-07) rectangle (axis cs:0.00274410778028398,0.000123067188483405);
\draw[draw=white,fill=black] (axis cs:0.00274410778028398,3.05377545894428e-07) rectangle (axis cs:0.00415146433037551,9.77208444341343e-05);
\draw[draw=white,fill=black] (axis cs:0.00415146433037551,3.05377545894428e-07) rectangle (axis cs:0.00628060465052019,7.8787430807003e-05);
\draw[draw=white,fill=black] (axis cs:0.00628060465052019,3.05377545894428e-07) rectangle (axis cs:0.0095017062985503,6.10755277363534e-05);
\draw[draw=white,fill=black] (axis cs:0.0095017062985503,3.05377545894428e-07) rectangle (axis cs:0.0143747979068277,4.85550445313615e-05);
\draw[draw=white,fill=black] (axis cs:0.0143747979068277,3.05377545894428e-07) rectangle (axis cs:0.0217471271337514,3.54238060479067e-05);
\draw[draw=white,fill=black] (axis cs:0.0217471271337514,3.05377545894428e-07) rectangle (axis cs:0.032900465219543,2.59570992343411e-05);
\draw[draw=white,fill=black] (axis cs:0.032900465219543,3.05377545894428e-07) rectangle (axis cs:0.0497739588776495,1.77119029773681e-05);
\draw[draw=white,fill=black] (axis cs:0.0497739588776495,3.05377545894428e-07) rectangle (axis cs:0.0753012750981506,1.52688818640717e-05);
\draw[draw=white,fill=black] (axis cs:0.0753012750981507,3.05377545894428e-07) rectangle (axis cs:0.113920655685549,1.03828396378119e-05);
\draw[draw=white,fill=black] (axis cs:0.113920655685549,3.05377545894428e-07) rectangle (axis cs:0.172346560863802,8.24519616380247e-06);
\draw[draw=white,fill=black] (axis cs:0.172346560863802,3.05377545894428e-07) rectangle (axis cs:0.260737061798252,4.88604213308929e-06);
\draw[draw=white,fill=black] (axis cs:0.260737061798252,3.05377545894428e-07) rectangle (axis cs:0.394459947761359,2.74839865907982e-06);
\draw[draw=white,fill=black] (axis cs:0.394459947761359,3.05377545894428e-07) rectangle (axis cs:0.596764607665519,1.52688810248713e-06);
\draw[draw=white,fill=black] (axis cs:0.596764607665519,3.05377545894428e-07) rectangle (axis cs:0.902824225839103,9.16132824135266e-07);
\draw[draw=white,fill=black] (axis cs:0.902824225839103,3.05377545894428e-07) rectangle (axis cs:1.36585107811693,1.52688810248713e-06);
\draw[draw=white,fill=black] (axis cs:1.36585107811693,3.05377545894428e-07) rectangle (axis cs:2.06634814862139,1.52688810248713e-06);
\draw[draw=white,fill=black] (axis cs:2.06634814862139,3.05377545894428e-07) rectangle (axis cs:3.12610557601766,1.52688810248713e-06);
\draw[draw=white,fill=black] (axis cs:3.12610557601766,3.05377545894428e-07) rectangle (axis cs:4.72937538571545,3.35915393743168e-06);
\draw[draw=white,fill=black] (axis cs:4.72937538571545,3.05377545894428e-07) rectangle (axis cs:7.15490599888967,3.05377629825575e-06);
\draw[draw=white,fill=black] (axis cs:7.15490599888967,3.05377545894428e-07) rectangle (axis cs:10.8244061166236,1.2215104633112e-06);
\draw[draw=white,fill=black] (axis cs:10.8244061166236,3.05377545894428e-07) rectangle (axis cs:16.3758640289307,3.05377545894428e-07);
\end{axis}

\end{tikzpicture}
     \end{minipage}
    \hspace{1.9ex}
    \begin{minipage}[t]{0.33\linewidth}
      \centering
\pgfkeys{/pgfplots/spectrumdefault/.style={
    width=1.38\linewidth,
    height=\goldenRatio*1.38\linewidth,
    every axis plot/.append style={line width = 1.2pt},
    every axis plot post/.append style={
      mark size=1, mark options={opacity=0.3}
    },
    tick pos = left,
    xmajorticks = true,
    ymajorticks = true,
    ylabel near ticks,
    xlabel near ticks,
    xtick align = inside,
    ytick align = inside,
    legend cell align = left,
    legend columns = 1,
    legend pos = north west,
    legend style = {
      fill opacity = 0.9,
      text opacity = 1,
      font = \small,
    },
    xticklabel style = {font = \small, inner xsep = 0ex},
    xlabel style = {font = \small},
    axis line style = {black},
    yticklabel style = {font = \small, inner ysep = 0ex},
    ylabel style = {font = \small, inner ysep = 0ex},
    title style = {font = \small, inner ysep = 0ex, yshift = -0.75ex},
    grid = major,
    grid style = {dashed}
  }
}
\pgfkeys{/pgfplots/spectrumdefaultleft/.style={
    spectrumdefault,
    title=\empty,
    ymax=3e-4,
    xlabel=\phantom{eigenvalues},
  }}
\pgfkeys{/pgfplots/spectrumdefaultcenter/.style={
    spectrumdefault,
    ylabel=\empty,
    yticklabels=\empty,
    title=\empty,
    ymax=3e-4,
  }}
\pgfkeys{/pgfplots/spectrumdefaultright/.style={
    spectrumdefault,
    ylabel=\empty,
    yticklabels=\empty,
    title=\empty,
    ymax=3e-4,
    xlabel=\phantom{eigenvalues},
  }}
 \pgfkeys{/pgfplots/zmystyle/.style={spectrumdefaultcenter, ymax = 5e-4,
          title={exact, sub}}}
      \begin{tikzpicture}

\definecolor{color0}{rgb}{0.274509803921569,0.6,0.564705882352941}

\begin{axis}[
axis line style={white!80!black},
log basis x={10},
tick pos=left,
title={frac\_batch\_exact, layerwise\_group, N=128, D=3274634},
xlabel={eigenvalues},
xmin=0.0001, xmax=16.3758640289307,
xmode=log,
ylabel={density},
ymin=3.05377545894428e-07, ymax=1,
ymode=log,
zmystyle
]
\draw[draw=white,fill=color0] (axis cs:0.0001,3.05377545894428e-07) rectangle (axis cs:0.000151286489554207,0.999824407856176);
\draw[draw=white,fill=color0] (axis cs:0.000151286489554207,3.05377545894428e-07) rectangle (axis cs:0.000228876019216353,3.05377545894428e-07);
\draw[draw=white,fill=color0] (axis cs:0.000228876019216353,3.05377545894428e-07) rectangle (axis cs:0.000346258494903833,3.05377545894428e-07);
\draw[draw=white,fill=color0] (axis cs:0.000346258494903833,3.05377545894428e-07) rectangle (axis cs:0.000523842321723243,3.05377545894428e-07);
\draw[draw=white,fill=color0] (axis cs:0.000523842321723243,3.05377545894428e-07) rectangle (axis cs:0.000792502659334351,3.05377545894428e-07);
\draw[draw=white,fill=color0] (axis cs:0.000792502659334351,3.05377545894428e-07) rectangle (axis cs:0.00119894945293068,3.05377545894428e-07);
\draw[draw=white,fill=color0] (axis cs:0.00119894945293068,3.05377545894428e-07) rectangle (axis cs:0.0018138485388682,3.05377545894428e-07);
\draw[draw=white,fill=color0] (axis cs:0.0018138485388682,3.05377545894428e-07) rectangle (axis cs:0.00274410778028398,9.16132824135266e-07);
\draw[draw=white,fill=color0] (axis cs:0.00274410778028398,3.05377545894428e-07) rectangle (axis cs:0.00415146433037551,5.19141977226522e-06);
\draw[draw=white,fill=color0] (axis cs:0.00415146433037551,3.05377545894428e-07) rectangle (axis cs:0.00628060465052019,1.03828396378119e-05);
\draw[draw=white,fill=color0] (axis cs:0.00628060465052019,3.05377545894428e-07) rectangle (axis cs:0.0095017062985503,1.34366160292382e-05);
\draw[draw=white,fill=color0] (axis cs:0.0095017062985503,3.05377545894428e-07) rectangle (axis cs:0.0143747979068277,1.98495464513776e-05);
\draw[draw=white,fill=color0] (axis cs:0.0143747979068277,3.05377545894428e-07) rectangle (axis cs:0.0217471271337514,2.32087004820908e-05);
\draw[draw=white,fill=color0] (axis cs:0.0217471271337514,3.05377545894428e-07) rectangle (axis cs:0.032900465219543,2.41248333993965e-05);
\draw[draw=white,fill=color0] (axis cs:0.032900465219543,3.05377545894428e-07) rectangle (axis cs:0.0497739588776495,1.92387911731368e-05);
\draw[draw=white,fill=color0] (axis cs:0.0497739588776495,3.05377545894428e-07) rectangle (axis cs:0.0753012750981506,1.71011476990163e-05);
\draw[draw=white,fill=color0] (axis cs:0.0753012750981507,3.05377545894428e-07) rectangle (axis cs:0.113920655685549,1.46581265858309e-05);
\draw[draw=white,fill=color0] (axis cs:0.113920655685549,3.05377545894428e-07) rectangle (axis cs:0.172346560863802,9.16132908121924e-06);
\draw[draw=white,fill=color0] (axis cs:0.172346560863802,3.05377545894428e-07) rectangle (axis cs:0.260737061798252,6.71830796803385e-06);
\draw[draw=white,fill=color0] (axis cs:0.260737061798252,3.05377545894428e-07) rectangle (axis cs:0.394459947761359,3.35915393732066e-06);
\draw[draw=white,fill=color0] (axis cs:0.394459947761359,3.05377545894428e-07) rectangle (axis cs:0.596764607665519,1.83226574166306e-06);
\draw[draw=white,fill=color0] (axis cs:0.596764607665519,3.05377545894428e-07) rectangle (axis cs:0.902824225839103,6.10755185070358e-07);
\draw[draw=white,fill=color0] (axis cs:0.902824225839103,3.05377545894428e-07) rectangle (axis cs:1.36585107811693,1.2215104633112e-06);
\draw[draw=white,fill=color0] (axis cs:1.36585107811693,3.05377545894428e-07) rectangle (axis cs:2.06634814862139,2.13764338072796e-06);
\draw[draw=white,fill=color0] (axis cs:2.06634814862139,3.05377545894428e-07) rectangle (axis cs:3.12610557601766,1.52688810248713e-06);
\draw[draw=white,fill=color0] (axis cs:3.12610557601766,3.05377545894428e-07) rectangle (axis cs:4.72937538571545,3.05377629825575e-06);
\draw[draw=white,fill=color0] (axis cs:4.72937538571545,3.05377545894428e-07) rectangle (axis cs:7.15490599888967,3.35915393743168e-06);
\draw[draw=white,fill=color0] (axis cs:7.15490599888967,3.05377545894428e-07) rectangle (axis cs:10.8244061166236,1.2215104633112e-06);
\draw[draw=white,fill=color0] (axis cs:10.8244061166236,3.05377545894428e-07) rectangle (axis cs:16.3758640289307,3.05377545894428e-07);
\end{axis}

\end{tikzpicture}
     \end{minipage}
    \hspace{-3.5ex}
    \begin{minipage}[t]{0.33\linewidth}
      \centering
\pgfkeys{/pgfplots/spectrumdefault/.style={
    width=1.38\linewidth,
    height=\goldenRatio*1.38\linewidth,
    every axis plot/.append style={line width = 1.2pt},
    every axis plot post/.append style={
      mark size=1, mark options={opacity=0.3}
    },
    tick pos = left,
    xmajorticks = true,
    ymajorticks = true,
    ylabel near ticks,
    xlabel near ticks,
    xtick align = inside,
    ytick align = inside,
    legend cell align = left,
    legend columns = 1,
    legend pos = north west,
    legend style = {
      fill opacity = 0.9,
      text opacity = 1,
      font = \small,
    },
    xticklabel style = {font = \small, inner xsep = 0ex},
    xlabel style = {font = \small},
    axis line style = {black},
    yticklabel style = {font = \small, inner ysep = 0ex},
    ylabel style = {font = \small, inner ysep = 0ex},
    title style = {font = \small, inner ysep = 0ex, yshift = -0.75ex},
    grid = major,
    grid style = {dashed}
  }
}
\pgfkeys{/pgfplots/spectrumdefaultleft/.style={
    spectrumdefault,
    title=\empty,
    ymax=3e-4,
    xlabel=\phantom{eigenvalues},
  }}
\pgfkeys{/pgfplots/spectrumdefaultcenter/.style={
    spectrumdefault,
    ylabel=\empty,
    yticklabels=\empty,
    title=\empty,
    ymax=3e-4,
  }}
\pgfkeys{/pgfplots/spectrumdefaultright/.style={
    spectrumdefault,
    ylabel=\empty,
    yticklabels=\empty,
    title=\empty,
    ymax=3e-4,
    xlabel=\phantom{eigenvalues},
  }}
 \pgfkeys{/pgfplots/zmystyle/.style={spectrumdefaultright, ymax = 5e-4,
          title={mc, full}}}
      \begin{tikzpicture}

\definecolor{color0}{rgb}{0.870588235294118,0.623529411764706,0.0862745098039216}

\begin{axis}[
axis line style={white!80!black},
log basis x={10},
tick pos=left,
title={full\_batch\_mc, layerwise\_group, N=128, D=3274634},
xlabel={eigenvalues},
xmin=0.0001, xmax=16.3758640289307,
xmode=log,
ylabel={density},
ymin=3.05377545894428e-07, ymax=1,
ymode=log,
zmystyle
]
\draw[draw=white,fill=color0] (axis cs:0.0001,3.05377545894428e-07) rectangle (axis cs:0.000151286489554207,0.999843952025082);
\draw[draw=white,fill=color0] (axis cs:0.000151286489554207,3.05377545894428e-07) rectangle (axis cs:0.000228876019216353,3.05377545894428e-07);
\draw[draw=white,fill=color0] (axis cs:0.000228876019216353,3.05377545894428e-07) rectangle (axis cs:0.000346258494903833,3.05377545894428e-07);
\draw[draw=white,fill=color0] (axis cs:0.000346258494903833,3.05377545894428e-07) rectangle (axis cs:0.000523842321723243,3.05377545894428e-07);
\draw[draw=white,fill=color0] (axis cs:0.000523842321723243,3.05377545894428e-07) rectangle (axis cs:0.000792502659334351,3.05377545894428e-07);
\draw[draw=white,fill=color0] (axis cs:0.000792502659334351,3.05377545894428e-07) rectangle (axis cs:0.00119894945293068,3.05377545894428e-07);
\draw[draw=white,fill=color0] (axis cs:0.00119894945293068,3.05377545894428e-07) rectangle (axis cs:0.0018138485388682,3.05377545894428e-07);
\draw[draw=white,fill=color0] (axis cs:0.0018138485388682,3.05377545894428e-07) rectangle (axis cs:0.00274410778028398,3.05377545894428e-07);
\draw[draw=white,fill=color0] (axis cs:0.00274410778028398,3.05377545894428e-07) rectangle (axis cs:0.00415146433037551,2.74839865907982e-06);
\draw[draw=white,fill=color0] (axis cs:0.00415146433037551,3.05377545894428e-07) rectangle (axis cs:0.00628060465052019,4.27528685484845e-06);
\draw[draw=white,fill=color0] (axis cs:0.00628060465052019,3.05377545894428e-07) rectangle (axis cs:0.0095017062985503,9.16132908110822e-06);
\draw[draw=white,fill=color0] (axis cs:0.0095017062985503,3.05377545894428e-07) rectangle (axis cs:0.0143747979068277,1.404737130759e-05);
\draw[draw=white,fill=color0] (axis cs:0.0143747979068277,3.05377545894428e-07) rectangle (axis cs:0.0217471271337514,2.16818122863221e-05);
\draw[draw=white,fill=color0] (axis cs:0.0217471271337514,3.05377545894428e-07) rectangle (axis cs:0.032900465219543,2.32087004819798e-05);
\draw[draw=white,fill=color0] (axis cs:0.032900465219543,3.05377545894428e-07) rectangle (axis cs:0.0497739588776495,2.16818122862111e-05);
\draw[draw=white,fill=color0] (axis cs:0.0497739588776495,3.05377545894428e-07) rectangle (axis cs:0.0753012750981506,1.71011476991273e-05);
\draw[draw=white,fill=color0] (axis cs:0.0753012750981507,3.05377545894428e-07) rectangle (axis cs:0.113920655685549,1.34366160292382e-05);
\draw[draw=white,fill=color0] (axis cs:0.113920655685549,3.05377545894428e-07) rectangle (axis cs:0.172346560863802,8.85595144204331e-06);
\draw[draw=white,fill=color0] (axis cs:0.172346560863802,3.05377545894428e-07) rectangle (axis cs:0.260737061798252,7.02368560709875e-06);
\draw[draw=white,fill=color0] (axis cs:0.260737061798252,3.05377545894428e-07) rectangle (axis cs:0.394459947761359,4.27528685484845e-06);
\draw[draw=white,fill=color0] (axis cs:0.394459947761359,3.05377545894428e-07) rectangle (axis cs:0.596764607665519,1.83226574166306e-06);
\draw[draw=white,fill=color0] (axis cs:0.596764607665519,3.05377545894428e-07) rectangle (axis cs:0.902824225839103,9.16132824135266e-07);
\draw[draw=white,fill=color0] (axis cs:0.902824225839103,3.05377545894428e-07) rectangle (axis cs:1.36585107811693,1.52688810248713e-06);
\draw[draw=white,fill=color0] (axis cs:1.36585107811693,3.05377545894428e-07) rectangle (axis cs:2.06634814862139,1.52688810248713e-06);
\draw[draw=white,fill=color0] (axis cs:2.06634814862139,3.05377545894428e-07) rectangle (axis cs:3.12610557601766,2.13764338083899e-06);
\draw[draw=white,fill=color0] (axis cs:3.12610557601766,3.05377545894428e-07) rectangle (axis cs:4.72937538571545,3.05377629825575e-06);
\draw[draw=white,fill=color0] (axis cs:4.72937538571545,3.05377545894428e-07) rectangle (axis cs:7.15490599888967,2.13764338072796e-06);
\draw[draw=white,fill=color0] (axis cs:7.15490599888967,3.05377545894428e-07) rectangle (axis cs:10.8244061166236,1.83226574166306e-06);
\draw[draw=white,fill=color0] (axis cs:10.8244061166236,3.05377545894428e-07) rectangle (axis cs:16.3758640289307,3.05377545894428e-07);
\end{axis}

\end{tikzpicture}
     \end{minipage}
  \end{minipage}

  \begin{small}
    \textbf{\cifarten \threecthreed}
  \end{small}

  \begin{minipage}{0.495\linewidth}
    \centering
    \begin{small}
      \textbf{Full network}
    \end{small}
    \vspace{1ex}

    \begin{minipage}[t]{0.33\linewidth}
      \centering
\pgfkeys{/pgfplots/spectrumdefault/.style={
    width=1.38\linewidth,
    height=\goldenRatio*1.38\linewidth,
    every axis plot/.append style={line width = 1.2pt},
    every axis plot post/.append style={
      mark size=1, mark options={opacity=0.3}
    },
    tick pos = left,
    xmajorticks = true,
    ymajorticks = true,
    ylabel near ticks,
    xlabel near ticks,
    xtick align = inside,
    ytick align = inside,
    legend cell align = left,
    legend columns = 1,
    legend pos = north west,
    legend style = {
      fill opacity = 0.9,
      text opacity = 1,
      font = \small,
    },
    xticklabel style = {font = \small, inner xsep = 0ex},
    xlabel style = {font = \small},
    axis line style = {black},
    yticklabel style = {font = \small, inner ysep = 0ex},
    ylabel style = {font = \small, inner ysep = 0ex},
    title style = {font = \small, inner ysep = 0ex, yshift = -0.75ex},
    grid = major,
    grid style = {dashed}
  }
}
\pgfkeys{/pgfplots/spectrumdefaultleft/.style={
    spectrumdefault,
    title=\empty,
    ymax=3e-4,
    xlabel=\phantom{eigenvalues},
  }}
\pgfkeys{/pgfplots/spectrumdefaultcenter/.style={
    spectrumdefault,
    ylabel=\empty,
    yticklabels=\empty,
    title=\empty,
    ymax=3e-4,
  }}
\pgfkeys{/pgfplots/spectrumdefaultright/.style={
    spectrumdefault,
    ylabel=\empty,
    yticklabels=\empty,
    title=\empty,
    ymax=3e-4,
    xlabel=\phantom{eigenvalues},
  }}
 \pgfkeys{/pgfplots/zmystyle/.style={spectrumdefaultleft,
          title={exact, mb}}}
      \begin{tikzpicture}

\begin{axis}[
axis line style={white!80!black},
log basis x={10},
tick pos=left,
title={full\_batch\_exact, one\_group, N=128, D=895210},
xlabel={eigenvalues},
xmin=0.0001, xmax=35.8794860839844,
xmode=log,
ylabel={density},
ymin=1.11705508533742e-06, ymax=1,
ymode=log,
zmystyle
]
\draw[draw=white,fill=black] (axis cs:0.0001,1.11705508533742e-06) rectangle (axis cs:0.000155434142340701,0.99871426815899);
\draw[draw=white,fill=black] (axis cs:0.000155434142340701,1.11705508533742e-06) rectangle (axis cs:0.000241597726051892,1.11705508533742e-06);
\draw[draw=white,fill=black] (axis cs:0.000241597726051892,1.11705508533742e-06) rectangle (axis cs:0.000375525353403394,1.11705508533742e-06);
\draw[draw=white,fill=black] (axis cs:0.000375525353403394,1.11705508533742e-06) rectangle (axis cs:0.00058369461233445,5.58528041795857e-06);
\draw[draw=white,fill=black] (axis cs:0.00058369461233445,1.11705508533742e-06) rectangle (axis cs:0.000907260714570931,2.68093507478535e-05);
\draw[draw=white,fill=black] (axis cs:0.000907260714570931,1.11705508533742e-06) rectangle (axis cs:0.00141019291048744,4.80334210778594e-05);
\draw[draw=white,fill=black] (axis cs:0.00141019291048744,1.11705508533742e-06) rectangle (axis cs:0.00219192125576551,7.26086604072757e-05);
\draw[draw=white,fill=black] (axis cs:0.00219192125576551,1.11705508533742e-06) rectangle (axis cs:0.00340699400468264,9.04815617377603e-05);
\draw[draw=white,fill=black] (axis cs:0.00340699400468264,1.11705508533742e-06) rectangle (axis cs:0.00529563191077756,0.000105003294068835);
\draw[draw=white,fill=black] (axis cs:0.00529563191077756,1.11705508533742e-06) rectangle (axis cs:0.00823122004203757,0.000123993251732419);
\draw[draw=white,fill=black] (axis cs:0.00823122004203757,1.11705508533742e-06) rectangle (axis cs:0.0127941262765169,0.000138514984063382);
\draw[draw=white,fill=black] (axis cs:0.0127941262765169,1.11705508533742e-06) rectangle (axis cs:0.0198864404478903,0.000137397927730282);
\draw[draw=white,fill=black] (axis cs:0.0198864404478903,1.11705508533742e-06) rectangle (axis cs:0.0309103181522725,0.000130695589731351);
\draw[draw=white,fill=black] (axis cs:0.0309103181522725,1.11705508533742e-06) rectangle (axis cs:0.0480451879147667,0.000115056801067177);
\draw[draw=white,fill=black] (axis cs:0.0480451879147667,1.11705508533742e-06) rectangle (axis cs:0.0746786257712956,9.27156744040709e-05);
\draw[draw=white,fill=black] (axis cs:0.0746786257712956,1.11705508533742e-06) rectangle (axis cs:0.116076081479435,7.26086604072757e-05);
\draw[draw=white,fill=black] (axis cs:0.116076081479435,1.11705508533742e-06) rectangle (axis cs:0.180421861710252,5.5852815410002e-05);
\draw[draw=white,fill=black] (axis cs:0.180421861710252,1.11705508533742e-06) rectangle (axis cs:0.280437173344456,3.79799140794064e-05);
\draw[draw=white,fill=black] (axis cs:0.280437173344456,1.11705508533742e-06) rectangle (axis cs:0.435895115192459,2.23411254153434e-05);
\draw[draw=white,fill=black] (axis cs:0.435895115192459,1.11705508533742e-06) rectangle (axis cs:0.677529833804408,1.45217310832009e-05);
\draw[draw=white,fill=black] (axis cs:0.677529833804408,1.11705508533742e-06) rectangle (axis cs:1.05311268627626,5.58528041795857e-06);
\draw[draw=white,fill=black] (axis cs:1.05311268627626,1.11705508533742e-06) rectangle (axis cs:1.63689667179461,2.2341114184372e-06);
\draw[draw=white,fill=black] (axis cs:1.63689667179461,1.11705508533742e-06) rectangle (axis cs:2.54429630280743,1.11705508533742e-06);
\draw[draw=white,fill=black] (axis cs:2.54429630280743,1.11705508533742e-06) rectangle (axis cs:3.95470513687488,2.23411141854822e-06);
\draw[draw=white,fill=black] (axis cs:3.95470513687488,1.11705508533742e-06) rectangle (axis cs:6.1469620116051,3.351167751648e-06);
\draw[draw=white,fill=black] (axis cs:6.14696201160511,1.11705508533742e-06) rectangle (axis cs:9.55447768274709,4.46822408474777e-06);
\draw[draw=white,fill=black] (axis cs:9.55447768274709,1.11705508533742e-06) rectangle (axis cs:14.8509204413116,4.46822408485879e-06);
\draw[draw=white,fill=black] (axis cs:14.8509204413116,1.11705508533742e-06) rectangle (axis cs:23.0834008176524,1.11705508533742e-06);
\draw[draw=white,fill=black] (axis cs:23.0834008176524,1.11705508533742e-06) rectangle (axis cs:35.8794860839844,1.11705508533742e-06);
\end{axis}

\end{tikzpicture}
     \end{minipage}
    \hspace{1.9ex}
    \begin{minipage}[t]{0.33\linewidth}
      \centering
\pgfkeys{/pgfplots/spectrumdefault/.style={
    width=1.38\linewidth,
    height=\goldenRatio*1.38\linewidth,
    every axis plot/.append style={line width = 1.2pt},
    every axis plot post/.append style={
      mark size=1, mark options={opacity=0.3}
    },
    tick pos = left,
    xmajorticks = true,
    ymajorticks = true,
    ylabel near ticks,
    xlabel near ticks,
    xtick align = inside,
    ytick align = inside,
    legend cell align = left,
    legend columns = 1,
    legend pos = north west,
    legend style = {
      fill opacity = 0.9,
      text opacity = 1,
      font = \small,
    },
    xticklabel style = {font = \small, inner xsep = 0ex},
    xlabel style = {font = \small},
    axis line style = {black},
    yticklabel style = {font = \small, inner ysep = 0ex},
    ylabel style = {font = \small, inner ysep = 0ex},
    title style = {font = \small, inner ysep = 0ex, yshift = -0.75ex},
    grid = major,
    grid style = {dashed}
  }
}
\pgfkeys{/pgfplots/spectrumdefaultleft/.style={
    spectrumdefault,
    title=\empty,
    ymax=3e-4,
    xlabel=\phantom{eigenvalues},
  }}
\pgfkeys{/pgfplots/spectrumdefaultcenter/.style={
    spectrumdefault,
    ylabel=\empty,
    yticklabels=\empty,
    title=\empty,
    ymax=3e-4,
  }}
\pgfkeys{/pgfplots/spectrumdefaultright/.style={
    spectrumdefault,
    ylabel=\empty,
    yticklabels=\empty,
    title=\empty,
    ymax=3e-4,
    xlabel=\phantom{eigenvalues},
  }}
 \pgfkeys{/pgfplots/zmystyle/.style={spectrumdefaultcenter,
          title={exact, sub}}}
      \begin{tikzpicture}

\definecolor{color0}{rgb}{0.274509803921569,0.6,0.564705882352941}

\begin{axis}[
axis line style={white!80!black},
log basis x={10},
tick pos=left,
title={frac\_batch\_exact, one\_group, N=128, D=895210},
xlabel={eigenvalues},
xmin=0.0001, xmax=35.8794860839844,
xmode=log,
ylabel={density},
ymin=1.11705508533742e-06, ymax=1,
ymode=log,
zmystyle
]
\draw[draw=white,fill=color0] (axis cs:0.0001,1.11705508533742e-06) rectangle (axis cs:0.000155434142340701,0.999840260942811);
\draw[draw=white,fill=color0] (axis cs:0.000155434142340701,1.11705508533742e-06) rectangle (axis cs:0.000241597726051892,1.11705508533742e-06);
\draw[draw=white,fill=color0] (axis cs:0.000241597726051892,1.11705508533742e-06) rectangle (axis cs:0.000375525353403394,1.11705508533742e-06);
\draw[draw=white,fill=color0] (axis cs:0.000375525353403394,1.11705508533742e-06) rectangle (axis cs:0.00058369461233445,1.11705508533742e-06);
\draw[draw=white,fill=color0] (axis cs:0.00058369461233445,1.11705508533742e-06) rectangle (axis cs:0.000907260714570931,1.11705508533742e-06);
\draw[draw=white,fill=color0] (axis cs:0.000907260714570931,1.11705508533742e-06) rectangle (axis cs:0.00141019291048744,1.11705508533742e-06);
\draw[draw=white,fill=color0] (axis cs:0.00141019291048744,1.11705508533742e-06) rectangle (axis cs:0.00219192125576551,1.11705508533742e-06);
\draw[draw=white,fill=color0] (axis cs:0.00219192125576551,1.11705508533742e-06) rectangle (axis cs:0.00340699400468264,1.11705508533742e-06);
\draw[draw=white,fill=color0] (axis cs:0.00340699400468264,1.11705508533742e-06) rectangle (axis cs:0.00529563191077756,1.11705508533742e-06);
\draw[draw=white,fill=color0] (axis cs:0.00529563191077756,1.11705508533742e-06) rectangle (axis cs:0.00823122004203757,1.11705508533742e-06);
\draw[draw=white,fill=color0] (axis cs:0.00823122004203757,1.11705508533742e-06) rectangle (axis cs:0.0127941262765169,1.11705508533742e-06);
\draw[draw=white,fill=color0] (axis cs:0.0127941262765169,1.11705508533742e-06) rectangle (axis cs:0.0198864404478903,5.58528041795857e-06);
\draw[draw=white,fill=color0] (axis cs:0.0198864404478903,1.11705508533742e-06) rectangle (axis cs:0.0309103181522725,6.70233675105834e-06);
\draw[draw=white,fill=color0] (axis cs:0.0309103181522725,1.11705508533742e-06) rectangle (axis cs:0.0480451879147667,1.00535057505797e-05);
\draw[draw=white,fill=color0] (axis cs:0.0480451879147667,1.11705508533742e-06) rectangle (axis cs:0.0746786257712956,1.11705620837905e-05);
\draw[draw=white,fill=color0] (axis cs:0.0746786257712956,1.11705508533742e-06) rectangle (axis cs:0.116076081479435,1.22876184168903e-05);
\draw[draw=white,fill=color0] (axis cs:0.116076081479435,1.11705508533742e-06) rectangle (axis cs:0.180421861710252,1.78729000826112e-05);
\draw[draw=white,fill=color0] (axis cs:0.180421861710252,1.11705508533742e-06) rectangle (axis cs:0.280437173344456,2.45752380816539e-05);
\draw[draw=white,fill=color0] (axis cs:0.280437173344456,1.11705508533742e-06) rectangle (axis cs:0.435895115192459,2.23411254152324e-05);
\draw[draw=white,fill=color0] (axis cs:0.435895115192459,1.11705508533742e-06) rectangle (axis cs:0.677529833804408,2.23411254153434e-05);
\draw[draw=white,fill=color0] (axis cs:0.677529833804408,1.11705508533742e-06) rectangle (axis cs:1.05311268627626,1.56387874163006e-05);
\draw[draw=white,fill=color0] (axis cs:1.05311268627626,1.11705508533742e-06) rectangle (axis cs:1.63689667179461,1.00535057505797e-05);
\draw[draw=white,fill=color0] (axis cs:1.63689667179461,1.11705508533742e-06) rectangle (axis cs:2.54429630280743,5.58528041795857e-06);
\draw[draw=white,fill=color0] (axis cs:2.54429630280743,1.11705508533742e-06) rectangle (axis cs:3.95470513687488,1.11705508533742e-06);
\draw[draw=white,fill=color0] (axis cs:3.95470513687488,1.11705508533742e-06) rectangle (axis cs:6.1469620116051,3.351167751648e-06);
\draw[draw=white,fill=color0] (axis cs:6.14696201160511,1.11705508533742e-06) rectangle (axis cs:9.55447768274709,4.46822408485879e-06);
\draw[draw=white,fill=color0] (axis cs:9.55447768274709,1.11705508533742e-06) rectangle (axis cs:14.8509204413116,3.351167751648e-06);
\draw[draw=white,fill=color0] (axis cs:14.8509204413116,1.11705508533742e-06) rectangle (axis cs:23.0834008176524,3.351167751648e-06);
\draw[draw=white,fill=color0] (axis cs:23.0834008176524,1.11705508533742e-06) rectangle (axis cs:35.8794860839844,1.11705508533742e-06);
\end{axis}

\end{tikzpicture}
     \end{minipage}
    \hspace{-3.5ex}
    \begin{minipage}[t]{0.33\linewidth}
      \centering
\pgfkeys{/pgfplots/spectrumdefault/.style={
    width=1.38\linewidth,
    height=\goldenRatio*1.38\linewidth,
    every axis plot/.append style={line width = 1.2pt},
    every axis plot post/.append style={
      mark size=1, mark options={opacity=0.3}
    },
    tick pos = left,
    xmajorticks = true,
    ymajorticks = true,
    ylabel near ticks,
    xlabel near ticks,
    xtick align = inside,
    ytick align = inside,
    legend cell align = left,
    legend columns = 1,
    legend pos = north west,
    legend style = {
      fill opacity = 0.9,
      text opacity = 1,
      font = \small,
    },
    xticklabel style = {font = \small, inner xsep = 0ex},
    xlabel style = {font = \small},
    axis line style = {black},
    yticklabel style = {font = \small, inner ysep = 0ex},
    ylabel style = {font = \small, inner ysep = 0ex},
    title style = {font = \small, inner ysep = 0ex, yshift = -0.75ex},
    grid = major,
    grid style = {dashed}
  }
}
\pgfkeys{/pgfplots/spectrumdefaultleft/.style={
    spectrumdefault,
    title=\empty,
    ymax=3e-4,
    xlabel=\phantom{eigenvalues},
  }}
\pgfkeys{/pgfplots/spectrumdefaultcenter/.style={
    spectrumdefault,
    ylabel=\empty,
    yticklabels=\empty,
    title=\empty,
    ymax=3e-4,
  }}
\pgfkeys{/pgfplots/spectrumdefaultright/.style={
    spectrumdefault,
    ylabel=\empty,
    yticklabels=\empty,
    title=\empty,
    ymax=3e-4,
    xlabel=\phantom{eigenvalues},
  }}
 \pgfkeys{/pgfplots/zmystyle/.style={spectrumdefaultright,
          title={mc, full}}}
      \begin{tikzpicture}

\definecolor{color0}{rgb}{0.870588235294118,0.623529411764706,0.0862745098039216}

\begin{axis}[
axis line style={white!80!black},
log basis x={10},
tick pos=left,
title={full\_batch\_mc, one\_group, N=128, D=895210},
xlabel={eigenvalues},
xmin=0.0001, xmax=35.8794860839844,
xmode=log,
ylabel={density},
ymin=1.11705508533742e-06, ymax=1,
ymode=log,
zmystyle
]
\draw[draw=white,fill=color0] (axis cs:0.0001,1.11705508533742e-06) rectangle (axis cs:0.000155434142340701,0.999858133844141);
\draw[draw=white,fill=color0] (axis cs:0.000155434142340701,1.11705508533742e-06) rectangle (axis cs:0.000241597726051892,1.11705508533742e-06);
\draw[draw=white,fill=color0] (axis cs:0.000241597726051892,1.11705508533742e-06) rectangle (axis cs:0.000375525353403394,1.11705508533742e-06);
\draw[draw=white,fill=color0] (axis cs:0.000375525353403394,1.11705508533742e-06) rectangle (axis cs:0.00058369461233445,1.11705508533742e-06);
\draw[draw=white,fill=color0] (axis cs:0.00058369461233445,1.11705508533742e-06) rectangle (axis cs:0.000907260714570931,1.11705508533742e-06);
\draw[draw=white,fill=color0] (axis cs:0.000907260714570931,1.11705508533742e-06) rectangle (axis cs:0.00141019291048744,1.11705508533742e-06);
\draw[draw=white,fill=color0] (axis cs:0.00141019291048744,1.11705508533742e-06) rectangle (axis cs:0.00219192125576551,1.11705508533742e-06);
\draw[draw=white,fill=color0] (axis cs:0.00219192125576551,1.11705508533742e-06) rectangle (axis cs:0.00340699400468264,1.11705508533742e-06);
\draw[draw=white,fill=color0] (axis cs:0.00340699400468264,1.11705508533742e-06) rectangle (axis cs:0.00529563191077756,1.11705508533742e-06);
\draw[draw=white,fill=color0] (axis cs:0.00529563191077756,1.11705508533742e-06) rectangle (axis cs:0.00823122004203757,1.11705508533742e-06);
\draw[draw=white,fill=color0] (axis cs:0.00823122004203757,1.11705508533742e-06) rectangle (axis cs:0.0127941262765169,1.11705508533742e-06);
\draw[draw=white,fill=color0] (axis cs:0.0127941262765169,1.11705508533742e-06) rectangle (axis cs:0.0198864404478903,1.11705508533742e-06);
\draw[draw=white,fill=color0] (axis cs:0.0198864404478903,1.11705508533742e-06) rectangle (axis cs:0.0309103181522725,4.46822408474777e-06);
\draw[draw=white,fill=color0] (axis cs:0.0309103181522725,1.11705508533742e-06) rectangle (axis cs:0.0480451879147667,6.70233675116937e-06);
\draw[draw=white,fill=color0] (axis cs:0.0480451879147667,1.11705508533742e-06) rectangle (axis cs:0.0746786257712956,1.34046747499901e-05);
\draw[draw=white,fill=color0] (axis cs:0.0746786257712956,1.11705508533742e-06) rectangle (axis cs:0.116076081479435,1.22876184168903e-05);
\draw[draw=white,fill=color0] (axis cs:0.116076081479435,1.11705508533742e-06) rectangle (axis cs:0.180421861710252,1.45217310832009e-05);
\draw[draw=white,fill=color0] (axis cs:0.180421861710252,1.11705508533742e-06) rectangle (axis cs:0.280437173344456,1.8989956415822e-05);
\draw[draw=white,fill=color0] (axis cs:0.280437173344456,1.11705508533742e-06) rectangle (axis cs:0.435895115192459,2.01070127490328e-05);
\draw[draw=white,fill=color0] (axis cs:0.435895115192459,1.11705508533742e-06) rectangle (axis cs:0.677529833804408,2.34581817484432e-05);
\draw[draw=white,fill=color0] (axis cs:0.677529833804408,1.11705508533742e-06) rectangle (axis cs:1.05311268627626,1.56387874163006e-05);
\draw[draw=white,fill=color0] (axis cs:1.05311268627626,1.11705508533742e-06) rectangle (axis cs:1.63689667179461,1.22876184168903e-05);
\draw[draw=white,fill=color0] (axis cs:1.63689667179461,1.11705508533742e-06) rectangle (axis cs:2.54429630280743,3.351167751648e-06);
\draw[draw=white,fill=color0] (axis cs:2.54429630280743,1.11705508533742e-06) rectangle (axis cs:3.95470513687488,1.11705508533742e-06);
\draw[draw=white,fill=color0] (axis cs:3.95470513687488,1.11705508533742e-06) rectangle (axis cs:6.1469620116051,3.351167751648e-06);
\draw[draw=white,fill=color0] (axis cs:6.14696201160511,1.11705508533742e-06) rectangle (axis cs:9.55447768274709,5.58528041795857e-06);
\draw[draw=white,fill=color0] (axis cs:9.55447768274709,1.11705508533742e-06) rectangle (axis cs:14.8509204413116,2.23411141854822e-06);
\draw[draw=white,fill=color0] (axis cs:14.8509204413116,1.11705508533742e-06) rectangle (axis cs:23.0834008176524,3.351167751648e-06);
\draw[draw=white,fill=color0] (axis cs:23.0834008176524,1.11705508533742e-06) rectangle (axis cs:35.8794860839844,1.11705508533742e-06);
\end{axis}

\end{tikzpicture}
     \end{minipage}
  \end{minipage}
  \hfill
  \begin{minipage}{0.495\linewidth}
    \centering
    \begin{small}
      \textbf{Block-diagonal approximation}
    \end{small}
    \vspace{1ex}

    \begin{minipage}[t]{0.33\linewidth}
      \centering
\pgfkeys{/pgfplots/spectrumdefault/.style={
    width=1.38\linewidth,
    height=\goldenRatio*1.38\linewidth,
    every axis plot/.append style={line width = 1.2pt},
    every axis plot post/.append style={
      mark size=1, mark options={opacity=0.3}
    },
    tick pos = left,
    xmajorticks = true,
    ymajorticks = true,
    ylabel near ticks,
    xlabel near ticks,
    xtick align = inside,
    ytick align = inside,
    legend cell align = left,
    legend columns = 1,
    legend pos = north west,
    legend style = {
      fill opacity = 0.9,
      text opacity = 1,
      font = \small,
    },
    xticklabel style = {font = \small, inner xsep = 0ex},
    xlabel style = {font = \small},
    axis line style = {black},
    yticklabel style = {font = \small, inner ysep = 0ex},
    ylabel style = {font = \small, inner ysep = 0ex},
    title style = {font = \small, inner ysep = 0ex, yshift = -0.75ex},
    grid = major,
    grid style = {dashed}
  }
}
\pgfkeys{/pgfplots/spectrumdefaultleft/.style={
    spectrumdefault,
    title=\empty,
    ymax=3e-4,
    xlabel=\phantom{eigenvalues},
  }}
\pgfkeys{/pgfplots/spectrumdefaultcenter/.style={
    spectrumdefault,
    ylabel=\empty,
    yticklabels=\empty,
    title=\empty,
    ymax=3e-4,
  }}
\pgfkeys{/pgfplots/spectrumdefaultright/.style={
    spectrumdefault,
    ylabel=\empty,
    yticklabels=\empty,
    title=\empty,
    ymax=3e-4,
    xlabel=\phantom{eigenvalues},
  }}
 \pgfkeys{/pgfplots/zmystyle/.style={spectrumdefaultleft, ymax = 3e-3,
          title={exact, mb}}}
      \begin{tikzpicture}

\begin{axis}[
axis line style={white!80!black},
log basis x={10},
tick pos=left,
title={full\_batch\_exact, layerwise\_group, N=128, D=895210},
xlabel={eigenvalues},
xmin=0.0001, xmax=7.47674083709717,
xmode=log,
ylabel={density},
ymin=1.11705508533742e-06, ymax=1,
ymode=log,
zmystyle
]
\draw[draw=white,fill=black] (axis cs:0.0001,1.11705508533742e-06) rectangle (axis cs:0.000147251268855062,0.993807039687439);
\draw[draw=white,fill=black] (axis cs:0.000147251268855062,1.11705508533742e-06) rectangle (axis cs:0.000216829361794257,0.000392086771689688);
\draw[draw=white,fill=black] (axis cs:0.000216829361794257,1.11705508533742e-06) rectangle (axis cs:0.000319283986492377,0.000470280715010558);
\draw[draw=white,fill=black] (axis cs:0.000319283986492377,1.11705508533742e-06) rectangle (axis cs:0.00047014972136105,0.000520548250002601);
\draw[draw=white,fill=black] (axis cs:0.00047014972136105,1.11705508533742e-06) rectangle (axis cs:0.000692301430222684,0.000547357601998217);
\draw[draw=white,fill=black] (axis cs:0.000692301430222684,1.11705508533742e-06) rectangle (axis cs:0.00101942264030464,0.000568581672328223);
\draw[draw=white,fill=black] (axis cs:0.00101942264030464,1.11705508533742e-06) rectangle (axis cs:0.00150111277284436,0.000546240545665117);
\draw[draw=white,fill=black] (axis cs:0.00150111277284436,1.11705508533742e-06) rectangle (axis cs:0.00221040760495872,0.000522782362668912);
\draw[draw=white,fill=black] (axis cs:0.00221040760495872,1.11705508533742e-06) rectangle (axis cs:0.00325485324517051,0.000471397771343657);
\draw[draw=white,fill=black] (axis cs:0.00325485324517051,1.11705508533742e-06) rectangle (axis cs:0.00479281270288374,0.000424481405351246);
\draw[draw=white,fill=black] (axis cs:0.00479281270288374,1.11705508533742e-06) rectangle (axis cs:0.00705747751884289,0.000359692138028129);
\draw[draw=white,fill=black] (axis cs:0.00705747751884289,1.11705508533742e-06) rectangle (axis cs:0.0103922251956569,0.000304956377703575);
\draw[draw=white,fill=black] (axis cs:0.0103922251956569,1.11705508533742e-06) rectangle (axis cs:0.0153026834628802,0.000260274124377364);
\draw[draw=white,fill=black] (axis cs:0.0153026834628802,1.11705508533742e-06) rectangle (axis cs:0.0225333955679649,0.000211123645718531);
\draw[draw=white,fill=black] (axis cs:0.0225333955679649,1.11705508533742e-06) rectangle (axis cs:0.0331807108899586,0.000164207279726009);
\draw[draw=white,fill=black] (axis cs:0.0331807108899586,1.11705508533742e-06) rectangle (axis cs:0.0488590178005937,0.000131812646064561);
\draw[draw=white,fill=black] (axis cs:0.0488590178005937,1.11705508533742e-06) rectangle (axis cs:0.0719455236614949,0.000100535068736102);
\draw[draw=white,fill=black] (axis cs:0.0719455236614949,1.11705508533742e-06) rectangle (axis cs:0.10594069647597,7.26086604072757e-05);
\draw[draw=white,fill=black] (axis cs:0.10594069647597,1.11705508533742e-06) rectangle (axis cs:0.155999019794756,4.69163647447597e-05);
\draw[draw=white,fill=black] (axis cs:0.155999019794756,1.11705508533742e-06) rectangle (axis cs:0.229710536049237,3.01605197473749e-05);
\draw[draw=white,fill=black] (axis cs:0.229710536049237,1.11705508533742e-06) rectangle (axis cs:0.338251679026266,1.67558437495114e-05);
\draw[draw=white,fill=black] (axis cs:0.338251679026266,1.11705508533742e-06) rectangle (axis cs:0.498079889289728,6.70233675105834e-06);
\draw[draw=white,fill=black] (axis cs:0.498079889289728,1.11705508533742e-06) rectangle (axis cs:0.733428956891012,1.22876184168903e-05);
\draw[draw=white,fill=black] (axis cs:0.733428956891012,1.11705508533742e-06) rectangle (axis cs:1.07998344517246,1.11705620837905e-05);
\draw[draw=white,fill=black] (axis cs:1.07998344517246,1.11705508533742e-06) rectangle (axis cs:1.59028932644106,1.11705620836795e-05);
\draw[draw=white,fill=black] (axis cs:1.59028932644106,1.11705508533742e-06) rectangle (axis cs:2.34172121165108,8.93644941747994e-06);
\draw[draw=white,fill=black] (axis cs:2.34172121165108,1.11705508533742e-06) rectangle (axis cs:3.44821419720435,5.58528041795857e-06);
\draw[draw=white,fill=black] (axis cs:3.44821419720435,1.11705508533742e-06) rectangle (axis cs:5.07753915822379,5.58528041795857e-06);
\draw[draw=white,fill=black] (axis cs:5.07753915822379,1.11705508533742e-06) rectangle (axis cs:7.47674083709717,1.11705508533742e-06);
\end{axis}

\end{tikzpicture}
     \end{minipage}
    \hspace{1.9ex}
    \begin{minipage}[t]{0.33\linewidth}
      \centering
\pgfkeys{/pgfplots/spectrumdefault/.style={
    width=1.38\linewidth,
    height=\goldenRatio*1.38\linewidth,
    every axis plot/.append style={line width = 1.2pt},
    every axis plot post/.append style={
      mark size=1, mark options={opacity=0.3}
    },
    tick pos = left,
    xmajorticks = true,
    ymajorticks = true,
    ylabel near ticks,
    xlabel near ticks,
    xtick align = inside,
    ytick align = inside,
    legend cell align = left,
    legend columns = 1,
    legend pos = north west,
    legend style = {
      fill opacity = 0.9,
      text opacity = 1,
      font = \small,
    },
    xticklabel style = {font = \small, inner xsep = 0ex},
    xlabel style = {font = \small},
    axis line style = {black},
    yticklabel style = {font = \small, inner ysep = 0ex},
    ylabel style = {font = \small, inner ysep = 0ex},
    title style = {font = \small, inner ysep = 0ex, yshift = -0.75ex},
    grid = major,
    grid style = {dashed}
  }
}
\pgfkeys{/pgfplots/spectrumdefaultleft/.style={
    spectrumdefault,
    title=\empty,
    ymax=3e-4,
    xlabel=\phantom{eigenvalues},
  }}
\pgfkeys{/pgfplots/spectrumdefaultcenter/.style={
    spectrumdefault,
    ylabel=\empty,
    yticklabels=\empty,
    title=\empty,
    ymax=3e-4,
  }}
\pgfkeys{/pgfplots/spectrumdefaultright/.style={
    spectrumdefault,
    ylabel=\empty,
    yticklabels=\empty,
    title=\empty,
    ymax=3e-4,
    xlabel=\phantom{eigenvalues},
  }}
 \pgfkeys{/pgfplots/zmystyle/.style={spectrumdefaultcenter, ymax = 3e-3,
          title={exact, sub}}}
      \begin{tikzpicture}

\definecolor{color0}{rgb}{0.274509803921569,0.6,0.564705882352941}

\begin{axis}[
axis line style={white!80!black},
log basis x={10},
tick pos=left,
title={frac\_batch\_exact, layerwise\_group, N=128, D=895210},
xlabel={eigenvalues},
xmin=0.0001, xmax=7.47674083709717,
xmode=log,
ylabel={density},
ymin=1.11705508533742e-06, ymax=1,
ymode=log,
zmystyle
]
\draw[draw=white,fill=color0] (axis cs:0.0001,1.11705508533742e-06) rectangle (axis cs:0.000147251268855062,0.999042682720938);
\draw[draw=white,fill=color0] (axis cs:0.000147251268855062,1.11705508533742e-06) rectangle (axis cs:0.000216829361794257,1.45217310832009e-05);
\draw[draw=white,fill=color0] (axis cs:0.000216829361794257,1.11705508533742e-06) rectangle (axis cs:0.000319283986492377,1.11705620836795e-05);
\draw[draw=white,fill=color0] (axis cs:0.000319283986492377,1.11705508533742e-06) rectangle (axis cs:0.00047014972136105,3.351167751648e-06);
\draw[draw=white,fill=color0] (axis cs:0.00047014972136105,1.11705508533742e-06) rectangle (axis cs:0.000692301430222684,1.8989956415822e-05);
\draw[draw=white,fill=color0] (axis cs:0.000692301430222684,1.11705508533742e-06) rectangle (axis cs:0.00101942264030464,1.34046747501011e-05);
\draw[draw=white,fill=color0] (axis cs:0.00101942264030464,1.11705508533742e-06) rectangle (axis cs:0.00150111277284436,3.23946324136854e-05);
\draw[draw=white,fill=color0] (axis cs:0.00150111277284436,1.11705508533742e-06) rectangle (axis cs:0.00221040760495872,3.23946324136854e-05);
\draw[draw=white,fill=color0] (axis cs:0.00221040760495872,1.11705508533742e-06) rectangle (axis cs:0.00325485324517051,4.57993084115489e-05);
\draw[draw=white,fill=color0] (axis cs:0.00325485324517051,1.11705508533742e-06) rectangle (axis cs:0.00479281270288374,5.92039844094123e-05);
\draw[draw=white,fill=color0] (axis cs:0.00479281270288374,1.11705508533742e-06) rectangle (axis cs:0.00705747751884289,6.70233787414438e-05);
\draw[draw=white,fill=color0] (axis cs:0.00705747751884289,1.11705508533742e-06) rectangle (axis cs:0.0103922251956569,7.59598294067971e-05);
\draw[draw=white,fill=color0] (axis cs:0.0103922251956569,1.11705508533742e-06) rectangle (axis cs:0.0153026834628802,7.81939420729967e-05);
\draw[draw=white,fill=color0] (axis cs:0.0153026834628802,1.11705508533742e-06) rectangle (axis cs:0.0225333955679649,8.1545111072518e-05);
\draw[draw=white,fill=color0] (axis cs:0.0225333955679649,1.11705508533742e-06) rectangle (axis cs:0.0331807108899586,8.26621674057288e-05);
\draw[draw=white,fill=color0] (axis cs:0.0331807108899586,1.11705508533742e-06) rectangle (axis cs:0.0488590178005937,7.14916040740649e-05);
\draw[draw=white,fill=color0] (axis cs:0.0488590178005937,1.11705508533742e-06) rectangle (axis cs:0.0719455236614949,6.70233787415548e-05);
\draw[draw=white,fill=color0] (axis cs:0.0719455236614949,1.11705508533742e-06) rectangle (axis cs:0.10594069647597,5.47357590767912e-05);
\draw[draw=white,fill=color0] (axis cs:0.10594069647597,1.11705508533742e-06) rectangle (axis cs:0.155999019794756,4.69163647446487e-05);
\draw[draw=white,fill=color0] (axis cs:0.155999019794756,1.11705508533742e-06) rectangle (axis cs:0.229710536049237,3.35116887468962e-05);
\draw[draw=white,fill=color0] (axis cs:0.229710536049237,1.11705508533742e-06) rectangle (axis cs:0.338251679026266,2.56922944147537e-05);
\draw[draw=white,fill=color0] (axis cs:0.338251679026266,1.11705508533742e-06) rectangle (axis cs:0.498079889289728,1.67558437495114e-05);
\draw[draw=white,fill=color0] (axis cs:0.498079889289728,1.11705508533742e-06) rectangle (axis cs:0.733428956891012,1.56387874163006e-05);
\draw[draw=white,fill=color0] (axis cs:0.733428956891012,1.11705508533742e-06) rectangle (axis cs:1.07998344517246,1.11705620837905e-05);
\draw[draw=white,fill=color0] (axis cs:1.07998344517246,1.11705508533742e-06) rectangle (axis cs:1.59028932644106,1.22876184168903e-05);
\draw[draw=white,fill=color0] (axis cs:1.59028932644106,1.11705508533742e-06) rectangle (axis cs:2.34172121165108,8.93644941736892e-06);
\draw[draw=white,fill=color0] (axis cs:2.34172121165108,1.11705508533742e-06) rectangle (axis cs:3.44821419720435,5.58528041795857e-06);
\draw[draw=white,fill=color0] (axis cs:3.44821419720435,1.11705508533742e-06) rectangle (axis cs:5.07753915822379,2.23411141854822e-06);
\draw[draw=white,fill=color0] (axis cs:5.07753915822379,1.11705508533742e-06) rectangle (axis cs:7.47674083709717,1.11705508533742e-06);
\end{axis}

\end{tikzpicture}
     \end{minipage}
    \hspace{-3.5ex}
    \begin{minipage}[t]{0.33\linewidth}
      \centering
\pgfkeys{/pgfplots/spectrumdefault/.style={
    width=1.38\linewidth,
    height=\goldenRatio*1.38\linewidth,
    every axis plot/.append style={line width = 1.2pt},
    every axis plot post/.append style={
      mark size=1, mark options={opacity=0.3}
    },
    tick pos = left,
    xmajorticks = true,
    ymajorticks = true,
    ylabel near ticks,
    xlabel near ticks,
    xtick align = inside,
    ytick align = inside,
    legend cell align = left,
    legend columns = 1,
    legend pos = north west,
    legend style = {
      fill opacity = 0.9,
      text opacity = 1,
      font = \small,
    },
    xticklabel style = {font = \small, inner xsep = 0ex},
    xlabel style = {font = \small},
    axis line style = {black},
    yticklabel style = {font = \small, inner ysep = 0ex},
    ylabel style = {font = \small, inner ysep = 0ex},
    title style = {font = \small, inner ysep = 0ex, yshift = -0.75ex},
    grid = major,
    grid style = {dashed}
  }
}
\pgfkeys{/pgfplots/spectrumdefaultleft/.style={
    spectrumdefault,
    title=\empty,
    ymax=3e-4,
    xlabel=\phantom{eigenvalues},
  }}
\pgfkeys{/pgfplots/spectrumdefaultcenter/.style={
    spectrumdefault,
    ylabel=\empty,
    yticklabels=\empty,
    title=\empty,
    ymax=3e-4,
  }}
\pgfkeys{/pgfplots/spectrumdefaultright/.style={
    spectrumdefault,
    ylabel=\empty,
    yticklabels=\empty,
    title=\empty,
    ymax=3e-4,
    xlabel=\phantom{eigenvalues},
  }}
 \pgfkeys{/pgfplots/zmystyle/.style={spectrumdefaultright, ymax = 3e-3,
          title={mc, full}}}
      \begin{tikzpicture}

\definecolor{color0}{rgb}{0.870588235294118,0.623529411764706,0.0862745098039216}

\begin{axis}[
axis line style={white!80!black},
log basis x={10},
tick pos=left,
title={full\_batch\_mc, layerwise\_group, N=128, D=895210},
xlabel={eigenvalues},
xmin=0.0001, xmax=7.47674083709717,
xmode=log,
ylabel={density},
ymin=1.11705508533742e-06, ymax=1,
ymode=log,
zmystyle
]
\draw[draw=white,fill=color0] (axis cs:0.0001,1.11705508533742e-06) rectangle (axis cs:0.000147251268855062,0.999145451903588);
\draw[draw=white,fill=color0] (axis cs:0.000147251268855062,1.11705508533742e-06) rectangle (axis cs:0.000216829361794257,4.46822408474777e-06);
\draw[draw=white,fill=color0] (axis cs:0.000216829361794257,1.11705508533742e-06) rectangle (axis cs:0.000319283986492377,5.58528041795857e-06);
\draw[draw=white,fill=color0] (axis cs:0.000319283986492377,1.11705508533742e-06) rectangle (axis cs:0.00047014972136105,1.11705620837905e-05);
\draw[draw=white,fill=color0] (axis cs:0.00047014972136105,1.11705508533742e-06) rectangle (axis cs:0.000692301430222684,1.34046747499901e-05);
\draw[draw=white,fill=color0] (axis cs:0.000692301430222684,1.11705508533742e-06) rectangle (axis cs:0.00101942264030464,1.11705620837905e-05);
\draw[draw=white,fill=color0] (axis cs:0.00101942264030464,1.11705508533742e-06) rectangle (axis cs:0.00150111277284436,1.56387874163006e-05);
\draw[draw=white,fill=color0] (axis cs:0.00150111277284436,1.11705508533742e-06) rectangle (axis cs:0.00221040760495872,1.8989956415822e-05);
\draw[draw=white,fill=color0] (axis cs:0.00221040760495872,1.11705508533742e-06) rectangle (axis cs:0.00325485324517051,3.35116887468962e-05);
\draw[draw=white,fill=color0] (axis cs:0.00325485324517051,1.11705508533742e-06) rectangle (axis cs:0.00479281270288374,3.68628577463066e-05);
\draw[draw=white,fill=color0] (axis cs:0.00479281270288374,1.11705508533742e-06) rectangle (axis cs:0.00705747751884289,5.13845900772698e-05);
\draw[draw=white,fill=color0] (axis cs:0.00705747751884289,1.11705508533742e-06) rectangle (axis cs:0.0103922251956569,5.47357590767912e-05);
\draw[draw=white,fill=color0] (axis cs:0.0103922251956569,1.11705508533742e-06) rectangle (axis cs:0.0153026834628802,7.03745477409652e-05);
\draw[draw=white,fill=color0] (axis cs:0.0153026834628802,1.11705508533742e-06) rectangle (axis cs:0.0225333955679649,7.03745477409652e-05);
\draw[draw=white,fill=color0] (axis cs:0.0225333955679649,1.11705508533742e-06) rectangle (axis cs:0.0331807108899586,6.92574914078654e-05);
\draw[draw=white,fill=color0] (axis cs:0.0331807108899586,1.11705508533742e-06) rectangle (axis cs:0.0488590178005937,6.47892660751332e-05);
\draw[draw=white,fill=color0] (axis cs:0.0488590178005937,1.11705508533742e-06) rectangle (axis cs:0.0719455236614949,7.1491604074176e-05);
\draw[draw=white,fill=color0] (axis cs:0.0719455236614949,1.11705508533742e-06) rectangle (axis cs:0.10594069647597,6.36722097420335e-05);
\draw[draw=white,fill=color0] (axis cs:0.10594069647597,1.11705508533742e-06) rectangle (axis cs:0.155999019794756,5.80869280762015e-05);
\draw[draw=white,fill=color0] (axis cs:0.155999019794756,1.11705508533742e-06) rectangle (axis cs:0.229710536049237,5.026753374417e-05);
\draw[draw=white,fill=color0] (axis cs:0.229710536049237,1.11705508533742e-06) rectangle (axis cs:0.338251679026266,3.35116887468962e-05);
\draw[draw=white,fill=color0] (axis cs:0.338251679026266,1.11705508533742e-06) rectangle (axis cs:0.498079889289728,2.12240690821326e-05);
\draw[draw=white,fill=color0] (axis cs:0.498079889289728,1.11705508533742e-06) rectangle (axis cs:0.733428956891012,1.22876184168903e-05);
\draw[draw=white,fill=color0] (axis cs:0.733428956891012,1.11705508533742e-06) rectangle (axis cs:1.07998344517246,1.11705620836795e-05);
\draw[draw=white,fill=color0] (axis cs:1.07998344517246,1.11705508533742e-06) rectangle (axis cs:1.59028932644106,1.11705620837905e-05);
\draw[draw=white,fill=color0] (axis cs:1.59028932644106,1.11705508533742e-06) rectangle (axis cs:2.34172121165108,1.00535057505797e-05);
\draw[draw=white,fill=color0] (axis cs:2.34172121165108,1.11705508533742e-06) rectangle (axis cs:3.44821419720435,5.58528041795857e-06);
\draw[draw=white,fill=color0] (axis cs:3.44821419720435,1.11705508533742e-06) rectangle (axis cs:5.07753915822379,4.46822408474777e-06);
\draw[draw=white,fill=color0] (axis cs:5.07753915822379,1.11705508533742e-06) rectangle (axis cs:7.47674083709717,2.23411141854822e-06);
\end{axis}

\end{tikzpicture}
     \end{minipage}
  \end{minipage}

  \begin{small}
    \textbf{\cifarten ResNet32}
  \end{small}

  \begin{minipage}{0.495\linewidth}
    \centering
    \begin{small}
      \textbf{Full network}
    \end{small}
    \vspace{1ex}

    \begin{minipage}[t]{0.33\linewidth}
      \centering
\pgfkeys{/pgfplots/spectrumdefault/.style={
    width=1.38\linewidth,
    height=\goldenRatio*1.38\linewidth,
    every axis plot/.append style={line width = 1.2pt},
    every axis plot post/.append style={
      mark size=1, mark options={opacity=0.3}
    },
    tick pos = left,
    xmajorticks = true,
    ymajorticks = true,
    ylabel near ticks,
    xlabel near ticks,
    xtick align = inside,
    ytick align = inside,
    legend cell align = left,
    legend columns = 1,
    legend pos = north west,
    legend style = {
      fill opacity = 0.9,
      text opacity = 1,
      font = \small,
    },
    xticklabel style = {font = \small, inner xsep = 0ex},
    xlabel style = {font = \small},
    axis line style = {black},
    yticklabel style = {font = \small, inner ysep = 0ex},
    ylabel style = {font = \small, inner ysep = 0ex},
    title style = {font = \small, inner ysep = 0ex, yshift = -0.75ex},
    grid = major,
    grid style = {dashed}
  }
}
\pgfkeys{/pgfplots/spectrumdefaultleft/.style={
    spectrumdefault,
    title=\empty,
    ymax=3e-4,
    xlabel=\phantom{eigenvalues},
  }}
\pgfkeys{/pgfplots/spectrumdefaultcenter/.style={
    spectrumdefault,
    ylabel=\empty,
    yticklabels=\empty,
    title=\empty,
    ymax=3e-4,
  }}
\pgfkeys{/pgfplots/spectrumdefaultright/.style={
    spectrumdefault,
    ylabel=\empty,
    yticklabels=\empty,
    title=\empty,
    ymax=3e-4,
    xlabel=\phantom{eigenvalues},
  }}
 \pgfkeys{/pgfplots/zmystyle/.style={spectrumdefaultleft, ymax = 1e-4,
          title={exact, mb}}}
      \begin{tikzpicture}

\begin{axis}[
axis line style={white!80!black},
log basis x={10},
tick pos=left,
title={full\_batch\_exact, one\_group, N=128, D=464154},
xlabel={eigenvalues},
xmin=0.0001, xmax=91.7985534667969,
xmode=log,
ylabel={density},
ymin=2.15445271514903e-06, ymax=1,
ymode=log,
zmystyle
]
\draw[draw=white,fill=black] (axis cs:0.0001,2.15445271514903e-06) rectangle (axis cs:0.000160551744524522,0.999317037012497);
\draw[draw=white,fill=black] (axis cs:0.000160551744524522,2.15445271514903e-06) rectangle (axis cs:0.000257768626698675,3.01623983538796e-05);
\draw[draw=white,fill=black] (axis cs:0.000257768626698675,2.15445271514903e-06) rectangle (axis cs:0.000413852027001628,2.58534836401775e-05);
\draw[draw=white,fill=black] (axis cs:0.000413852027001628,2.15445271514903e-06) rectangle (axis cs:0.00066444664910121,3.66257704243217e-05);
\draw[draw=white,fill=black] (axis cs:0.00066444664910121,2.15445271514903e-06) rectangle (axis cs:0.00106678068656673,4.09346851379128e-05);
\draw[draw=white,fill=black] (axis cs:0.00106678068656673,2.15445271514903e-06) rectangle (axis cs:0.00171273500253356,3.87802277811727e-05);
\draw[draw=white,fill=black] (axis cs:0.00171273500253356,2.15445271514903e-06) rectangle (axis cs:0.00274982592564975,3.87802277811727e-05);
\draw[draw=white,fill=black] (axis cs:0.00274982592564975,2.15445271514903e-06) rectangle (axis cs:0.00441489349501828,3.66257704243217e-05);
\draw[draw=white,fill=black] (axis cs:0.00441489349501828,2.15445271514903e-06) rectangle (axis cs:0.0070881885251515,3.44713130674707e-05);
\draw[draw=white,fill=black] (axis cs:0.0070881885251515,2.15445271514903e-06) rectangle (axis cs:0.0113802103323178,4.09346851380238e-05);
\draw[draw=white,fill=black] (axis cs:0.0113802103323178,2.15445271514903e-06) rectangle (axis cs:0.0182711262190961,4.09346851379128e-05);
\draw[draw=white,fill=black] (axis cs:0.0182711262190961,2.15445271514903e-06) rectangle (axis cs:0.0293346118890363,4.09346851380238e-05);
\draw[draw=white,fill=black] (axis cs:0.0293346118890363,2.15445271514903e-06) rectangle (axis cs:0.0470972311373457,4.73980572083549e-05);
\draw[draw=white,fill=black] (axis cs:0.0470972311373457,2.15445271514903e-06) rectangle (axis cs:0.0756154262137552,5.3861429278908e-05);
\draw[draw=white,fill=black] (axis cs:0.0756154262137552,2.15445271514903e-06) rectangle (axis cs:0.121401885915837,4.9552514565317e-05);
\draw[draw=white,fill=black] (axis cs:0.121401885915837,2.15445271514903e-06) rectangle (axis cs:0.194912845723547,4.52435998516149e-05);
\draw[draw=white,fill=black] (axis cs:0.194912845723547,2.15445271514903e-06) rectangle (axis cs:0.312935974111546,2.80079409970285e-05);
\draw[draw=white,fill=black] (axis cs:0.312935974111546,2.15445271514903e-06) rectangle (axis cs:0.502424165680895,2.58534836401775e-05);
\draw[draw=white,fill=black] (axis cs:0.502424165680895,2.15445271514903e-06) rectangle (axis cs:0.806650762913453,2.36990262834375e-05);
\draw[draw=white,fill=black] (axis cs:0.806650762913454,2.15445271514903e-06) rectangle (axis cs:1.29509187207792,1.72356542128843e-05);
\draw[draw=white,fill=black] (axis cs:1.29509187207792,2.15445271514903e-06) rectangle (axis cs:2.0792925938164,1.07722821424422e-05);
\draw[draw=white,fill=black] (axis cs:2.0792925938164,2.15445271514903e-06) rectangle (axis cs:3.33834053314142,1.07722821424422e-05);
\draw[draw=white,fill=black] (axis cs:3.33834053314142,2.15445271514903e-06) rectangle (axis cs:5.3597639641278,6.46336742874011e-06);
\draw[draw=white,fill=black] (axis cs:5.3597639641278,2.15445271514903e-06) rectangle (axis cs:8.60519454680389,8.61782478570218e-06);
\draw[draw=white,fill=black] (axis cs:8.60519454680389,2.15445271514903e-06) rectangle (axis cs:13.8157899646227,2.15445271514903e-06);
\draw[draw=white,fill=black] (axis cs:13.8157899646227,2.15445271514903e-06) rectangle (axis cs:22.1814918080457,2.15445271514903e-06);
\draw[draw=white,fill=black] (axis cs:22.1814918080457,2.15445271514903e-06) rectangle (axis cs:35.6127720593814,2.15445271514903e-06);
\draw[draw=white,fill=black] (axis cs:35.6127720593813,2.15445271514903e-06) rectangle (axis cs:57.1769268148785,4.30891007188906e-06);
\draw[draw=white,fill=black] (axis cs:57.1769268148785,2.15445271514903e-06) rectangle (axis cs:91.7985534667969,2.15445271514903e-06);
\end{axis}

\end{tikzpicture}
     \end{minipage}
    \hspace{1.9ex}
    \begin{minipage}[t]{0.33\linewidth}
      \centering
\pgfkeys{/pgfplots/spectrumdefault/.style={
    width=1.38\linewidth,
    height=\goldenRatio*1.38\linewidth,
    every axis plot/.append style={line width = 1.2pt},
    every axis plot post/.append style={
      mark size=1, mark options={opacity=0.3}
    },
    tick pos = left,
    xmajorticks = true,
    ymajorticks = true,
    ylabel near ticks,
    xlabel near ticks,
    xtick align = inside,
    ytick align = inside,
    legend cell align = left,
    legend columns = 1,
    legend pos = north west,
    legend style = {
      fill opacity = 0.9,
      text opacity = 1,
      font = \small,
    },
    xticklabel style = {font = \small, inner xsep = 0ex},
    xlabel style = {font = \small},
    axis line style = {black},
    yticklabel style = {font = \small, inner ysep = 0ex},
    ylabel style = {font = \small, inner ysep = 0ex},
    title style = {font = \small, inner ysep = 0ex, yshift = -0.75ex},
    grid = major,
    grid style = {dashed}
  }
}
\pgfkeys{/pgfplots/spectrumdefaultleft/.style={
    spectrumdefault,
    title=\empty,
    ymax=3e-4,
    xlabel=\phantom{eigenvalues},
  }}
\pgfkeys{/pgfplots/spectrumdefaultcenter/.style={
    spectrumdefault,
    ylabel=\empty,
    yticklabels=\empty,
    title=\empty,
    ymax=3e-4,
  }}
\pgfkeys{/pgfplots/spectrumdefaultright/.style={
    spectrumdefault,
    ylabel=\empty,
    yticklabels=\empty,
    title=\empty,
    ymax=3e-4,
    xlabel=\phantom{eigenvalues},
  }}
 \pgfkeys{/pgfplots/zmystyle/.style={spectrumdefaultcenter, ymax = 1e-4,
          title={exact, sub}}}
      \begin{tikzpicture}

\definecolor{color0}{rgb}{0.274509803921569,0.6,0.564705882352941}

\begin{axis}[
axis line style={white!80!black},
log basis x={10},
tick pos=left,
title={frac\_batch\_exact, one\_group, N=128, D=464154},
xlabel={eigenvalues},
xmin=0.0001, xmax=91.7985534667969,
xmode=log,
ylabel={density},
ymin=2.15445271514903e-06, ymax=1,
ymode=log,
zmystyle
]
\draw[draw=white,fill=color0] (axis cs:0.0001,2.15445271514903e-06) rectangle (axis cs:0.000160551744524522,0.999939675188618);
\draw[draw=white,fill=color0] (axis cs:0.000160551744524522,2.15445271514903e-06) rectangle (axis cs:0.000257768626698675,6.46336742874011e-06);
\draw[draw=white,fill=color0] (axis cs:0.000257768626698675,2.15445271514903e-06) rectangle (axis cs:0.000413852027001628,2.15445271514903e-06);
\draw[draw=white,fill=color0] (axis cs:0.000413852027001628,2.15445271514903e-06) rectangle (axis cs:0.00066444664910121,4.30891007200008e-06);
\draw[draw=white,fill=color0] (axis cs:0.00066444664910121,2.15445271514903e-06) rectangle (axis cs:0.00106678068656673,8.61782478559115e-06);
\draw[draw=white,fill=color0] (axis cs:0.00106678068656673,2.15445271514903e-06) rectangle (axis cs:0.00171273500253356,8.61782478559115e-06);
\draw[draw=white,fill=color0] (axis cs:0.00171273500253356,2.15445271514903e-06) rectangle (axis cs:0.00274982592564975,4.30891007200008e-06);
\draw[draw=white,fill=color0] (axis cs:0.00274982592564975,2.15445271514903e-06) rectangle (axis cs:0.00441489349501828,4.30891007200008e-06);
\draw[draw=white,fill=color0] (axis cs:0.00441489349501828,2.15445271514903e-06) rectangle (axis cs:0.0070881885251515,4.30891007200008e-06);
\draw[draw=white,fill=color0] (axis cs:0.0070881885251515,2.15445271514903e-06) rectangle (axis cs:0.0113802103323178,4.30891007188906e-06);
\draw[draw=white,fill=color0] (axis cs:0.0113802103323178,2.15445271514903e-06) rectangle (axis cs:0.0182711262190961,8.61782478570218e-06);
\draw[draw=white,fill=color0] (axis cs:0.0182711262190961,2.15445271514903e-06) rectangle (axis cs:0.0293346118890363,4.30891007188906e-06);
\draw[draw=white,fill=color0] (axis cs:0.0293346118890363,2.15445271514903e-06) rectangle (axis cs:0.0470972311373457,4.30891007200008e-06);
\draw[draw=white,fill=color0] (axis cs:0.0470972311373457,2.15445271514903e-06) rectangle (axis cs:0.0756154262137552,4.30891007200008e-06);
\draw[draw=white,fill=color0] (axis cs:0.0756154262137552,2.15445271514903e-06) rectangle (axis cs:0.121401885915837,4.30891007200008e-06);
\draw[draw=white,fill=color0] (axis cs:0.121401885915837,2.15445271514903e-06) rectangle (axis cs:0.194912845723547,6.46336742874011e-06);
\draw[draw=white,fill=color0] (axis cs:0.194912845723547,2.15445271514903e-06) rectangle (axis cs:0.312935974111546,4.30891007200008e-06);
\draw[draw=white,fill=color0] (axis cs:0.312935974111546,2.15445271514903e-06) rectangle (axis cs:0.502424165680895,2.15445271514903e-06);
\draw[draw=white,fill=color0] (axis cs:0.502424165680895,2.15445271514903e-06) rectangle (axis cs:0.806650762913453,6.46336742874011e-06);
\draw[draw=white,fill=color0] (axis cs:0.806650762913454,2.15445271514903e-06) rectangle (axis cs:1.29509187207792,2.15445271514903e-06);
\draw[draw=white,fill=color0] (axis cs:1.29509187207792,2.15445271514903e-06) rectangle (axis cs:2.0792925938164,4.30891007200008e-06);
\draw[draw=white,fill=color0] (axis cs:2.0792925938164,2.15445271514903e-06) rectangle (axis cs:3.33834053314142,4.30891007200008e-06);
\draw[draw=white,fill=color0] (axis cs:3.33834053314142,2.15445271514903e-06) rectangle (axis cs:5.3597639641278,4.30891007200008e-06);
\draw[draw=white,fill=color0] (axis cs:5.3597639641278,2.15445271514903e-06) rectangle (axis cs:8.60519454680389,2.15445271514903e-06);
\draw[draw=white,fill=color0] (axis cs:8.60519454680389,2.15445271514903e-06) rectangle (axis cs:13.8157899646227,2.15445271514903e-06);
\draw[draw=white,fill=color0] (axis cs:13.8157899646227,2.15445271514903e-06) rectangle (axis cs:22.1814918080457,2.15445271514903e-06);
\draw[draw=white,fill=color0] (axis cs:22.1814918080457,2.15445271514903e-06) rectangle (axis cs:35.6127720593814,4.30891007188906e-06);
\draw[draw=white,fill=color0] (axis cs:35.6127720593813,2.15445271514903e-06) rectangle (axis cs:57.1769268148785,2.15445271514903e-06);
\draw[draw=white,fill=color0] (axis cs:57.1769268148785,2.15445271514903e-06) rectangle (axis cs:91.7985534667969,2.15445271514903e-06);
\end{axis}

\end{tikzpicture}
     \end{minipage}
    \hspace{-3.5ex}
    \begin{minipage}[t]{0.33\linewidth}
      \centering
\pgfkeys{/pgfplots/spectrumdefault/.style={
    width=1.38\linewidth,
    height=\goldenRatio*1.38\linewidth,
    every axis plot/.append style={line width = 1.2pt},
    every axis plot post/.append style={
      mark size=1, mark options={opacity=0.3}
    },
    tick pos = left,
    xmajorticks = true,
    ymajorticks = true,
    ylabel near ticks,
    xlabel near ticks,
    xtick align = inside,
    ytick align = inside,
    legend cell align = left,
    legend columns = 1,
    legend pos = north west,
    legend style = {
      fill opacity = 0.9,
      text opacity = 1,
      font = \small,
    },
    xticklabel style = {font = \small, inner xsep = 0ex},
    xlabel style = {font = \small},
    axis line style = {black},
    yticklabel style = {font = \small, inner ysep = 0ex},
    ylabel style = {font = \small, inner ysep = 0ex},
    title style = {font = \small, inner ysep = 0ex, yshift = -0.75ex},
    grid = major,
    grid style = {dashed}
  }
}
\pgfkeys{/pgfplots/spectrumdefaultleft/.style={
    spectrumdefault,
    title=\empty,
    ymax=3e-4,
    xlabel=\phantom{eigenvalues},
  }}
\pgfkeys{/pgfplots/spectrumdefaultcenter/.style={
    spectrumdefault,
    ylabel=\empty,
    yticklabels=\empty,
    title=\empty,
    ymax=3e-4,
  }}
\pgfkeys{/pgfplots/spectrumdefaultright/.style={
    spectrumdefault,
    ylabel=\empty,
    yticklabels=\empty,
    title=\empty,
    ymax=3e-4,
    xlabel=\phantom{eigenvalues},
  }}
 \pgfkeys{/pgfplots/zmystyle/.style={spectrumdefaultright, ymax = 1e-4,
          title={mc, full}}}
      \begin{tikzpicture}

\definecolor{color0}{rgb}{0.870588235294118,0.623529411764706,0.0862745098039216}

\begin{axis}[
axis line style={white!80!black},
log basis x={10},
tick pos=left,
title={full\_batch\_mc, one\_group, N=128, D=464154},
xlabel={eigenvalues},
xmin=0.0001, xmax=91.7985534667969,
xmode=log,
ylabel={density},
ymin=2.15445271514903e-06, ymax=1,
ymode=log,
zmystyle
]
\draw[draw=white,fill=color0] (axis cs:0.0001,2.15445271514903e-06) rectangle (axis cs:0.000160551744524522,0.999928902901834);
\draw[draw=white,fill=color0] (axis cs:0.000160551744524522,2.15445271514903e-06) rectangle (axis cs:0.000257768626698675,4.30891007200008e-06);
\draw[draw=white,fill=color0] (axis cs:0.000257768626698675,2.15445271514903e-06) rectangle (axis cs:0.000413852027001628,2.15445271514903e-06);
\draw[draw=white,fill=color0] (axis cs:0.000413852027001628,2.15445271514903e-06) rectangle (axis cs:0.00066444664910121,4.30891007188906e-06);
\draw[draw=white,fill=color0] (axis cs:0.00066444664910121,2.15445271514903e-06) rectangle (axis cs:0.00106678068656673,6.46336742885113e-06);
\draw[draw=white,fill=color0] (axis cs:0.00106678068656673,2.15445271514903e-06) rectangle (axis cs:0.00171273500253356,4.30891007200008e-06);
\draw[draw=white,fill=color0] (axis cs:0.00171273500253356,2.15445271514903e-06) rectangle (axis cs:0.00274982592564975,4.30891007188906e-06);
\draw[draw=white,fill=color0] (axis cs:0.00274982592564975,2.15445271514903e-06) rectangle (axis cs:0.00441489349501828,6.46336742885113e-06);
\draw[draw=white,fill=color0] (axis cs:0.00441489349501828,2.15445271514903e-06) rectangle (axis cs:0.0070881885251515,2.15445271514903e-06);
\draw[draw=white,fill=color0] (axis cs:0.0070881885251515,2.15445271514903e-06) rectangle (axis cs:0.0113802103323178,6.46336742874011e-06);
\draw[draw=white,fill=color0] (axis cs:0.0113802103323178,2.15445271514903e-06) rectangle (axis cs:0.0182711262190961,4.30891007200008e-06);
\draw[draw=white,fill=color0] (axis cs:0.0182711262190961,2.15445271514903e-06) rectangle (axis cs:0.0293346118890363,6.46336742885113e-06);
\draw[draw=white,fill=color0] (axis cs:0.0293346118890363,2.15445271514903e-06) rectangle (axis cs:0.0470972311373457,8.61782478559115e-06);
\draw[draw=white,fill=color0] (axis cs:0.0470972311373457,2.15445271514903e-06) rectangle (axis cs:0.0756154262137552,6.46336742874011e-06);
\draw[draw=white,fill=color0] (axis cs:0.0756154262137552,2.15445271514903e-06) rectangle (axis cs:0.121401885915837,6.46336742885113e-06);
\draw[draw=white,fill=color0] (axis cs:0.121401885915837,2.15445271514903e-06) rectangle (axis cs:0.194912845723547,6.46336742874011e-06);
\draw[draw=white,fill=color0] (axis cs:0.194912845723547,2.15445271514903e-06) rectangle (axis cs:0.312935974111546,4.30891007200008e-06);
\draw[draw=white,fill=color0] (axis cs:0.312935974111546,2.15445271514903e-06) rectangle (axis cs:0.502424165680895,2.15445271514903e-06);
\draw[draw=white,fill=color0] (axis cs:0.502424165680895,2.15445271514903e-06) rectangle (axis cs:0.806650762913453,6.46336742885113e-06);
\draw[draw=white,fill=color0] (axis cs:0.806650762913454,2.15445271514903e-06) rectangle (axis cs:1.29509187207792,6.46336742874011e-06);
\draw[draw=white,fill=color0] (axis cs:1.29509187207792,2.15445271514903e-06) rectangle (axis cs:2.0792925938164,4.30891007200008e-06);
\draw[draw=white,fill=color0] (axis cs:2.0792925938164,2.15445271514903e-06) rectangle (axis cs:3.33834053314142,4.30891007200008e-06);
\draw[draw=white,fill=color0] (axis cs:3.33834053314142,2.15445271514903e-06) rectangle (axis cs:5.3597639641278,2.15445271514903e-06);
\draw[draw=white,fill=color0] (axis cs:5.3597639641278,2.15445271514903e-06) rectangle (axis cs:8.60519454680389,6.46336742874011e-06);
\draw[draw=white,fill=color0] (axis cs:8.60519454680389,2.15445271514903e-06) rectangle (axis cs:13.8157899646227,2.15445271514903e-06);
\draw[draw=white,fill=color0] (axis cs:13.8157899646227,2.15445271514903e-06) rectangle (axis cs:22.1814918080457,4.30891007200008e-06);
\draw[draw=white,fill=color0] (axis cs:22.1814918080457,2.15445271514903e-06) rectangle (axis cs:35.6127720593814,4.30891007200008e-06);
\draw[draw=white,fill=color0] (axis cs:35.6127720593813,2.15445271514903e-06) rectangle (axis cs:57.1769268148785,2.15445271514903e-06);
\draw[draw=white,fill=color0] (axis cs:57.1769268148785,2.15445271514903e-06) rectangle (axis cs:91.7985534667969,4.30891007188906e-06);
\end{axis}

\end{tikzpicture}
     \end{minipage}
  \end{minipage}
  \hfill
  \begin{minipage}{0.495\linewidth}
    \centering
    \begin{small}
      \textbf{Block-diagonal approximation}
    \end{small}
    \vspace{1ex}

    \begin{minipage}[t]{0.33\linewidth}
      \centering
\pgfkeys{/pgfplots/spectrumdefault/.style={
    width=1.38\linewidth,
    height=\goldenRatio*1.38\linewidth,
    every axis plot/.append style={line width = 1.2pt},
    every axis plot post/.append style={
      mark size=1, mark options={opacity=0.3}
    },
    tick pos = left,
    xmajorticks = true,
    ymajorticks = true,
    ylabel near ticks,
    xlabel near ticks,
    xtick align = inside,
    ytick align = inside,
    legend cell align = left,
    legend columns = 1,
    legend pos = north west,
    legend style = {
      fill opacity = 0.9,
      text opacity = 1,
      font = \small,
    },
    xticklabel style = {font = \small, inner xsep = 0ex},
    xlabel style = {font = \small},
    axis line style = {black},
    yticklabel style = {font = \small, inner ysep = 0ex},
    ylabel style = {font = \small, inner ysep = 0ex},
    title style = {font = \small, inner ysep = 0ex, yshift = -0.75ex},
    grid = major,
    grid style = {dashed}
  }
}
\pgfkeys{/pgfplots/spectrumdefaultleft/.style={
    spectrumdefault,
    title=\empty,
    ymax=3e-4,
    xlabel=\phantom{eigenvalues},
  }}
\pgfkeys{/pgfplots/spectrumdefaultcenter/.style={
    spectrumdefault,
    ylabel=\empty,
    yticklabels=\empty,
    title=\empty,
    ymax=3e-4,
  }}
\pgfkeys{/pgfplots/spectrumdefaultright/.style={
    spectrumdefault,
    ylabel=\empty,
    yticklabels=\empty,
    title=\empty,
    ymax=3e-4,
    xlabel=\phantom{eigenvalues},
  }}
 \pgfkeys{/pgfplots/zmystyle/.style={spectrumdefaultleft, ymax = 3e-3,
          title={exact, mb}}}
      \begin{tikzpicture}

\begin{axis}[
axis line style={white!80!black},
log basis x={10},
tick pos=left,
title={full\_batch\_exact, layerwise\_group, N=128, D=464154},
xlabel={eigenvalues},
xmin=0.0001, xmax=7.02149963378906,
xmode=log,
ylabel={density},
ymin=2.15445271514903e-06, ymax=1,
ymode=log,
zmystyle
]
\draw[draw=white,fill=black] (axis cs:0.0001,2.15445271514903e-06) rectangle (axis cs:0.000146932636556619,0.986321350236138);
\draw[draw=white,fill=black] (axis cs:0.000146932636556619,2.15445271514903e-06) rectangle (axis cs:0.000215891996854795,0.00114617130918829);
\draw[draw=white,fill=black] (axis cs:0.000215891996854795,2.15445271514903e-06) rectangle (axis cs:0.000317215803093484,0.0011418623944747);
\draw[draw=white,fill=black] (axis cs:0.000317215803093484,2.15445271514903e-06) rectangle (axis cs:0.000466093543059509,0.0011957238283952);
\draw[draw=white,fill=black] (axis cs:0.000466093543059509,2.15445271514903e-06) rectangle (axis cs:0.000684843531637498,0.00130991006830697);
\draw[draw=white,fill=black] (axis cs:0.000684843531637498,2.15445271514903e-06) rectangle (axis cs:0.00100625865732244,0.00129267440945227);
\draw[draw=white,fill=black] (axis cs:0.00100625865732244,2.15445271514903e-06) rectangle (axis cs:0.0014785223757831,0.00127974766531139);
\draw[draw=white,fill=black] (axis cs:0.0014785223757831,2.15445271514903e-06) rectangle (axis cs:0.00217243190881767,0.00115263468125873);
\draw[draw=white,fill=black] (axis cs:0.00217243190881767,2.15445271514903e-06) rectangle (axis cs:0.00319201148102309,0.00095873351914469);
\draw[draw=white,fill=black] (axis cs:0.00319201148102309,2.15445271514903e-06) rectangle (axis cs:0.00469010662825721,0.000801458132096562);
\draw[draw=white,fill=black] (axis cs:0.00469010662825721,2.15445271514903e-06) rectangle (axis cs:0.00689129732621506,0.000687271892185011);
\draw[draw=white,fill=black] (axis cs:0.00689129732621506,2.15445271514903e-06) rectangle (axis cs:0.0101255648543636,0.000570931194916498);
\draw[draw=white,fill=black] (axis cs:0.0101255648543636,2.15445271514903e-06) rectangle (axis cs:0.0148777594067668,0.000473980613859422);
\draw[draw=white,fill=black] (axis cs:0.0148777594067668,2.15445271514903e-06) rectangle (axis cs:0.0218602841569129,0.00041365580786837);
\draw[draw=white,fill=black] (axis cs:0.0218602841569129,2.15445271514903e-06) rectangle (axis cs:0.032119891870521,0.000329631970952289);
\draw[draw=white,fill=black] (axis cs:0.032119891870521,2.15445271514903e-06) rectangle (axis cs:0.0471946039844917,0.00027577053703179);
\draw[draw=white,fill=black] (axis cs:0.0471946039844917,2.15445271514903e-06) rectangle (axis cs:0.0693442759468689,0.000185283328045046);
\draw[draw=white,fill=black] (axis cs:0.0693442759468689,2.15445271514903e-06) rectangle (axis cs:0.101889372949832,0.00014003972355184);
\draw[draw=white,fill=black] (axis cs:0.101889372949832,2.15445271514903e-06) rectangle (axis cs:0.149708742046195,0.000107722863199518);
\draw[draw=white,fill=black] (axis cs:0.149708742046195,2.15445271514903e-06) rectangle (axis cs:0.219971001844222,7.75604602039365e-05);
\draw[draw=white,fill=black] (axis cs:0.219971001844222,2.15445271514903e-06) rectangle (axis cs:0.323209192669724,5.60158866357591e-05);
\draw[draw=white,fill=black] (axis cs:0.323209192669724,2.15445271514903e-06) rectangle (axis cs:0.474899788382989,3.01623983538796e-05);
\draw[draw=white,fill=black] (axis cs:0.474899788382989,2.15445271514903e-06) rectangle (axis cs:0.69778278007293,3.66257704243217e-05);
\draw[draw=white,fill=black] (axis cs:0.69778278007293,2.15445271514903e-06) rectangle (axis cs:1.02527063619923,2.15445689265864e-05);
\draw[draw=white,fill=black] (axis cs:1.02527063619923,2.15445271514903e-06) rectangle (axis cs:1.50645717760835,1.72356542128843e-05);
\draw[draw=white,fill=black] (axis cs:1.50645717760835,2.15445271514903e-06) rectangle (axis cs:2.21347724965639,1.07722821424422e-05);
\draw[draw=white,fill=black] (axis cs:2.21347724965639,2.15445271514903e-06) rectangle (axis cs:3.25232048250106,1.93901115697354e-05);
\draw[draw=white,fill=black] (axis cs:3.25232048250106,2.15445271514903e-06) rectangle (axis cs:4.77872023420977,6.46336742874011e-06);
\draw[draw=white,fill=black] (axis cs:4.77872023420977,2.15445271514903e-06) rectangle (axis cs:7.02149963378906,2.15445271514903e-06);
\end{axis}

\end{tikzpicture}
     \end{minipage}
    \hspace{1.9ex}
    \begin{minipage}[t]{0.33\linewidth}
      \centering
\pgfkeys{/pgfplots/spectrumdefault/.style={
    width=1.38\linewidth,
    height=\goldenRatio*1.38\linewidth,
    every axis plot/.append style={line width = 1.2pt},
    every axis plot post/.append style={
      mark size=1, mark options={opacity=0.3}
    },
    tick pos = left,
    xmajorticks = true,
    ymajorticks = true,
    ylabel near ticks,
    xlabel near ticks,
    xtick align = inside,
    ytick align = inside,
    legend cell align = left,
    legend columns = 1,
    legend pos = north west,
    legend style = {
      fill opacity = 0.9,
      text opacity = 1,
      font = \small,
    },
    xticklabel style = {font = \small, inner xsep = 0ex},
    xlabel style = {font = \small},
    axis line style = {black},
    yticklabel style = {font = \small, inner ysep = 0ex},
    ylabel style = {font = \small, inner ysep = 0ex},
    title style = {font = \small, inner ysep = 0ex, yshift = -0.75ex},
    grid = major,
    grid style = {dashed}
  }
}
\pgfkeys{/pgfplots/spectrumdefaultleft/.style={
    spectrumdefault,
    title=\empty,
    ymax=3e-4,
    xlabel=\phantom{eigenvalues},
  }}
\pgfkeys{/pgfplots/spectrumdefaultcenter/.style={
    spectrumdefault,
    ylabel=\empty,
    yticklabels=\empty,
    title=\empty,
    ymax=3e-4,
  }}
\pgfkeys{/pgfplots/spectrumdefaultright/.style={
    spectrumdefault,
    ylabel=\empty,
    yticklabels=\empty,
    title=\empty,
    ymax=3e-4,
    xlabel=\phantom{eigenvalues},
  }}
 \pgfkeys{/pgfplots/zmystyle/.style={spectrumdefaultcenter, ymax = 3e-3,
          title={exact, sub}}}
      \begin{tikzpicture}

\definecolor{color0}{rgb}{0.274509803921569,0.6,0.564705882352941}

\begin{axis}[
axis line style={white!80!black},
log basis x={10},
tick pos=left,
title={frac\_batch\_exact, layerwise\_group, N=128, D=464154},
xlabel={eigenvalues},
xmin=0.0001, xmax=7.02149963378906,
xmode=log,
ylabel={density},
ymin=2.15445271514903e-06, ymax=1,
ymode=log,
zmystyle
]
\draw[draw=white,fill=color0] (axis cs:0.0001,2.15445271514903e-06) rectangle (axis cs:0.000146932636556619,0.998595293797961);
\draw[draw=white,fill=color0] (axis cs:0.000146932636556619,2.15445271514903e-06) rectangle (axis cs:0.000215891996854795,0.0001163406926267);
\draw[draw=white,fill=color0] (axis cs:0.000215891996854795,2.15445271514903e-06) rectangle (axis cs:0.000317215803093484,0.000109877320556369);
\draw[draw=white,fill=color0] (axis cs:0.000317215803093484,2.15445271514903e-06) rectangle (axis cs:0.000466093543059509,0.000112031777913109);
\draw[draw=white,fill=color0] (axis cs:0.000466093543059509,2.15445271514903e-06) rectangle (axis cs:0.000684843531637498,0.000116340692626811);
\draw[draw=white,fill=color0] (axis cs:0.000684843531637498,2.15445271514903e-06) rectangle (axis cs:0.00100625865732244,9.69505764153739e-05);
\draw[draw=white,fill=color0] (axis cs:0.00100625865732244,2.15445271514903e-06) rectangle (axis cs:0.0014785223757831,9.91050337722249e-05);
\draw[draw=white,fill=color0] (axis cs:0.0014785223757831,2.15445271514903e-06) rectangle (axis cs:0.00217243190881767,9.26416617016718e-05);
\draw[draw=white,fill=color0] (axis cs:0.00217243190881767,2.15445271514903e-06) rectangle (axis cs:0.00319201148102309,9.04872043449318e-05);
\draw[draw=white,fill=color0] (axis cs:0.00319201148102309,2.15445271514903e-06) rectangle (axis cs:0.00469010662825721,7.32515454903454e-05);
\draw[draw=white,fill=color0] (axis cs:0.00469010662825721,2.15445271514903e-06) rectangle (axis cs:0.00689129732621506,7.32515454902344e-05);
\draw[draw=white,fill=color0] (axis cs:0.00689129732621506,2.15445271514903e-06) rectangle (axis cs:0.0101255648543636,7.10970881334944e-05);
\draw[draw=white,fill=color0] (axis cs:0.0101255648543636,2.15445271514903e-06) rectangle (axis cs:0.0148777594067668,6.03248013494612e-05);
\draw[draw=white,fill=color0] (axis cs:0.0148777594067668,2.15445271514903e-06) rectangle (axis cs:0.0218602841569129,4.95525145652059e-05);
\draw[draw=white,fill=color0] (axis cs:0.0218602841569129,2.15445271514903e-06) rectangle (axis cs:0.032119891870521,4.73980572084659e-05);
\draw[draw=white,fill=color0] (axis cs:0.032119891870521,2.15445271514903e-06) rectangle (axis cs:0.0471946039844917,4.95525145652059e-05);
\draw[draw=white,fill=color0] (axis cs:0.0471946039844917,2.15445271514903e-06) rectangle (axis cs:0.0693442759468689,4.73980572084659e-05);
\draw[draw=white,fill=color0] (axis cs:0.0693442759468689,2.15445271514903e-06) rectangle (axis cs:0.101889372949832,1.72356542128843e-05);
\draw[draw=white,fill=color0] (axis cs:0.101889372949832,2.15445271514903e-06) rectangle (axis cs:0.149708742046195,2.15445689265864e-05);
\draw[draw=white,fill=color0] (axis cs:0.149708742046195,2.15445271514903e-06) rectangle (axis cs:0.219971001844222,1.72356542128843e-05);
\draw[draw=white,fill=color0] (axis cs:0.219971001844222,2.15445271514903e-06) rectangle (axis cs:0.323209192669724,2.36990262834375e-05);
\draw[draw=white,fill=color0] (axis cs:0.323209192669724,2.15445271514903e-06) rectangle (axis cs:0.474899788382989,2.58534836401775e-05);
\draw[draw=white,fill=color0] (axis cs:0.474899788382989,2.15445271514903e-06) rectangle (axis cs:0.69778278007293,1.07722821424422e-05);
\draw[draw=white,fill=color0] (axis cs:0.69778278007293,2.15445271514903e-06) rectangle (axis cs:1.02527063619923,1.72356542128843e-05);
\draw[draw=white,fill=color0] (axis cs:1.02527063619923,2.15445271514903e-06) rectangle (axis cs:1.50645717760835,1.50811968561443e-05);
\draw[draw=white,fill=color0] (axis cs:1.50645717760835,2.15445271514903e-06) rectangle (axis cs:2.21347724965639,6.46336742874011e-06);
\draw[draw=white,fill=color0] (axis cs:2.21347724965639,2.15445271514903e-06) rectangle (axis cs:3.25232048250106,2.15445271514903e-06);
\draw[draw=white,fill=color0] (axis cs:3.25232048250106,2.15445271514903e-06) rectangle (axis cs:4.77872023420977,2.15445271514903e-06);
\draw[draw=white,fill=color0] (axis cs:4.77872023420977,2.15445271514903e-06) rectangle (axis cs:7.02149963378906,2.15445271514903e-06);
\end{axis}

\end{tikzpicture}
     \end{minipage}
    \hspace{-3.5ex}
    \begin{minipage}[t]{0.33\linewidth}
      \centering
\pgfkeys{/pgfplots/spectrumdefault/.style={
    width=1.38\linewidth,
    height=\goldenRatio*1.38\linewidth,
    every axis plot/.append style={line width = 1.2pt},
    every axis plot post/.append style={
      mark size=1, mark options={opacity=0.3}
    },
    tick pos = left,
    xmajorticks = true,
    ymajorticks = true,
    ylabel near ticks,
    xlabel near ticks,
    xtick align = inside,
    ytick align = inside,
    legend cell align = left,
    legend columns = 1,
    legend pos = north west,
    legend style = {
      fill opacity = 0.9,
      text opacity = 1,
      font = \small,
    },
    xticklabel style = {font = \small, inner xsep = 0ex},
    xlabel style = {font = \small},
    axis line style = {black},
    yticklabel style = {font = \small, inner ysep = 0ex},
    ylabel style = {font = \small, inner ysep = 0ex},
    title style = {font = \small, inner ysep = 0ex, yshift = -0.75ex},
    grid = major,
    grid style = {dashed}
  }
}
\pgfkeys{/pgfplots/spectrumdefaultleft/.style={
    spectrumdefault,
    title=\empty,
    ymax=3e-4,
    xlabel=\phantom{eigenvalues},
  }}
\pgfkeys{/pgfplots/spectrumdefaultcenter/.style={
    spectrumdefault,
    ylabel=\empty,
    yticklabels=\empty,
    title=\empty,
    ymax=3e-4,
  }}
\pgfkeys{/pgfplots/spectrumdefaultright/.style={
    spectrumdefault,
    ylabel=\empty,
    yticklabels=\empty,
    title=\empty,
    ymax=3e-4,
    xlabel=\phantom{eigenvalues},
  }}
 \pgfkeys{/pgfplots/zmystyle/.style={spectrumdefaultright, ymax = 3e-3,
          title={mc, full}}}
      \begin{tikzpicture}

\definecolor{color0}{rgb}{0.870588235294118,0.623529411764706,0.0862745098039216}

\begin{axis}[
axis line style={white!80!black},
log basis x={10},
tick pos=left,
title={full\_batch\_mc, layerwise\_group, N=128, D=464154},
xlabel={eigenvalues},
xmin=0.0001, xmax=7.02149963378906,
xmode=log,
ylabel={density},
ymin=2.15445271514903e-06, ymax=1,
ymode=log,
zmystyle
]
\draw[draw=white,fill=color0] (axis cs:0.0001,2.15445271514903e-06) rectangle (axis cs:0.000146932636556619,0.997651641475672);
\draw[draw=white,fill=color0] (axis cs:0.000146932636556619,2.15445271514903e-06) rectangle (axis cs:0.000215891996854795,0.000118495149983662);
\draw[draw=white,fill=color0] (axis cs:0.000215891996854795,2.15445271514903e-06) rectangle (axis cs:0.000317215803093484,0.000122804064697253);
\draw[draw=white,fill=color0] (axis cs:0.000317215803093484,2.15445271514903e-06) rectangle (axis cs:0.000466093543059509,0.000161584297120017);
\draw[draw=white,fill=color0] (axis cs:0.000466093543059509,2.15445271514903e-06) rectangle (axis cs:0.000684843531637498,0.00014003972355184);
\draw[draw=white,fill=color0] (axis cs:0.000684843531637498,2.15445271514903e-06) rectangle (axis cs:0.00100625865732244,0.00016804766919057);
\draw[draw=white,fill=color0] (axis cs:0.00100625865732244,2.15445271514903e-06) rectangle (axis cs:0.0014785223757831,0.000118495149983551);
\draw[draw=white,fill=color0] (axis cs:0.0014785223757831,2.15445271514903e-06) rectangle (axis cs:0.00217243190881767,0.000146503095622393);
\draw[draw=white,fill=color0] (axis cs:0.00217243190881767,2.15445271514903e-06) rectangle (axis cs:0.00319201148102309,0.000131421894124547);
\draw[draw=white,fill=color0] (axis cs:0.00319201148102309,2.15445271514903e-06) rectangle (axis cs:0.00469010662825721,0.000127112979410844);
\draw[draw=white,fill=color0] (axis cs:0.00469010662825721,2.15445271514903e-06) rectangle (axis cs:0.00689129732621506,0.000135730808838249);
\draw[draw=white,fill=color0] (axis cs:0.00689129732621506,2.15445271514903e-06) rectangle (axis cs:0.0101255648543636,0.000146503095622282);
\draw[draw=white,fill=color0] (axis cs:0.0101255648543636,2.15445271514903e-06) rectangle (axis cs:0.0148777594067668,0.000103413948485816);
\draw[draw=white,fill=color0] (axis cs:0.0148777594067668,2.15445271514903e-06) rectangle (axis cs:0.0218602841569129,0.000107722863199518);
\draw[draw=white,fill=color0] (axis cs:0.0218602841569129,2.15445271514903e-06) rectangle (axis cs:0.032119891870521,7.54060028470854e-05);
\draw[draw=white,fill=color0] (axis cs:0.032119891870521,2.15445271514903e-06) rectangle (axis cs:0.0471946039844917,9.47961190586338e-05);
\draw[draw=white,fill=color0] (axis cs:0.0471946039844917,2.15445271514903e-06) rectangle (axis cs:0.0693442759468689,7.75604602039365e-05);
\draw[draw=white,fill=color0] (axis cs:0.0693442759468689,2.15445271514903e-06) rectangle (axis cs:0.101889372949832,4.95525145652059e-05);
\draw[draw=white,fill=color0] (axis cs:0.101889372949832,2.15445271514903e-06) rectangle (axis cs:0.149708742046195,7.54060028471965e-05);
\draw[draw=white,fill=color0] (axis cs:0.149708742046195,2.15445271514903e-06) rectangle (axis cs:0.219971001844222,5.1706971922057e-05);
\draw[draw=white,fill=color0] (axis cs:0.219971001844222,2.15445271514903e-06) rectangle (axis cs:0.323209192669724,5.3861429278908e-05);
\draw[draw=white,fill=color0] (axis cs:0.323209192669724,2.15445271514903e-06) rectangle (axis cs:0.474899788382989,4.52435998516149e-05);
\draw[draw=white,fill=color0] (axis cs:0.474899788382989,2.15445271514903e-06) rectangle (axis cs:0.69778278007293,3.01623983538796e-05);
\draw[draw=white,fill=color0] (axis cs:0.69778278007293,2.15445271514903e-06) rectangle (axis cs:1.02527063619923,2.80079409970285e-05);
\draw[draw=white,fill=color0] (axis cs:1.02527063619923,2.15445271514903e-06) rectangle (axis cs:1.50645717760835,3.66257704243217e-05);
\draw[draw=white,fill=color0] (axis cs:1.50645717760835,2.15445271514903e-06) rectangle (axis cs:2.21347724965639,2.15445689265864e-05);
\draw[draw=white,fill=color0] (axis cs:2.21347724965639,2.15445271514903e-06) rectangle (axis cs:3.25232048250106,8.61782478559115e-06);
\draw[draw=white,fill=color0] (axis cs:3.25232048250106,2.15445271514903e-06) rectangle (axis cs:4.77872023420977,2.36990262833264e-05);
\draw[draw=white,fill=color0] (axis cs:4.77872023420977,2.15445271514903e-06) rectangle (axis cs:7.02149963378906,1.07722821424422e-05);
\end{axis}

\end{tikzpicture}
     \end{minipage}
  \end{minipage}
   
\caption{\textbf{\ggn spectra of different architectures under \vivittitle's
      approximations:} Left and right columns contain results with the full
    network's \ggn and a per-layer block-diagonal approximation, respectively.}
  \label{fig:spectrum_1}   
   
\end{figure}
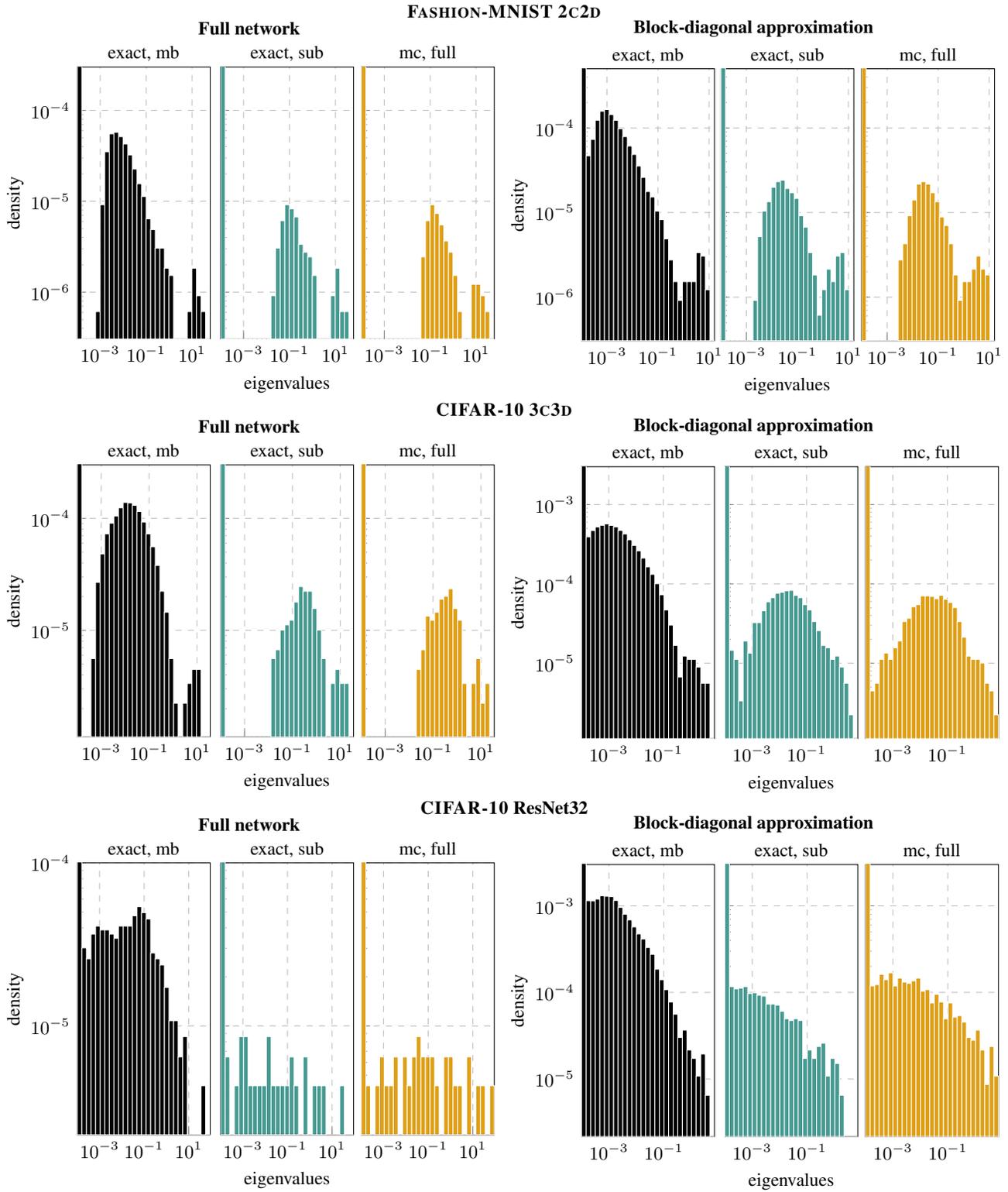

\begin{figure}[t]
  \centering

  \begin{small}
    \textbf{\cifarten ResNet56}
  \end{small}

  \begin{minipage}{0.495\linewidth}
    \centering
    \begin{small}
      \textbf{Full network}
    \end{small}
    \vspace{1ex}

    \begin{minipage}[t]{0.33\linewidth}
      \centering
\pgfkeys{/pgfplots/spectrumdefault/.style={
    width=1.38\linewidth,
    height=\goldenRatio*1.38\linewidth,
    every axis plot/.append style={line width = 1.2pt},
    every axis plot post/.append style={
      mark size=1, mark options={opacity=0.3}
    },
    tick pos = left,
    xmajorticks = true,
    ymajorticks = true,
    ylabel near ticks,
    xlabel near ticks,
    xtick align = inside,
    ytick align = inside,
    legend cell align = left,
    legend columns = 1,
    legend pos = north west,
    legend style = {
      fill opacity = 0.9,
      text opacity = 1,
      font = \small,
    },
    xticklabel style = {font = \small, inner xsep = 0ex},
    xlabel style = {font = \small},
    axis line style = {black},
    yticklabel style = {font = \small, inner ysep = 0ex},
    ylabel style = {font = \small, inner ysep = 0ex},
    title style = {font = \small, inner ysep = 0ex, yshift = -0.75ex},
    grid = major,
    grid style = {dashed}
  }
}
\pgfkeys{/pgfplots/spectrumdefaultleft/.style={
    spectrumdefault,
    title=\empty,
    ymax=3e-4,
    xlabel=\phantom{eigenvalues},
  }}
\pgfkeys{/pgfplots/spectrumdefaultcenter/.style={
    spectrumdefault,
    ylabel=\empty,
    yticklabels=\empty,
    title=\empty,
    ymax=3e-4,
  }}
\pgfkeys{/pgfplots/spectrumdefaultright/.style={
    spectrumdefault,
    ylabel=\empty,
    yticklabels=\empty,
    title=\empty,
    ymax=3e-4,
    xlabel=\phantom{eigenvalues},
  }}
 \pgfkeys{/pgfplots/zmystyle/.style={spectrumdefaultleft, ymax = 1e-4,
          title={exact, mb}}}
      \begin{tikzpicture}

\begin{axis}[
axis line style={white!80!black},
log basis x={10},
tick pos=left,
title={full\_batch\_exact, one\_group, N=128, D=853018},
xlabel={eigenvalues},
xmin=0.0001, xmax=1265.48181152344,
xmode=log,
ylabel={density},
ymin=1.17230683021128e-06, ymax=1,
ymode=log,
zmystyle
]
\draw[draw=white,fill=black] (axis cs:0.0001,1.17230683021128e-06) rectangle (axis cs:0.00017575406401071,0.999459565915986);
\draw[draw=white,fill=black] (axis cs:0.00017575406401071,1.17230683021128e-06) rectangle (axis cs:0.000308894910162809,2.57907791249464e-05);
\draw[draw=white,fill=black] (axis cs:0.000308894910162809,1.17230683021128e-06) rectangle (axis cs:0.000542895358133368,2.69630873294576e-05);
\draw[draw=white,fill=black] (axis cs:0.000542895358133369,1.17230683021128e-06) rectangle (axis cs:0.000954160655244895,2.46184709204352e-05);
\draw[draw=white,fill=black] (axis cs:0.000954160655244895,1.17230683021128e-06) rectangle (axis cs:0.00167697612878413,3.04800119429912e-05);
\draw[draw=white,fill=black] (axis cs:0.00167697612878413,1.17230683021128e-06) rectangle (axis cs:0.00294735369882759,3.16523201475024e-05);
\draw[draw=white,fill=black] (axis cs:0.00294735369882759,1.17230683021128e-06) rectangle (axis cs:0.00518009390645948,2.81353955339688e-05);
\draw[draw=white,fill=black] (axis cs:0.00518009390645948,1.17230683021128e-06) rectangle (axis cs:0.00910422556017368,2.930770373848e-05);
\draw[draw=white,fill=black] (axis cs:0.00910422556017368,1.17230683021128e-06) rectangle (axis cs:0.0160010464187071,2.81353955339688e-05);
\draw[draw=white,fill=black] (axis cs:0.0160010464187071,1.17230683021128e-06) rectangle (axis cs:0.028122489365118,3.28246283520136e-05);
\draw[draw=white,fill=black] (axis cs:0.028122489365118,1.17230683021128e-06) rectangle (axis cs:0.0494264179601746,2.930770373848e-05);
\draw[draw=white,fill=black] (axis cs:0.0494264179601746,1.17230683021128e-06) rectangle (axis cs:0.0868689382599265,2.81353955339688e-05);
\draw[draw=white,fill=black] (axis cs:0.0868689382599265,1.17230683021128e-06) rectangle (axis cs:0.152675689354776,2.69630873294576e-05);
\draw[draw=white,fill=black] (axis cs:0.152675689354776,1.17230683021128e-06) rectangle (axis cs:0.268333728797386,2.930770373848e-05);
\draw[draw=white,fill=black] (axis cs:0.268333728797385,1.17230683021128e-06) rectangle (axis cs:0.471607433472883,2.930770373848e-05);
\draw[draw=white,fill=black] (axis cs:0.471607433472883,1.17230683021128e-06) rectangle (axis cs:0.828869230505199,3.63415529655472e-05);
\draw[draw=white,fill=black] (axis cs:0.828869230505199,1.17230683021128e-06) rectangle (axis cs:1.45677135794719,2.81353955339688e-05);
\draw[draw=white,fill=black] (axis cs:1.45677135794719,1.17230683021128e-06) rectangle (axis cs:2.56033486493619,2.46184709204352e-05);
\draw[draw=white,fill=black] (axis cs:2.56033486493619,1.17230683021128e-06) rectangle (axis cs:4.4998925774085,1.87569298978792e-05);
\draw[draw=white,fill=black] (axis cs:4.4998925774085,1.17230683021128e-06) rectangle (axis cs:7.90874408091173,1.7584621693368e-05);
\draw[draw=white,fill=black] (axis cs:7.90874408091173,1.17230683021128e-06) rectangle (axis cs:13.8999391344089,1.64123134888568e-05);
\draw[draw=white,fill=black] (axis cs:13.8999391344089,1.17230683021128e-06) rectangle (axis cs:24.4297079237387,1.28953888753233e-05);
\draw[draw=white,fill=black] (axis cs:24.4297079237387,1.17230683021128e-06) rectangle (axis cs:42.9362045019173,8.20615605727846e-06);
\draw[draw=white,fill=black] (axis cs:42.9362045019173,1.17230683021128e-06) rectangle (axis cs:75.4621243440693,2.34461503472248e-06);
\draw[draw=white,fill=black] (axis cs:75.4621243440693,1.17230683021128e-06) rectangle (axis cs:132.627750323517,1.17230683021128e-06);
\draw[draw=white,fill=black] (axis cs:132.627750323517,1.17230683021128e-06) rectangle (axis cs:233.09866119956,2.34461503472248e-06);
\draw[draw=white,fill=black] (axis cs:233.09866119956,1.17230683021128e-06) rectangle (axis cs:409.680370212783,1.17230683021128e-06);
\draw[draw=white,fill=black] (axis cs:409.680370212783,1.17230683021128e-06) rectangle (axis cs:720.029900103088,1.17230683021128e-06);
\draw[draw=white,fill=black] (axis cs:720.029900103088,1.17230683021128e-06) rectangle (axis cs:1265.48181152344,2.34461503472248e-06);
\end{axis}

\end{tikzpicture}
     \end{minipage}
    \hspace{1.9ex}
    \begin{minipage}[t]{0.33\linewidth}
      \centering
\pgfkeys{/pgfplots/spectrumdefault/.style={
    width=1.38\linewidth,
    height=\goldenRatio*1.38\linewidth,
    every axis plot/.append style={line width = 1.2pt},
    every axis plot post/.append style={
      mark size=1, mark options={opacity=0.3}
    },
    tick pos = left,
    xmajorticks = true,
    ymajorticks = true,
    ylabel near ticks,
    xlabel near ticks,
    xtick align = inside,
    ytick align = inside,
    legend cell align = left,
    legend columns = 1,
    legend pos = north west,
    legend style = {
      fill opacity = 0.9,
      text opacity = 1,
      font = \small,
    },
    xticklabel style = {font = \small, inner xsep = 0ex},
    xlabel style = {font = \small},
    axis line style = {black},
    yticklabel style = {font = \small, inner ysep = 0ex},
    ylabel style = {font = \small, inner ysep = 0ex},
    title style = {font = \small, inner ysep = 0ex, yshift = -0.75ex},
    grid = major,
    grid style = {dashed}
  }
}
\pgfkeys{/pgfplots/spectrumdefaultleft/.style={
    spectrumdefault,
    title=\empty,
    ymax=3e-4,
    xlabel=\phantom{eigenvalues},
  }}
\pgfkeys{/pgfplots/spectrumdefaultcenter/.style={
    spectrumdefault,
    ylabel=\empty,
    yticklabels=\empty,
    title=\empty,
    ymax=3e-4,
  }}
\pgfkeys{/pgfplots/spectrumdefaultright/.style={
    spectrumdefault,
    ylabel=\empty,
    yticklabels=\empty,
    title=\empty,
    ymax=3e-4,
    xlabel=\phantom{eigenvalues},
  }}
 \pgfkeys{/pgfplots/zmystyle/.style={spectrumdefaultcenter, ymax = 1e-4,
          title={exact, sub}}}
      \begin{tikzpicture}

\definecolor{color0}{rgb}{0.274509803921569,0.6,0.564705882352941}

\begin{axis}[
axis line style={white!80!black},
log basis x={10},
tick pos=left,
title={frac\_batch\_exact, one\_group, N=128, D=853018},
xlabel={eigenvalues},
xmin=0.0001, xmax=1265.48181152344,
xmode=log,
ylabel={density},
ymin=1.17230683021128e-06, ymax=1,
ymode=log,
zmystyle
]
\draw[draw=white,fill=color0] (axis cs:0.0001,1.17230683021128e-06) rectangle (axis cs:0.00017575406401071,0.999915593807541);
\draw[draw=white,fill=color0] (axis cs:0.00017575406401071,1.17230683021128e-06) rectangle (axis cs:0.000308894910162809,2.34461503472248e-06);
\draw[draw=white,fill=color0] (axis cs:0.000308894910162809,1.17230683021128e-06) rectangle (axis cs:0.000542895358133368,3.51692323923368e-06);
\draw[draw=white,fill=color0] (axis cs:0.000542895358133369,1.17230683021128e-06) rectangle (axis cs:0.000954160655244895,1.17230683021128e-06);
\draw[draw=white,fill=color0] (axis cs:0.000954160655244895,1.17230683021128e-06) rectangle (axis cs:0.00167697612878413,4.68923144374487e-06);
\draw[draw=white,fill=color0] (axis cs:0.00167697612878413,1.17230683021128e-06) rectangle (axis cs:0.00294735369882759,3.51692323923368e-06);
\draw[draw=white,fill=color0] (axis cs:0.00294735369882759,1.17230683021128e-06) rectangle (axis cs:0.00518009390645948,5.86153964825607e-06);
\draw[draw=white,fill=color0] (axis cs:0.00518009390645948,1.17230683021128e-06) rectangle (axis cs:0.00910422556017368,4.68923144374487e-06);
\draw[draw=white,fill=color0] (axis cs:0.00910422556017368,1.17230683021128e-06) rectangle (axis cs:0.0160010464187071,5.86153964825607e-06);
\draw[draw=white,fill=color0] (axis cs:0.0160010464187071,1.17230683021128e-06) rectangle (axis cs:0.028122489365118,3.51692323923368e-06);
\draw[draw=white,fill=color0] (axis cs:0.028122489365118,1.17230683021128e-06) rectangle (axis cs:0.0494264179601746,4.68923144374487e-06);
\draw[draw=white,fill=color0] (axis cs:0.0494264179601746,1.17230683021128e-06) rectangle (axis cs:0.0868689382599265,5.86153964825607e-06);
\draw[draw=white,fill=color0] (axis cs:0.0868689382599265,1.17230683021128e-06) rectangle (axis cs:0.152675689354776,4.68923144374487e-06);
\draw[draw=white,fill=color0] (axis cs:0.152675689354776,1.17230683021128e-06) rectangle (axis cs:0.268333728797386,7.03384785276727e-06);
\draw[draw=white,fill=color0] (axis cs:0.268333728797385,1.17230683021128e-06) rectangle (axis cs:0.471607433472883,4.68923144374487e-06);
\draw[draw=white,fill=color0] (axis cs:0.471607433472883,1.17230683021128e-06) rectangle (axis cs:0.828869230505199,4.68923144374487e-06);
\draw[draw=white,fill=color0] (axis cs:0.828869230505199,1.17230683021128e-06) rectangle (axis cs:1.45677135794719,3.51692323923368e-06);
\draw[draw=white,fill=color0] (axis cs:1.45677135794719,1.17230683021128e-06) rectangle (axis cs:2.56033486493619,3.51692323923368e-06);
\draw[draw=white,fill=color0] (axis cs:2.56033486493619,1.17230683021128e-06) rectangle (axis cs:4.4998925774085,7.03384785276727e-06);
\draw[draw=white,fill=color0] (axis cs:4.4998925774085,1.17230683021128e-06) rectangle (axis cs:7.90874408091173,4.68923144374487e-06);
\draw[draw=white,fill=color0] (axis cs:7.90874408091173,1.17230683021128e-06) rectangle (axis cs:13.8999391344089,5.86153964825607e-06);
\draw[draw=white,fill=color0] (axis cs:13.8999391344089,1.17230683021128e-06) rectangle (axis cs:24.4297079237387,4.68923144374487e-06);
\draw[draw=white,fill=color0] (axis cs:24.4297079237387,1.17230683021128e-06) rectangle (axis cs:42.9362045019173,5.86153964825607e-06);
\draw[draw=white,fill=color0] (axis cs:42.9362045019173,1.17230683021128e-06) rectangle (axis cs:75.4621243440693,4.68923144374487e-06);
\draw[draw=white,fill=color0] (axis cs:75.4621243440693,1.17230683021128e-06) rectangle (axis cs:132.627750323517,3.51692323923368e-06);
\draw[draw=white,fill=color0] (axis cs:132.627750323517,1.17230683021128e-06) rectangle (axis cs:233.09866119956,2.34461503472248e-06);
\draw[draw=white,fill=color0] (axis cs:233.09866119956,1.17230683021128e-06) rectangle (axis cs:409.680370212783,2.34461503472248e-06);
\draw[draw=white,fill=color0] (axis cs:409.680370212783,1.17230683021128e-06) rectangle (axis cs:720.029900103088,1.17230683021128e-06);
\draw[draw=white,fill=color0] (axis cs:720.029900103088,1.17230683021128e-06) rectangle (axis cs:1265.48181152344,1.17230683021128e-06);
\end{axis}

\end{tikzpicture}
     \end{minipage}
    \hspace{-3.5ex}
    \begin{minipage}[t]{0.33\linewidth}
      \centering
\pgfkeys{/pgfplots/spectrumdefault/.style={
    width=1.38\linewidth,
    height=\goldenRatio*1.38\linewidth,
    every axis plot/.append style={line width = 1.2pt},
    every axis plot post/.append style={
      mark size=1, mark options={opacity=0.3}
    },
    tick pos = left,
    xmajorticks = true,
    ymajorticks = true,
    ylabel near ticks,
    xlabel near ticks,
    xtick align = inside,
    ytick align = inside,
    legend cell align = left,
    legend columns = 1,
    legend pos = north west,
    legend style = {
      fill opacity = 0.9,
      text opacity = 1,
      font = \small,
    },
    xticklabel style = {font = \small, inner xsep = 0ex},
    xlabel style = {font = \small},
    axis line style = {black},
    yticklabel style = {font = \small, inner ysep = 0ex},
    ylabel style = {font = \small, inner ysep = 0ex},
    title style = {font = \small, inner ysep = 0ex, yshift = -0.75ex},
    grid = major,
    grid style = {dashed}
  }
}
\pgfkeys{/pgfplots/spectrumdefaultleft/.style={
    spectrumdefault,
    title=\empty,
    ymax=3e-4,
    xlabel=\phantom{eigenvalues},
  }}
\pgfkeys{/pgfplots/spectrumdefaultcenter/.style={
    spectrumdefault,
    ylabel=\empty,
    yticklabels=\empty,
    title=\empty,
    ymax=3e-4,
  }}
\pgfkeys{/pgfplots/spectrumdefaultright/.style={
    spectrumdefault,
    ylabel=\empty,
    yticklabels=\empty,
    title=\empty,
    ymax=3e-4,
    xlabel=\phantom{eigenvalues},
  }}
 \pgfkeys{/pgfplots/zmystyle/.style={spectrumdefaultright, ymax = 1e-4,
          title={mc, full}}}
      \begin{tikzpicture}

\definecolor{color0}{rgb}{0.870588235294118,0.623529411764706,0.0862745098039216}

\begin{axis}[
axis line style={white!80!black},
log basis x={10},
tick pos=left,
title={full\_batch\_mc, one\_group, N=128, D=853018},
xlabel={eigenvalues},
xmin=0.0001, xmax=1265.48181152344,
xmode=log,
ylabel={density},
ymin=1.17230683021128e-06, ymax=1,
ymode=log,
zmystyle
]
\draw[draw=white,fill=color0] (axis cs:0.0001,1.17230683021128e-06) rectangle (axis cs:0.00017575406401071,0.999920283040359);
\draw[draw=white,fill=color0] (axis cs:0.00017575406401071,1.17230683021128e-06) rectangle (axis cs:0.000308894910162809,3.51692323923368e-06);
\draw[draw=white,fill=color0] (axis cs:0.000308894910162809,1.17230683021128e-06) rectangle (axis cs:0.000542895358133368,3.51692323923368e-06);
\draw[draw=white,fill=color0] (axis cs:0.000542895358133369,1.17230683021128e-06) rectangle (axis cs:0.000954160655244895,3.51692323923368e-06);
\draw[draw=white,fill=color0] (axis cs:0.000954160655244895,1.17230683021128e-06) rectangle (axis cs:0.00167697612878413,4.68923144374487e-06);
\draw[draw=white,fill=color0] (axis cs:0.00167697612878413,1.17230683021128e-06) rectangle (axis cs:0.00294735369882759,1.17230683021128e-06);
\draw[draw=white,fill=color0] (axis cs:0.00294735369882759,1.17230683021128e-06) rectangle (axis cs:0.00518009390645948,5.86153964825607e-06);
\draw[draw=white,fill=color0] (axis cs:0.00518009390645948,1.17230683021128e-06) rectangle (axis cs:0.00910422556017368,4.68923144374487e-06);
\draw[draw=white,fill=color0] (axis cs:0.00910422556017368,1.17230683021128e-06) rectangle (axis cs:0.0160010464187071,5.86153964825607e-06);
\draw[draw=white,fill=color0] (axis cs:0.0160010464187071,1.17230683021128e-06) rectangle (axis cs:0.028122489365118,4.68923144374487e-06);
\draw[draw=white,fill=color0] (axis cs:0.028122489365118,1.17230683021128e-06) rectangle (axis cs:0.0494264179601746,4.68923144374487e-06);
\draw[draw=white,fill=color0] (axis cs:0.0494264179601746,1.17230683021128e-06) rectangle (axis cs:0.0868689382599265,3.51692323923368e-06);
\draw[draw=white,fill=color0] (axis cs:0.0868689382599265,1.17230683021128e-06) rectangle (axis cs:0.152675689354776,3.51692323923368e-06);
\draw[draw=white,fill=color0] (axis cs:0.152675689354776,1.17230683021128e-06) rectangle (axis cs:0.268333728797386,3.51692323923368e-06);
\draw[draw=white,fill=color0] (axis cs:0.268333728797385,1.17230683021128e-06) rectangle (axis cs:0.471607433472883,4.68923144374487e-06);
\draw[draw=white,fill=color0] (axis cs:0.471607433472883,1.17230683021128e-06) rectangle (axis cs:0.828869230505199,4.68923144374487e-06);
\draw[draw=white,fill=color0] (axis cs:0.828869230505199,1.17230683021128e-06) rectangle (axis cs:1.45677135794719,5.86153964825607e-06);
\draw[draw=white,fill=color0] (axis cs:1.45677135794719,1.17230683021128e-06) rectangle (axis cs:2.56033486493619,5.86153964825607e-06);
\draw[draw=white,fill=color0] (axis cs:2.56033486493619,1.17230683021128e-06) rectangle (axis cs:4.4998925774085,5.86153964825607e-06);
\draw[draw=white,fill=color0] (axis cs:4.4998925774085,1.17230683021128e-06) rectangle (axis cs:7.90874408091173,5.86153964825607e-06);
\draw[draw=white,fill=color0] (axis cs:7.90874408091173,1.17230683021128e-06) rectangle (axis cs:13.8999391344089,5.86153964825607e-06);
\draw[draw=white,fill=color0] (axis cs:13.8999391344089,1.17230683021128e-06) rectangle (axis cs:24.4297079237387,3.51692323923368e-06);
\draw[draw=white,fill=color0] (axis cs:24.4297079237387,1.17230683021128e-06) rectangle (axis cs:42.9362045019173,4.68923144374487e-06);
\draw[draw=white,fill=color0] (axis cs:42.9362045019173,1.17230683021128e-06) rectangle (axis cs:75.4621243440693,4.68923144374487e-06);
\draw[draw=white,fill=color0] (axis cs:75.4621243440693,1.17230683021128e-06) rectangle (axis cs:132.627750323517,3.51692323923368e-06);
\draw[draw=white,fill=color0] (axis cs:132.627750323517,1.17230683021128e-06) rectangle (axis cs:233.09866119956,1.17230683021128e-06);
\draw[draw=white,fill=color0] (axis cs:233.09866119956,1.17230683021128e-06) rectangle (axis cs:409.680370212783,1.17230683021128e-06);
\draw[draw=white,fill=color0] (axis cs:409.680370212783,1.17230683021128e-06) rectangle (axis cs:720.029900103088,1.17230683021128e-06);
\draw[draw=white,fill=color0] (axis cs:720.029900103088,1.17230683021128e-06) rectangle (axis cs:1265.48181152344,2.34461503472248e-06);
\end{axis}

\end{tikzpicture}
     \end{minipage}
  \end{minipage}
  \hfill
  \begin{minipage}{0.495\linewidth}
    \centering
    \begin{small}
      \textbf{Block-diagonal approximation}
    \end{small}
    \vspace{1ex}

    \begin{minipage}[t]{0.33\linewidth}
      \centering
\pgfkeys{/pgfplots/spectrumdefault/.style={
    width=1.38\linewidth,
    height=\goldenRatio*1.38\linewidth,
    every axis plot/.append style={line width = 1.2pt},
    every axis plot post/.append style={
      mark size=1, mark options={opacity=0.3}
    },
    tick pos = left,
    xmajorticks = true,
    ymajorticks = true,
    ylabel near ticks,
    xlabel near ticks,
    xtick align = inside,
    ytick align = inside,
    legend cell align = left,
    legend columns = 1,
    legend pos = north west,
    legend style = {
      fill opacity = 0.9,
      text opacity = 1,
      font = \small,
    },
    xticklabel style = {font = \small, inner xsep = 0ex},
    xlabel style = {font = \small},
    axis line style = {black},
    yticklabel style = {font = \small, inner ysep = 0ex},
    ylabel style = {font = \small, inner ysep = 0ex},
    title style = {font = \small, inner ysep = 0ex, yshift = -0.75ex},
    grid = major,
    grid style = {dashed}
  }
}
\pgfkeys{/pgfplots/spectrumdefaultleft/.style={
    spectrumdefault,
    title=\empty,
    ymax=3e-4,
    xlabel=\phantom{eigenvalues},
  }}
\pgfkeys{/pgfplots/spectrumdefaultcenter/.style={
    spectrumdefault,
    ylabel=\empty,
    yticklabels=\empty,
    title=\empty,
    ymax=3e-4,
  }}
\pgfkeys{/pgfplots/spectrumdefaultright/.style={
    spectrumdefault,
    ylabel=\empty,
    yticklabels=\empty,
    title=\empty,
    ymax=3e-4,
    xlabel=\phantom{eigenvalues},
  }}
 \pgfkeys{/pgfplots/zmystyle/.style={spectrumdefaultleft, ymax = 3e-3,
          title={exact, mb}}}
      \begin{tikzpicture}

\begin{axis}[
axis line style={white!80!black},
log basis x={10},
tick pos=left,
title={full\_batch\_exact, layerwise\_group, N=128, D=853018},
xlabel={eigenvalues},
xmin=0.0001, xmax=105.220306396484,
xmode=log,
ylabel={density},
ymin=1.17230683021128e-06, ymax=1,
ymode=log,
zmystyle
]
\draw[draw=white,fill=black] (axis cs:0.0001,1.17230683021128e-06) rectangle (axis cs:0.000161309002488124,0.981610001194098);
\draw[draw=white,fill=black] (axis cs:0.000161309002488124,1.17230683021128e-06) rectangle (axis cs:0.000260205942837136,0.00157792684189777);
\draw[draw=white,fill=black] (axis cs:0.000260205942837136,1.17230683021128e-06) rectangle (axis cs:0.000419735610805402,0.0015181391234677);
\draw[draw=white,fill=black] (axis cs:0.000419735610805402,1.17230683021128e-06) rectangle (axis cs:0.000677071326877628,0.00147828064451432);
\draw[draw=white,fill=black] (axis cs:0.000677071326877628,1.17230683021128e-06) rectangle (axis cs:0.00109217700351941,0.00149234834296845);
\draw[draw=white,fill=black] (axis cs:0.00109217700351941,1.17230683021128e-06) rectangle (axis cs:0.00176177982978184,0.0014864868019459);
\draw[draw=white,fill=black] (axis cs:0.00176177982978184,1.17230683021128e-06) rectangle (axis cs:0.00284190946945805,0.00153572374653537);
\draw[draw=white,fill=black] (axis cs:0.00284190946945805,1.17230683021128e-06) rectangle (axis cs:0.00458425581679832,0.00147828064451432);
\draw[draw=white,fill=black] (axis cs:0.00458425581679832,1.17230683021128e-06) rectangle (axis cs:0.00739481732958117,0.00134698212560906);
\draw[draw=white,fill=black] (axis cs:0.00739481732958117,1.17230683021128e-06) rectangle (axis cs:0.0119285060701663,0.00120161590824968);
\draw[draw=white,fill=black] (axis cs:0.0119285060701663,1.17230683021128e-06) rectangle (axis cs:0.0192417541535206,0.00104335430064066);
\draw[draw=white,fill=black] (axis cs:0.0192417541535206,1.17230683021128e-06) rectangle (axis cs:0.0310386816862612,0.000929640404803079);
\draw[draw=white,fill=black] (axis cs:0.0310386816862612,1.17230683021128e-06) rectangle (axis cs:0.050068187813572,0.000771378797194067);
\draw[draw=white,fill=black] (axis cs:0.050068187813572,1.17230683021128e-06) rectangle (axis cs:0.0807644943259534,0.000655320284947459);
\draw[draw=white,fill=black] (axis cs:0.0807644943259534,1.17230683021128e-06) rectangle (axis cs:0.130280400161773,0.000541606389109873);
\draw[draw=white,fill=black] (axis cs:0.130280400161773,1.17230683021128e-06) rectangle (axis cs:0.210154013938492,0.00041148017840913);
\draw[draw=white,fill=black] (axis cs:0.210154013938492,1.17230683021128e-06) rectangle (axis cs:0.338997343572935,0.000286043200526432);
\draw[draw=white,fill=black] (axis cs:0.338997343572935,1.17230683021128e-06) rectangle (axis cs:0.546833233378739,0.000219221632869294);
\draw[draw=white,fill=black] (axis cs:0.546833233378739,1.17230683021128e-06) rectangle (axis cs:0.882091234036799,0.000144193907780577);
\draw[draw=white,fill=black] (axis cs:0.882091234036799,1.17230683021128e-06) rectangle (axis cs:1.42289257065994,9.73015796001294e-05);
\draw[draw=white,fill=black] (axis cs:1.42289257065994,1.17230683021128e-06) rectangle (axis cs:2.29525381220918,7.03384908963719e-05);
\draw[draw=white,fill=black] (axis cs:2.29525381220918,1.17230683021128e-06) rectangle (axis cs:3.70245102904526,3.98584775790808e-05);
\draw[draw=white,fill=black] (axis cs:3.70245102904526,1.17230683021128e-06) rectangle (axis cs:5.97238682256419,3.16523201475024e-05);
\draw[draw=white,fill=black] (axis cs:5.97238682256419,1.17230683021128e-06) rectangle (axis cs:9.63399760821048,1.87569298978792e-05);
\draw[draw=white,fill=black] (axis cs:9.63399760821048,1.17230683021128e-06) rectangle (axis cs:15.540505441534,1.7584621693368e-05);
\draw[draw=white,fill=black] (axis cs:15.540505441534,1.17230683021128e-06) rectangle (axis cs:25.0682343093512,1.40676970798344e-05);
\draw[draw=white,fill=black] (axis cs:25.0682343093512,1.17230683021128e-06) rectangle (axis cs:40.4373187058001,5.86153964825607e-06);
\draw[draw=white,fill=black] (axis cs:40.4373187058001,1.17230683021128e-06) rectangle (axis cs:65.2290354372696,5.86153964825607e-06);
\draw[draw=white,fill=black] (axis cs:65.2290354372696,1.17230683021128e-06) rectangle (axis cs:105.220306396484,4.68923144374487e-06);
\end{axis}

\end{tikzpicture}
     \end{minipage}
    \hspace{1.9ex}
    \begin{minipage}[t]{0.33\linewidth}
      \centering
\pgfkeys{/pgfplots/spectrumdefault/.style={
    width=1.38\linewidth,
    height=\goldenRatio*1.38\linewidth,
    every axis plot/.append style={line width = 1.2pt},
    every axis plot post/.append style={
      mark size=1, mark options={opacity=0.3}
    },
    tick pos = left,
    xmajorticks = true,
    ymajorticks = true,
    ylabel near ticks,
    xlabel near ticks,
    xtick align = inside,
    ytick align = inside,
    legend cell align = left,
    legend columns = 1,
    legend pos = north west,
    legend style = {
      fill opacity = 0.9,
      text opacity = 1,
      font = \small,
    },
    xticklabel style = {font = \small, inner xsep = 0ex},
    xlabel style = {font = \small},
    axis line style = {black},
    yticklabel style = {font = \small, inner ysep = 0ex},
    ylabel style = {font = \small, inner ysep = 0ex},
    title style = {font = \small, inner ysep = 0ex, yshift = -0.75ex},
    grid = major,
    grid style = {dashed}
  }
}
\pgfkeys{/pgfplots/spectrumdefaultleft/.style={
    spectrumdefault,
    title=\empty,
    ymax=3e-4,
    xlabel=\phantom{eigenvalues},
  }}
\pgfkeys{/pgfplots/spectrumdefaultcenter/.style={
    spectrumdefault,
    ylabel=\empty,
    yticklabels=\empty,
    title=\empty,
    ymax=3e-4,
  }}
\pgfkeys{/pgfplots/spectrumdefaultright/.style={
    spectrumdefault,
    ylabel=\empty,
    yticklabels=\empty,
    title=\empty,
    ymax=3e-4,
    xlabel=\phantom{eigenvalues},
  }}
 \pgfkeys{/pgfplots/zmystyle/.style={spectrumdefaultcenter, ymax = 3e-3,
          title={exact, sub}}}
      \begin{tikzpicture}

\definecolor{color0}{rgb}{0.274509803921569,0.6,0.564705882352941}

\begin{axis}[
axis line style={white!80!black},
log basis x={10},
tick pos=left,
title={frac\_batch\_exact, layerwise\_group, N=128, D=853018},
xlabel={eigenvalues},
xmin=0.0001, xmax=105.220306396484,
xmode=log,
ylabel={density},
ymin=1.17230683021128e-06, ymax=1,
ymode=log,
zmystyle
]
\draw[draw=white,fill=color0] (axis cs:0.0001,1.17230683021128e-06) rectangle (axis cs:0.000161309002488124,0.99516774557927);
\draw[draw=white,fill=color0] (axis cs:0.000161309002488124,1.17230683021128e-06) rectangle (axis cs:0.000260205942837136,0.000339969377933947);
\draw[draw=white,fill=color0] (axis cs:0.000260205942837136,1.17230683021128e-06) rectangle (axis cs:0.000419735610805402,0.00031300628923019);
\draw[draw=white,fill=color0] (axis cs:0.000419735610805402,1.17230683021128e-06) rectangle (axis cs:0.000677071326877628,0.000308317056412145);
\draw[draw=white,fill=color0] (axis cs:0.000677071326877628,1.17230683021128e-06) rectangle (axis cs:0.00109217700351941,0.000334107836911391);
\draw[draw=white,fill=color0] (axis cs:0.00109217700351941,1.17230683021128e-06) rectangle (axis cs:0.00176177982978184,0.000316523213843723);
\draw[draw=white,fill=color0] (axis cs:0.00176177982978184,1.17230683021128e-06) rectangle (axis cs:0.00284190946945805,0.000295421666162522);
\draw[draw=white,fill=color0] (axis cs:0.00284190946945805,1.17230683021128e-06) rectangle (axis cs:0.00458425581679832,0.000256735495413652);
\draw[draw=white,fill=color0] (axis cs:0.00458425581679832,1.17230683021128e-06) rectangle (axis cs:0.00739481732958117,0.000216877016460271);
\draw[draw=white,fill=color0] (axis cs:0.00739481732958117,1.17230683021128e-06) rectangle (axis cs:0.0119285060701663,0.00026494165284523);
\draw[draw=white,fill=color0] (axis cs:0.0119285060701663,1.17230683021128e-06) rectangle (axis cs:0.0192417541535206,0.000253218570800119);
\draw[draw=white,fill=color0] (axis cs:0.0192417541535206,1.17230683021128e-06) rectangle (axis cs:0.0310386816862612,0.00026494165284523);
\draw[draw=white,fill=color0] (axis cs:0.0310386816862612,1.17230683021128e-06) rectangle (axis cs:0.050068187813572,0.000236806255936962);
\draw[draw=white,fill=color0] (axis cs:0.050068187813572,1.17230683021128e-06) rectangle (axis cs:0.0807644943259534,0.000240323180550495);
\draw[draw=white,fill=color0] (axis cs:0.0807644943259534,1.17230683021128e-06) rectangle (axis cs:0.130280400161773,0.000216877016460271);
\draw[draw=white,fill=color0] (axis cs:0.130280400161773,1.17230683021128e-06) rectangle (axis cs:0.210154013938492,0.00019577546877907);
\draw[draw=white,fill=color0] (axis cs:0.210154013938492,1.17230683021128e-06) rectangle (axis cs:0.338997343572935,0.000169984688279824);
\draw[draw=white,fill=color0] (axis cs:0.338997343572935,1.17230683021128e-06) rectangle (axis cs:0.546833233378739,0.000141849291371555);
\draw[draw=white,fill=color0] (axis cs:0.546833233378739,1.17230683021128e-06) rectangle (axis cs:0.882091234036799,0.000134815442144488);
\draw[draw=white,fill=color0] (axis cs:0.882091234036799,1.17230683021128e-06) rectangle (axis cs:1.42289257065994,9.84738878046406e-05);
\draw[draw=white,fill=color0] (axis cs:1.42289257065994,1.17230683021128e-06) rectangle (axis cs:2.29525381220918,7.03384908963719e-05);
\draw[draw=white,fill=color0] (axis cs:2.29525381220918,1.17230683021128e-06) rectangle (axis cs:3.70245102904526,4.6892326806148e-05);
\draw[draw=white,fill=color0] (axis cs:3.70245102904526,1.17230683021128e-06) rectangle (axis cs:5.97238682256419,4.92369432151703e-05);
\draw[draw=white,fill=color0] (axis cs:5.97238682256419,1.17230683021128e-06) rectangle (axis cs:9.63399760821048,3.75138611700584e-05);
\draw[draw=white,fill=color0] (axis cs:9.63399760821048,1.17230683021128e-06) rectangle (axis cs:15.540505441534,1.87569298978792e-05);
\draw[draw=white,fill=color0] (axis cs:15.540505441534,1.17230683021128e-06) rectangle (axis cs:25.0682343093512,2.22738545114128e-05);
\draw[draw=white,fill=color0] (axis cs:25.0682343093512,1.17230683021128e-06) rectangle (axis cs:40.4373187058001,9.37846426178966e-06);
\draw[draw=white,fill=color0] (axis cs:40.4373187058001,1.17230683021128e-06) rectangle (axis cs:65.2290354372696,8.20615605727846e-06);
\draw[draw=white,fill=color0] (axis cs:65.2290354372696,1.17230683021128e-06) rectangle (axis cs:105.220306396484,4.68923144374487e-06);
\end{axis}

\end{tikzpicture}
     \end{minipage}
    \hspace{-3.5ex}
    \begin{minipage}[t]{0.33\linewidth}
      \centering
\pgfkeys{/pgfplots/spectrumdefault/.style={
    width=1.38\linewidth,
    height=\goldenRatio*1.38\linewidth,
    every axis plot/.append style={line width = 1.2pt},
    every axis plot post/.append style={
      mark size=1, mark options={opacity=0.3}
    },
    tick pos = left,
    xmajorticks = true,
    ymajorticks = true,
    ylabel near ticks,
    xlabel near ticks,
    xtick align = inside,
    ytick align = inside,
    legend cell align = left,
    legend columns = 1,
    legend pos = north west,
    legend style = {
      fill opacity = 0.9,
      text opacity = 1,
      font = \small,
    },
    xticklabel style = {font = \small, inner xsep = 0ex},
    xlabel style = {font = \small},
    axis line style = {black},
    yticklabel style = {font = \small, inner ysep = 0ex},
    ylabel style = {font = \small, inner ysep = 0ex},
    title style = {font = \small, inner ysep = 0ex, yshift = -0.75ex},
    grid = major,
    grid style = {dashed}
  }
}
\pgfkeys{/pgfplots/spectrumdefaultleft/.style={
    spectrumdefault,
    title=\empty,
    ymax=3e-4,
    xlabel=\phantom{eigenvalues},
  }}
\pgfkeys{/pgfplots/spectrumdefaultcenter/.style={
    spectrumdefault,
    ylabel=\empty,
    yticklabels=\empty,
    title=\empty,
    ymax=3e-4,
  }}
\pgfkeys{/pgfplots/spectrumdefaultright/.style={
    spectrumdefault,
    ylabel=\empty,
    yticklabels=\empty,
    title=\empty,
    ymax=3e-4,
    xlabel=\phantom{eigenvalues},
  }}
 \pgfkeys{/pgfplots/zmystyle/.style={spectrumdefaultright, ymax = 3e-3,
          title={mc, full}}}
      \begin{tikzpicture}

\definecolor{color0}{rgb}{0.870588235294118,0.623529411764706,0.0862745098039216}

\begin{axis}[
axis line style={white!80!black},
log basis x={10},
tick pos=left,
title={full\_batch\_mc, layerwise\_group, N=128, D=853018},
xlabel={eigenvalues},
xmin=0.0001, xmax=105.220306396484,
xmode=log,
ylabel={density},
ymin=1.17230683021128e-06, ymax=1,
ymode=log,
zmystyle
]
\draw[draw=white,fill=color0] (axis cs:0.0001,1.17230683021128e-06) rectangle (axis cs:0.000161309002488124,0.995678871956437);
\draw[draw=white,fill=color0] (axis cs:0.000161309002488124,1.17230683021128e-06) rectangle (axis cs:0.000260205942837136,0.00027432011848132);
\draw[draw=white,fill=color0] (axis cs:0.000260205942837136,1.17230683021128e-06) rectangle (axis cs:0.000419735610805402,0.000255563187209141);
\draw[draw=white,fill=color0] (axis cs:0.000419735610805402,1.17230683021128e-06) rectangle (axis cs:0.000677071326877628,0.000263769344640719);
\draw[draw=white,fill=color0] (axis cs:0.000677071326877628,1.17230683021128e-06) rectangle (axis cs:0.00109217700351941,0.000246184721573051);
\draw[draw=white,fill=color0] (axis cs:0.00109217700351941,1.17230683021128e-06) rectangle (axis cs:0.00176177982978184,0.000252046262595607);
\draw[draw=white,fill=color0] (axis cs:0.00176177982978184,1.17230683021128e-06) rectangle (axis cs:0.00284190946945805,0.000237978564141473);
\draw[draw=white,fill=color0] (axis cs:0.00284190946945805,1.17230683021128e-06) rectangle (axis cs:0.00458425581679832,0.000262597036436208);
\draw[draw=white,fill=color0] (axis cs:0.00458425581679832,1.17230683021128e-06) rectangle (axis cs:0.00739481732958117,0.000288387816935454);
\draw[draw=white,fill=color0] (axis cs:0.00739481732958117,1.17230683021128e-06) rectangle (axis cs:0.0119285060701663,0.00026494165284523);
\draw[draw=white,fill=color0] (axis cs:0.0119285060701663,1.17230683021128e-06) rectangle (axis cs:0.0192417541535206,0.000253218570800119);
\draw[draw=white,fill=color0] (axis cs:0.0192417541535206,1.17230683021128e-06) rectangle (axis cs:0.0310386816862612,0.000250873954391096);
\draw[draw=white,fill=color0] (axis cs:0.0310386816862612,1.17230683021128e-06) rectangle (axis cs:0.050068187813572,0.000242667796959518);
\draw[draw=white,fill=color0] (axis cs:0.050068187813572,1.17230683021128e-06) rectangle (axis cs:0.0807644943259534,0.000211015475437715);
\draw[draw=white,fill=color0] (axis cs:0.0807644943259534,1.17230683021128e-06) rectangle (axis cs:0.130280400161773,0.000194603160574559);
\draw[draw=white,fill=color0] (axis cs:0.130280400161773,1.17230683021128e-06) rectangle (axis cs:0.210154013938492,0.00018639700314298);
\draw[draw=white,fill=color0] (axis cs:0.210154013938492,1.17230683021128e-06) rectangle (axis cs:0.338997343572935,0.000143021599576066);
\draw[draw=white,fill=color0] (axis cs:0.338997343572935,1.17230683021128e-06) rectangle (axis cs:0.546833233378739,0.000143021599576066);
\draw[draw=white,fill=color0] (axis cs:0.546833233378739,1.17230683021128e-06) rectangle (axis cs:0.882091234036799,0.00010785235344073);
\draw[draw=white,fill=color0] (axis cs:0.882091234036799,1.17230683021128e-06) rectangle (axis cs:1.42289257065994,8.44061893505062e-05);
\draw[draw=white,fill=color0] (axis cs:1.42289257065994,1.17230683021128e-06) rectangle (axis cs:2.29525381220918,6.33046416693047e-05);
\draw[draw=white,fill=color0] (axis cs:2.29525381220918,1.17230683021128e-06) rectangle (axis cs:3.70245102904526,3.5169244761036e-05);
\draw[draw=white,fill=color0] (axis cs:3.70245102904526,1.17230683021128e-06) rectangle (axis cs:5.97238682256419,2.57907791249464e-05);
\draw[draw=white,fill=color0] (axis cs:5.97238682256419,1.17230683021128e-06) rectangle (axis cs:9.63399760821048,2.22738545114128e-05);
\draw[draw=white,fill=color0] (axis cs:9.63399760821048,1.17230683021128e-06) rectangle (axis cs:15.540505441534,1.64123134888568e-05);
\draw[draw=white,fill=color0] (axis cs:15.540505441534,1.17230683021128e-06) rectangle (axis cs:25.0682343093512,1.28953888753233e-05);
\draw[draw=white,fill=color0] (axis cs:25.0682343093512,1.17230683021128e-06) rectangle (axis cs:40.4373187058001,7.03384785276727e-06);
\draw[draw=white,fill=color0] (axis cs:40.4373187058001,1.17230683021128e-06) rectangle (axis cs:65.2290354372696,4.68923144374487e-06);
\draw[draw=white,fill=color0] (axis cs:65.2290354372696,1.17230683021128e-06) rectangle (axis cs:105.220306396484,4.68923144374487e-06);
\end{axis}

\end{tikzpicture}
     \end{minipage}
  \end{minipage}

  \begin{small}
    \textbf{\cifarhun \allcnnc}
  \end{small}

  \begin{minipage}{0.495\linewidth}
    \centering
    \begin{small}
      \textbf{Full network}
    \end{small}
    \vspace{1ex}

    \begin{minipage}[t]{0.33\linewidth}
      \centering
\pgfkeys{/pgfplots/spectrumdefault/.style={
    width=1.38\linewidth,
    height=\goldenRatio*1.38\linewidth,
    every axis plot/.append style={line width = 1.2pt},
    every axis plot post/.append style={
      mark size=1, mark options={opacity=0.3}
    },
    tick pos = left,
    xmajorticks = true,
    ymajorticks = true,
    ylabel near ticks,
    xlabel near ticks,
    xtick align = inside,
    ytick align = inside,
    legend cell align = left,
    legend columns = 1,
    legend pos = north west,
    legend style = {
      fill opacity = 0.9,
      text opacity = 1,
      font = \small,
    },
    xticklabel style = {font = \small, inner xsep = 0ex},
    xlabel style = {font = \small},
    axis line style = {black},
    yticklabel style = {font = \small, inner ysep = 0ex},
    ylabel style = {font = \small, inner ysep = 0ex},
    title style = {font = \small, inner ysep = 0ex, yshift = -0.75ex},
    grid = major,
    grid style = {dashed}
  }
}
\pgfkeys{/pgfplots/spectrumdefaultleft/.style={
    spectrumdefault,
    title=\empty,
    ymax=3e-4,
    xlabel=\phantom{eigenvalues},
  }}
\pgfkeys{/pgfplots/spectrumdefaultcenter/.style={
    spectrumdefault,
    ylabel=\empty,
    yticklabels=\empty,
    title=\empty,
    ymax=3e-4,
  }}
\pgfkeys{/pgfplots/spectrumdefaultright/.style={
    spectrumdefault,
    ylabel=\empty,
    yticklabels=\empty,
    title=\empty,
    ymax=3e-4,
    xlabel=\phantom{eigenvalues},
  }}
 \pgfkeys{/pgfplots/zmystyle/.style={spectrumdefaultleft, ymax = 1e-3,
          title={exact, mb}}}
      \begin{tikzpicture}

\begin{axis}[
axis line style={white!80!black},
log basis x={10},
tick pos=left,
title={full\_batch\_exact, one\_group, N=64, D=1387108},
xlabel={eigenvalues},
xmin=0.0001, xmax=18.325138092041,
xmode=log,
ylabel={density},
ymin=7.20923878368607e-07, ymax=1,
ymode=log,
zmystyle
]
\draw[draw=white,fill=black] (axis cs:0.0001,7.20923878368607e-07) rectangle (axis cs:0.000151874334581277,0.997220115520873);
\draw[draw=white,fill=black] (axis cs:0.000151874334581277,7.20923878368607e-07) rectangle (axis cs:0.000230658135045056,0.00055727455921328);
\draw[draw=white,fill=black] (axis cs:0.000230658135045056,7.20923878368607e-07) rectangle (axis cs:0.000350310507757262,0.000496716909772928);
\draw[draw=white,fill=black] (axis cs:0.000350310507757262,7.20923878368607e-07) rectangle (axis cs:0.000532031752624634,0.000445531277507568);
\draw[draw=white,fill=black] (axis cs:0.000532031752624633,7.20923878368607e-07) rectangle (axis cs:0.000808019684059767,0.000394345645242207);
\draw[draw=white,fill=black] (axis cs:0.000808019684059767,7.20923878368607e-07) rectangle (axis cs:0.00122717451845151,0.000309997490664401);
\draw[draw=white,fill=black] (axis cs:0.00122717451845151,7.20923878368607e-07) rectangle (axis cs:0.00186376313404921,0.00020762622613368);
\draw[draw=white,fill=black] (axis cs:0.00186376313404921,7.20923878368607e-07) rectangle (axis cs:0.00283057785800839,0.000127603617944279);
\draw[draw=white,fill=black] (axis cs:0.00283057785800839,7.20923878368607e-07) rectangle (axis cs:0.0042989212866552,7.06505904943773e-05);
\draw[draw=white,fill=black] (axis cs:0.0042989212866552,7.20923878368607e-07) rectangle (axis cs:0.00652895809828045,4.90228585512361e-05);
\draw[draw=white,fill=black] (axis cs:0.00652895809828045,7.20923878368607e-07) rectangle (axis cs:0.00991581166685381,3.74880681814867e-05);
\draw[draw=white,fill=black] (axis cs:0.00991581166685381,7.20923878368607e-07) rectangle (axis cs:0.0150595729873668,2.52323534140027e-05);
\draw[draw=white,fill=black] (axis cs:0.0150595729873668,7.20923878368607e-07) rectangle (axis cs:0.0228716262653451,1.08138654517605e-05);
\draw[draw=white,fill=black] (axis cs:0.0228716262653451,7.20923878368607e-07) rectangle (axis cs:0.0347361301984094,5.04647026699686e-06);
\draw[draw=white,fill=black] (axis cs:0.0347361301984094,7.20923878368607e-07) rectangle (axis cs:0.0527552665981202,2.88369707268273e-06);
\draw[draw=white,fill=black] (axis cs:0.0527552665981202,7.20923878368607e-07) rectangle (axis cs:0.0801217101024737,3.60462147086146e-06);
\draw[draw=white,fill=black] (axis cs:0.0801217101024737,7.20923878368607e-07) rectangle (axis cs:0.121684314073272,5.04647026677482e-06);
\draw[draw=white,fill=black] (axis cs:0.121684314073272,7.20923878368607e-07) rectangle (axis cs:0.184807242288572,7.93016785948971e-06);
\draw[draw=white,fill=black] (axis cs:0.184807242288572,7.20923878368607e-07) rectangle (axis cs:0.280674769483777,9.37201665562512e-06);
\draw[draw=white,fill=black] (axis cs:0.280674769483777,7.20923878368607e-07) rectangle (axis cs:0.426272938491019,6.48831906313226e-06);
\draw[draw=white,fill=black] (axis cs:0.426272938491019,7.20923878368607e-07) rectangle (axis cs:0.64739918883329,7.20924346131099e-06);
\draw[draw=white,fill=black] (axis cs:0.64739918883329,7.20923878368607e-07) rectangle (axis cs:0.983233210125142,7.20924346131099e-06);
\draw[draw=white,fill=black] (axis cs:0.983233210125142,7.20923878368607e-07) rectangle (axis cs:1.49327889525969,5.76739466517559e-06);
\draw[draw=white,fill=black] (axis cs:1.49327889525969,7.20923878368607e-07) rectangle (axis cs:2.26790738561829,4.32554586881814e-06);
\draw[draw=white,fill=black] (axis cs:2.26790738561829,7.20923878368607e-07) rectangle (axis cs:3.44436925082741,7.20923878368607e-07);
\draw[draw=white,fill=black] (axis cs:3.44436925082741,7.20923878368607e-07) rectangle (axis cs:5.23111288021623,7.20923878368607e-07);
\draw[draw=white,fill=black] (axis cs:5.23111288021623,7.20923878368607e-07) rectangle (axis cs:7.94471787802387,7.20923878368607e-07);
\draw[draw=white,fill=black] (axis cs:7.94471787802387,7.20923878368607e-07) rectangle (axis cs:12.0659874116085,7.20923878368607e-07);
\draw[draw=white,fill=black] (axis cs:12.0659874116085,7.20923878368607e-07) rectangle (axis cs:18.325138092041,7.20923878368607e-07);
\end{axis}

\end{tikzpicture}
     \end{minipage}
    \hspace{1.9ex}
    \begin{minipage}[t]{0.33\linewidth}
      \centering
\pgfkeys{/pgfplots/spectrumdefault/.style={
    width=1.38\linewidth,
    height=\goldenRatio*1.38\linewidth,
    every axis plot/.append style={line width = 1.2pt},
    every axis plot post/.append style={
      mark size=1, mark options={opacity=0.3}
    },
    tick pos = left,
    xmajorticks = true,
    ymajorticks = true,
    ylabel near ticks,
    xlabel near ticks,
    xtick align = inside,
    ytick align = inside,
    legend cell align = left,
    legend columns = 1,
    legend pos = north west,
    legend style = {
      fill opacity = 0.9,
      text opacity = 1,
      font = \small,
    },
    xticklabel style = {font = \small, inner xsep = 0ex},
    xlabel style = {font = \small},
    axis line style = {black},
    yticklabel style = {font = \small, inner ysep = 0ex},
    ylabel style = {font = \small, inner ysep = 0ex},
    title style = {font = \small, inner ysep = 0ex, yshift = -0.75ex},
    grid = major,
    grid style = {dashed}
  }
}
\pgfkeys{/pgfplots/spectrumdefaultleft/.style={
    spectrumdefault,
    title=\empty,
    ymax=3e-4,
    xlabel=\phantom{eigenvalues},
  }}
\pgfkeys{/pgfplots/spectrumdefaultcenter/.style={
    spectrumdefault,
    ylabel=\empty,
    yticklabels=\empty,
    title=\empty,
    ymax=3e-4,
  }}
\pgfkeys{/pgfplots/spectrumdefaultright/.style={
    spectrumdefault,
    ylabel=\empty,
    yticklabels=\empty,
    title=\empty,
    ymax=3e-4,
    xlabel=\phantom{eigenvalues},
  }}
 \pgfkeys{/pgfplots/zmystyle/.style={spectrumdefaultcenter, ymax = 1e-3,
          title={exact, sub}}}
      \begin{tikzpicture}

\definecolor{color0}{rgb}{0.274509803921569,0.6,0.564705882352941}

\begin{axis}[
axis line style={white!80!black},
log basis x={10},
tick pos=left,
title={frac\_batch\_exact, one\_group, N=64, D=1387108},
xlabel={eigenvalues},
xmin=0.0001, xmax=18.325138092041,
xmode=log,
ylabel={density},
ymin=7.20923878368607e-07, ymax=1,
ymode=log,
zmystyle
]
\draw[draw=white,fill=color0] (axis cs:0.0001,7.20923878368607e-07) rectangle (axis cs:0.000151874334581277,0.999491027374895);
\draw[draw=white,fill=color0] (axis cs:0.000151874334581277,7.20923878368607e-07) rectangle (axis cs:0.000230658135045056,1.441848744221e-05);
\draw[draw=white,fill=color0] (axis cs:0.000230658135045056,7.20923878368607e-07) rectangle (axis cs:0.000350310507757262,1.44184874424321e-05);
\draw[draw=white,fill=color0] (axis cs:0.000350310507757262,7.20923878368607e-07) rectangle (axis cs:0.000532031752624634,1.51394118403888e-05);
\draw[draw=white,fill=color0] (axis cs:0.000532031752624633,7.20923878368607e-07) rectangle (axis cs:0.000808019684059767,2.01858826271958e-05);
\draw[draw=white,fill=color0] (axis cs:0.000808019684059767,7.20923878368607e-07) rectangle (axis cs:0.00122717451845151,4.54182365605645e-05);
\draw[draw=white,fill=color0] (axis cs:0.00122717451845151,7.20923878368607e-07) rectangle (axis cs:0.00186376313404921,6.27204221152996e-05);
\draw[draw=white,fill=color0] (axis cs:0.00186376313404921,7.20923878368607e-07) rectangle (axis cs:0.00283057785800839,5.9115800124628e-05);
\draw[draw=white,fill=color0] (axis cs:0.00283057785800839,7.20923878368607e-07) rectangle (axis cs:0.0042989212866552,5.33484049398643e-05);
\draw[draw=white,fill=color0] (axis cs:0.0042989212866552,7.20923878368607e-07) rectangle (axis cs:0.00652895809828045,4.90228585512361e-05);
\draw[draw=white,fill=color0] (axis cs:0.00652895809828045,7.20923878368607e-07) rectangle (axis cs:0.00991581166685381,4.54182365605645e-05);
\draw[draw=white,fill=color0] (axis cs:0.00991581166685381,7.20923878368607e-07) rectangle (axis cs:0.0150595729873668,3.74880681817088e-05);
\draw[draw=white,fill=color0] (axis cs:0.0150595729873668,7.20923878368607e-07) rectangle (axis cs:0.0228716262653451,2.66742022101381e-05);
\draw[draw=white,fill=color0] (axis cs:0.0228716262653451,7.20923878368607e-07) rectangle (axis cs:0.0347361301984094,1.80231094326596e-05);
\draw[draw=white,fill=color0] (axis cs:0.0347361301984094,7.20923878368607e-07) rectangle (axis cs:0.0527552665981202,7.93016785948971e-06);
\draw[draw=white,fill=color0] (axis cs:0.0527552665981202,7.20923878368607e-07) rectangle (axis cs:0.0801217101024737,3.60462147086146e-06);
\draw[draw=white,fill=color0] (axis cs:0.0801217101024737,7.20923878368607e-07) rectangle (axis cs:0.121684314073272,5.04647026677482e-06);
\draw[draw=white,fill=color0] (axis cs:0.121684314073272,7.20923878368607e-07) rectangle (axis cs:0.184807242288572,7.93016785948971e-06);
\draw[draw=white,fill=color0] (axis cs:0.184807242288572,7.20923878368607e-07) rectangle (axis cs:0.280674769483777,9.37201665562512e-06);
\draw[draw=white,fill=color0] (axis cs:0.280674769483777,7.20923878368607e-07) rectangle (axis cs:0.426272938491019,6.48831906313226e-06);
\draw[draw=white,fill=color0] (axis cs:0.426272938491019,7.20923878368607e-07) rectangle (axis cs:0.64739918883329,7.20924346131099e-06);
\draw[draw=white,fill=color0] (axis cs:0.64739918883329,7.20923878368607e-07) rectangle (axis cs:0.983233210125142,7.20924346131099e-06);
\draw[draw=white,fill=color0] (axis cs:0.983233210125142,7.20923878368607e-07) rectangle (axis cs:1.49327889525969,5.76739466517559e-06);
\draw[draw=white,fill=color0] (axis cs:1.49327889525969,7.20923878368607e-07) rectangle (axis cs:2.26790738561829,4.32554586881814e-06);
\draw[draw=white,fill=color0] (axis cs:2.26790738561829,7.20923878368607e-07) rectangle (axis cs:3.44436925082741,7.20923878368607e-07);
\draw[draw=white,fill=color0] (axis cs:3.44436925082741,7.20923878368607e-07) rectangle (axis cs:5.23111288021623,7.20923878368607e-07);
\draw[draw=white,fill=color0] (axis cs:5.23111288021623,7.20923878368607e-07) rectangle (axis cs:7.94471787802387,7.20923878368607e-07);
\draw[draw=white,fill=color0] (axis cs:7.94471787802387,7.20923878368607e-07) rectangle (axis cs:12.0659874116085,7.20923878368607e-07);
\draw[draw=white,fill=color0] (axis cs:12.0659874116085,7.20923878368607e-07) rectangle (axis cs:18.325138092041,7.20923878368607e-07);
\end{axis}

\end{tikzpicture}
     \end{minipage}
    \hspace{-3.5ex}
    \begin{minipage}[t]{0.33\linewidth}
      \centering
\pgfkeys{/pgfplots/spectrumdefault/.style={
    width=1.38\linewidth,
    height=\goldenRatio*1.38\linewidth,
    every axis plot/.append style={line width = 1.2pt},
    every axis plot post/.append style={
      mark size=1, mark options={opacity=0.3}
    },
    tick pos = left,
    xmajorticks = true,
    ymajorticks = true,
    ylabel near ticks,
    xlabel near ticks,
    xtick align = inside,
    ytick align = inside,
    legend cell align = left,
    legend columns = 1,
    legend pos = north west,
    legend style = {
      fill opacity = 0.9,
      text opacity = 1,
      font = \small,
    },
    xticklabel style = {font = \small, inner xsep = 0ex},
    xlabel style = {font = \small},
    axis line style = {black},
    yticklabel style = {font = \small, inner ysep = 0ex},
    ylabel style = {font = \small, inner ysep = 0ex},
    title style = {font = \small, inner ysep = 0ex, yshift = -0.75ex},
    grid = major,
    grid style = {dashed}
  }
}
\pgfkeys{/pgfplots/spectrumdefaultleft/.style={
    spectrumdefault,
    title=\empty,
    ymax=3e-4,
    xlabel=\phantom{eigenvalues},
  }}
\pgfkeys{/pgfplots/spectrumdefaultcenter/.style={
    spectrumdefault,
    ylabel=\empty,
    yticklabels=\empty,
    title=\empty,
    ymax=3e-4,
  }}
\pgfkeys{/pgfplots/spectrumdefaultright/.style={
    spectrumdefault,
    ylabel=\empty,
    yticklabels=\empty,
    title=\empty,
    ymax=3e-4,
    xlabel=\phantom{eigenvalues},
  }}
 \pgfkeys{/pgfplots/zmystyle/.style={spectrumdefaultright, ymax = 1e-3,
          title={mc, full}}}
      \begin{tikzpicture}

\definecolor{color0}{rgb}{0.870588235294118,0.623529411764706,0.0862745098039216}

\begin{axis}[
axis line style={white!80!black},
log basis x={10},
tick pos=left,
title={full\_batch\_mc, one\_group, N=64, D=1387108},
xlabel={eigenvalues},
xmin=0.0001, xmax=18.325138092041,
xmode=log,
ylabel={density},
ymin=7.20923878368607e-07, ymax=1,
ymode=log,
zmystyle
]
\draw[draw=white,fill=color0] (axis cs:0.0001,7.20923878368607e-07) rectangle (axis cs:0.000151874334581277,0.999954581762875);
\draw[draw=white,fill=color0] (axis cs:0.000151874334581277,7.20923878368607e-07) rectangle (axis cs:0.000230658135045056,7.20923878368607e-07);
\draw[draw=white,fill=color0] (axis cs:0.000230658135045056,7.20923878368607e-07) rectangle (axis cs:0.000350310507757262,7.20923878368607e-07);
\draw[draw=white,fill=color0] (axis cs:0.000350310507757262,7.20923878368607e-07) rectangle (axis cs:0.000532031752624634,7.20923878368607e-07);
\draw[draw=white,fill=color0] (axis cs:0.000532031752624633,7.20923878368607e-07) rectangle (axis cs:0.000808019684059767,7.20923878368607e-07);
\draw[draw=white,fill=color0] (axis cs:0.000808019684059767,7.20923878368607e-07) rectangle (axis cs:0.00122717451845151,7.20923878368607e-07);
\draw[draw=white,fill=color0] (axis cs:0.00122717451845151,7.20923878368607e-07) rectangle (axis cs:0.00186376313404921,7.20923878368607e-07);
\draw[draw=white,fill=color0] (axis cs:0.00186376313404921,7.20923878368607e-07) rectangle (axis cs:0.00283057785800839,7.20923878368607e-07);
\draw[draw=white,fill=color0] (axis cs:0.00283057785800839,7.20923878368607e-07) rectangle (axis cs:0.0042989212866552,1.44184827632529e-06);
\draw[draw=white,fill=color0] (axis cs:0.0042989212866552,7.20923878368607e-07) rectangle (axis cs:0.00652895809828045,1.44184827654733e-06);
\draw[draw=white,fill=color0] (axis cs:0.00652895809828045,7.20923878368607e-07) rectangle (axis cs:0.00991581166685381,2.16277267450401e-06);
\draw[draw=white,fill=color0] (axis cs:0.00991581166685381,7.20923878368607e-07) rectangle (axis cs:0.0150595729873668,7.20923878368607e-07);
\draw[draw=white,fill=color0] (axis cs:0.0150595729873668,7.20923878368607e-07) rectangle (axis cs:0.0228716262653451,2.16277267472605e-06);
\draw[draw=white,fill=color0] (axis cs:0.0228716262653451,7.20923878368607e-07) rectangle (axis cs:0.0347361301984094,1.44184827632529e-06);
\draw[draw=white,fill=color0] (axis cs:0.0347361301984094,7.20923878368607e-07) rectangle (axis cs:0.0527552665981202,2.88369707268273e-06);
\draw[draw=white,fill=color0] (axis cs:0.0527552665981202,7.20923878368607e-07) rectangle (axis cs:0.0801217101024737,3.60462147086146e-06);
\draw[draw=white,fill=color0] (axis cs:0.0801217101024737,7.20923878368607e-07) rectangle (axis cs:0.121684314073272,6.48831906313226e-06);
\draw[draw=white,fill=color0] (axis cs:0.121684314073272,7.20923878368607e-07) rectangle (axis cs:0.184807242288572,3.60462147086146e-06);
\draw[draw=white,fill=color0] (axis cs:0.184807242288572,7.20923878368607e-07) rectangle (axis cs:0.280674769483777,2.88369707268273e-06);
\draw[draw=white,fill=color0] (axis cs:0.280674769483777,7.20923878368607e-07) rectangle (axis cs:0.426272938491019,5.04647026699686e-06);
\draw[draw=white,fill=color0] (axis cs:0.426272938491019,7.20923878368607e-07) rectangle (axis cs:0.64739918883329,5.04647026699686e-06);
\draw[draw=white,fill=color0] (axis cs:0.64739918883329,7.20923878368607e-07) rectangle (axis cs:0.983233210125142,5.76739466495354e-06);
\draw[draw=white,fill=color0] (axis cs:0.983233210125142,7.20923878368607e-07) rectangle (axis cs:1.49327889525969,5.04647026699686e-06);
\draw[draw=white,fill=color0] (axis cs:1.49327889525969,7.20923878368607e-07) rectangle (axis cs:2.26790738561829,4.32554586904018e-06);
\draw[draw=white,fill=color0] (axis cs:2.26790738561829,7.20923878368607e-07) rectangle (axis cs:3.44436925082741,2.88369707268273e-06);
\draw[draw=white,fill=color0] (axis cs:3.44436925082741,7.20923878368607e-07) rectangle (axis cs:5.23111288021623,2.16277267450401e-06);
\draw[draw=white,fill=color0] (axis cs:5.23111288021623,7.20923878368607e-07) rectangle (axis cs:7.94471787802387,7.20923878368607e-07);
\draw[draw=white,fill=color0] (axis cs:7.94471787802387,7.20923878368607e-07) rectangle (axis cs:12.0659874116085,7.20923878368607e-07);
\draw[draw=white,fill=color0] (axis cs:12.0659874116085,7.20923878368607e-07) rectangle (axis cs:18.325138092041,7.20923878368607e-07);
\end{axis}

\end{tikzpicture}
     \end{minipage}
  \end{minipage}
  \hfill
  \begin{minipage}{0.495\linewidth}
    \centering
    \begin{small}
      \textbf{Block-diagonal approximation}
    \end{small}
    \vspace{1ex}

    \begin{minipage}[t]{0.33\linewidth}
      \centering
\pgfkeys{/pgfplots/spectrumdefault/.style={
    width=1.38\linewidth,
    height=\goldenRatio*1.38\linewidth,
    every axis plot/.append style={line width = 1.2pt},
    every axis plot post/.append style={
      mark size=1, mark options={opacity=0.3}
    },
    tick pos = left,
    xmajorticks = true,
    ymajorticks = true,
    ylabel near ticks,
    xlabel near ticks,
    xtick align = inside,
    ytick align = inside,
    legend cell align = left,
    legend columns = 1,
    legend pos = north west,
    legend style = {
      fill opacity = 0.9,
      text opacity = 1,
      font = \small,
    },
    xticklabel style = {font = \small, inner xsep = 0ex},
    xlabel style = {font = \small},
    axis line style = {black},
    yticklabel style = {font = \small, inner ysep = 0ex},
    ylabel style = {font = \small, inner ysep = 0ex},
    title style = {font = \small, inner ysep = 0ex, yshift = -0.75ex},
    grid = major,
    grid style = {dashed}
  }
}
\pgfkeys{/pgfplots/spectrumdefaultleft/.style={
    spectrumdefault,
    title=\empty,
    ymax=3e-4,
    xlabel=\phantom{eigenvalues},
  }}
\pgfkeys{/pgfplots/spectrumdefaultcenter/.style={
    spectrumdefault,
    ylabel=\empty,
    yticklabels=\empty,
    title=\empty,
    ymax=3e-4,
  }}
\pgfkeys{/pgfplots/spectrumdefaultright/.style={
    spectrumdefault,
    ylabel=\empty,
    yticklabels=\empty,
    title=\empty,
    ymax=3e-4,
    xlabel=\phantom{eigenvalues},
  }}
 \pgfkeys{/pgfplots/zmystyle/.style={spectrumdefaultleft, ymax = 5e-3,
          title={exact, mb}}}
      \begin{tikzpicture}

\begin{axis}[
axis line style={white!80!black},
log basis x={10},
tick pos=left,
title={full\_batch\_exact, layerwise\_group, N=64, D=1387108},
xlabel={eigenvalues},
xmin=0.0001, xmax=6.4880485534668,
xmode=log,
ylabel={density},
ymin=7.20923878368607e-07, ymax=1,
ymode=log,
zmystyle
]
\draw[draw=white,fill=black] (axis cs:0.0001,7.20923878368607e-07) rectangle (axis cs:0.000146532840598227,0.996411238346202);
\draw[draw=white,fill=black] (axis cs:0.000146532840598227,7.20923878368607e-07) rectangle (axis cs:0.000214718733737854,0.00111454911894659);
\draw[draw=white,fill=black] (axis cs:0.000214718733737854,7.20923878368607e-07) rectangle (axis cs:0.000314633459842622,0.000771389105449934);
\draw[draw=white,fill=black] (axis cs:0.000314633459842622,7.20923878368607e-07) rectangle (axis cs:0.000461041346179876,0.000488786741393628);
\draw[draw=white,fill=black] (axis cs:0.000461041346179876,7.20923878368607e-07) rectangle (axis cs:0.000675576980889677,0.000309276566266222);
\draw[draw=white,fill=black] (axis cs:0.000675576980889677,7.20923878368607e-07) rectangle (axis cs:0.000989942140525386,0.000193928662569839);
\draw[draw=white,fill=black] (axis cs:0.000989942140525386,7.20923878368607e-07) rectangle (axis cs:0.00145059033879074,0.0001348128619254);
\draw[draw=white,fill=black] (axis cs:0.00145059033879074,7.20923878368607e-07) rectangle (axis cs:0.00212559122887352,0.000103813112807046);
\draw[draw=white,fill=black] (axis cs:0.00212559122887352,7.20923878368607e-07) rectangle (axis cs:0.00311468920717513,8.43481540582188e-05);
\draw[draw=white,fill=black] (axis cs:0.00311468920717513,7.20923878368607e-07) rectangle (axis cs:0.00456404257108012,6.63250441057491e-05);
\draw[draw=white,fill=black] (axis cs:0.00456404257108012,7.20923878368607e-07) rectangle (axis cs:0.00668782122551605,4.75810097548786e-05);
\draw[draw=white,fill=black] (axis cs:0.00668782122551605,7.20923878368607e-07) rectangle (axis cs:0.00979985441587983,2.95578998024089e-05);
\draw[draw=white,fill=black] (axis cs:0.00979985441587983,7.20923878368607e-07) rectangle (axis cs:0.0143600050500795,3.09997485987664e-05);
\draw[draw=white,fill=black] (axis cs:0.0143600050500795,7.20923878368607e-07) rectangle (axis cs:0.0210421233099304,2.66742022101381e-05);
\draw[draw=white,fill=black] (axis cs:0.0210421233099304,7.20923878368607e-07) rectangle (axis cs:0.0308336210082227,2.37905046176453e-05);
\draw[draw=white,fill=black] (axis cs:0.0308336210082227,7.20923878368607e-07) rectangle (axis cs:0.0451813807226404,2.4511429015602e-05);
\draw[draw=white,fill=black] (axis cs:0.0451813807226404,7.20923878368607e-07) rectangle (axis cs:0.0662055605943847,2.59532778119594e-05);
\draw[draw=white,fill=black] (axis cs:0.0662055605943847,7.20923878368607e-07) rectangle (axis cs:0.0970128885729323,3.46043705892159e-05);
\draw[draw=white,fill=black] (axis cs:0.0970128885729323,7.20923878368607e-07) rectangle (axis cs:0.142155741372311,3.74880681817088e-05);
\draw[draw=white,fill=black] (axis cs:0.142155741372311,7.20923878368607e-07) rectangle (axis cs:0.208304845906316,2.52323534137807e-05);
\draw[draw=white,fill=black] (axis cs:0.208304845906316,7.20923878368607e-07) rectangle (axis cs:0.305235007810284,1.08138654517605e-05);
\draw[draw=white,fill=black] (axis cs:0.305235007810284,7.20923878368607e-07) rectangle (axis cs:0.44726952744463,9.37201665562512e-06);
\draw[draw=white,fill=black] (axis cs:0.44726952744463,7.20923878368607e-07) rectangle (axis cs:0.655396743694883,7.93016785948971e-06);
\draw[draw=white,fill=black] (axis cs:0.655396743694883,7.20923878368607e-07) rectangle (axis cs:0.960371465724394,4.32554586881814e-06);
\draw[draw=white,fill=black] (axis cs:0.960371465724394,7.20923878368607e-07) rectangle (axis cs:1.40725958902078,7.20923878368607e-07);
\draw[draw=white,fill=black] (axis cs:1.40725958902078,7.20923878368607e-07) rectangle (axis cs:2.06209745038309,7.20923878368607e-07);
\draw[draw=white,fill=black] (axis cs:2.06209745038309,7.20923878368607e-07) rectangle (axis cs:3.02164996994996,7.20923878368607e-07);
\draw[draw=white,fill=black] (axis cs:3.02164996994996,7.20923878368607e-07) rectangle (axis cs:4.42770953390315,7.20923878368607e-07);
\draw[draw=white,fill=black] (axis cs:4.42770953390315,7.20923878368607e-07) rectangle (axis cs:6.4880485534668,7.20923878368607e-07);
\end{axis}

\end{tikzpicture}
     \end{minipage}
    \hspace{1.9ex}
    \begin{minipage}[t]{0.33\linewidth}
      \centering
\pgfkeys{/pgfplots/spectrumdefault/.style={
    width=1.38\linewidth,
    height=\goldenRatio*1.38\linewidth,
    every axis plot/.append style={line width = 1.2pt},
    every axis plot post/.append style={
      mark size=1, mark options={opacity=0.3}
    },
    tick pos = left,
    xmajorticks = true,
    ymajorticks = true,
    ylabel near ticks,
    xlabel near ticks,
    xtick align = inside,
    ytick align = inside,
    legend cell align = left,
    legend columns = 1,
    legend pos = north west,
    legend style = {
      fill opacity = 0.9,
      text opacity = 1,
      font = \small,
    },
    xticklabel style = {font = \small, inner xsep = 0ex},
    xlabel style = {font = \small},
    axis line style = {black},
    yticklabel style = {font = \small, inner ysep = 0ex},
    ylabel style = {font = \small, inner ysep = 0ex},
    title style = {font = \small, inner ysep = 0ex, yshift = -0.75ex},
    grid = major,
    grid style = {dashed}
  }
}
\pgfkeys{/pgfplots/spectrumdefaultleft/.style={
    spectrumdefault,
    title=\empty,
    ymax=3e-4,
    xlabel=\phantom{eigenvalues},
  }}
\pgfkeys{/pgfplots/spectrumdefaultcenter/.style={
    spectrumdefault,
    ylabel=\empty,
    yticklabels=\empty,
    title=\empty,
    ymax=3e-4,
  }}
\pgfkeys{/pgfplots/spectrumdefaultright/.style={
    spectrumdefault,
    ylabel=\empty,
    yticklabels=\empty,
    title=\empty,
    ymax=3e-4,
    xlabel=\phantom{eigenvalues},
  }}
 \pgfkeys{/pgfplots/zmystyle/.style={spectrumdefaultcenter, ymax = 5e-3,
          title={exact, sub}}}
      \begin{tikzpicture}

\definecolor{color0}{rgb}{0.274509803921569,0.6,0.564705882352941}

\begin{axis}[
axis line style={white!80!black},
log basis x={10},
tick pos=left,
title={frac\_batch\_exact, layerwise\_group, N=64, D=1387108},
xlabel={eigenvalues},
xmin=0.0001, xmax=6.4880485534668,
xmode=log,
ylabel={density},
ymin=7.20923878368607e-07, ymax=1,
ymode=log,
zmystyle
]
\draw[draw=white,fill=color0] (axis cs:0.0001,7.20923878368607e-07) rectangle (axis cs:0.000146532840598227,0.997698088396815);
\draw[draw=white,fill=color0] (axis cs:0.000146532840598227,7.20923878368607e-07) rectangle (axis cs:0.000214718733737854,0.000378485308484052);
\draw[draw=white,fill=color0] (axis cs:0.000214718733737854,7.20923878368607e-07) rectangle (axis cs:0.000314633459842622,0.00029125345631353);
\draw[draw=white,fill=color0] (axis cs:0.000314633459842622,7.20923878368607e-07) rectangle (axis cs:0.000461041346179876,0.000274672195157196);
\draw[draw=white,fill=color0] (axis cs:0.000461041346179876,7.20923878368607e-07) rectangle (axis cs:0.000675576980889677,0.00024439337043702);
\draw[draw=white,fill=color0] (axis cs:0.000675576980889677,7.20923878368607e-07) rectangle (axis cs:0.000989942140525386,0.000208347150531859);
\draw[draw=white,fill=color0] (axis cs:0.000989942140525386,7.20923878368607e-07) rectangle (axis cs:0.00145059033879074,0.000173742779423055);
\draw[draw=white,fill=color0] (axis cs:0.00145059033879074,7.20923878368607e-07) rectangle (axis cs:0.00212559122887352,0.000143463954702657);
\draw[draw=white,fill=color0] (axis cs:0.00212559122887352,7.20923878368607e-07) rectangle (axis cs:0.00311468920717513,0.00012039437396338);
\draw[draw=white,fill=color0] (axis cs:0.00311468920717513,7.20923878368607e-07) rectangle (axis cs:0.00456404257108012,9.44410956316107e-05);
\draw[draw=white,fill=color0] (axis cs:0.00456404257108012,7.20923878368607e-07) rectangle (axis cs:0.00668782122551605,7.28133636886915e-05);
\draw[draw=white,fill=color0] (axis cs:0.00668782122551605,7.20923878368607e-07) rectangle (axis cs:0.00979985441587983,4.75810097548786e-05);
\draw[draw=white,fill=color0] (axis cs:0.00979985441587983,7.20923878368607e-07) rectangle (axis cs:0.0143600050500795,3.53252949873946e-05);
\draw[draw=white,fill=color0] (axis cs:0.0143600050500795,7.20923878368607e-07) rectangle (axis cs:0.0210421233099304,2.73951266080948e-05);
\draw[draw=white,fill=color0] (axis cs:0.0210421233099304,7.20923878368607e-07) rectangle (axis cs:0.0308336210082227,2.66742022101381e-05);
\draw[draw=white,fill=color0] (axis cs:0.0308336210082227,7.20923878368607e-07) rectangle (axis cs:0.0451813807226404,2.59532778119594e-05);
\draw[draw=white,fill=color0] (axis cs:0.0451813807226404,7.20923878368607e-07) rectangle (axis cs:0.0662055605943847,2.66742022099161e-05);
\draw[draw=white,fill=color0] (axis cs:0.0662055605943847,7.20923878368607e-07) rectangle (axis cs:0.0970128885729323,3.38834461910372e-05);
\draw[draw=white,fill=color0] (axis cs:0.0970128885729323,7.20923878368607e-07) rectangle (axis cs:0.142155741372311,3.60462193855734e-05);
\draw[draw=white,fill=color0] (axis cs:0.142155741372311,7.20923878368607e-07) rectangle (axis cs:0.208304845906316,2.52323534137807e-05);
\draw[draw=white,fill=color0] (axis cs:0.208304845906316,7.20923878368607e-07) rectangle (axis cs:0.305235007810284,1.15347898499392e-05);
\draw[draw=white,fill=color0] (axis cs:0.305235007810284,7.20923878368607e-07) rectangle (axis cs:0.44726952744463,8.65109225744639e-06);
\draw[draw=white,fill=color0] (axis cs:0.44726952744463,7.20923878368607e-07) rectangle (axis cs:0.655396743694883,7.93016785948971e-06);
\draw[draw=white,fill=color0] (axis cs:0.655396743694883,7.20923878368607e-07) rectangle (axis cs:0.960371465724394,4.32554586881814e-06);
\draw[draw=white,fill=color0] (axis cs:0.960371465724394,7.20923878368607e-07) rectangle (axis cs:1.40725958902078,7.20923878368607e-07);
\draw[draw=white,fill=color0] (axis cs:1.40725958902078,7.20923878368607e-07) rectangle (axis cs:2.06209745038309,7.20923878368607e-07);
\draw[draw=white,fill=color0] (axis cs:2.06209745038309,7.20923878368607e-07) rectangle (axis cs:3.02164996994996,7.20923878368607e-07);
\draw[draw=white,fill=color0] (axis cs:3.02164996994996,7.20923878368607e-07) rectangle (axis cs:4.42770953390315,7.20923878368607e-07);
\draw[draw=white,fill=color0] (axis cs:4.42770953390315,7.20923878368607e-07) rectangle (axis cs:6.4880485534668,7.20923878368607e-07);
\end{axis}

\end{tikzpicture}
     \end{minipage}
    \hspace{-3.5ex}
    \begin{minipage}[t]{0.33\linewidth}
      \centering
\pgfkeys{/pgfplots/spectrumdefault/.style={
    width=1.38\linewidth,
    height=\goldenRatio*1.38\linewidth,
    every axis plot/.append style={line width = 1.2pt},
    every axis plot post/.append style={
      mark size=1, mark options={opacity=0.3}
    },
    tick pos = left,
    xmajorticks = true,
    ymajorticks = true,
    ylabel near ticks,
    xlabel near ticks,
    xtick align = inside,
    ytick align = inside,
    legend cell align = left,
    legend columns = 1,
    legend pos = north west,
    legend style = {
      fill opacity = 0.9,
      text opacity = 1,
      font = \small,
    },
    xticklabel style = {font = \small, inner xsep = 0ex},
    xlabel style = {font = \small},
    axis line style = {black},
    yticklabel style = {font = \small, inner ysep = 0ex},
    ylabel style = {font = \small, inner ysep = 0ex},
    title style = {font = \small, inner ysep = 0ex, yshift = -0.75ex},
    grid = major,
    grid style = {dashed}
  }
}
\pgfkeys{/pgfplots/spectrumdefaultleft/.style={
    spectrumdefault,
    title=\empty,
    ymax=3e-4,
    xlabel=\phantom{eigenvalues},
  }}
\pgfkeys{/pgfplots/spectrumdefaultcenter/.style={
    spectrumdefault,
    ylabel=\empty,
    yticklabels=\empty,
    title=\empty,
    ymax=3e-4,
  }}
\pgfkeys{/pgfplots/spectrumdefaultright/.style={
    spectrumdefault,
    ylabel=\empty,
    yticklabels=\empty,
    title=\empty,
    ymax=3e-4,
    xlabel=\phantom{eigenvalues},
  }}
 \pgfkeys{/pgfplots/zmystyle/.style={spectrumdefaultright, ymax = 5e-3,
          title={mc, full}}}
      \begin{tikzpicture}

\definecolor{color0}{rgb}{0.870588235294118,0.623529411764706,0.0862745098039216}

\begin{axis}[
axis line style={white!80!black},
log basis x={10},
tick pos=left,
title={full\_batch\_mc, layerwise\_group, N=64, D=1387108},
xlabel={eigenvalues},
xmin=0.0001, xmax=6.4880485534668,
xmode=log,
ylabel={density},
ymin=7.20923878368607e-07, ymax=1,
ymode=log,
zmystyle
]
\draw[draw=white,fill=color0] (axis cs:0.0001,7.20923878368607e-07) rectangle (axis cs:0.000146532840598227,0.999596282337018);
\draw[draw=white,fill=color0] (axis cs:0.000146532840598227,7.20923878368607e-07) rectangle (axis cs:0.000214718733737854,5.04647026699686e-06);
\draw[draw=white,fill=color0] (axis cs:0.000214718733737854,7.20923878368607e-07) rectangle (axis cs:0.000314633459842622,8.65109225744639e-06);
\draw[draw=white,fill=color0] (axis cs:0.000314633459842622,7.20923878368607e-07) rectangle (axis cs:0.000461041346179876,7.93016785926767e-06);
\draw[draw=white,fill=color0] (axis cs:0.000461041346179876,7.20923878368607e-07) rectangle (axis cs:0.000675576980889677,1.44184874424321e-05);
\draw[draw=white,fill=color0] (axis cs:0.000675576980889677,7.20923878368607e-07) rectangle (axis cs:0.000989942140525386,1.65812606365242e-05);
\draw[draw=white,fill=color0] (axis cs:0.000989942140525386,7.20923878368607e-07) rectangle (axis cs:0.00145059033879074,2.23486558215099e-05);
\draw[draw=white,fill=color0] (axis cs:0.00145059033879074,7.20923878368607e-07) rectangle (axis cs:0.00212559122887352,2.4511429015824e-05);
\draw[draw=white,fill=color0] (axis cs:0.00212559122887352,7.20923878368607e-07) rectangle (axis cs:0.00311468920717513,2.66742022099161e-05);
\draw[draw=white,fill=color0] (axis cs:0.00311468920717513,7.20923878368607e-07) rectangle (axis cs:0.00456404257108012,2.30695802196886e-05);
\draw[draw=white,fill=color0] (axis cs:0.00456404257108012,7.20923878368607e-07) rectangle (axis cs:0.00668782122551605,2.37905046176453e-05);
\draw[draw=white,fill=color0] (axis cs:0.00668782122551605,7.20923878368607e-07) rectangle (axis cs:0.00979985441587983,2.37905046176453e-05);
\draw[draw=white,fill=color0] (axis cs:0.00979985441587983,7.20923878368607e-07) rectangle (axis cs:0.0143600050500795,2.59532778119594e-05);
\draw[draw=white,fill=color0] (axis cs:0.0143600050500795,7.20923878368607e-07) rectangle (axis cs:0.0210421233099304,2.52323534137807e-05);
\draw[draw=white,fill=color0] (axis cs:0.0210421233099304,7.20923878368607e-07) rectangle (axis cs:0.0308336210082227,2.4511429015824e-05);
\draw[draw=white,fill=color0] (axis cs:0.0308336210082227,7.20923878368607e-07) rectangle (axis cs:0.0451813807226404,2.16277314233312e-05);
\draw[draw=white,fill=color0] (axis cs:0.0451813807226404,7.20923878368607e-07) rectangle (axis cs:0.0662055605943847,1.65812606365242e-05);
\draw[draw=white,fill=color0] (axis cs:0.0662055605943847,7.20923878368607e-07) rectangle (axis cs:0.0970128885729323,1.80231094328816e-05);
\draw[draw=white,fill=color0] (axis cs:0.0970128885729323,7.20923878368607e-07) rectangle (axis cs:0.142155741372311,2.16277314233312e-05);
\draw[draw=white,fill=color0] (axis cs:0.142155741372311,7.20923878368607e-07) rectangle (axis cs:0.208304845906316,2.66742022099161e-05);
\draw[draw=white,fill=color0] (axis cs:0.208304845906316,7.20923878368607e-07) rectangle (axis cs:0.305235007810284,1.15347898499392e-05);
\draw[draw=white,fill=color0] (axis cs:0.305235007810284,7.20923878368607e-07) rectangle (axis cs:0.44726952744463,1.51394118403888e-05);
\draw[draw=white,fill=color0] (axis cs:0.44726952744463,7.20923878368607e-07) rectangle (axis cs:0.655396743694883,7.93016785948971e-06);
\draw[draw=white,fill=color0] (axis cs:0.655396743694883,7.20923878368607e-07) rectangle (axis cs:0.960371465724394,5.76739466517559e-06);
\draw[draw=white,fill=color0] (axis cs:0.960371465724394,7.20923878368607e-07) rectangle (axis cs:1.40725958902078,3.60462147063941e-06);
\draw[draw=white,fill=color0] (axis cs:1.40725958902078,7.20923878368607e-07) rectangle (axis cs:2.06209745038309,1.44184827654733e-06);
\draw[draw=white,fill=color0] (axis cs:2.06209745038309,7.20923878368607e-07) rectangle (axis cs:3.02164996994996,7.20923878368607e-07);
\draw[draw=white,fill=color0] (axis cs:3.02164996994996,7.20923878368607e-07) rectangle (axis cs:4.42770953390315,7.20923878368607e-07);
\draw[draw=white,fill=color0] (axis cs:4.42770953390315,7.20923878368607e-07) rectangle (axis cs:6.4880485534668,7.20923878368607e-07);
\end{axis}

\end{tikzpicture}
     \end{minipage}
  \end{minipage}

  \caption{\textbf{\ggn spectra of different architectures under \vivittitle's
      approximations:} Left and right columns contain results with the full
    network's \ggn and a per-layer block-diagonal approximation, respectively.}
  \label{fig:spectrum_2}
\end{figure}

\subsection{Performance evaluation}
\label{sec:performance-experiments}
\paragraph{Hardware specifications:} Results in this section were generated on a workstation with an Intel Core i7-8700K CPU (32\,GB)
and one NVIDIA GeForce RTX 2080 Ti GPU (11\,GB). 

\paragraph{Note:} \vivit{}'s quantities are implemented through \backpack, which is triggered by
\pytorch's gradient computation. Consequently, they can only be computed
together with \pytorch{}'s mini-batch gradient.

\paragraph{Architectures:} We use untrained deep convolutional and residual
networks from \deepobs \cite{schneider2019deepobs} and
\cite{idelbayev2018proper}. If a net has batch normalization layers, we set them
to evaluation mode. Otherwise, the loss would not obey the sum structure of
\Cref{eq:objective-function}. The batch normalization layers' internal moving
averages, required for evaluation mode, are initialized by performing five
forward passes with the current mini-batch in training mode before. 

In experiments with fixed mini-batches the batch sizes correspond to \deepobs'
default value for training where possible (\cifarten: $N=128$, \fmnist:
$N=128$). The residual networks use a batch size of $N=128$. 
On \cifarhun (trained with $N=256$), we reduce the batch size to
$N=64$ to fit the exact computation on the full mini-batch, used as baseline,
into memory. 
If the \ggn approximation is evaluated on a subset of the mini-batch (\textbf{sub}), 
$\floor{\nicefrac{N}{8}}$ of the samples are used (as in \cite{zhang2017blockdiagonal}). The \mc approximation is always evaluated with a single sample ($M=1$).

\paragraph{Memory performance (critical batch sizes):} 
Two tasks are considered (see \Cref{subsec:scalability}):
\begin{enumerate}
\item \textbf{Computing eigenvalues:} Compute the nontrivial eigenvalues $\{\lambda_{k}\,|\,
  (\lambda_{k}, \vetilde_{k}) \in \tilde{\sS}_+\}$ . 
\item \textbf{Computing the top eigenpair:} Compute the top eigenpair $(\lambda_{1}, \ve_{1})$.
\end{enumerate}

We repeat the tasks above and vary the mini-batch size until the device runs out
of memory. 
The largest mini-batch size that can be handled by our system is denoted as
$N_{\text{crit}}$, the critical batch size.
We determine this number by
bisection on the interval $[1; 32768]$.

\Cref{fig:performance-cifar10-3c3d-cpu,fig:performance-cifar10-3c3d-cuda,fig:performance-fmnist-2c2d-cpu,fig:performance-fmnist-2c2d-cuda,fig:performance-cifar100-allcnnc-cpu,fig:performance-cifar100-allcnnc-cuda,fig:performance-cifar10-resnet32-cpu,fig:performance-cifar10-resnet32-cuda,fig:performance-cifar10-resnet56-cpu,fig:performance-cifar10-resnet56-cuda}a,b
present the results. As described in \Cref{sec:method-complexity},
computing eigenvalues is more memory-efficient than computing eigenvectors and
exhibits larger critical batch sizes. In line with the description in
\Cref{sec:approximations}, a block-diagonal approximation is usually more
memory-efficient and results in a larger critical batch size. Curvature
sub-sampling and \mc approximation further increase the applicable batch sizes.

In summary, we find that there always exists a combination of approximations
which allows for critical batch sizes larger than the traditional size used for
training (some architectures even permit exact computation). Different
accuracy-cost trade-offs may be preferred, depending on the application and the
computational budget. By the presented approximations, \vivit's representation
is capable to adapt over a wide range.

\paragraph{Runtime performance:} 
Here, we consider the task of computing the $k$ leading eigenvectors and eigenvalues of a matrix.
\vivit{}'s eigenpair computation is compared
with a power iteration that computes eigenpairs iteratively via matrix-vector products. 
The power iteration baseline is based on the
\pyhessian library \cite{yao2019pyhessian} and uses the same termination
criterion (at most 100 matrix-vector products per eigenvalue; stop if the
eigenvalue estimate's relative change is less than $10^{-3}$). In contrast to
\pyhessian, we use a different data format and stack the computed eigenvectors.
This reduces the number of \texttt{for}-loops in the orthonormalization step. We
repeat each run time measurement $20$ times and report the shortest execution time
as result.

\Cref{fig:performance-cifar10-3c3d-cpu,fig:performance-cifar10-3c3d-cuda,fig:performance-fmnist-2c2d-cpu,fig:performance-fmnist-2c2d-cuda,fig:performance-cifar100-allcnnc-cpu,fig:performance-cifar100-allcnnc-cuda,fig:performance-cifar10-resnet32-cpu,fig:performance-cifar10-resnet32-cuda,fig:performance-cifar10-resnet56-cpu,fig:performance-cifar10-resnet56-cuda}c,d
show the results. For most architectures, our exact method outperforms
the power iteration for $k>1$ and increases only marginally in runtime as the
number of requested eigenvectors grows. The proposed approximations share this
property, and further reduce run time.

\paragraph{Note on \cifarhun (large $\mathbf{C}$):} For data sets with a large number of
classes, like \cifarhun ($C=100$), computations with the exact \ggn are costly.
In particular, constructing the Gram matrix $\mGtilde$ has quadratic memory cost
in $C$, and its eigendecomposition has cubic cost in time with $C$ (see
\Cref{sec:method-complexity}).

As a result, the exact computation only works with batch sizes smaller than
\deepobs' default ($N=256$ for \cifarhun, see
\Cref{fig:performance-cifar100-allcnnc-cpu,fig:performance-cifar100-allcnnc-cuda}a,b).
For the \ggn block-diagonal approximation, which fits into CPU memory for
$N=64$, the exact computation of top eigenpairs is slower than a power iteration
and only becomes comparable if a large number of eigenpairs is requested, see
\Cref{fig:performance-cifar100-allcnnc-cpu}d.

For such data sets, the approximations proposed in \Cref{sec:approximations} are
essential to reduce costs. The most effective approximation to eliminate the
scaling with $C$ is using an \mc approximation.
\Cref{fig:performance-cifar100-allcnnc-cpu,fig:performance-cifar100-allcnnc-cuda}
confirm that the approximate computations scale to batch sizes used for training
and that computing eigenpairs takes less time than a power iteration.

\paragraph{Computing damped Newton steps:} A Newton step $-(\mG + \delta
\mI)^{-1} \vg$ with damping $\delta > 0$ can be decomposed into updates along the eigenvectors of the \ggn $\mG$, 
\begin{equation}
  \label{eq:newton-step}
  -(\mG + \delta \mI)^{-1} \vg
=
   \sum_{k=1}^{K} \frac{- \gamma_{k}}{\lambda_{k} + \delta} \ve_{k}
  + \sum_{k = K + 1}^{D} \frac{-\gamma_{k}}{\delta} \ve_{k}\,.
\end{equation}
It corresponds to a Newton update along nontrivial eigendirections that uses the
first- and second-order directional derivatives described in
\Cref{sec:comp-direct-deriv} and a gradient descent step with learning rate
$\nicefrac{1}{\delta}$ along trivial directions (with $\lambda_k = 0$). In the following, we refer to
the first summand of \Cref{eq:newton-step} as Newton step. As described in
\Cref{sec:method-complexity}, we can perform the weighted sum in the Gram matrix
space, rather than the parameter space, by computing
\begin{equation*}
  \sum_{k=1}^{K} \frac{- \gamma_{k}}{\lambda_{k} + \delta} \ve_{k}
  =
  \sum_{k=1}^{K} \frac{- \gamma_{k}}{\lambda_{k} + \delta} \frac{1}{\sqrt{\lambda_{k}}} \mV \vetilde_{k}
  =\mV \left(
    \sum_{k=1}^{K} \frac{- \gamma_{k}}{(\lambda_{k} + \delta)\sqrt{\lambda_{k}}} \vetilde_{k}
    \right)\,.
\end{equation*}
This way, only a single vector needs to be transformed from Gram space into parameter space. 

\Cref{tab:performance} shows the critical batch sizes for the Newton step
computation (first term on the right-hand side of \Cref{eq:newton-step}), using Gram matrix eigenvalues
larger than $10^{-4}$ and constant damping $\delta=1$. Second-order directional
derivatives $\lambda_{k}$ are evaluated on the same samples as the \ggn
eigenvectors, but we \emph{always} use all mini-batch samples to compute the
directional gradients $\gamma_{n}$. Using our approximations, the Newton step
computation scales to batch sizes beyond the traditional sizes used for training.

\clearpage
\begin{figure*}[tb]
  \centering
  \begin{minipage}[t]{0.49\linewidth}
    \centering
    \textbf{Full network}
    \vspace{-3ex}

    \begin{minipage}[t]{0.49\linewidth}
      \centering

      \begin{flushleft}
        \vspace{1ex}
        (a)
      \end{flushleft}

      \vspace{-1.0\baselineskip}
      \begin{small}
        $N_{\text{crit}}$ (eigenvalues)
      \end{small}
      \vspace{0.15\baselineskip}

      \begin{small}
        \begin{tabular}{lll}
    \toprule
    $_{\text{\tiny{\ggn}}}$$^{\text{\tiny{Data}}}$ & mb & sub \\
    \midrule
    exact & 1753
              & 8235 \\
    mc   & 7585
              & 12434 \\
    \bottomrule
\end{tabular}       \end{small}
    \end{minipage}
    \hfill
    \begin{minipage}[t]{0.49\linewidth}
      \centering

      \begin{flushleft}
        \vspace{1ex}
        \phantom{(a)}
      \end{flushleft}
      \vspace{-1.0\baselineskip}

      \begin{small}
        $N_{\text{crit}}$ (top eigenpair)
      \end{small}
      \vspace{0.15\baselineskip}

      \begin{small}
        \begin{tabular}{lll}
    \toprule
    $_{\text{\tiny{\ggn}}}$$^{\text{\tiny{Data}}}$ & mb & sub \\
    \midrule
    exact & 872
              & 6439 \\
    mc   & 6708
              & 12162 \\
    \bottomrule
\end{tabular}       \end{small}
    \end{minipage}

    \begin{flushleft}
      \vspace{1ex}
      (c)
    \end{flushleft}

    \vspace{-1.3\baselineskip}

\pgfkeys{/pgfplots/performancedefault/.style={
    width=1.04\linewidth,
    height=\goldenRatioInv*1.04\linewidth,
    every axis plot/.append style={line width = 1.2pt},
    every axis plot post/.append style={
      mark size=2, mark options={opacity=0.9, solid, line width = 1pt}
    },
    tick pos = left,
    xmajorticks = true,
    ymajorticks = true,
    ylabel near ticks,
    xlabel near ticks,
    xtick align = inside,
    ytick align = inside,
    legend cell align = left,
    legend columns = 3,
legend style = {
      fill opacity = 0.9,
      text opacity = 1,
      font = \small,
      at={(1, 1.025)},
      anchor=south east,
    },
    xticklabel style = {font = \small, inner xsep = 0ex},
    xlabel style = {font = \small},
    axis line style = {black},
    yticklabel style = {font = \small, inner ysep = 0ex},
    ylabel style = {font = \small, inner ysep = 0ex},
    title style = {font = \small, inner ysep = 0ex, yshift = -0.75ex},
    grid = major,
    grid style = {dashed},
    title = {},
  }
}
 \pgfkeys{/pgfplots/zmystyle/.style={performancedefault}}
    \begin{tikzpicture}

\definecolor{color0}{rgb}{0.937254901960784,0.231372549019608,0.172549019607843}
\definecolor{color1}{rgb}{0.274509803921569,0.6,0.564705882352941}
\definecolor{color2}{rgb}{0.870588235294118,0.623529411764706,0.0862745098039216}
\definecolor{color3}{rgb}{0.501960784313725,0.184313725490196,0.6}

\begin{axis}[
axis line style={white!80!black},
legend style={fill opacity=0.8, draw opacity=1, text opacity=1, at={(0.03,0.97)}, anchor=north west, draw=white!80!black},
tick pos=left,
title={fmnist\_2c2d, N=128, cuda, one\_group},
xlabel={top eigenpairs (\(\displaystyle k\))},
xmin=0.55, xmax=10.45,
ylabel={time [s]},
ymin=0.00716881992461236, ymax=1.64798554063795,
ymode=log,
zmystyle
]
\addplot [, color0, dashed, mark=pentagon*, mark size=3, mark options={solid}]
table {1 0.0389841239957605
2 0.215932399005396
3 0.36564161098795
4 0.508233977016062
5 0.779013090010267
6 0.847288260003552
7 1.00752733499394
8 1.03728621799382
9 1.07439484001952
10 1.28709981401335
};
\addlegendentry{power iteration}
\addplot [, black, dashed, mark=*, mark size=3, mark options={solid}]
table {1 0.104063767001207
2 0.104606729000807
3 0.105111179000232
4 0.105373603997577
5 0.106095311995887
6 0.10573411200312
7 0.106235621999076
8 0.105651376004971
9 0.106636896001874
10 0.106543941001291
};
\addlegendentry{mb, exact}
\addplot [, color1, dashed, mark=diamond*, mark size=3, mark options={solid}]
table {1 0.0169207959988853
2 0.0169227890000911
3 0.0169238109956495
4 0.0171220839984016
5 0.0175330389974988
6 0.0174389730018447
7 0.0176948610023828
8 0.0178204210023978
9 0.0182167439998011
10 0.0180330259972834
};
\addlegendentry{sub, exact}
\addplot [, color2, dashed, mark=square*, mark size=3, mark options={solid}]
table {1 0.013006890003453
2 0.0119940609947662
3 0.0113651979991118
4 0.0114681339982781
5 0.0134660580006312
6 0.0117812719981885
7 0.0118628670024918
8 0.0131400940008461
9 0.0123189779988024
10 0.0123788509954466
};
\addlegendentry{mb, mc}
\addplot [, color3, dashed, mark=triangle*, mark size=3, mark options={solid,rotate=180}]
table {1 0.00918164000177057
2 0.00917886200477369
3 0.00932389000081457
4 0.00943565999477869
5 0.00955776900082128
6 0.00954905400431016
7 0.00984142100060126
8 0.00992608799424488
9 0.0100936599992565
10 0.0102828860035515
};
\addlegendentry{sub, mc}
\end{axis}

\end{tikzpicture}
   \end{minipage}
  \hfill
  \begin{minipage}[t]{0.49\linewidth}
    \centering
    \textbf{Block-diagonal approximation}
    \vspace{-3ex}

    \begin{minipage}[t]{0.49\linewidth}
      \centering

      \begin{flushleft}
        \vspace{1ex}
        (b)
      \end{flushleft}

      \vspace{-1.0\baselineskip}
      \begin{small}
        $N_{\text{crit}}$ (eigenvalues)
      \end{small}
      \vspace{0.15\baselineskip}

      \begin{small}
        \begin{tabular}{lll}
    \toprule
    $_{\text{\tiny{\ggn}}}$$^{\text{\tiny{Data}}}$ & mb & sub \\
    \midrule
    exact & 1012
              & 8317 \\
    mc   & 8448
              & 12455 \\
    \bottomrule
\end{tabular}       \end{small}
    \end{minipage}
    \hfill
    \begin{minipage}[t]{0.49\linewidth}
      \centering

      \begin{flushleft}
        \vspace{1ex}
        \phantom{(a)}
      \end{flushleft}
      \vspace{-1.0\baselineskip}

      \begin{small}
        $N_{\text{crit}}$ (top eigenpair)
      \end{small}
      \vspace{0.15\baselineskip}

      \begin{small}
        \begin{tabular}{lll}
    \toprule
    $_{\text{\tiny{\ggn}}}$$^{\text{\tiny{Data}}}$ & mb & sub \\
    \midrule
    exact & 993
              & 7479 \\
    mc   & 8336
              & 12413 \\
    \bottomrule
\end{tabular}       \end{small}
    \end{minipage}

    \begin{flushleft}
      \vspace{1ex}
      (d)
    \end{flushleft}

    \vspace{-1.3\baselineskip}

\pgfkeys{/pgfplots/performancedefault/.style={
    width=1.04\linewidth,
    height=\goldenRatioInv*1.04\linewidth,
    every axis plot/.append style={line width = 1.2pt},
    every axis plot post/.append style={
      mark size=2, mark options={opacity=0.9, solid, line width = 1pt}
    },
    tick pos = left,
    xmajorticks = true,
    ymajorticks = true,
    ylabel near ticks,
    xlabel near ticks,
    xtick align = inside,
    ytick align = inside,
    legend cell align = left,
    legend columns = 3,
legend style = {
      fill opacity = 0.9,
      text opacity = 1,
      font = \small,
      at={(1, 1.025)},
      anchor=south east,
    },
    xticklabel style = {font = \small, inner xsep = 0ex},
    xlabel style = {font = \small},
    axis line style = {black},
    yticklabel style = {font = \small, inner ysep = 0ex},
    ylabel style = {font = \small, inner ysep = 0ex},
    title style = {font = \small, inner ysep = 0ex, yshift = -0.75ex},
    grid = major,
    grid style = {dashed},
    title = {},
  }
}
 \pgfkeys{/pgfplots/zmystyle/.style={performancedefault}}
    \begin{tikzpicture}

\definecolor{color0}{rgb}{0.937254901960784,0.231372549019608,0.172549019607843}
\definecolor{color1}{rgb}{0.274509803921569,0.6,0.564705882352941}
\definecolor{color2}{rgb}{0.870588235294118,0.623529411764706,0.0862745098039216}
\definecolor{color3}{rgb}{0.501960784313725,0.184313725490196,0.6}

\begin{axis}[
axis line style={white!80!black},
legend style={fill opacity=0.8, draw opacity=1, text opacity=1, at={(0.03,0.97)}, anchor=north west, draw=white!80!black},
tick pos=left,
title={fmnist\_2c2d, N=128, cuda, layerwise\_group},
xlabel={top eigenpairs (\(\displaystyle k\))},
xmin=0.55, xmax=10.45,
ylabel={time [s]},
ymin=0.00915948282752666, ymax=2.11071547112854,
ymode=log,
zmystyle
]
\addplot [, color0, dashed, mark=pentagon*, mark size=3, mark options={solid}]
table {1 0.0952086269971915
2 0.30820911299088
3 0.446097901003668
4 0.654512957989937
5 0.881432018009946
6 1.01119179200032
7 1.16028599897982
8 1.37328214399167
9 1.48039353499189
10 1.64831670399872
};
\addlegendentry{power iteration}
\addplot [, black, dashed, mark=*, mark size=3, mark options={solid}]
table {1 0.231905585998902
2 0.230233552996651
3 0.23013138800161
4 0.233178911999858
5 0.232878572998743
6 0.232029974999023
7 0.231374043003598
8 0.232957799002179
9 0.230938428001537
10 0.233476699999301
};
\addlegendentry{mb, exact}
\addplot [, color1, dashed, mark=diamond*, mark size=3, mark options={solid}]
table {1 0.0311024019974866
2 0.0308066540019354
3 0.0311830930004362
4 0.0309782749973238
5 0.0313144540050416
6 0.031231934997777
7 0.0313270209953771
8 0.0314724230047432
9 0.0313592019956559
10 0.0314672520034947
};
\addlegendentry{sub, exact}
\addplot [, color2, dashed, mark=square*, mark size=3, mark options={solid}]
table {1 0.0155177899941918
2 0.0154152489994885
3 0.0160517190015526
4 0.0162854049995076
5 0.0158937140004127
6 0.0168336790011381
7 0.0163636810029857
8 0.01703397899837
9 0.016520387995115
10 0.0169850530000986
};
\addlegendentry{mb, mc}
\addplot [, color3, dashed, mark=triangle*, mark size=3, mark options={solid,rotate=180}]
table {1 0.0120809770014603
2 0.0117289729969343
3 0.0123634920018958
4 0.0123846329952357
5 0.0127582870045444
6 0.0129996379982913
7 0.0124986509981682
8 0.0130726139977924
9 0.0123342719962238
10 0.0132257009972818
};
\addlegendentry{sub, mc}
\end{axis}

\end{tikzpicture}
   \end{minipage}

  \vspace{-2ex}
  \caption{\textbf{GPU memory and run time performance for the \twoctwod
      architecture on \fmnist:} Left and right columns show results with the
    full network's \ggn ($D = 3,\!274,\!634$, $C=10$) and a per-layer
    block-diagonal approximation, respectively. (a, b) Critical batch sizes
    $N_{\text{crit}}$ for computing eigenvalues and the top eigenpair. (c, d)
    Run time comparison with a power iteration for extracting the $k$ leading
    eigenpairs using a mini-batch of size $N=128$.}
  \label{fig:performance-fmnist-2c2d-cuda}
\end{figure*}
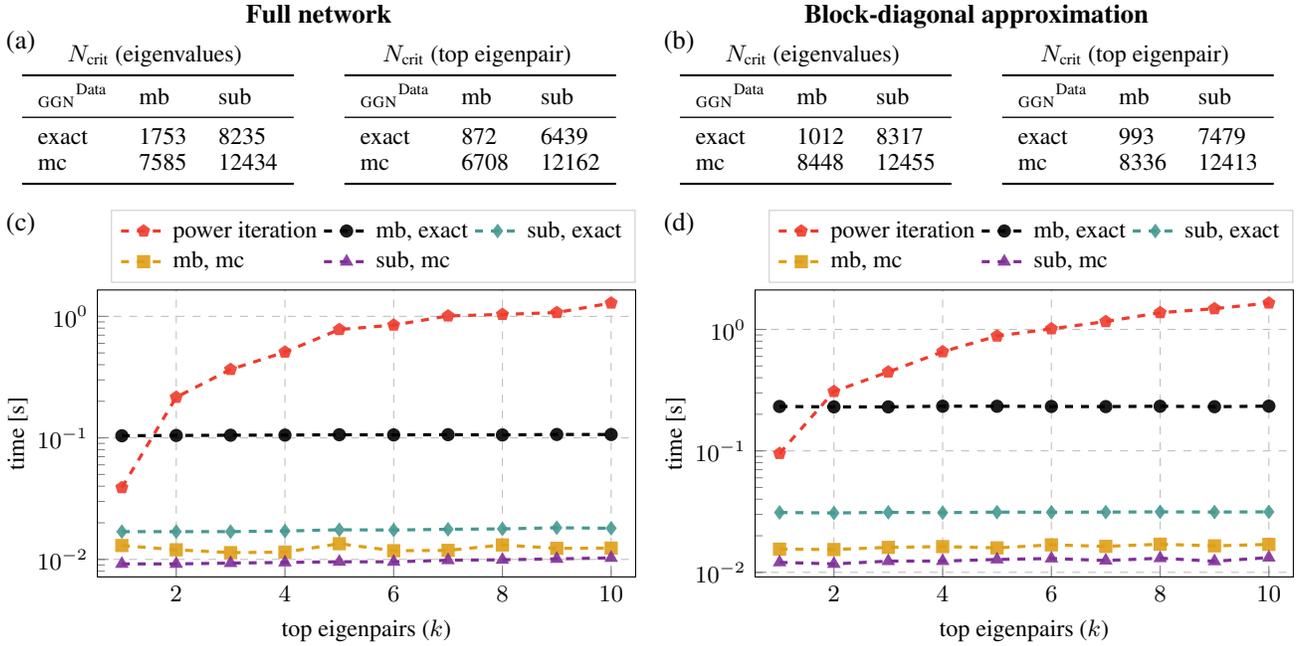

 \begin{figure*}[tb]
  \centering
  \begin{minipage}[t]{0.49\linewidth}
    \centering
    \textbf{Full network}
    \vspace{-3ex}

    \begin{minipage}[t]{0.49\linewidth}
      \centering

      \begin{flushleft}
        \vspace{1ex}
        (a)
      \end{flushleft}

      \vspace{-1.0\baselineskip}
      \begin{small}
        $N_{\text{crit}}$ (eigenvalues)
      \end{small}
      \vspace{0.15\baselineskip}

      \begin{small}
        \begin{tabular}{lll}
    \toprule
    $_{\text{\tiny{\ggn}}}$$^{\text{\tiny{Data}}}$ & mb & sub \\
    \midrule
    exact & 4224
              & 21164 \\
    mc   & 24064
              & > 32768 \\
    \bottomrule
\end{tabular}       \end{small}
    \end{minipage}
    \hfill
    \begin{minipage}[t]{0.49\linewidth}
      \centering

      \begin{flushleft}
        \vspace{1ex}
        \phantom{(a)}
      \end{flushleft}
      \vspace{-1.0\baselineskip}

      \begin{small}
        $N_{\text{crit}}$ (top eigenpair)
      \end{small}
      \vspace{0.15\baselineskip}

      \begin{small}
        \begin{tabular}{lll}
    \toprule
    $_{\text{\tiny{\ggn}}}$$^{\text{\tiny{Data}}}$ & mb & sub \\
    \midrule
    exact & 3276
              & 21295 \\
    mc   & 24064
              & > 32768 \\
    \bottomrule
\end{tabular}       \end{small}
    \end{minipage}

    \begin{flushleft}
      \vspace{1ex}
      (c)
    \end{flushleft}

    \vspace{-1.3\baselineskip}

\pgfkeys{/pgfplots/performancedefault/.style={
    width=1.04\linewidth,
    height=\goldenRatioInv*1.04\linewidth,
    every axis plot/.append style={line width = 1.2pt},
    every axis plot post/.append style={
      mark size=2, mark options={opacity=0.9, solid, line width = 1pt}
    },
    tick pos = left,
    xmajorticks = true,
    ymajorticks = true,
    ylabel near ticks,
    xlabel near ticks,
    xtick align = inside,
    ytick align = inside,
    legend cell align = left,
    legend columns = 3,
legend style = {
      fill opacity = 0.9,
      text opacity = 1,
      font = \small,
      at={(1, 1.025)},
      anchor=south east,
    },
    xticklabel style = {font = \small, inner xsep = 0ex},
    xlabel style = {font = \small},
    axis line style = {black},
    yticklabel style = {font = \small, inner ysep = 0ex},
    ylabel style = {font = \small, inner ysep = 0ex},
    title style = {font = \small, inner ysep = 0ex, yshift = -0.75ex},
    grid = major,
    grid style = {dashed},
    title = {},
  }
}
 \pgfkeys{/pgfplots/zmystyle/.style={performancedefault}}
    \begin{tikzpicture}

\definecolor{color0}{rgb}{0.937254901960784,0.231372549019608,0.172549019607843}
\definecolor{color1}{rgb}{0.274509803921569,0.6,0.564705882352941}
\definecolor{color2}{rgb}{0.870588235294118,0.623529411764706,0.0862745098039216}
\definecolor{color3}{rgb}{0.501960784313725,0.184313725490196,0.6}

\begin{axis}[
axis line style={white!80!black},
legend style={fill opacity=0.8, draw opacity=1, text opacity=1, at={(0.03,0.97)}, anchor=north west, draw=white!80!black},
tick pos=left,
title={fmnist\_2c2d, N=128, cpu, one\_group},
xlabel={top eigenpairs (\(\displaystyle k\))},
xmin=0.55, xmax=10.45,
ylabel={time [s]},
ymin=0.0428967413451367, ymax=33.6545126089106,
ymode=log,
zmystyle
]
\addplot [, color0, dashed, mark=pentagon*, mark size=3, mark options={solid}]
table {1 0.760177509859204
2 4.09210771904327
3 7.0791558760684
4 9.67988434806466
5 14.8155923979357
6 16.0011548358016
7 20.3713114329148
8 20.886346177198
9 21.5930130518973
10 24.8582020879257
};
\addlegendentry{power iteration}
\addplot [, black, dashed, mark=*, mark size=3, mark options={solid}]
table {1 0.885590265970677
2 0.908894127001986
3 0.913377678021789
4 0.912511325906962
5 0.927521636011079
6 0.922896699979901
7 0.940916499122977
8 0.933356910012662
9 0.950288704829291
10 0.945480581838638
};
\addlegendentry{mb, exact}
\addplot [, color1, dashed, mark=diamond*, mark size=3, mark options={solid}]
table {1 0.118213178124279
2 0.122155995108187
3 0.126360181951895
4 0.131313235964626
5 0.135502902092412
6 0.140353764872998
7 0.145170437870547
8 0.146590950898826
9 0.149795203004032
10 0.155959815019742
};
\addlegendentry{sub, exact}
\addplot [, color2, dashed, mark=square*, mark size=3, mark options={solid}]
table {1 0.107025705976412
2 0.114174135960639
3 0.117819188861176
4 0.121060512959957
5 0.131325802998617
6 0.134794336045161
7 0.141315412940457
8 0.144923571031541
9 0.15176796913147
10 0.154811842832714
};
\addlegendentry{mb, mc}
\addplot [, color3, dashed, mark=triangle*, mark size=3, mark options={solid,rotate=180}]
table {1 0.0580761600285769
2 0.0628426959738135
3 0.0675858769100159
4 0.0714079360477626
5 0.0752897569909692
6 0.0791625399142504
7 0.0834236771333963
8 0.0876421630382538
9 0.0916584860533476
10 0.0950959629844874
};
\addlegendentry{sub, mc}
\end{axis}

\end{tikzpicture}
   \end{minipage}
  \hfill
  \begin{minipage}[t]{0.49\linewidth}
    \centering
    \textbf{Block-diagonal approximation}
    \vspace{-3ex}

    \begin{minipage}[t]{0.49\linewidth}
      \centering

      \begin{flushleft}
        \vspace{1ex}
        (b)
      \end{flushleft}

      \vspace{-1.0\baselineskip}
      \begin{small}
        $N_{\text{crit}}$ (eigenvalues)
      \end{small}
      \vspace{0.15\baselineskip}

      \begin{small}
        \begin{tabular}{lll}
    \toprule
    $_{\text{\tiny{\ggn}}}$$^{\text{\tiny{Data}}}$ & mb & sub \\
    \midrule
    exact & 4416
              & 21453 \\
    mc   & 23723
              & > 32768 \\
    \bottomrule
\end{tabular}       \end{small}
    \end{minipage}
    \hfill
    \begin{minipage}[t]{0.49\linewidth}
      \centering

      \begin{flushleft}
        \vspace{1ex}
        \phantom{(a)}
      \end{flushleft}
      \vspace{-1.0\baselineskip}

      \begin{small}
        $N_{\text{crit}}$ (top eigenpair)
      \end{small}
      \vspace{0.15\baselineskip}

      \begin{small}
        \begin{tabular}{lll}
    \toprule
    $_{\text{\tiny{\ggn}}}$$^{\text{\tiny{Data}}}$ & mb & sub \\
    \midrule
    exact & 3276
              & 20378 \\
    mc   & 23457
              & > 32768 \\
    \bottomrule
\end{tabular}       \end{small}
    \end{minipage}

    \begin{flushleft}
      \vspace{1ex}
      (d)
    \end{flushleft}

    \vspace{-1.3\baselineskip}

\pgfkeys{/pgfplots/performancedefault/.style={
    width=1.04\linewidth,
    height=\goldenRatioInv*1.04\linewidth,
    every axis plot/.append style={line width = 1.2pt},
    every axis plot post/.append style={
      mark size=2, mark options={opacity=0.9, solid, line width = 1pt}
    },
    tick pos = left,
    xmajorticks = true,
    ymajorticks = true,
    ylabel near ticks,
    xlabel near ticks,
    xtick align = inside,
    ytick align = inside,
    legend cell align = left,
    legend columns = 3,
legend style = {
      fill opacity = 0.9,
      text opacity = 1,
      font = \small,
      at={(1, 1.025)},
      anchor=south east,
    },
    xticklabel style = {font = \small, inner xsep = 0ex},
    xlabel style = {font = \small},
    axis line style = {black},
    yticklabel style = {font = \small, inner ysep = 0ex},
    ylabel style = {font = \small, inner ysep = 0ex},
    title style = {font = \small, inner ysep = 0ex, yshift = -0.75ex},
    grid = major,
    grid style = {dashed},
    title = {},
  }
}
 \pgfkeys{/pgfplots/zmystyle/.style={performancedefault}}
    \begin{tikzpicture}

\definecolor{color0}{rgb}{0.937254901960784,0.231372549019608,0.172549019607843}
\definecolor{color1}{rgb}{0.274509803921569,0.6,0.564705882352941}
\definecolor{color2}{rgb}{0.870588235294118,0.623529411764706,0.0862745098039216}
\definecolor{color3}{rgb}{0.501960784313725,0.184313725490196,0.6}

\begin{axis}[
axis line style={white!80!black},
legend style={fill opacity=0.8, draw opacity=1, text opacity=1, at={(0.03,0.97)}, anchor=north west, draw=white!80!black},
tick pos=left,
title={fmnist\_2c2d, N=128, cpu, layerwise\_group},
xlabel={top eigenpairs (\(\displaystyle k\))},
xmin=0.55, xmax=10.45,
ylabel={time [s]},
ymin=0.0436073950617573, ymax=33.9641460722438,
ymode=log,
zmystyle
]
\addplot [, color0, dashed, mark=pentagon*, mark size=3, mark options={solid}]
table {1 1.17931268899702
2 4.43780238507316
3 6.90319510921836
4 9.32137685711496
5 13.2482134720776
6 15.1912524348591
7 17.1338205600623
8 20.000691100955
9 21.6443038480356
10 25.0952007831074
};
\addlegendentry{power iteration}
\addplot [, black, dashed, mark=*, mark size=3, mark options={solid}]
table {1 0.993757080985233
2 1.01725016906857
3 0.969855146016926
4 1.02375503093936
5 1.01738667604513
6 1.02294186805375
7 1.04676783294417
8 1.05367807485163
9 1.04395340289921
10 1.04209575802088
};
\addlegendentry{mb, exact}
\addplot [, color1, dashed, mark=diamond*, mark size=3, mark options={solid}]
table {1 0.12058909679763
2 0.125573492841795
3 0.131428444059566
4 0.134894742164761
5 0.138687375001609
6 0.143550229026005
7 0.149569579865783
8 0.152659858809784
9 0.156955083133653
10 0.159642005106434
};
\addlegendentry{sub, exact}
\addplot [, color2, dashed, mark=square*, mark size=3, mark options={solid}]
table {1 0.111478687031195
2 0.116113855969161
3 0.119950169930235
4 0.124852742999792
5 0.127372208051383
6 0.138100695097819
7 0.14323065103963
8 0.147628817008808
9 0.151176857994869
10 0.157302788924426
};
\addlegendentry{mb, mc}
\addplot [, color3, dashed, mark=triangle*, mark size=3, mark options={solid,rotate=180}]
table {1 0.0590187720954418
2 0.0629965041298419
3 0.0676103350706398
4 0.0721131549216807
5 0.0765076370444149
6 0.0802871128544211
7 0.0840252880007029
8 0.0884458560030907
9 0.0917519330978394
10 0.0963397091254592
};
\addlegendentry{sub, mc}
\end{axis}

\end{tikzpicture}
   \end{minipage}

  \vspace{-2ex}
  \caption{\textbf{CPU memory and run time performance for the \twoctwod
      architecture on \fmnist:} Left and right columns show results with the
    full network's \ggn ($D = 3,\!274,\!634$, $C=10$) and a per-layer
    block-diagonal approximation, respectively. (a, b) Critical batch sizes
    $N_{\text{crit}}$ for computing eigenvalues and the top eigenpair. (c, d)
    Run time comparison with a power iteration for extracting the $k$ leading
    eigenpairs using a mini-batch of size $N=128$.}
  \label{fig:performance-fmnist-2c2d-cpu}
\end{figure*}

\begin{figure*}[tb]
  \centering
  \begin{minipage}[t]{0.49\linewidth}
    \centering
    \textbf{Full network}
    \vspace{-3ex}

    \begin{minipage}[t]{0.49\linewidth}
      \centering

      \begin{flushleft}
        \vspace{1ex}
        (a)
      \end{flushleft}

      \vspace{-1.0\baselineskip}
      \begin{small}
        $N_{\text{crit}}$ (eigenvalues)
      \end{small}
      \vspace{0.15\baselineskip}

      \begin{small}
        \begin{tabular}{lll}
    \toprule
    $_{\text{\tiny{\ggn}}}$$^{\text{\tiny{Data}}}$ & mb & sub \\
    \midrule
    exact & 909
              & 4375 \\
    mc   & 3840
              & 6626 \\
    \bottomrule
\end{tabular}       \end{small}
    \end{minipage}
    \hfill
    \begin{minipage}[t]{0.49\linewidth}
      \centering

      \begin{flushleft}
        \vspace{1ex}
        \phantom{(a)}
      \end{flushleft}
      \vspace{-1.0\baselineskip}

      \begin{small}
        $N_{\text{crit}}$ (top eigenpair)
      \end{small}
      \vspace{0.15\baselineskip}

      \begin{small}
        \begin{tabular}{lll}
    \toprule
    $_{\text{\tiny{\ggn}}}$$^{\text{\tiny{Data}}}$ & mb & sub \\
    \midrule
    exact & 677
              & 3184 \\
    mc   & 3060
              & 6029 \\
    \bottomrule
\end{tabular}       \end{small}
    \end{minipage}

    \begin{flushleft}
      \vspace{1ex}
      (c)
    \end{flushleft}

    \vspace{-1.3\baselineskip}

\pgfkeys{/pgfplots/performancedefault/.style={
    width=1.04\linewidth,
    height=\goldenRatioInv*1.04\linewidth,
    every axis plot/.append style={line width = 1.2pt},
    every axis plot post/.append style={
      mark size=2, mark options={opacity=0.9, solid, line width = 1pt}
    },
    tick pos = left,
    xmajorticks = true,
    ymajorticks = true,
    ylabel near ticks,
    xlabel near ticks,
    xtick align = inside,
    ytick align = inside,
    legend cell align = left,
    legend columns = 3,
legend style = {
      fill opacity = 0.9,
      text opacity = 1,
      font = \small,
      at={(1, 1.025)},
      anchor=south east,
    },
    xticklabel style = {font = \small, inner xsep = 0ex},
    xlabel style = {font = \small},
    axis line style = {black},
    yticklabel style = {font = \small, inner ysep = 0ex},
    ylabel style = {font = \small, inner ysep = 0ex},
    title style = {font = \small, inner ysep = 0ex, yshift = -0.75ex},
    grid = major,
    grid style = {dashed},
    title = {},
  }
}
 \pgfkeys{/pgfplots/zmystyle/.style={performancedefault}}
    \begin{tikzpicture}

\definecolor{color0}{rgb}{0.937254901960784,0.231372549019608,0.172549019607843}
\definecolor{color1}{rgb}{0.274509803921569,0.6,0.564705882352941}
\definecolor{color2}{rgb}{0.870588235294118,0.623529411764706,0.0862745098039216}
\definecolor{color3}{rgb}{0.501960784313725,0.184313725490196,0.6}

\begin{axis}[
axis line style={white!80!black},
legend style={fill opacity=0.8, draw opacity=1, text opacity=1, at={(0.03,0.97)}, anchor=north west, draw=white!80!black},
tick pos=left,
title={cifar10\_3c3d, N=128, cuda, one\_group},
xlabel={top eigenpairs (\(\displaystyle k\))},
xmin=0.55, xmax=10.45,
ylabel={time [s]},
ymin=0.00910443848057122, ymax=1.89239674986061,
ymode=log,
zmystyle
]
\addplot [, color0, dashed, mark=pentagon*, mark size=3, mark options={solid}]
table {1 0.10646684000676
2 0.263332082999113
3 0.41296657500061
4 0.562371585998335
5 0.961040356996818
6 1.08828064100089
7 1.20301098700293
8 1.29626815899974
9 1.36217651501647
10 1.48477111599641
};
\addlegendentry{power iteration}
\addplot [, black, dashed, mark=*, mark size=3, mark options={solid}]
table {1 0.185383651005395
2 0.186310245997447
3 0.186981072998606
4 0.187526319001336
5 0.18765962299949
6 0.18803448399558
7 0.187778580999293
8 0.186061766995408
9 0.189352801004134
10 0.188033979997272
};
\addlegendentry{mb, exact}
\addplot [, color1, dashed, mark=diamond*, mark size=3, mark options={solid}]
table {1 0.0260587410011794
2 0.0255732080040616
3 0.0259397419940797
4 0.0259047899962752
5 0.0255934300002991
6 0.0258599759981735
7 0.0261660430041957
8 0.02605828599917
9 0.0260880819987506
10 0.0260066910050227
};
\addlegendentry{sub, exact}
\addplot [, color2, dashed, mark=square*, mark size=3, mark options={solid}]
table {1 0.0172249570023268
2 0.0170504369962146
3 0.017168151003716
4 0.0170781389970216
5 0.0171006899981876
6 0.0171182029953343
7 0.0170811919961125
8 0.0169726210006047
9 0.017144368001027
10 0.0171162970000296
};
\addlegendentry{mb, mc}
\addplot [, color3, dashed, mark=triangle*, mark size=3, mark options={solid,rotate=180}]
table {1 0.0127098220036714
2 0.0126403089961968
3 0.0126599400027771
4 0.0123849169976893
5 0.011915481001779
6 0.0116139689998818
7 0.0116039499989711
8 0.0117499150001095
9 0.0118982629937818
10 0.0121629770001164
};
\addlegendentry{sub, mc}
\end{axis}

\end{tikzpicture}
   \end{minipage}
  \hfill
  \begin{minipage}[t]{0.49\linewidth}
    \centering
    \textbf{Block-diagonal approximation}
    \vspace{-3ex}

    \begin{minipage}[t]{0.49\linewidth}
      \centering

      \begin{flushleft}
        \vspace{1ex}
        (b)
      \end{flushleft}

      \vspace{-1.0\baselineskip}
      \begin{small}
        $N_{\text{crit}}$ (eigenvalues)
      \end{small}
      \vspace{0.15\baselineskip}

      \begin{small}
        \begin{tabular}{lll}
    \toprule
    $_{\text{\tiny{\ggn}}}$$^{\text{\tiny{Data}}}$ & mb & sub \\
    \midrule
    exact & 1379
              & 4464 \\
    mc   & 3854
              & 6626 \\
    \bottomrule
\end{tabular}       \end{small}
    \end{minipage}
    \hfill
    \begin{minipage}[t]{0.49\linewidth}
      \centering

      \begin{flushleft}
        \vspace{1ex}
        \phantom{(a)}
      \end{flushleft}
      \vspace{-1.0\baselineskip}

      \begin{small}
        $N_{\text{crit}}$ (top eigenpair)
      \end{small}
      \vspace{0.15\baselineskip}

      \begin{small}
        \begin{tabular}{lll}
    \toprule
    $_{\text{\tiny{\ggn}}}$$^{\text{\tiny{Data}}}$ & mb & sub \\
    \midrule
    exact & 1085
              & 4415 \\
    mc   & 3854
              & 6624 \\
    \bottomrule
\end{tabular}       \end{small}
    \end{minipage}

    \begin{flushleft}
      \vspace{1ex}
      (d)
    \end{flushleft}

    \vspace{-1.3\baselineskip}

\pgfkeys{/pgfplots/performancedefault/.style={
    width=1.04\linewidth,
    height=\goldenRatioInv*1.04\linewidth,
    every axis plot/.append style={line width = 1.2pt},
    every axis plot post/.append style={
      mark size=2, mark options={opacity=0.9, solid, line width = 1pt}
    },
    tick pos = left,
    xmajorticks = true,
    ymajorticks = true,
    ylabel near ticks,
    xlabel near ticks,
    xtick align = inside,
    ytick align = inside,
    legend cell align = left,
    legend columns = 3,
legend style = {
      fill opacity = 0.9,
      text opacity = 1,
      font = \small,
      at={(1, 1.025)},
      anchor=south east,
    },
    xticklabel style = {font = \small, inner xsep = 0ex},
    xlabel style = {font = \small},
    axis line style = {black},
    yticklabel style = {font = \small, inner ysep = 0ex},
    ylabel style = {font = \small, inner ysep = 0ex},
    title style = {font = \small, inner ysep = 0ex, yshift = -0.75ex},
    grid = major,
    grid style = {dashed},
    title = {},
  }
}
 \pgfkeys{/pgfplots/zmystyle/.style={performancedefault}}
    \begin{tikzpicture}

\definecolor{color0}{rgb}{0.937254901960784,0.231372549019608,0.172549019607843}
\definecolor{color1}{rgb}{0.274509803921569,0.6,0.564705882352941}
\definecolor{color2}{rgb}{0.870588235294118,0.623529411764706,0.0862745098039216}
\definecolor{color3}{rgb}{0.501960784313725,0.184313725490196,0.6}

\begin{axis}[
axis line style={white!80!black},
legend style={fill opacity=0.8, draw opacity=1, text opacity=1, at={(0.03,0.97)}, anchor=north west, draw=white!80!black},
tick pos=left,
title={cifar10\_3c3d, N=128, cuda, layerwise\_group},
xlabel={top eigenpairs (\(\displaystyle k\))},
xmin=0.55, xmax=10.45,
ylabel={time [s]},
ymin=0.0127662167828905, ymax=3.83430292879011,
ymode=log,
zmystyle
]
\addplot [, color0, dashed, mark=pentagon*, mark size=3, mark options={solid}]
table {1 0.205712862982182
2 0.417346514004748
3 0.855886295990786
4 1.13910098798806
5 1.4351699949766
6 1.80530388798798
7 2.08452072698856
8 2.47517625699402
9 2.85886634301278
10 2.95846979197813
};
\addlegendentry{power iteration}
\addplot [, black, dashed, mark=*, mark size=3, mark options={solid}]
table {1 0.424219748005271
2 0.425042736998876
3 0.427045016003831
4 0.426793671998894
5 0.427465906002908
6 0.427657038002508
7 0.422722199997224
8 0.426844451001671
9 0.424288540998532
10 0.427101075998507
};
\addlegendentry{mb, exact}
\addplot [, color1, dashed, mark=diamond*, mark size=3, mark options={solid}]
table {1 0.0478439490034361
2 0.0479736979978043
3 0.0480268640021677
4 0.0476782599944272
5 0.0479310669979895
6 0.04799145799916
7 0.0479564880006365
8 0.0481291249961942
9 0.047855205993983
10 0.0475601929938421
};
\addlegendentry{sub, exact}
\addplot [, color2, dashed, mark=square*, mark size=3, mark options={solid}]
table {1 0.0244440180031233
2 0.0245332970007439
3 0.0241551399958553
4 0.0246412620035699
5 0.0246672599969315
6 0.0246290909999516
7 0.0245163069994305
8 0.0246254179946845
9 0.0244739060071879
10 0.0245790059998399
};
\addlegendentry{mb, mc}
\addplot [, color3, dashed, mark=triangle*, mark size=3, mark options={solid,rotate=180}]
table {1 0.0168340720047127
2 0.0165955599950394
3 0.0165455609967466
4 0.0177279749987065
5 0.0170032040041406
6 0.0178205410047667
7 0.0165819899993949
8 0.0170743950002361
9 0.0166633120024926
10 0.0166679559988552
};
\addlegendentry{sub, mc}
\end{axis}

\end{tikzpicture}
   \end{minipage}

  \vspace{-2ex}
  \caption{\textbf{GPU memory and run time performance for the \threecthreed
      architecture on \cifarten:} Left and right columns show results with the
    full network's \ggn ($D = 895,\!210$, $C=10$) and a per-layer block-diagonal
    approximation, respectively. \textbf{(a, b)} Critical batch sizes $N_{\text{crit}}$
    for computing eigenvalues and the top eigenpair. \textbf{(c, d)} Run time comparison
    with a power iteration for extracting the $k$ leading eigenpairs using a
    mini-batch of size $N=128$.}
  \label{fig:performance-cifar10-3c3d-cuda}
\end{figure*}

 \begin{figure*}[!b]
  \centering
  \begin{minipage}[t]{0.49\linewidth}
    \centering
    \textbf{Full network}
    \vspace{-3ex}

    \begin{minipage}[t]{0.49\linewidth}
      \centering

      \begin{flushleft}
        \vspace{1ex}
        (a)
      \end{flushleft}

      \vspace{-1.0\baselineskip}
      \begin{small}
        $N_{\text{crit}}$ (eigenvalues)
      \end{small}
      \vspace{0.15\baselineskip}

      \begin{small}
        \begin{tabular}{lll}
    \toprule
    $_{\text{\tiny{\ggn}}}$$^{\text{\tiny{Data}}}$ & mb & sub \\
    \midrule
    exact & 2688
              & 12768 \\
    mc   & 14330
              & 20828 \\
    \bottomrule
\end{tabular}       \end{small}
    \end{minipage}
    \hfill
    \begin{minipage}[t]{0.49\linewidth}
      \centering

      \begin{flushleft}
        \vspace{1ex}
        \phantom{(a)}
      \end{flushleft}
      \vspace{-1.0\baselineskip}

      \begin{small}
        $N_{\text{crit}}$ (top eigenpair)
      \end{small}
      \vspace{0.15\baselineskip}

      \begin{small}
        \begin{tabular}{lll}
    \toprule
    $_{\text{\tiny{\ggn}}}$$^{\text{\tiny{Data}}}$ & mb & sub \\
    \midrule
    exact & 2479
              & 11138 \\
    mc   & 11499
              & 19703 \\
    \bottomrule
\end{tabular}       \end{small}
    \end{minipage}

    \begin{flushleft}
      \vspace{1ex}
      (c)
    \end{flushleft}

    \vspace{-1.3\baselineskip}

\pgfkeys{/pgfplots/performancedefault/.style={
    width=1.04\linewidth,
    height=\goldenRatioInv*1.04\linewidth,
    every axis plot/.append style={line width = 1.2pt},
    every axis plot post/.append style={
      mark size=2, mark options={opacity=0.9, solid, line width = 1pt}
    },
    tick pos = left,
    xmajorticks = true,
    ymajorticks = true,
    ylabel near ticks,
    xlabel near ticks,
    xtick align = inside,
    ytick align = inside,
    legend cell align = left,
    legend columns = 3,
legend style = {
      fill opacity = 0.9,
      text opacity = 1,
      font = \small,
      at={(1, 1.025)},
      anchor=south east,
    },
    xticklabel style = {font = \small, inner xsep = 0ex},
    xlabel style = {font = \small},
    axis line style = {black},
    yticklabel style = {font = \small, inner ysep = 0ex},
    ylabel style = {font = \small, inner ysep = 0ex},
    title style = {font = \small, inner ysep = 0ex, yshift = -0.75ex},
    grid = major,
    grid style = {dashed},
    title = {},
  }
}
 \pgfkeys{/pgfplots/zmystyle/.style={performancedefault}}
    \begin{tikzpicture}

\definecolor{color0}{rgb}{0.937254901960784,0.231372549019608,0.172549019607843}
\definecolor{color1}{rgb}{0.274509803921569,0.6,0.564705882352941}
\definecolor{color2}{rgb}{0.870588235294118,0.623529411764706,0.0862745098039216}
\definecolor{color3}{rgb}{0.501960784313725,0.184313725490196,0.6}

\begin{axis}[
axis line style={white!80!black},
legend style={fill opacity=0.8, draw opacity=1, text opacity=1, at={(0.03,0.97)}, anchor=north west, draw=white!80!black},
tick pos=left,
title={cifar10\_3c3d, N=128, cpu, one\_group},
xlabel={top eigenpairs (\(\displaystyle k\))},
xmin=0.55, xmax=10.45,
ylabel={time [s]},
ymin=0.0634046808937379, ymax=31.5741607213845,
ymode=log,
zmystyle
]
\addplot [, color0, dashed, mark=pentagon*, mark size=3, mark options={solid}]
table {1 1.87858159909956
2 4.37033005594276
3 6.69596778205596
4 9.04794550314546
5 15.6270945928991
6 17.7007441001479
7 19.4971010941081
8 20.7641035739798
9 21.535390316043
10 23.8084618321154
};
\addlegendentry{power iteration}
\addplot [, black, dashed, mark=*, mark size=3, mark options={solid}]
table {1 1.75790749303997
2 1.79026688495651
3 1.78108807699755
4 1.78850234905258
5 1.78921697800979
6 1.79007035493851
7 1.78624628204852
8 1.79183469992131
9 1.78554186294787
10 1.78973950515501
};
\addlegendentry{mb, exact}
\addplot [, color1, dashed, mark=diamond*, mark size=3, mark options={solid}]
table {1 0.174480294808745
2 0.177483749110252
3 0.181896100984886
4 0.182066101813689
5 0.18386583798565
6 0.182626577094197
7 0.182279932079837
8 0.185124204028398
9 0.185907420003787
10 0.18588300794363
};
\addlegendentry{sub, exact}
\addplot [, color2, dashed, mark=square*, mark size=3, mark options={solid}]
table {1 0.160975076956674
2 0.163370409049094
3 0.164221629034728
4 0.168904393911362
5 0.166820840910077
6 0.169543879106641
7 0.169996321899816
8 0.170935553032905
9 0.170961502939463
10 0.172233132878318
};
\addlegendentry{mb, mc}
\addplot [, color3, dashed, mark=triangle*, mark size=3, mark options={solid,rotate=180}]
table {1 0.0840856330469251
2 0.0847525950521231
3 0.0842940059956163
4 0.0870152788702399
5 0.0875452680047601
6 0.088767166947946
7 0.0889978739432991
8 0.0915548719931394
9 0.0921855601482093
10 0.0921470001339912
};
\addlegendentry{sub, mc}
\end{axis}

\end{tikzpicture}
   \end{minipage}
  \hfill
  \begin{minipage}[t]{0.49\linewidth}
    \centering
    \textbf{Block-diagonal approximation}
    \vspace{-3ex}

    \begin{minipage}[t]{0.49\linewidth}
      \centering

      \begin{flushleft}
        \vspace{1ex}
        (b)
      \end{flushleft}

      \vspace{-1.0\baselineskip}
      \begin{small}
        $N_{\text{crit}}$ (eigenvalues)
      \end{small}
      \vspace{0.15\baselineskip}

      \begin{small}
        \begin{tabular}{lll}
    \toprule
    $_{\text{\tiny{\ggn}}}$$^{\text{\tiny{Data}}}$ & mb & sub \\
    \midrule
    exact & 2978
              & 13312 \\
    mc   & 14848
              & 20732 \\
    \bottomrule
\end{tabular}       \end{small}
    \end{minipage}
    \hfill
    \begin{minipage}[t]{0.49\linewidth}
      \centering

      \begin{flushleft}
        \vspace{1ex}
        \phantom{(a)}
      \end{flushleft}
      \vspace{-1.0\baselineskip}

      \begin{small}
        $N_{\text{crit}}$ (top eigenpair)
      \end{small}
      \vspace{0.15\baselineskip}

      \begin{small}
        \begin{tabular}{lll}
    \toprule
    $_{\text{\tiny{\ggn}}}$$^{\text{\tiny{Data}}}$ & mb & sub \\
    \midrule
    exact & 2911
              & 13186 \\
    mc   & 14656
              & 20757 \\
    \bottomrule
\end{tabular}       \end{small}
    \end{minipage}

    \begin{flushleft}
      \vspace{1ex}
      (d)
    \end{flushleft}

    \vspace{-1.3\baselineskip}

\pgfkeys{/pgfplots/performancedefault/.style={
    width=1.04\linewidth,
    height=\goldenRatioInv*1.04\linewidth,
    every axis plot/.append style={line width = 1.2pt},
    every axis plot post/.append style={
      mark size=2, mark options={opacity=0.9, solid, line width = 1pt}
    },
    tick pos = left,
    xmajorticks = true,
    ymajorticks = true,
    ylabel near ticks,
    xlabel near ticks,
    xtick align = inside,
    ytick align = inside,
    legend cell align = left,
    legend columns = 3,
legend style = {
      fill opacity = 0.9,
      text opacity = 1,
      font = \small,
      at={(1, 1.025)},
      anchor=south east,
    },
    xticklabel style = {font = \small, inner xsep = 0ex},
    xlabel style = {font = \small},
    axis line style = {black},
    yticklabel style = {font = \small, inner ysep = 0ex},
    ylabel style = {font = \small, inner ysep = 0ex},
    title style = {font = \small, inner ysep = 0ex, yshift = -0.75ex},
    grid = major,
    grid style = {dashed},
    title = {},
  }
}
 \pgfkeys{/pgfplots/zmystyle/.style={performancedefault}}
    \begin{tikzpicture}

\definecolor{color0}{rgb}{0.937254901960784,0.231372549019608,0.172549019607843}
\definecolor{color1}{rgb}{0.274509803921569,0.6,0.564705882352941}
\definecolor{color2}{rgb}{0.870588235294118,0.623529411764706,0.0862745098039216}
\definecolor{color3}{rgb}{0.501960784313725,0.184313725490196,0.6}

\begin{axis}[
axis line style={white!80!black},
legend style={fill opacity=0.8, draw opacity=1, text opacity=1, at={(0.03,0.97)}, anchor=north west, draw=white!80!black},
tick pos=left,
title={cifar10\_3c3d, N=128, cpu, layerwise\_group},
xlabel={top eigenpairs (\(\displaystyle k\))},
xmin=0.55, xmax=10.45,
ylabel={time [s]},
ymin=0.0618645003514787, ymax=50.5729839649641,
ymode=log,
zmystyle
]
\addplot [, color0, dashed, mark=pentagon*, mark size=3, mark options={solid}]
table {1 2.47491615894251
2 4.7517678369768
3 9.37998736090958
4 12.901587727014
5 16.7241862050723
6 23.2307811980136
7 25.8022972478066
8 29.1975529468618
9 32.8837849451229
10 37.2849236230832
};
\addlegendentry{power iteration}
\addplot [, black, dashed, mark=*, mark size=3, mark options={solid}]
table {1 1.82142753503285
2 1.83885124209337
3 1.85235056211241
4 1.83410064200871
5 1.84360350901261
6 1.85989074688405
7 1.84343316801824
8 1.84274471900426
9 1.85886549414136
10 1.84833160997368
};
\addlegendentry{mb, exact}
\addplot [, color1, dashed, mark=diamond*, mark size=3, mark options={solid}]
table {1 0.184518829919398
2 0.180415353970602
3 0.187068244908005
4 0.185979808913544
5 0.187472068006173
6 0.188167828135192
7 0.191501739900559
8 0.192352197831497
9 0.190203272970393
10 0.190426491899416
};
\addlegendentry{sub, exact}
\addplot [, color2, dashed, mark=square*, mark size=3, mark options={solid}]
table {1 0.163756247144192
2 0.166101312031969
3 0.167324421927333
4 0.168611278990284
5 0.169896885054186
6 0.170876645017415
7 0.171972966985777
8 0.172984215896577
9 0.173909130971879
10 0.175805911887437
};
\addlegendentry{mb, mc}
\addplot [, color3, dashed, mark=triangle*, mark size=3, mark options={solid,rotate=180}]
table {1 0.0839125330094248
2 0.0859126290306449
3 0.0862711591180414
4 0.0869505030568689
5 0.0891102550085634
6 0.0882730158045888
7 0.0897101711016148
8 0.0908182128332555
9 0.092485309112817
10 0.093585150083527
};
\addlegendentry{sub, mc}
\end{axis}

\end{tikzpicture}
   \end{minipage}

  \vspace{-2ex}
  \caption{\textbf{CPU memory and run time performance for the \threecthreed
      architecture on \cifarten:} Left and right columns show results with the
    full network's \ggn ($D = 895,\!210$, $C=10$) and a per-layer block-diagonal
    approximation, respectively. (a, b) Critical batch sizes $N_{\text{crit}}$
    for computing eigenvalues and the top eigenpair. (c, d) Run time comparison
    with a power iteration for extracting the $k$ leading eigenpairs using a
    mini-batch of size $N=128$.}
  \label{fig:performance-cifar10-3c3d-cpu}
\end{figure*}

\begin{figure*}[tb]
  \centering
  \begin{minipage}[t]{0.49\linewidth}
    \centering
    \textbf{Full network}
    \vspace{-3ex}

    \begin{minipage}[t]{0.49\linewidth}
      \centering

      \begin{flushleft}
        \vspace{1ex}
        (a)
      \end{flushleft}

      \vspace{-1.0\baselineskip}
      \begin{small}
        $N_{\text{crit}}$ (eigenvalues)
      \end{small}
      \vspace{0.15\baselineskip}

      \begin{small}
        \begin{tabular}{lll}
    \toprule
    $_{\text{\tiny{\ggn}}}$$^{\text{\tiny{Data}}}$ & mb & sub \\
    \midrule
    exact & 1054
              & 2052 \\
    mc   & 2064
              & 2271 \\
    \bottomrule
\end{tabular}       \end{small}
    \end{minipage}
    \hfill
    \begin{minipage}[t]{0.49\linewidth}
      \centering

      \begin{flushleft}
        \vspace{1ex}
        \phantom{(a)}
      \end{flushleft}
      \vspace{-1.0\baselineskip}

      \begin{small}
        $N_{\text{crit}}$ (top eigenpair)
      \end{small}
      \vspace{0.15\baselineskip}

      \begin{small}
        \begin{tabular}{lll}
    \toprule
    $_{\text{\tiny{\ggn}}}$$^{\text{\tiny{Data}}}$ & mb & sub \\
    \midrule
    exact & 364
              & 1360 \\
    mc   & 1536
              & 2176 \\
    \bottomrule
\end{tabular}       \end{small}
    \end{minipage}

    \begin{flushleft}
      \vspace{1ex}
      (c)
    \end{flushleft}

    \vspace{-1.3\baselineskip}

\pgfkeys{/pgfplots/performancedefault/.style={
    width=1.04\linewidth,
    height=\goldenRatioInv*1.04\linewidth,
    every axis plot/.append style={line width = 1.2pt},
    every axis plot post/.append style={
      mark size=2, mark options={opacity=0.9, solid, line width = 1pt}
    },
    tick pos = left,
    xmajorticks = true,
    ymajorticks = true,
    ylabel near ticks,
    xlabel near ticks,
    xtick align = inside,
    ytick align = inside,
    legend cell align = left,
    legend columns = 3,
legend style = {
      fill opacity = 0.9,
      text opacity = 1,
      font = \small,
      at={(1, 1.025)},
      anchor=south east,
    },
    xticklabel style = {font = \small, inner xsep = 0ex},
    xlabel style = {font = \small},
    axis line style = {black},
    yticklabel style = {font = \small, inner ysep = 0ex},
    ylabel style = {font = \small, inner ysep = 0ex},
    title style = {font = \small, inner ysep = 0ex, yshift = -0.75ex},
    grid = major,
    grid style = {dashed},
    title = {},
  }
}
 \pgfkeys{/pgfplots/zmystyle/.style={performancedefault}}
    \begin{tikzpicture}

\definecolor{color0}{rgb}{0.937254901960784,0.231372549019608,0.172549019607843}
\definecolor{color1}{rgb}{0.274509803921569,0.6,0.564705882352941}
\definecolor{color2}{rgb}{0.870588235294118,0.623529411764706,0.0862745098039216}
\definecolor{color3}{rgb}{0.501960784313725,0.184313725490196,0.6}

\begin{axis}[
axis line style={white!80!black},
legend style={fill opacity=0.8, draw opacity=1, text opacity=1, at={(0.03,0.97)}, anchor=north west, draw=white!80!black},
tick pos=left,
title={cifar10\_resnet32, N=128, cuda, one\_group},
xlabel={top eigenpairs (\(\displaystyle k\))},
xmin=0.55, xmax=10.45,
ylabel={time [s]},
ymin=0.0581892945661985, ymax=17.4373662749464,
ymode=log,
zmystyle
]
\addplot [, color0, dashed, mark=pentagon*, mark size=3, mark options={solid}]
table {1 0.317886747012381
2 2.78040053299628
3 3.99538965401007
4 4.95181790099014
5 5.74813608097611
6 7.20148194098147
7 9.61561046703719
8 10.50072550599
9 12.5891499459976
10 13.4557046140544
};
\addlegendentry{power iteration}
\addplot [, black, dashed, mark=*, mark size=3, mark options={solid}]
table {1 0.646277644962538
2 0.651712424994912
3 0.654213111964054
4 0.655864007014316
5 0.640351209964138
6 0.659304711036384
7 0.661320103972685
8 0.661268591007683
9 0.662068961013574
10 0.660987386014313
};
\addlegendentry{mb, exact}
\addplot [, color1, dashed, mark=diamond*, mark size=3, mark options={solid}]
table {1 0.131122773047537
2 0.131278580985963
3 0.130970751051791
4 0.130664691037964
5 0.131307907984592
6 0.131342619017232
7 0.131067821988836
8 0.13155019201804
9 0.130366104014684
10 0.131142130005173
};
\addlegendentry{sub, exact}
\addplot [, color2, dashed, mark=square*, mark size=3, mark options={solid}]
table {1 0.0901080530020408
2 0.0898521009949036
3 0.0898170699947514
4 0.0900665009976365
5 0.0898881499888375
6 0.0896032140008174
7 0.089393840986304
8 0.0897041920106858
9 0.0895591279841028
10 0.0895693419734016
};
\addlegendentry{mb, mc}
\addplot [, color3, dashed, mark=triangle*, mark size=3, mark options={solid,rotate=180}]
table {1 0.0770514360046946
2 0.0768451059702784
3 0.0754080199985765
4 0.0771796050248668
5 0.0773291389923543
6 0.0773030149866827
7 0.0774055179790594
8 0.0770304700126871
9 0.0773624330176972
10 0.0773117459611967
};
\addlegendentry{sub, mc}
\end{axis}

\end{tikzpicture}
   \end{minipage}
  \hfill
  \begin{minipage}[t]{0.49\linewidth}
    \centering
    \textbf{Block-diagonal approximation}
    \vspace{-3ex}

    \begin{minipage}[t]{0.49\linewidth}
      \centering

      \begin{flushleft}
        \vspace{1ex}
        (b)
      \end{flushleft}

      \vspace{-1.0\baselineskip}
      \begin{small}
        $N_{\text{crit}}$ (eigenvalues)
      \end{small}
      \vspace{0.15\baselineskip}

      \begin{small}
        \begin{tabular}{lll}
    \toprule
    $_{\text{\tiny{\ggn}}}$$^{\text{\tiny{Data}}}$ & mb & sub \\
    \midrule
    exact & 1061
              & 2048 \\
    mc   & 2064
              & 2271 \\
    \bottomrule
\end{tabular}       \end{small}
    \end{minipage}
    \hfill
    \begin{minipage}[t]{0.49\linewidth}
      \centering

      \begin{flushleft}
        \vspace{1ex}
        \phantom{(a)}
      \end{flushleft}
      \vspace{-1.0\baselineskip}

      \begin{small}
        $N_{\text{crit}}$ (top eigenpair)
      \end{small}
      \vspace{0.15\baselineskip}

      \begin{small}
        \begin{tabular}{lll}
    \toprule
    $_{\text{\tiny{\ggn}}}$$^{\text{\tiny{Data}}}$ & mb & sub \\
    \midrule
    exact & 1040
              & 2048 \\
    mc   & 2064
              & 2271 \\
    \bottomrule
\end{tabular}       \end{small}
    \end{minipage}

    \begin{flushleft}
      \vspace{1ex}
      (d)
    \end{flushleft}

    \vspace{-1.3\baselineskip}

\pgfkeys{/pgfplots/performancedefault/.style={
    width=1.04\linewidth,
    height=\goldenRatioInv*1.04\linewidth,
    every axis plot/.append style={line width = 1.2pt},
    every axis plot post/.append style={
      mark size=2, mark options={opacity=0.9, solid, line width = 1pt}
    },
    tick pos = left,
    xmajorticks = true,
    ymajorticks = true,
    ylabel near ticks,
    xlabel near ticks,
    xtick align = inside,
    ytick align = inside,
    legend cell align = left,
    legend columns = 3,
legend style = {
      fill opacity = 0.9,
      text opacity = 1,
      font = \small,
      at={(1, 1.025)},
      anchor=south east,
    },
    xticklabel style = {font = \small, inner xsep = 0ex},
    xlabel style = {font = \small},
    axis line style = {black},
    yticklabel style = {font = \small, inner ysep = 0ex},
    ylabel style = {font = \small, inner ysep = 0ex},
    title style = {font = \small, inner ysep = 0ex, yshift = -0.75ex},
    grid = major,
    grid style = {dashed},
    title = {},
  }
}
 \pgfkeys{/pgfplots/zmystyle/.style={performancedefault}}
    \begin{tikzpicture}

\definecolor{color0}{rgb}{0.937254901960784,0.231372549019608,0.172549019607843}
\definecolor{color1}{rgb}{0.274509803921569,0.6,0.564705882352941}
\definecolor{color2}{rgb}{0.870588235294118,0.623529411764706,0.0862745098039216}
\definecolor{color3}{rgb}{0.501960784313725,0.184313725490196,0.6}

\begin{axis}[
axis line style={white!80!black},
legend style={fill opacity=0.8, draw opacity=1, text opacity=1, at={(0.03,0.97)}, anchor=north west, draw=white!80!black},
tick pos=left,
title={cifar10\_resnet32, N=128, cuda, layerwise\_group},
xlabel={top eigenpairs (\(\displaystyle k\))},
xmin=0.55, xmax=10.45,
ylabel={time [s]},
ymin=0.091871695151768, ymax=316.722173232366,
ymode=log,
zmystyle
]
\addplot [, color0, dashed, mark=pentagon*, mark size=3, mark options={solid}]
table {1 8.46525032399222
2 27.3670097150025
3 46.3895484379609
4 65.2337697780458
5 86.9196544280276
6 110.461126691021
7 136.136389262974
8 156.849942249013
9 197.690285135992
10 218.717313289002
};
\addlegendentry{power iteration}
\addplot [, black, dashed, mark=*, mark size=3, mark options={solid}]
table {1 3.45496036403347
2 3.4472869829624
3 3.45691761100898
4 3.44431749999058
5 3.44981490401551
6 3.44834257900948
7 3.45422143995529
8 3.44626589800464
9 3.45662904798519
10 3.44807815802051
};
\addlegendentry{mb, exact}
\addplot [, color1, dashed, mark=diamond*, mark size=3, mark options={solid}]
table {1 0.359202596999239
2 0.359984207025263
3 0.356307542999275
4 0.360894134035334
5 0.360749487008434
6 0.361699934990611
7 0.360871836950537
8 0.361921988951508
9 0.358979958982673
10 0.362240851041861
};
\addlegendentry{sub, exact}
\addplot [, color2, dashed, mark=square*, mark size=3, mark options={solid}]
table {1 0.180109129985794
2 0.1773541449802
3 0.180314343015198
4 0.179963884002063
5 0.178170341008808
6 0.179666399024427
7 0.176738966023549
8 0.180119647004176
9 0.179837154981215
10 0.181404962961096
};
\addlegendentry{mb, mc}
\addplot [, color3, dashed, mark=triangle*, mark size=3, mark options={solid,rotate=180}]
table {1 0.133668831957038
2 0.135696217010263
3 0.133714537019841
4 0.13608517596731
5 0.133038407017011
6 0.134667831996921
7 0.134144404961262
8 0.134421745955478
9 0.135790712025482
10 0.133976881974377
};
\addlegendentry{sub, mc}
\end{axis}

\end{tikzpicture}
   \end{minipage}

  \vspace{-2ex}
  \caption{\textbf{GPU memory and run time performance for the ResNet32
      architecture on \cifarten:} Left and right columns show results with the
    full network's \ggn ($D = 464,\!154$, $C=10$) and a per-layer block-diagonal
    approximation, respectively. (a, b) Critical batch sizes $N_{\text{crit}}$
    for computing eigenvalues and the top eigenpair. (c, d) Run time comparison
    with a power iteration for extracting the $k$ leading eigenpairs using a
    mini-batch of size $N=128$.}
  \label{fig:performance-cifar10-resnet32-cuda}
\end{figure*}

 \begin{figure*}[tb]
  \centering
  \begin{minipage}[t]{0.49\linewidth}
    \centering
    \textbf{Full network}

    \begin{flushleft}
      \vspace{1ex}
      (c)
    \end{flushleft}

    \vspace{-1.3\baselineskip}

\pgfkeys{/pgfplots/performancedefault/.style={
    width=1.04\linewidth,
    height=\goldenRatioInv*1.04\linewidth,
    every axis plot/.append style={line width = 1.2pt},
    every axis plot post/.append style={
      mark size=2, mark options={opacity=0.9, solid, line width = 1pt}
    },
    tick pos = left,
    xmajorticks = true,
    ymajorticks = true,
    ylabel near ticks,
    xlabel near ticks,
    xtick align = inside,
    ytick align = inside,
    legend cell align = left,
    legend columns = 3,
legend style = {
      fill opacity = 0.9,
      text opacity = 1,
      font = \small,
      at={(1, 1.025)},
      anchor=south east,
    },
    xticklabel style = {font = \small, inner xsep = 0ex},
    xlabel style = {font = \small},
    axis line style = {black},
    yticklabel style = {font = \small, inner ysep = 0ex},
    ylabel style = {font = \small, inner ysep = 0ex},
    title style = {font = \small, inner ysep = 0ex, yshift = -0.75ex},
    grid = major,
    grid style = {dashed},
    title = {},
  }
}
 \pgfkeys{/pgfplots/zmystyle/.style={performancedefault}}
    \begin{tikzpicture}

\definecolor{color0}{rgb}{0.937254901960784,0.231372549019608,0.172549019607843}
\definecolor{color1}{rgb}{0.274509803921569,0.6,0.564705882352941}
\definecolor{color2}{rgb}{0.870588235294118,0.623529411764706,0.0862745098039216}
\definecolor{color3}{rgb}{0.501960784313725,0.184313725490196,0.6}

\begin{axis}[
axis line style={white!80!black},
legend style={fill opacity=0.8, draw opacity=1, text opacity=1, at={(0.03,0.97)}, anchor=north west, draw=white!80!black},
tick pos=left,
title={cifar10\_resnet32, N=128, cpu, one\_group},
xlabel={top eigenpairs (\(\displaystyle k\))},
xmin=0.55, xmax=10.45,
ylabel={time [s]},
ymin=0.322448679025841, ymax=286.722825590122,
ymode=log,
zmystyle
]
\addplot [, color0, dashed, mark=pentagon*, mark size=3, mark options={solid}]
table {1 4.69150080200052
2 36.8720023770002
3 53.4382686989848
4 65.636716287001
5 77.6254957000492
6 97.2201161999837
7 131.302255121002
8 142.576093982032
9 170.970027567993
10 210.579813935969
};
\addlegendentry{power iteration}
\addplot [, black, dashed, mark=*, mark size=3, mark options={solid}]
table {1 9.69427862798329
2 9.76183838100405
3 9.77180047798902
4 9.73586596298264
5 9.74195547198178
6 9.69951318402309
7 9.72692853299668
8 9.78560261899838
9 9.75923570297891
10 9.86185442603892
};
\addlegendentry{mb, exact}
\addplot [, color1, dashed, mark=diamond*, mark size=3, mark options={solid}]
table {1 1.16867451497819
2 1.1692463660147
3 1.17600206600036
4 1.17554128804477
5 1.17515433300287
6 1.17705015995307
7 1.17492631997447
8 1.17266865301644
9 1.17465022497345
10 1.1756574040046
};
\addlegendentry{sub, exact}
\addplot [, color2, dashed, mark=square*, mark size=3, mark options={solid}]
table {1 1.0103122050059
2 1.01216759899398
3 1.01439721503993
4 1.01509648602223
5 1.01313028804725
6 1.01610854204046
7 1.01340901997173
8 1.01141302497126
9 1.01093772199238
10 1.01403657399351
};
\addlegendentry{mb, mc}
\addplot [, color3, dashed, mark=triangle*, mark size=3, mark options={solid,rotate=180}]
table {1 0.439042065001559
2 0.441753674007487
3 0.441901749989484
4 0.440927318995818
5 0.445356941025238
6 0.441066055034753
7 0.442289864004124
8 0.442448605026584
9 0.442854888970032
10 0.442922357993666
};
\addlegendentry{sub, mc}
\end{axis}

\end{tikzpicture}
   \end{minipage}
  \hfill
  \begin{minipage}[t]{0.49\linewidth}
    \centering
    \textbf{Block-diagonal approximation}

    \begin{flushleft}
      \vspace{1ex}
      (d)
    \end{flushleft}

    \vspace{-1.3\baselineskip}

\pgfkeys{/pgfplots/performancedefault/.style={
    width=1.04\linewidth,
    height=\goldenRatioInv*1.04\linewidth,
    every axis plot/.append style={line width = 1.2pt},
    every axis plot post/.append style={
      mark size=2, mark options={opacity=0.9, solid, line width = 1pt}
    },
    tick pos = left,
    xmajorticks = true,
    ymajorticks = true,
    ylabel near ticks,
    xlabel near ticks,
    xtick align = inside,
    ytick align = inside,
    legend cell align = left,
    legend columns = 3,
legend style = {
      fill opacity = 0.9,
      text opacity = 1,
      font = \small,
      at={(1, 1.025)},
      anchor=south east,
    },
    xticklabel style = {font = \small, inner xsep = 0ex},
    xlabel style = {font = \small},
    axis line style = {black},
    yticklabel style = {font = \small, inner ysep = 0ex},
    ylabel style = {font = \small, inner ysep = 0ex},
    title style = {font = \small, inner ysep = 0ex, yshift = -0.75ex},
    grid = major,
    grid style = {dashed},
    title = {},
  }
}
 \pgfkeys{/pgfplots/zmystyle/.style={performancedefault}}
    \begin{tikzpicture}

\definecolor{color0}{rgb}{0.937254901960784,0.231372549019608,0.172549019607843}
\definecolor{color1}{rgb}{0.274509803921569,0.6,0.564705882352941}
\definecolor{color2}{rgb}{0.870588235294118,0.623529411764706,0.0862745098039216}
\definecolor{color3}{rgb}{0.501960784313725,0.184313725490196,0.6}

\begin{axis}[
axis line style={white!80!black},
legend style={fill opacity=0.8, draw opacity=1, text opacity=1, at={(0.03,0.97)}, anchor=north west, draw=white!80!black},
tick pos=left,
title={cifar10\_resnet32, N=128, cpu, layerwise\_group},
xlabel={top eigenpairs (\(\displaystyle k\))},
xmin=0.55, xmax=10.45,
ylabel={time [s]},
ymin=0.298585075257221, ymax=3487.72377788741,
ymode=log,
zmystyle
]
\addplot [, color0, dashed, mark=pentagon*, mark size=3, mark options={solid}]
table {1 82.239523777971
2 270.908339895948
3 484.327261643
4 644.810126798002
5 852.327436160002
6 1123.985210399
7 1348.653315286
8 1644.00099218001
9 1935.62427283701
10 2278.54127977599
};
\addlegendentry{power iteration}
\addplot [, black, dashed, mark=*, mark size=3, mark options={solid}]
table {1 17.6473416539957
2 17.5117687029997
3 17.4497449320043
4 17.5625249170116
5 17.6918856290285
6 17.6068161940202
7 17.5850780210458
8 17.6808727129828
9 17.643008020008
10 17.6491284200456
};
\addlegendentry{mb, exact}
\addplot [, color1, dashed, mark=diamond*, mark size=3, mark options={solid}]
table {1 1.29750978096854
2 1.30240013595903
3 1.30101109098177
4 1.30367434996879
5 1.30371632799506
6 1.30508769798325
7 1.29983986000298
8 1.29909317899728
9 1.29982969904086
10 1.30457733600633
};
\addlegendentry{sub, exact}
\addplot [, color2, dashed, mark=square*, mark size=3, mark options={solid}]
table {1 1.08640187000856
2 1.09759522695094
3 1.09323001600569
4 1.09653249400435
5 1.09533179097343
6 1.09786832204554
7 1.09042823000345
8 1.09402726194821
9 1.09787571796915
10 1.09742126398487
};
\addlegendentry{mb, mc}
\addplot [, color3, dashed, mark=triangle*, mark size=3, mark options={solid,rotate=180}]
table {1 0.45824151096167
2 0.457039016997442
3 0.459355822997168
4 0.458698283997364
5 0.460072189976927
6 0.460437513014767
7 0.463227465981618
8 0.462044437008444
9 0.462066260981373
10 0.460934479953721
};
\addlegendentry{sub, mc}
\end{axis}

\end{tikzpicture}
   \end{minipage}

  \vspace{-2ex}
  \caption{\textbf{CPU memory and run time performance for the ResNet32
      architecture on \cifarten:} Left and right columns show results with the
    full network's \ggn ($D = 464,\!154$, $C=10$) and a per-layer block-diagonal
    approximation, respectively.
(c, d) Run time comparison
    with a power iteration for extracting the $k$ leading eigenpairs using a
    mini-batch of size $N=128$.}
  \label{fig:performance-cifar10-resnet32-cpu}
\end{figure*}

\begin{figure*}[tb]
  \centering
  \begin{minipage}[t]{0.49\linewidth}
    \centering
    \textbf{Full network}
    \vspace{-3ex}

    \begin{minipage}[t]{0.49\linewidth}
      \centering

      \begin{flushleft}
        \vspace{1ex}
        (a)
      \end{flushleft}

      \vspace{-1.0\baselineskip}
      \begin{small}
        $N_{\text{crit}}$ (eigenvalues)
      \end{small}
      \vspace{0.15\baselineskip}

      \begin{small}
        \begin{tabular}{lll}
    \toprule
    $_{\text{\tiny{\ggn}}}$$^{\text{\tiny{Data}}}$ & mb & sub \\
    \midrule
    exact & 837
              & 1247 \\
    mc   & 1262
              & 1312 \\
    \bottomrule
\end{tabular}       \end{small}
    \end{minipage}
    \hfill
    \begin{minipage}[t]{0.49\linewidth}
      \centering

      \begin{flushleft}
        \vspace{1ex}
        \phantom{(a)}
      \end{flushleft}
      \vspace{-1.0\baselineskip}

      \begin{small}
        $N_{\text{crit}}$ (top eigenpair)
      \end{small}
      \vspace{0.15\baselineskip}

      \begin{small}
        \begin{tabular}{lll}
    \toprule
    $_{\text{\tiny{\ggn}}}$$^{\text{\tiny{Data}}}$ & mb & sub \\
    \midrule
    exact & 217
              & 765 \\
    mc   & 896
              & 1240 \\
    \bottomrule
\end{tabular}       \end{small}
    \end{minipage}

    \begin{flushleft}
      \vspace{1ex}
      (c)
    \end{flushleft}

    \vspace{-1.3\baselineskip}

\pgfkeys{/pgfplots/performancedefault/.style={
    width=1.04\linewidth,
    height=\goldenRatioInv*1.04\linewidth,
    every axis plot/.append style={line width = 1.2pt},
    every axis plot post/.append style={
      mark size=2, mark options={opacity=0.9, solid, line width = 1pt}
    },
    tick pos = left,
    xmajorticks = true,
    ymajorticks = true,
    ylabel near ticks,
    xlabel near ticks,
    xtick align = inside,
    ytick align = inside,
    legend cell align = left,
    legend columns = 3,
legend style = {
      fill opacity = 0.9,
      text opacity = 1,
      font = \small,
      at={(1, 1.025)},
      anchor=south east,
    },
    xticklabel style = {font = \small, inner xsep = 0ex},
    xlabel style = {font = \small},
    axis line style = {black},
    yticklabel style = {font = \small, inner ysep = 0ex},
    ylabel style = {font = \small, inner ysep = 0ex},
    title style = {font = \small, inner ysep = 0ex, yshift = -0.75ex},
    grid = major,
    grid style = {dashed},
    title = {},
  }
}
 \pgfkeys{/pgfplots/zmystyle/.style={performancedefault}}
    \begin{tikzpicture}

\definecolor{color0}{rgb}{0.937254901960784,0.231372549019608,0.172549019607843}
\definecolor{color1}{rgb}{0.274509803921569,0.6,0.564705882352941}
\definecolor{color2}{rgb}{0.870588235294118,0.623529411764706,0.0862745098039216}
\definecolor{color3}{rgb}{0.501960784313725,0.184313725490196,0.6}

\begin{axis}[
axis line style={white!80!black},
legend style={fill opacity=0.8, draw opacity=1, text opacity=1, at={(0.03,0.97)}, anchor=north west, draw=white!80!black},
tick pos=left,
title={cifar10\_resnet56, N=128, cuda, one\_group},
xlabel={top eigenpairs (\(\displaystyle k\))},
xmin=0.55, xmax=10.45,
ylabel={time [s]},
ymin=0.102998284529591, ymax=28.5178682639293,
ymode=log,
zmystyle
]
\addplot [, color0, dashed, mark=pentagon*, mark size=3, mark options={solid}]
table {1 0.567231939989142
2 1.32513407903025
3 3.00365596899064
4 5.58835422497941
5 7.71582919498906
6 10.6219418949913
7 13.8665194219793
8 16.4367114219931
9 19.023253086023
10 22.0853359549656
};
\addlegendentry{power iteration}
\addplot [, black, dashed, mark=*, mark size=3, mark options={solid}]
table {1 1.14294771297136
2 1.15610786399338
3 1.15863115398679
4 1.16373479203321
5 1.16933552897535
6 1.16682839399436
7 1.16421735601034
8 1.17255795799429
9 1.17344609496649
10 1.17550399003085
};
\addlegendentry{mb, exact}
\addplot [, color1, dashed, mark=diamond*, mark size=3, mark options={solid}]
table {1 0.22930659604026
2 0.228953950980213
3 0.228651682031341
4 0.229910403024405
5 0.229112853994593
6 0.229568459035363
7 0.22877453797264
8 0.228081217035651
9 0.229358198994305
10 0.228330979007296
};
\addlegendentry{sub, exact}
\addplot [, color2, dashed, mark=square*, mark size=3, mark options={solid}]
table {1 0.158681350003462
2 0.158850944018923
3 0.157865513989236
4 0.157779189001303
5 0.157393634028267
6 0.157388880033977
7 0.157099967007525
8 0.156777521013282
9 0.156840430980083
10 0.156645099981688
};
\addlegendentry{mb, mc}
\addplot [, color3, dashed, mark=triangle*, mark size=3, mark options={solid,rotate=180}]
table {1 0.134739618981257
2 0.135305938019883
3 0.13399555295473
4 0.132997366017662
5 0.134802999033127
6 0.135520600015298
7 0.135634100995958
8 0.135561620991211
9 0.135482086974662
10 0.133431055990513
};
\addlegendentry{sub, mc}
\end{axis}

\end{tikzpicture}
   \end{minipage}
  \hfill
  \begin{minipage}[t]{0.49\linewidth}
    \centering
    \textbf{Block-diagonal approximation}
    \vspace{-3ex}

    \begin{minipage}[t]{0.49\linewidth}
      \centering

      \begin{flushleft}
        \vspace{1ex}
        (b)
      \end{flushleft}

      \vspace{-1.0\baselineskip}
      \begin{small}
        $N_{\text{crit}}$ (eigenvalues)
      \end{small}
      \vspace{0.15\baselineskip}

      \begin{small}
        \begin{tabular}{lll}
    \toprule
    $_{\text{\tiny{\ggn}}}$$^{\text{\tiny{Data}}}$ & mb & sub \\
    \midrule
    exact & 688
              & 1247 \\
    mc   & 1232
              & 1259 \\
    \bottomrule
\end{tabular}       \end{small}
    \end{minipage}
    \hfill
    \begin{minipage}[t]{0.49\linewidth}
      \centering

      \begin{flushleft}
        \vspace{1ex}
        \phantom{(a)}
      \end{flushleft}
      \vspace{-1.0\baselineskip}

      \begin{small}
        $N_{\text{crit}}$ (top eigenpair)
      \end{small}
      \vspace{0.15\baselineskip}

      \begin{small}
        \begin{tabular}{lll}
    \toprule
    $_{\text{\tiny{\ggn}}}$$^{\text{\tiny{Data}}}$ & mb & sub \\
    \midrule
    exact & 383
              & 1247 \\
    mc   & 1232
              & 1255 \\
    \bottomrule
\end{tabular}       \end{small}
    \end{minipage}

    \begin{flushleft}
      \vspace{1ex}
      (d)
    \end{flushleft}

    \vspace{-1.3\baselineskip}

\pgfkeys{/pgfplots/performancedefault/.style={
    width=1.04\linewidth,
    height=\goldenRatioInv*1.04\linewidth,
    every axis plot/.append style={line width = 1.2pt},
    every axis plot post/.append style={
      mark size=2, mark options={opacity=0.9, solid, line width = 1pt}
    },
    tick pos = left,
    xmajorticks = true,
    ymajorticks = true,
    ylabel near ticks,
    xlabel near ticks,
    xtick align = inside,
    ytick align = inside,
    legend cell align = left,
    legend columns = 3,
legend style = {
      fill opacity = 0.9,
      text opacity = 1,
      font = \small,
      at={(1, 1.025)},
      anchor=south east,
    },
    xticklabel style = {font = \small, inner xsep = 0ex},
    xlabel style = {font = \small},
    axis line style = {black},
    yticklabel style = {font = \small, inner ysep = 0ex},
    ylabel style = {font = \small, inner ysep = 0ex},
    title style = {font = \small, inner ysep = 0ex, yshift = -0.75ex},
    grid = major,
    grid style = {dashed},
    title = {},
  }
}
 \pgfkeys{/pgfplots/zmystyle/.style={performancedefault}}
    \begin{tikzpicture}

\definecolor{color0}{rgb}{0.937254901960784,0.231372549019608,0.172549019607843}
\definecolor{color1}{rgb}{0.274509803921569,0.6,0.564705882352941}
\definecolor{color2}{rgb}{0.870588235294118,0.623529411764706,0.0862745098039216}
\definecolor{color3}{rgb}{0.501960784313725,0.184313725490196,0.6}

\begin{axis}[
axis line style={white!80!black},
legend style={fill opacity=0.8, draw opacity=1, text opacity=1, at={(0.03,0.97)}, anchor=north west, draw=white!80!black},
tick pos=left,
title={cifar10\_resnet56, N=128, cuda, layerwise\_group},
xlabel={top eigenpairs (\(\displaystyle k\))},
xmin=0.55, xmax=10.45,
ylabel={time [s]},
ymin=0.156500145423357, ymax=1053.50274131747,
ymode=log,
zmystyle
]
\addplot [, color0, dashed, mark=pentagon*, mark size=3, mark options={solid}]
table {1 35.9465068150312
2 84.3593857569504
3 146.126302307006
4 208.232056976005
5 281.931824769999
6 351.736283422972
7 440.110202115029
8 530.698788636946
9 631.689302489045
10 705.716340554995
};
\addlegendentry{power iteration}
\addplot [, black, dashed, mark=*, mark size=3, mark options={solid}]
table {1 12.8452389099984
2 12.7850116339978
3 12.8436084350105
4 12.7580787240295
5 12.8455084349844
6 12.8450681219692
7 12.8454774019774
8 12.8485926609719
9 12.8370250520529
10 12.8516433629557
};
\addlegendentry{mb, exact}
\addplot [, color1, dashed, mark=diamond*, mark size=3, mark options={solid}]
table {1 0.661565309972502
2 0.662593260989524
3 0.658750811999198
4 0.661287554947194
5 0.659255231963471
6 0.66349258099217
7 0.664640838978812
8 0.663353624986485
9 0.663450749008916
10 0.657783497998025
};
\addlegendentry{sub, exact}
\addplot [, color2, dashed, mark=square*, mark size=3, mark options={solid}]
table {1 0.317982607986778
2 0.317705555993598
3 0.318392805987969
4 0.318006807996426
5 0.320627753040753
6 0.318880411970895
7 0.317771400033962
8 0.318173856008798
9 0.316047711996362
10 0.319348148012068
};
\addlegendentry{mb, mc}
\addplot [, color3, dashed, mark=triangle*, mark size=3, mark options={solid,rotate=180}]
table {1 0.237116579024587
2 0.233625499007758
3 0.235234628024045
4 0.237652326992247
5 0.238258454948664
6 0.234770693990868
7 0.239681356004439
8 0.238967239973135
9 0.236119278008118
10 0.237969779991545
};
\addlegendentry{sub, mc}
\end{axis}

\end{tikzpicture}
   \end{minipage}

  \vspace{-2ex}
  \caption{\textbf{GPU memory and run time performance for the ResNet56
      architecture on \cifarten:} Left and right columns show results with the
    full network's \ggn ($D = 853,\!018$, $C=10$) and a per-layer block-diagonal
    approximation, respectively. (a, b) Critical batch sizes $N_{\text{crit}}$
    for computing eigenvalues and the top eigenpair. (c, d) Run time comparison
    with a power iteration for extracting the $k$ leading eigenpairs using a
    mini-batch of size $N=128$.}
  \label{fig:performance-cifar10-resnet56-cuda}
\end{figure*}

 \begin{figure*}[tb]
  \centering
  \begin{minipage}[t]{0.49\linewidth}
    \centering
    \textbf{Full network}

    \begin{flushleft}
      \vspace{1ex}
      (c)
    \end{flushleft}

    \vspace{-1.3\baselineskip}

\pgfkeys{/pgfplots/performancedefault/.style={
    width=1.04\linewidth,
    height=\goldenRatioInv*1.04\linewidth,
    every axis plot/.append style={line width = 1.2pt},
    every axis plot post/.append style={
      mark size=2, mark options={opacity=0.9, solid, line width = 1pt}
    },
    tick pos = left,
    xmajorticks = true,
    ymajorticks = true,
    ylabel near ticks,
    xlabel near ticks,
    xtick align = inside,
    ytick align = inside,
    legend cell align = left,
    legend columns = 3,
legend style = {
      fill opacity = 0.9,
      text opacity = 1,
      font = \small,
      at={(1, 1.025)},
      anchor=south east,
    },
    xticklabel style = {font = \small, inner xsep = 0ex},
    xlabel style = {font = \small},
    axis line style = {black},
    yticklabel style = {font = \small, inner ysep = 0ex},
    ylabel style = {font = \small, inner ysep = 0ex},
    title style = {font = \small, inner ysep = 0ex, yshift = -0.75ex},
    grid = major,
    grid style = {dashed},
    title = {},
  }
}
 \pgfkeys{/pgfplots/zmystyle/.style={performancedefault}}
    \begin{tikzpicture}

\definecolor{color0}{rgb}{0.937254901960784,0.231372549019608,0.172549019607843}
\definecolor{color1}{rgb}{0.274509803921569,0.6,0.564705882352941}
\definecolor{color2}{rgb}{0.870588235294118,0.623529411764706,0.0862745098039216}
\definecolor{color3}{rgb}{0.501960784313725,0.184313725490196,0.6}

\begin{axis}[
axis line style={white!80!black},
legend style={fill opacity=0.8, draw opacity=1, text opacity=1, at={(0.03,0.97)}, anchor=north west, draw=white!80!black},
tick pos=left,
title={cifar10\_resnet56, N=128, cpu, one\_group},
xlabel={top eigenpairs (\(\displaystyle k\))},
xmin=0.55, xmax=10.45,
ylabel={time [s]},
ymin=0.572407282573814, ymax=401.189565066433,
ymode=log,
zmystyle
]
\addplot [, color0, dashed, mark=pentagon*, mark size=3, mark options={solid}]
table {1 8.35777472300106
2 18.6413801919844
3 41.4093861880247
4 76.333671001019
5 106.744450980012
6 146.414122062008
7 188.386037142016
8 222.413451536995
9 256.507766285009
10 297.853119009
};
\addlegendentry{power iteration}
\addplot [, black, dashed, mark=*, mark size=3, mark options={solid}]
table {1 29.1653703899938
2 28.8747609449783
3 29.1059129119967
4 29.1530069530127
5 29.1820225879783
6 29.2213417749736
7 28.9870188800269
8 28.8675321909832
9 28.9813199110213
10 29.2272541190032
};
\addlegendentry{mb, exact}
\addplot [, color1, dashed, mark=diamond*, mark size=3, mark options={solid}]
table {1 2.13776933902409
2 2.14839096099604
3 2.14504697103985
4 2.14613211801043
5 2.14565351599595
6 2.14823607704602
7 2.14798551303102
8 2.14914905501064
9 2.15066661097808
10 2.14978381298715
};
\addlegendentry{sub, exact}
\addplot [, color2, dashed, mark=square*, mark size=3, mark options={solid}]
table {1 1.78197495598579
2 1.78938027698314
3 1.78942982602166
4 1.78973478102125
5 1.79528750898317
6 1.79466987302294
7 1.78695243701804
8 1.79162195295794
9 1.78782479400979
10 1.78907388896914
};
\addlegendentry{mb, mc}
\addplot [, color3, dashed, mark=triangle*, mark size=3, mark options={solid,rotate=180}]
table {1 0.773117487959098
2 0.770996891020332
3 0.779191085952334
4 0.779337409010623
5 0.781925552990288
6 0.775649828021415
7 0.777794838999398
8 0.7788330540061
9 0.781085689028259
10 0.782178880006541
};
\addlegendentry{sub, mc}
\end{axis}

\end{tikzpicture}
   \end{minipage}
  \hfill
  \begin{minipage}[t]{0.49\linewidth}
    \centering
    \textbf{Block-diagonal approximation}

    \begin{flushleft}
      \vspace{1ex}
      (d)
    \end{flushleft}

    \vspace{-1.3\baselineskip}

\pgfkeys{/pgfplots/performancedefault/.style={
    width=1.04\linewidth,
    height=\goldenRatioInv*1.04\linewidth,
    every axis plot/.append style={line width = 1.2pt},
    every axis plot post/.append style={
      mark size=2, mark options={opacity=0.9, solid, line width = 1pt}
    },
    tick pos = left,
    xmajorticks = true,
    ymajorticks = true,
    ylabel near ticks,
    xlabel near ticks,
    xtick align = inside,
    ytick align = inside,
    legend cell align = left,
    legend columns = 3,
legend style = {
      fill opacity = 0.9,
      text opacity = 1,
      font = \small,
      at={(1, 1.025)},
      anchor=south east,
    },
    xticklabel style = {font = \small, inner xsep = 0ex},
    xlabel style = {font = \small},
    axis line style = {black},
    yticklabel style = {font = \small, inner ysep = 0ex},
    ylabel style = {font = \small, inner ysep = 0ex},
    title style = {font = \small, inner ysep = 0ex, yshift = -0.75ex},
    grid = major,
    grid style = {dashed},
    title = {},
  }
}
 \pgfkeys{/pgfplots/zmystyle/.style={performancedefault}}
    \begin{tikzpicture}

\definecolor{color0}{rgb}{0.937254901960784,0.231372549019608,0.172549019607843}
\definecolor{color1}{rgb}{0.274509803921569,0.6,0.564705882352941}
\definecolor{color2}{rgb}{0.870588235294118,0.623529411764706,0.0862745098039216}
\definecolor{color3}{rgb}{0.501960784313725,0.184313725490196,0.6}

\begin{axis}[
axis line style={white!80!black},
legend style={fill opacity=0.8, draw opacity=1, text opacity=1, draw=white!80!black},
tick pos=left,
title={cifar10\_resnet56, N=128, cpu, layerwise\_group},
xlabel={top eigenpairs (\(\displaystyle k\))},
xmin=0.55, xmax=10.45,
ylabel={time [s]},
ymin=0.538042656384927, ymax=3089.40343748281,
ymode=log,
zmystyle
]
\addplot [, color0, dashed, mark=pentagon*, mark size=3, mark options={solid}]
table {1 367.807169881999
2 865.484152641002
3 1425.53856814501
4 2084.53099218599
};
\addlegendentry{power iteration}
\addplot [, black, dashed, mark=*, mark size=3, mark options={solid}]
table {1 55.2219943170203
2 55.4106552819721
3 55.4422056429903
4 55.3368064350216
5 55.3026694619912
6 55.6052085760166
7 55.4318958160002
8 55.2706480870256
9 55.2320651979535
10 55.4164835099946
};
\addlegendentry{mb, exact}
\addplot [, color1, dashed, mark=diamond*, mark size=3, mark options={solid}]
table {1 2.4204149650177
2 2.42182243202114
3 2.41888933698647
4 2.42889962799381
5 2.42284238501452
6 2.42879462597193
7 2.43069233500864
8 2.4292853269726
9 2.42816939100157
10 2.42926492501283
};
\addlegendentry{sub, exact}
\addplot [, color2, dashed, mark=square*, mark size=3, mark options={solid}]
table {1 1.96913653495722
2 1.96856531599769
3 1.9719867130043
4 1.97184026800096
5 1.97333031601738
6 1.97629103303188
7 1.97658809903078
8 1.96919273101958
9 1.97212190798018
10 1.97545947204344
};
\addlegendentry{mb, mc}
\addplot [, color3, dashed, mark=triangle*, mark size=3, mark options={solid,rotate=180}]
table {1 0.804750907002017
2 0.797412385989446
3 0.80686875001993
4 0.808035249996465
5 0.811571111960802
6 0.808792640978936
7 0.809119220008142
8 0.810877898009494
9 0.81188109703362
10 0.813371626951266
};
\addlegendentry{sub, mc}
\end{axis}

\end{tikzpicture}
   \end{minipage}

  \vspace{-2ex}
  \caption{\textbf{CPU memory and run time performance for the ResNet56
      architecture on \cifarten:} Left and right columns show results with the
    full network's \ggn ($D = 853,\!018$, $C=10$) and a per-layer block-diagonal
    approximation, respectively.
(c, d) Run time comparison with a power iteration for extracting the $k$
    leading eigenpairs using a mini-batch of size $N=128$.}
  \label{fig:performance-cifar10-resnet56-cpu}
\end{figure*}

\begin{figure*}[tb]
  \centering
  \begin{minipage}[t]{0.49\linewidth}
    \centering
    \textbf{Full network}
    \vspace{-3ex}

    \begin{minipage}[t]{0.49\linewidth}
      \centering

      \begin{flushleft}
        \vspace{1ex}
        (a)
      \end{flushleft}

      \vspace{-1.0\baselineskip}
      \begin{small}
        $N_{\text{crit}}$ (eigenvalues)
      \end{small}
      \vspace{0.15\baselineskip}

      \begin{small}
        \begin{tabular}{lll}
    \toprule
    $_{\text{\tiny{\ggn}}}$$^{\text{\tiny{Data}}}$ & mb & sub \\
    \midrule
    exact & 35
              & 255 \\
    mc   & 1119
              & 1536 \\
    \bottomrule
\end{tabular}       \end{small}
    \end{minipage}
    \hfill
    \begin{minipage}[t]{0.49\linewidth}
      \centering

      \begin{flushleft}
        \vspace{1ex}
        \phantom{(a)}
      \end{flushleft}
      \vspace{-1.0\baselineskip}

      \begin{small}
        $N_{\text{crit}}$ (top eigenpair)
      \end{small}
      \vspace{0.15\baselineskip}

      \begin{small}
        \begin{tabular}{lll}
    \toprule
    $_{\text{\tiny{\ggn}}}$$^{\text{\tiny{Data}}}$ & mb & sub \\
    \midrule
    exact & 14
              & 111 \\
    mc   & 745
              & 1402 \\
    \bottomrule
\end{tabular}       \end{small}
    \end{minipage}

    \begin{flushleft}
      \vspace{1ex}
      (c)
    \end{flushleft}

    \vspace{-1.3\baselineskip}

\pgfkeys{/pgfplots/performancedefault/.style={
    width=1.04\linewidth,
    height=\goldenRatioInv*1.04\linewidth,
    every axis plot/.append style={line width = 1.2pt},
    every axis plot post/.append style={
      mark size=2, mark options={opacity=0.9, solid, line width = 1pt}
    },
    tick pos = left,
    xmajorticks = true,
    ymajorticks = true,
    ylabel near ticks,
    xlabel near ticks,
    xtick align = inside,
    ytick align = inside,
    legend cell align = left,
    legend columns = 3,
legend style = {
      fill opacity = 0.9,
      text opacity = 1,
      font = \small,
      at={(1, 1.025)},
      anchor=south east,
    },
    xticklabel style = {font = \small, inner xsep = 0ex},
    xlabel style = {font = \small},
    axis line style = {black},
    yticklabel style = {font = \small, inner ysep = 0ex},
    ylabel style = {font = \small, inner ysep = 0ex},
    title style = {font = \small, inner ysep = 0ex, yshift = -0.75ex},
    grid = major,
    grid style = {dashed},
    title = {},
  }
}
 \pgfkeys{/pgfplots/zmystyle/.style={performancedefault}}
    \begin{tikzpicture}

\definecolor{color0}{rgb}{0.937254901960784,0.231372549019608,0.172549019607843}
\definecolor{color1}{rgb}{0.274509803921569,0.6,0.564705882352941}
\definecolor{color2}{rgb}{0.870588235294118,0.623529411764706,0.0862745098039216}
\definecolor{color3}{rgb}{0.501960784313725,0.184313725490196,0.6}

\begin{axis}[
axis line style={white!80!black},
legend style={fill opacity=0.8, draw opacity=1, text opacity=1, at={(0.03,0.97)}, anchor=north west, draw=white!80!black},
tick pos=left,
title={cifar100\_allcnnc, N=64, cuda, one\_group},
xlabel={top eigenpairs (\(\displaystyle k\))},
xmin=0.55, xmax=10.45,
ylabel={time [s]},
ymin=0.0230255965505877, ymax=14.6918532017484,
ymode=log,
zmystyle
]
\addplot [, color0, dashed, mark=pentagon*, mark size=3, mark options={solid}]
table {1 0.886074103007559
2 2.02152736901189
3 3.37827011797344
4 3.91675637400476
5 5.6643504719832
6 6.52234170999145
7 7.16257282497827
8 9.24401111400221
9 10.1827108059952
10 10.9542509349994
};
\addlegendentry{power iteration}
\addplot [, color1, dashed, mark=diamond*, mark size=3, mark options={solid}]
table {1 1.14568673400208
2 1.15923632399063
3 1.16384675900918
4 1.17219515002216
5 1.17467796101118
6 1.17597774000023
7 1.17535984099959
8 1.17781126400223
9 1.18129663201398
10 1.18212396098534
};
\addlegendentry{sub, exact}
\addplot [, color2, dashed, mark=square*, mark size=3, mark options={solid}]
table {1 0.088385869981721
2 0.0892419499868993
3 0.0891673169971909
4 0.0894071240036283
5 0.089427873987006
6 0.0895107639953494
7 0.089238487998955
8 0.0895071359991562
9 0.0897746040136553
10 0.0892908759997226
};
\addlegendentry{mb, mc}
\addplot [, color3, dashed, mark=triangle*, mark size=3, mark options={solid,rotate=180}]
table {1 0.0311014159815386
2 0.0311253659892827
3 0.0309164009813685
4 0.0309275010076817
5 0.0310676799854264
6 0.0310028200037777
7 0.0309121290047187
8 0.0308819549973123
};
\addlegendentry{sub, mc}
\end{axis}

\end{tikzpicture}
   \end{minipage}
  \hfill
  \begin{minipage}[t]{0.49\linewidth}
    \centering
    \textbf{Block-diagonal approximation}
    \vspace{-3ex}

    \begin{minipage}[t]{0.49\linewidth}
      \centering

      \begin{flushleft}
        \vspace{1ex}
        (b)
      \end{flushleft}

      \vspace{-1.0\baselineskip}
      \begin{small}
        $N_{\text{crit}}$ (eigenvalues)
      \end{small}
      \vspace{0.15\baselineskip}

      \begin{small}
        \begin{tabular}{lll}
    \toprule
    $_{\text{\tiny{\ggn}}}$$^{\text{\tiny{Data}}}$ & mb & sub \\
    \midrule
    exact & 36
              & 256 \\
    mc   & 1119
              & 1536 \\
    \bottomrule
\end{tabular}       \end{small}
    \end{minipage}
    \hfill
    \begin{minipage}[t]{0.49\linewidth}
      \centering

      \begin{flushleft}
        \vspace{1ex}
        \phantom{(a)}
      \end{flushleft}
      \vspace{-1.0\baselineskip}

      \begin{small}
        $N_{\text{crit}}$ (top eigenpair)
      \end{small}
      \vspace{0.15\baselineskip}

      \begin{small}
        \begin{tabular}{lll}
    \toprule
    $_{\text{\tiny{\ggn}}}$$^{\text{\tiny{Data}}}$ & mb & sub \\
    \midrule
    exact & 36
              & 255 \\
    mc   & 1119
              & 1536 \\
    \bottomrule
\end{tabular}       \end{small}
    \end{minipage}

    \begin{flushleft}
      \vspace{1ex}
      (d)
    \end{flushleft}

    \vspace{-1.3\baselineskip}

\pgfkeys{/pgfplots/performancedefault/.style={
    width=1.04\linewidth,
    height=\goldenRatioInv*1.04\linewidth,
    every axis plot/.append style={line width = 1.2pt},
    every axis plot post/.append style={
      mark size=2, mark options={opacity=0.9, solid, line width = 1pt}
    },
    tick pos = left,
    xmajorticks = true,
    ymajorticks = true,
    ylabel near ticks,
    xlabel near ticks,
    xtick align = inside,
    ytick align = inside,
    legend cell align = left,
    legend columns = 3,
legend style = {
      fill opacity = 0.9,
      text opacity = 1,
      font = \small,
      at={(1, 1.025)},
      anchor=south east,
    },
    xticklabel style = {font = \small, inner xsep = 0ex},
    xlabel style = {font = \small},
    axis line style = {black},
    yticklabel style = {font = \small, inner ysep = 0ex},
    ylabel style = {font = \small, inner ysep = 0ex},
    title style = {font = \small, inner ysep = 0ex, yshift = -0.75ex},
    grid = major,
    grid style = {dashed},
    title = {},
  }
}
 \pgfkeys{/pgfplots/zmystyle/.style={performancedefault}}
    \begin{tikzpicture}

\definecolor{color0}{rgb}{0.937254901960784,0.231372549019608,0.172549019607843}
\definecolor{color1}{rgb}{0.274509803921569,0.6,0.564705882352941}
\definecolor{color2}{rgb}{0.870588235294118,0.623529411764706,0.0862745098039216}
\definecolor{color3}{rgb}{0.501960784313725,0.184313725490196,0.6}

\begin{axis}[
axis line style={white!80!black},
legend style={fill opacity=0.8, draw opacity=1, text opacity=1, at={(0.03,0.97)}, anchor=north west, draw=white!80!black},
tick pos=left,
title={cifar100\_allcnnc, N=64, cuda, layerwise\_group},
xlabel={top eigenpairs (\(\displaystyle k\))},
xmin=0.55, xmax=10.45,
ylabel={time [s]},
ymin=0.0276978035242289, ymax=30.9392352817777,
ymode=log,
zmystyle
]
\addplot [, color0, dashed, mark=pentagon*, mark size=3, mark options={solid}]
table {1 1.53709048399469
2 3.29550092600402
3 6.01564522000263
4 8.66417485298007
5 11.024156790023
6 14.1937458989851
7 15.7953649429837
8 18.4152254970104
9 19.669088796014
10 22.4885368550022
};
\addlegendentry{power iteration}
\addplot [, color1, dashed, mark=diamond*, mark size=3, mark options={solid}]
table {1 1.33679832299822
2 1.35294976900332
3 1.35486114700325
4 1.35586281897849
5 1.36057051201351
6 1.35904047399526
7 1.35944268701132
8 1.35816546800197
9 1.36110231999191
10 1.35731867700815
};
\addlegendentry{sub, exact}
\addplot [, color2, dashed, mark=square*, mark size=3, mark options={solid}]
table {1 0.0988316820003092
2 0.0994032140006311
3 0.0991363560024183
4 0.0993853470135946
5 0.0987775640096515
6 0.0989322730165441
7 0.0989872500067577
8 0.0990799540013541
9 0.0992108779901173
10 0.0990415730047971
};
\addlegendentry{mb, mc}
\addplot [, color3, dashed, mark=triangle*, mark size=3, mark options={solid,rotate=180}]
table {1 0.0382870699977502
2 0.0384355849819258
3 0.0383684310072567
4 0.038122364989249
5 0.038201666000532
6 0.038215864013182
7 0.0381060299987439
8 0.0382287740067113
};
\addlegendentry{sub, mc}
\end{axis}

\end{tikzpicture}
   \end{minipage}

  \vspace{-2ex}
  \caption{\textbf{GPU memory and run time performance for the \allcnnc
      architecture on \cifarhun:} Left and right columns show results with the
    full network's \ggn ($D = 1,\!387,\!108, C=100$) and a per-layer
    block-diagonal approximation, respectively. (a, b) Critical batch sizes
    $N_{\text{crit}}$ for computing eigenvalues and the top eigenpair. (c, d)
    Run time comparison with a power iteration for extracting the $k$ leading
    eigenpairs using a mini-batch of size $N=64$.}
  \label{fig:performance-cifar100-allcnnc-cuda}
\end{figure*}

 \begin{figure*}[tb]
  \centering
  \begin{minipage}[t]{0.49\linewidth}
    \centering
    \textbf{Full network}
    \vspace{-3ex}

    \begin{minipage}[t]{0.49\linewidth}
      \centering

      \begin{flushleft}
        \vspace{1ex}
        (a)
      \end{flushleft}

      \vspace{-1.0\baselineskip}
      \begin{small}
        $N_{\text{crit}}$ (eigenvalues)
      \end{small}
      \vspace{0.15\baselineskip}

      \begin{small}
        \begin{tabular}{lll}
    \toprule
    $_{\text{\tiny{\ggn}}}$$^{\text{\tiny{Data}}}$ & mb & sub \\
    \midrule
    exact & 96
              & 695 \\
    mc   & 3771
              & 4941 \\
    \bottomrule
\end{tabular}       \end{small}
    \end{minipage}
    \hfill
    \begin{minipage}[t]{0.49\linewidth}
      \centering

      \begin{flushleft}
        \vspace{1ex}
        \phantom{(a)}
      \end{flushleft}
      \vspace{-1.0\baselineskip}

      \begin{small}
        $N_{\text{crit}}$ (top eigenpair)
      \end{small}
      \vspace{0.15\baselineskip}

      \begin{small}
        \begin{tabular}{lll}
    \toprule
    $_{\text{\tiny{\ggn}}}$$^{\text{\tiny{Data}}}$ & mb & sub \\
    \midrule
    exact & 45
              & 351 \\
    mc   & 2558
              & 4454 \\
    \bottomrule
\end{tabular}       \end{small}
    \end{minipage}

    \begin{flushleft}
      \vspace{1ex}
      (c)
    \end{flushleft}

    \vspace{-1.3\baselineskip}

\pgfkeys{/pgfplots/performancedefault/.style={
    width=1.04\linewidth,
    height=\goldenRatioInv*1.04\linewidth,
    every axis plot/.append style={line width = 1.2pt},
    every axis plot post/.append style={
      mark size=2, mark options={opacity=0.9, solid, line width = 1pt}
    },
    tick pos = left,
    xmajorticks = true,
    ymajorticks = true,
    ylabel near ticks,
    xlabel near ticks,
    xtick align = inside,
    ytick align = inside,
    legend cell align = left,
    legend columns = 3,
legend style = {
      fill opacity = 0.9,
      text opacity = 1,
      font = \small,
      at={(1, 1.025)},
      anchor=south east,
    },
    xticklabel style = {font = \small, inner xsep = 0ex},
    xlabel style = {font = \small},
    axis line style = {black},
    yticklabel style = {font = \small, inner ysep = 0ex},
    ylabel style = {font = \small, inner ysep = 0ex},
    title style = {font = \small, inner ysep = 0ex, yshift = -0.75ex},
    grid = major,
    grid style = {dashed},
    title = {},
  }
}
 \pgfkeys{/pgfplots/zmystyle/.style={performancedefault}}
    \begin{tikzpicture}

\definecolor{color0}{rgb}{0.937254901960784,0.231372549019608,0.172549019607843}
\definecolor{color1}{rgb}{0.274509803921569,0.6,0.564705882352941}
\definecolor{color2}{rgb}{0.870588235294118,0.623529411764706,0.0862745098039216}
\definecolor{color3}{rgb}{0.501960784313725,0.184313725490196,0.6}

\begin{axis}[
axis line style={white!80!black},
legend style={fill opacity=0.8, draw opacity=1, text opacity=1, at={(0.03,0.97)}, anchor=north west, draw=white!80!black},
tick pos=left,
title={cifar100\_allcnnc, N=64, cpu, one\_group},
xlabel={top eigenpairs (\(\displaystyle k\))},
xmin=0.55, xmax=10.45,
ylabel={time [s]},
ymin=0.245792677714802, ymax=306.435199901807,
ymode=log,
zmystyle
]
\addplot [, color0, dashed, mark=pentagon*, mark size=3, mark options={solid}]
table {1 18.6224392989825
2 45.2476005610079
3 71.6506203919998
4 83.5294155470037
5 114.604826588999
6 134.578176974988
7 145.192165841989
8 183.542261836003
9 206.574076799996
10 221.626580533979
};
\addlegendentry{power iteration}
\addplot [, color1, dashed, mark=diamond*, mark size=3, mark options={solid}]
table {1 8.27111610400607
2 8.34843894500227
3 8.32659733799665
4 8.34333644800063
5 8.34644191100233
6 8.34699494200322
7 8.32181938899885
8 8.31104396299634
9 8.3247039889975
10 8.31278976799513
};
\addlegendentry{sub, exact}
\addplot [, color2, dashed, mark=square*, mark size=3, mark options={solid}]
table {1 0.713133348996053
2 0.717113475002407
3 0.720882066998456
4 0.721453428996028
5 0.726131636998616
6 0.720149373999448
7 0.729221790003066
8 0.741393238000455
9 0.727941582001222
10 0.722816093002621
};
\addlegendentry{mb, mc}
\addplot [, color3, dashed, mark=triangle*, mark size=3, mark options={solid,rotate=180}]
table {1 0.33984880400385
2 0.34264848499879
3 0.349108467999031
4 0.350434840998787
5 0.349057795996487
6 0.350499678999768
7 0.351185671002895
8 0.351644793001469
};
\addlegendentry{sub, mc}
\end{axis}

\end{tikzpicture}
   \end{minipage}
  \hfill
  \begin{minipage}[t]{0.49\linewidth}
    \centering
    \textbf{Block-diagonal approximation}
    \vspace{-3ex}

    \begin{minipage}[t]{0.49\linewidth}
      \centering

      \begin{flushleft}
        \vspace{1ex}
        (b)
      \end{flushleft}

      \vspace{-1.0\baselineskip}
      \begin{small}
        $N_{\text{crit}}$ (eigenvalues)
      \end{small}
      \vspace{0.15\baselineskip}

      \begin{small}
        \begin{tabular}{lll}
    \toprule
    $_{\text{\tiny{\ggn}}}$$^{\text{\tiny{Data}}}$ & mb & sub \\
    \midrule
    exact & 97
              & 703 \\
    mc   & 3783
              & 4929 \\
    \bottomrule
\end{tabular}       \end{small}
    \end{minipage}
    \hfill
    \begin{minipage}[t]{0.49\linewidth}
      \centering

      \begin{flushleft}
        \vspace{1ex}
        \phantom{(a)}
      \end{flushleft}
      \vspace{-1.0\baselineskip}

      \begin{small}
        $N_{\text{crit}}$ (top eigenpair)
      \end{small}
      \vspace{0.15\baselineskip}

      \begin{small}
        \begin{tabular}{lll}
    \toprule
    $_{\text{\tiny{\ggn}}}$$^{\text{\tiny{Data}}}$ & mb & sub \\
    \midrule
    exact & 97
              & 703 \\
    mc   & 3759
              & 4918 \\
    \bottomrule
\end{tabular}       \end{small}
    \end{minipage}

    \begin{flushleft}
      \vspace{1ex}
      (d)
    \end{flushleft}

    \vspace{-1.3\baselineskip}

\pgfkeys{/pgfplots/performancedefault/.style={
    width=1.04\linewidth,
    height=\goldenRatioInv*1.04\linewidth,
    every axis plot/.append style={line width = 1.2pt},
    every axis plot post/.append style={
      mark size=2, mark options={opacity=0.9, solid, line width = 1pt}
    },
    tick pos = left,
    xmajorticks = true,
    ymajorticks = true,
    ylabel near ticks,
    xlabel near ticks,
    xtick align = inside,
    ytick align = inside,
    legend cell align = left,
    legend columns = 3,
legend style = {
      fill opacity = 0.9,
      text opacity = 1,
      font = \small,
      at={(1, 1.025)},
      anchor=south east,
    },
    xticklabel style = {font = \small, inner xsep = 0ex},
    xlabel style = {font = \small},
    axis line style = {black},
    yticklabel style = {font = \small, inner ysep = 0ex},
    ylabel style = {font = \small, inner ysep = 0ex},
    title style = {font = \small, inner ysep = 0ex, yshift = -0.75ex},
    grid = major,
    grid style = {dashed},
    title = {},
  }
}
 \pgfkeys{/pgfplots/zmystyle/.style={performancedefault}}
    \begin{tikzpicture}

\definecolor{color0}{rgb}{0.937254901960784,0.231372549019608,0.172549019607843}
\definecolor{color1}{rgb}{0.274509803921569,0.6,0.564705882352941}
\definecolor{color2}{rgb}{0.870588235294118,0.623529411764706,0.0862745098039216}
\definecolor{color3}{rgb}{0.501960784313725,0.184313725490196,0.6}

\begin{axis}[
axis line style={white!80!black},
legend style={fill opacity=0.8, draw opacity=1, text opacity=1, at={(0.91,0.5)}, anchor=east, draw=white!80!black},
tick pos=left,
title={cifar100\_allcnnc, N=64, cpu, layerwise\_group},
xlabel={top eigenpairs (\(\displaystyle k\))},
xmin=0.55, xmax=10.45,
ylabel={time [s]},
ymin=0.240381438551032, ymax=614.724467469512,
ymode=log,
zmystyle
]
\addplot [, color0, dashed, mark=pentagon*, mark size=3, mark options={solid}]
table {1 30.890429195977
2 72.4770839109842
3 126.664585363993
4 161.828190853004
5 205.971328154992
6 248.368187843997
7 290.767375240976
8 332.694239347009
9 388.587202214985
10 430.309989359987
};
\addlegendentry{power iteration}
\addplot [, black, dashed, mark=*, mark size=3, mark options={solid}]
table {1 375.53582859
2 376.194721504995
3 376.118281371004
4 375.323636784
5 375.422504825008
6 376.193631990012
7 375.975429985003
8 375.834260588003
9 375.633623753994
10 375.637536057999
};
\addlegendentry{mb, exact}
\addplot [, color1, dashed, mark=diamond*, mark size=3, mark options={solid}]
table {1 8.91458579801838
2 8.97201327097719
3 8.98541935198591
4 8.97522488300456
5 8.9699610620155
6 8.98417820601026
7 8.95071924000513
8 8.96681337498012
9 8.96218986698659
10 8.97993912099628
};
\addlegendentry{sub, exact}
\addplot [, color2, dashed, mark=square*, mark size=3, mark options={solid}]
table {1 0.715564451005775
2 0.721840932994382
3 0.734913141001016
4 0.720706533989869
5 0.725127355981385
6 0.719817852019332
7 0.722395085002063
8 0.730500807985663
9 0.746279801009223
10 0.739214531000471
};
\addlegendentry{mb, mc}
\addplot [, color3, dashed, mark=triangle*, mark size=3, mark options={solid,rotate=180}]
table {1 0.343399770994438
2 0.346202403015923
3 0.347909268981311
4 0.34835068800021
5 0.350254623015644
6 0.352106687991181
7 0.351124015025562
8 0.35080232899054
};
\addlegendentry{sub, mc}
\end{axis}

\end{tikzpicture}
   \end{minipage}

  \vspace{-2ex}
  \caption{\textbf{CPU memory and run time performance for the \allcnnc
      architecture on \cifarhun:} Left and right columns show results with the
    full network's \ggn ($D = 1,\!387,\!108, C=100$) and a per-layer
    block-diagonal approximation, respectively. (a, b) Critical batch sizes
    $N_{\text{crit}}$ for computing eigenvalues and the top eigenpair. (c, d)
    Run time comparison with a power iteration for extracting the $k$ leading
    eigenpairs using a mini-batch of size $N=64$.}
  \label{fig:performance-cifar100-allcnnc-cpu}
\end{figure*}

\begin{table}[tb]
  \centering
  \caption{\textbf{Memory performance for computing damped Newton steps:} Left
    and right columns show the critical batch sizes with the full network's \ggn and
    a per-layer block-diagonal approximation, respectively.}
  \label{tab:performance}

  \begin{small}
    \textbf{\fmnist \twoctwod}
  \end{small}

  \begin{minipage}{0.49\linewidth}
    \centering
    \begin{small}
      \textbf{Full network}
    \end{small}
  \end{minipage}
  \hfill
  \begin{minipage}{0.49\linewidth}
    \centering
    \begin{small}
      \textbf{Block-diagonal approximation}
    \end{small}
  \end{minipage}
  \vspace{1ex}

  \begin{minipage}{0.245\linewidth}
    \centering
    \begin{small}
      $N_{\text{crit}}$ (GPU)
    \end{small}
    \vspace{0.15\baselineskip}

    \begin{small}
      \begin{tabular}{lll}
    \toprule
    $_{\text{\tiny{\ggn}}}$$^{\text{\tiny{Data}}}$ & mb & sub \\
    \midrule
    exact & 66
              & 159 \\
    mc   & 362
              & 528 \\
    \bottomrule
\end{tabular}     \end{small}
  \end{minipage}
  \hfill
  \begin{minipage}{0.245\linewidth}
    \centering
    \begin{small}
      $N_{\text{crit}}$ (CPU)
    \end{small}
    \vspace{0.15\baselineskip}

    \begin{small}
      \begin{tabular}{lll}
    \toprule
    $_{\text{\tiny{\ggn}}}$$^{\text{\tiny{Data}}}$ & mb & sub \\
    \midrule
    exact & 202
              & 487 \\
    mc   & 1107
              & 1639 \\
    \bottomrule
\end{tabular}     \end{small}
  \end{minipage}
  \hfill
  \begin{minipage}{0.245\linewidth}
    \centering
    \begin{small}
      $N_{\text{crit}}$ (GPU)
    \end{small}
    \vspace{0.15\baselineskip}

    \begin{small}
      \begin{tabular}{lll}
    \toprule
    $_{\text{\tiny{\ggn}}}$$^{\text{\tiny{Data}}}$ & mb & sub \\
    \midrule
    exact & 68
              & 159 \\
    mc   & 368
              & 528 \\
    \bottomrule
\end{tabular}     \end{small}
  \end{minipage}
  \hfill
  \begin{minipage}{0.245\linewidth}
    \centering
    \begin{small}
      $N_{\text{crit}}$ (CPU)
    \end{small}
    \vspace{0.15\baselineskip}

    \begin{small}
      \begin{tabular}{lll}
    \toprule
    $_{\text{\tiny{\ggn}}}$$^{\text{\tiny{Data}}}$ & mb & sub \\
    \midrule
    exact & 210
              & 487 \\
    mc   & 1137
              & 1643 \\
    \bottomrule
\end{tabular}     \end{small}
  \end{minipage}

  \vspace{5ex}

\begin{small}
    \textbf{\cifarten \threecthreed}
  \end{small}

  \begin{minipage}{0.49\linewidth}
    \centering
    \begin{small}
      \textbf{Full network}
    \end{small}
  \end{minipage}
  \hfill
  \begin{minipage}{0.49\linewidth}
    \centering
    \begin{small}
      \textbf{Block-diagonal approximation}
    \end{small}
  \end{minipage}
  \vspace{1ex}

  \begin{minipage}{0.245\linewidth}
    \centering
    \begin{small}
      $N_{\text{crit}}$ (GPU)
    \end{small}
    \vspace{0.15\baselineskip}

    \begin{small}
      \begin{tabular}{lll}
    \toprule
    $_{\text{\tiny{\ggn}}}$$^{\text{\tiny{Data}}}$ & mb & sub \\
    \midrule
    exact & 208
              & 727 \\
    mc   & 1055
              & 1816 \\
    \bottomrule
\end{tabular}     \end{small}
  \end{minipage}
  \hfill
  \begin{minipage}{0.245\linewidth}
    \centering
    \begin{small}
      $N_{\text{crit}}$ (CPU)
    \end{small}
    \vspace{0.15\baselineskip}

    \begin{small}
      \begin{tabular}{lll}
    \toprule
    $_{\text{\tiny{\ggn}}}$$^{\text{\tiny{Data}}}$ & mb & sub \\
    \midrule
    exact & 667
              & 2215 \\
    mc   & 3473
              & 5632 \\
    \bottomrule
\end{tabular}     \end{small}
  \end{minipage}
  \hfill
  \begin{minipage}{0.245\linewidth}
    \centering
    \begin{small}
      $N_{\text{crit}}$ (GPU)
    \end{small}
    \vspace{0.15\baselineskip}

    \begin{small}
      \begin{tabular}{lll}
    \toprule
    $_{\text{\tiny{\ggn}}}$$^{\text{\tiny{Data}}}$ & mb & sub \\
    \midrule
    exact & 349
              & 795 \\
    mc   & 1659
              & 2112 \\
    \bottomrule
\end{tabular}     \end{small}
  \end{minipage}
  \hfill
  \begin{minipage}{0.245\linewidth}
    \centering
    \begin{small}
      $N_{\text{crit}}$ (CPU)
    \end{small}
    \vspace{0.15\baselineskip}

    \begin{small}
      \begin{tabular}{lll}
    \toprule
    $_{\text{\tiny{\ggn}}}$$^{\text{\tiny{Data}}}$ & mb & sub \\
    \midrule
    exact & 1046
              & 2423 \\
    mc   & 4997
              & 6838 \\
    \bottomrule
\end{tabular}     \end{small}
  \end{minipage}

  \vspace{5ex}

  \begin{small}
    \textbf{\cifarten ResNet32}
  \end{small}

  \begin{minipage}{0.49\linewidth}
    \centering
    \begin{small}
      \textbf{Full network}
    \end{small}
  \end{minipage}
  \hfill
  \begin{minipage}{0.49\linewidth}
    \centering
    \begin{small}
      \textbf{Block-diagonal approximation}
    \end{small}
  \end{minipage}
  \vspace{1ex}

  \begin{minipage}{0.245\linewidth}
    \centering
    \begin{small}
      $N_{\text{crit}}$ (GPU)
    \end{small}
    \vspace{0.15\baselineskip}

    \begin{small}
      \begin{tabular}{lll}
    \toprule
    $_{\text{\tiny{\ggn}}}$$^{\text{\tiny{Data}}}$ & mb & sub \\
    \midrule
    exact & 344
              & 1119 \\
    mc   & 1205
              & 1535 \\
    \bottomrule
\end{tabular}     \end{small}
  \end{minipage}
  \hfill
  \begin{minipage}{0.245\linewidth}
  \textcolor{white}{-}

\end{minipage}
  \hfill
  \begin{minipage}{0.245\linewidth}
    \centering
    \begin{small}
      $N_{\text{crit}}$ (GPU)
    \end{small}
    \vspace{0.15\baselineskip}

    \begin{small}
      \begin{tabular}{lll}
    \toprule
    $_{\text{\tiny{\ggn}}}$$^{\text{\tiny{Data}}}$ & mb & sub \\
    \midrule
    exact & 1051
              & 1851 \\
    mc   & 2048
              & 2208 \\
    \bottomrule
\end{tabular}     \end{small}
  \end{minipage}
  \hfill
  \begin{minipage}{0.245\linewidth}
  \textcolor{white}{.}

\end{minipage}

  \vspace{5ex}

  \begin{small}
    \textbf{\cifarten ResNet56}
  \end{small}

  \begin{minipage}{0.49\linewidth}
    \centering
    \begin{small}
      \textbf{Full network}
    \end{small}
  \end{minipage}
  \hfill
  \begin{minipage}{0.49\linewidth}
    \centering
    \begin{small}
      \textbf{Block-diagonal approximation}
    \end{small}
  \end{minipage}
  \vspace{1ex}

  \begin{minipage}{0.245\linewidth}
    \centering
    \begin{small}
      $N_{\text{crit}}$ (GPU)
    \end{small}
    \vspace{0.15\baselineskip}

    \begin{small}
      \begin{tabular}{lll}
    \toprule
    $_{\text{\tiny{\ggn}}}$$^{\text{\tiny{Data}}}$ & mb & sub \\
    \midrule
    exact & 209
              & 640 \\
    mc   & 687
              & 890 \\
    \bottomrule
\end{tabular}     \end{small}
  \end{minipage}
  \hfill
  \begin{minipage}{0.245\linewidth}
  \textcolor{white}{.}

\end{minipage}
  \hfill
  \begin{minipage}{0.245\linewidth}
    \centering
    \begin{small}
      $N_{\text{crit}}$ (GPU)
    \end{small}
    \vspace{0.15\baselineskip}

    \begin{small}
      \begin{tabular}{lll}
    \toprule
    $_{\text{\tiny{\ggn}}}$$^{\text{\tiny{Data}}}$ & mb & sub \\
    \midrule
    exact & 767
              & 1165 \\
    mc   & 1232
              & 1255 \\
    \bottomrule
\end{tabular}     \end{small}
  \end{minipage}
  \hfill
  \begin{minipage}{0.245\linewidth}
  \textcolor{white}{.}

\end{minipage}

\vspace{5ex}

  \begin{small}
    \textbf{\cifarhun \allcnnc}
  \end{small}

  \begin{minipage}{0.49\linewidth}
    \centering
    \begin{small}
      \textbf{Full network}
    \end{small}
  \end{minipage}
  \hfill
  \begin{minipage}{0.49\linewidth}
    \centering
    \begin{small}
      \textbf{Block-diagonal approximation}
    \end{small}
  \end{minipage}
  \vspace{1ex}

  \begin{minipage}{0.245\linewidth}
    \centering
    \begin{small}
      $N_{\text{crit}}$ (GPU)
    \end{small}
    \vspace{0.15\baselineskip}

    \begin{small}
      \begin{tabular}{lll}
    \toprule
    $_{\text{\tiny{\ggn}}}$$^{\text{\tiny{Data}}}$ & mb & sub \\
    \midrule
    exact & 13
              & 87 \\
    mc   & 640
              & 959 \\
    \bottomrule
\end{tabular}     \end{small}
  \end{minipage}
  \hfill
  \begin{minipage}{0.245\linewidth}
    \centering
    \begin{small}
      $N_{\text{crit}}$ (CPU)
    \end{small}
    \vspace{0.15\baselineskip}

    \begin{small}
      \begin{tabular}{lll}
    \toprule
    $_{\text{\tiny{\ggn}}}$$^{\text{\tiny{Data}}}$ & mb & sub \\
    \midrule
    exact & 43
              & 309 \\
    mc   & 2015
              & 2865 \\
    \bottomrule
\end{tabular}     \end{small}
  \end{minipage}
  \hfill
  \begin{minipage}{0.245\linewidth}
    \centering
    \begin{small}
      $N_{\text{crit}}$ (GPU)
    \end{small}
    \vspace{0.15\baselineskip}

    \begin{small}
      \begin{tabular}{lll}
    \toprule
    $_{\text{\tiny{\ggn}}}$$^{\text{\tiny{Data}}}$ & mb & sub \\
    \midrule
    exact & 35
              & 135 \\
    mc   & 1079
              & 1536 \\
    \bottomrule
\end{tabular}     \end{small}
  \end{minipage}
  \hfill
  \begin{minipage}{0.245\linewidth}
    \centering
    \begin{small}
      $N_{\text{crit}}$ (CPU)
    \end{small}
    \vspace{0.15\baselineskip}

    \begin{small}
      \begin{tabular}{lll}
    \toprule
    $_{\text{\tiny{\ggn}}}$$^{\text{\tiny{Data}}}$ & mb & sub \\
    \midrule
    exact & 95
              & 504 \\
    mc   & 3360
              & 3920 \\
    \bottomrule
\end{tabular}     \end{small}
  \end{minipage}

\end{table}

 \clearpage

\clearpage

\subsection{Training of neural networks}
\label{sec:training_of_nns}
\paragraph{Procedure:} We train the following \deepobs
\citep{schneider2019deepobs} architectures with \sgd and \adam: 
\threecthreed on \cifarten{}, 
\twoctwod on \fmnist{} and 
\allcnnc on \cifarhun{} -- all are equipped with cross-entropy loss. 
To ensure successful training, we use the hyperparameters from \cite{dangel2020backpack} (see \Cref{tab:noise-hyperparameters}).

We also train a residual network \resnetthirtytwo \cite{he2016deep} with cross-entropy loss on \cifarten{} with both \sgd and \adam{}. For this, we use a batch size of $128$ and train for $180$ epochs.  Momentum for \sgd was fixed to $0.9$, and \adam uses the default parameters ($\beta_1 = 0.9$, $\beta_2 = 0.999$, $\epsilon = 10^{-8}$). For both optimizers, the learning rate was determined via grid search. Following \citep{schneider2019deepobs}, we use a log-equidistant grid from $10^{-5}$ to $10^2$ and $36$ grid points. As performance metric, the best test accuracy during training (evaluated once every epoch) is used.  

\begin{figure}[p]
    \centering
    \textbf{\fmnist \twoctwod \sgd}\\[1mm]
    \begin{minipage}{0.50\textwidth}
        \centering
        \includegraphics[scale=1.0]{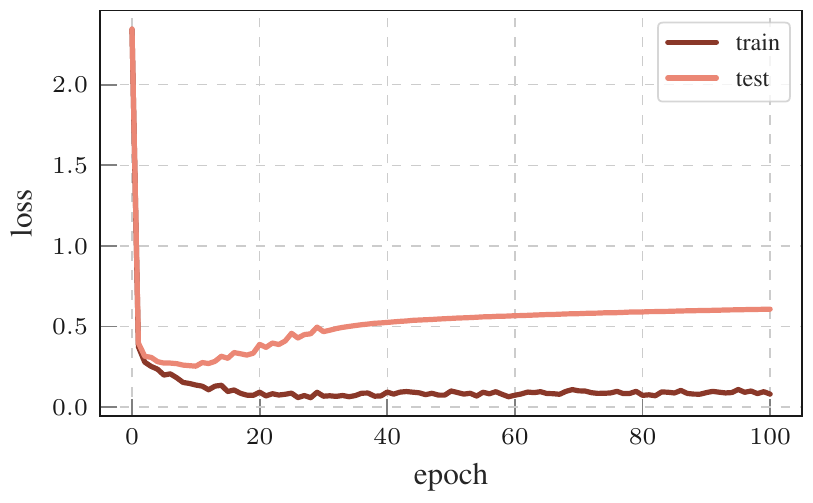}
    \end{minipage}\hfill
    \begin{minipage}{0.50\textwidth}
        \centering
        \includegraphics[scale=1.0]{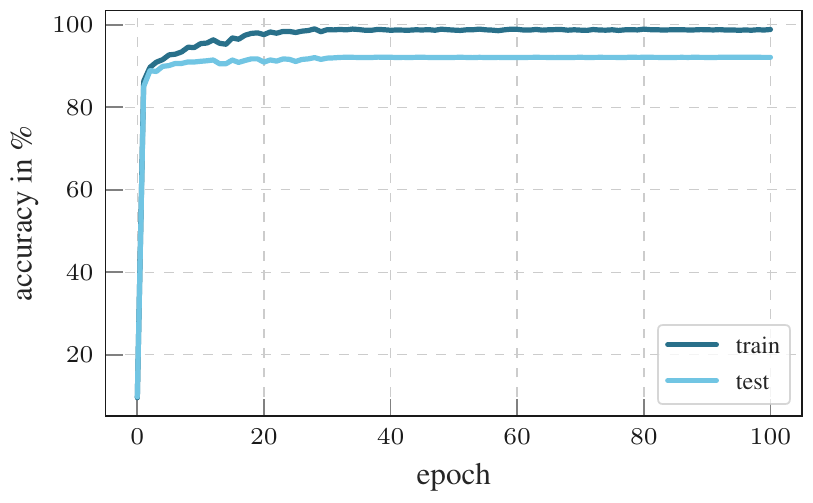}
    \end{minipage}

    \textbf{\fmnist \twoctwod \adam}\\[1mm]
    \begin{minipage}{0.50\textwidth}
        \centering
        \includegraphics[scale=1.0]{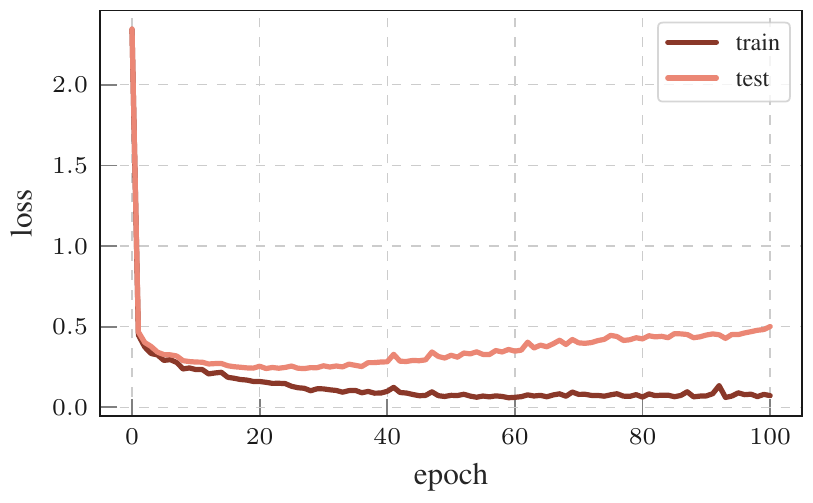}
    \end{minipage}\hfill
    \begin{minipage}{0.50\textwidth}
        \centering
        \includegraphics[scale=1.0]{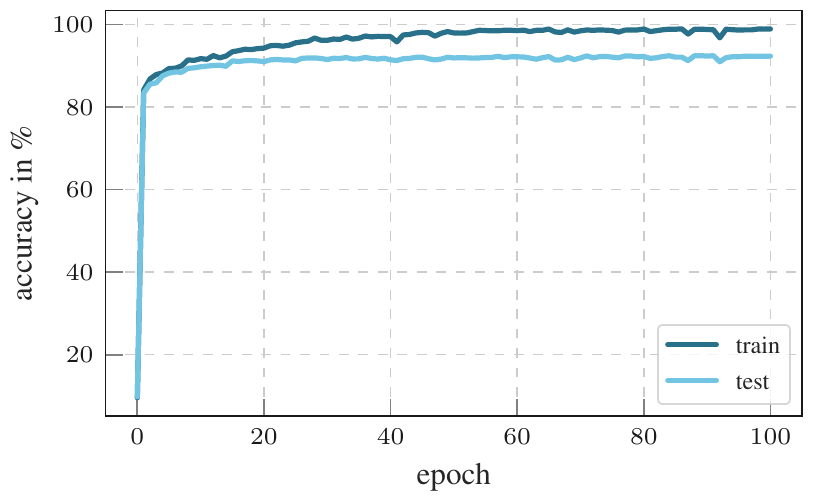}
    \end{minipage}

    \textbf{\cifarten \threecthreed \sgd}\\[1mm]
    \begin{minipage}{0.50\textwidth}
        \centering
        \includegraphics[scale=1.0]{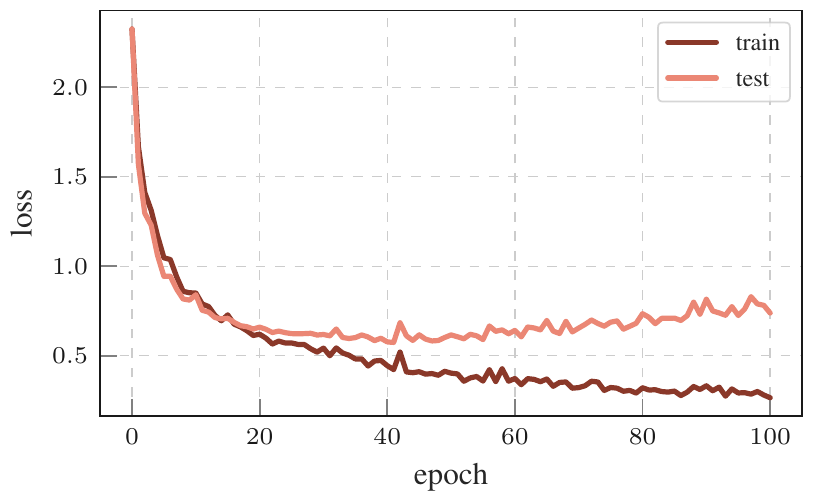}
    \end{minipage}\hfill
    \begin{minipage}{0.50\textwidth}
        \centering
        \includegraphics[scale=1.0]{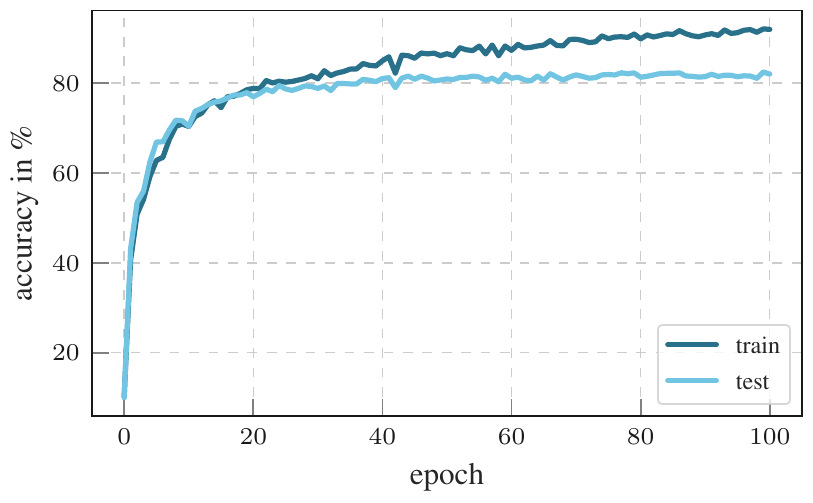}
    \end{minipage}

    \textbf{\cifarten \threecthreed \adam}\\[1mm]
    \begin{minipage}{0.50\textwidth}
        \centering
        \includegraphics[scale=1.0]{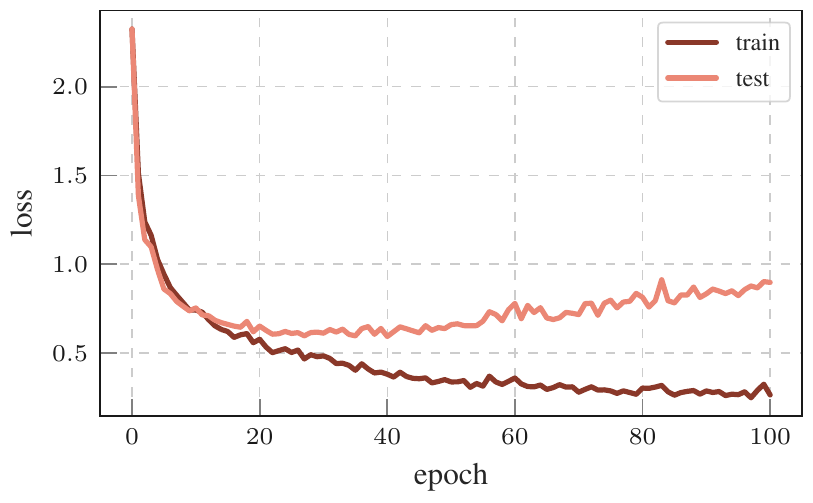}
    \end{minipage}\hfill
    \begin{minipage}{0.50\textwidth}
        \centering
        \includegraphics[scale=1.0]{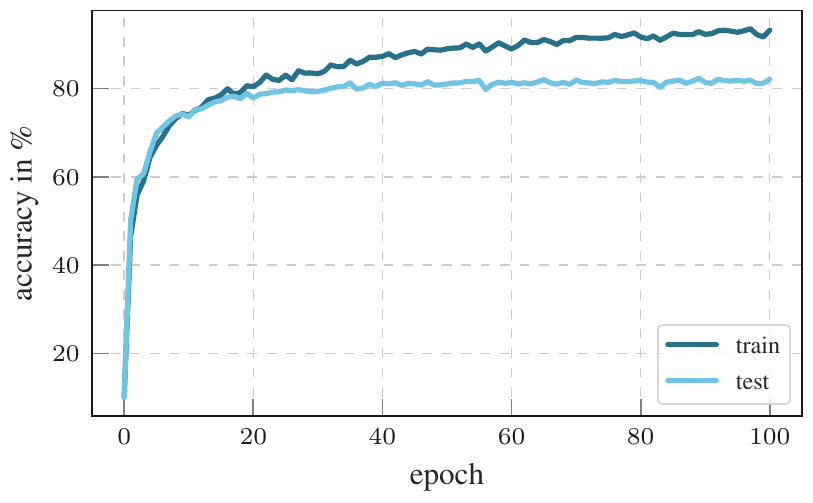}
    \end{minipage}

    \caption{\textbf{Training metrics (1):}
        Training/test loss/accuracy for all test problems.}
    \label{fig:training_metrics_1}
\end{figure}

\begin{figure}[p]
    \centering
    \textbf{\cifarten \resnetthirtytwo \sgd}\\[1mm]
    \begin{minipage}{0.50\textwidth}
        \centering
        \includegraphics[scale=1.0]{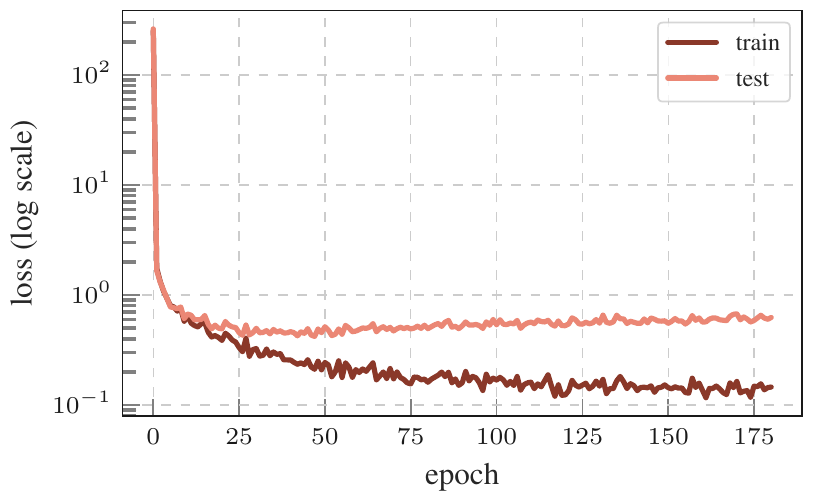}
    \end{minipage}\hfill
    \begin{minipage}{0.50\textwidth}
        \centering
        \includegraphics[scale=1.0]{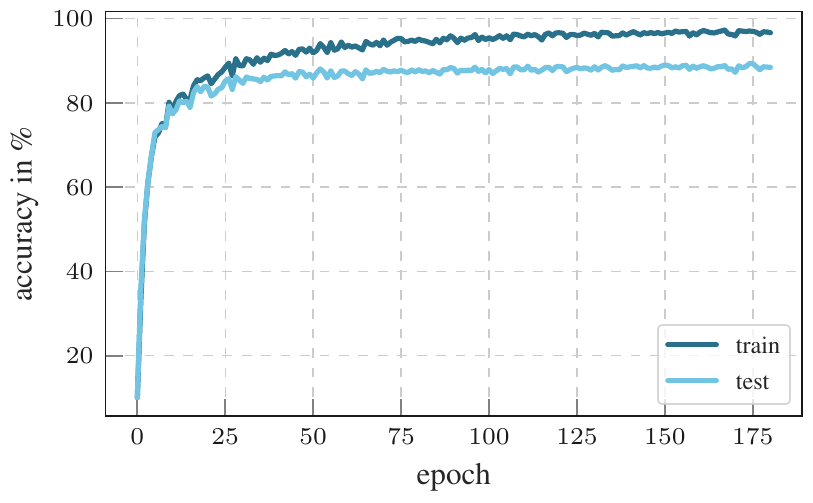}
    \end{minipage}

    \textbf{\cifarten \resnetthirtytwo \adam}\\[1mm]
    \begin{minipage}{0.50\textwidth}
        \centering
        \includegraphics[scale=1.0]{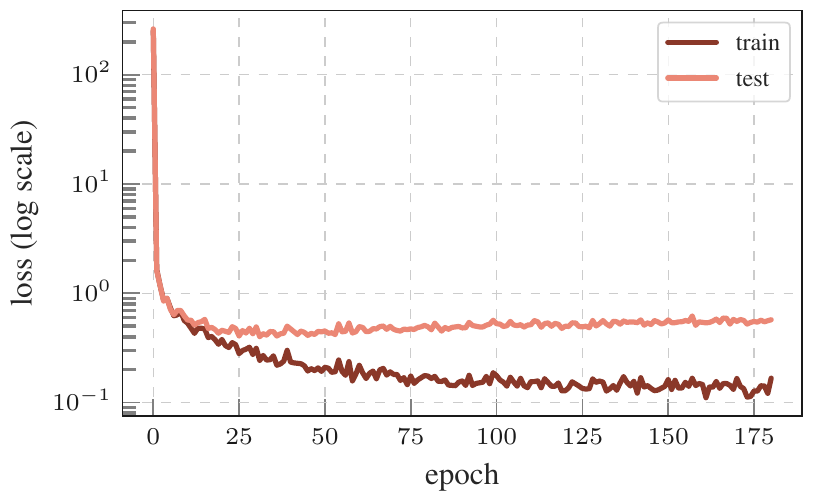}
    \end{minipage}\hfill
    \begin{minipage}{0.50\textwidth}
        \centering
        \includegraphics[scale=1.0]{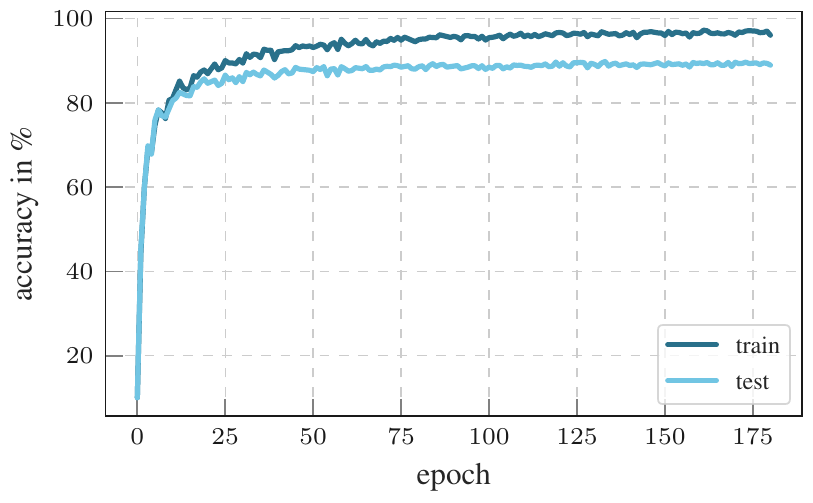}
    \end{minipage}

    \textbf{\cifarhun \allcnnc \sgd}\\[1mm]
    \begin{minipage}{0.50\textwidth}
        \centering
        \includegraphics[scale=1.0]{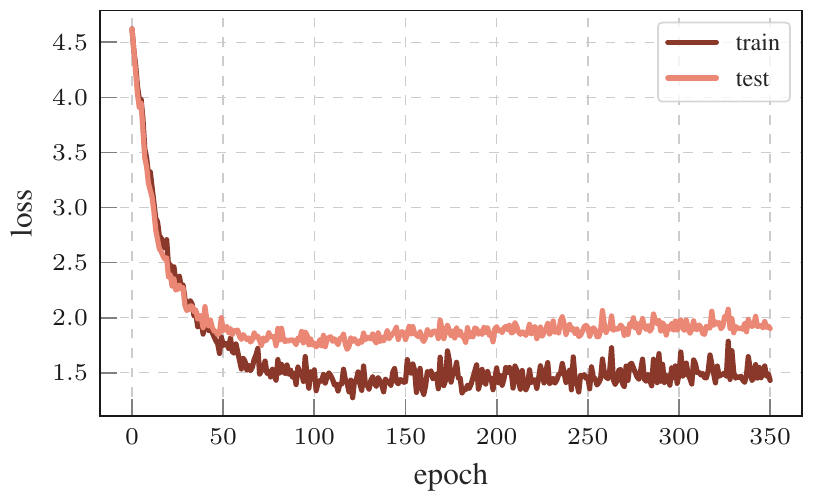}
    \end{minipage}\hfill
    \begin{minipage}{0.50\textwidth}
        \centering
        \includegraphics[scale=1.0]{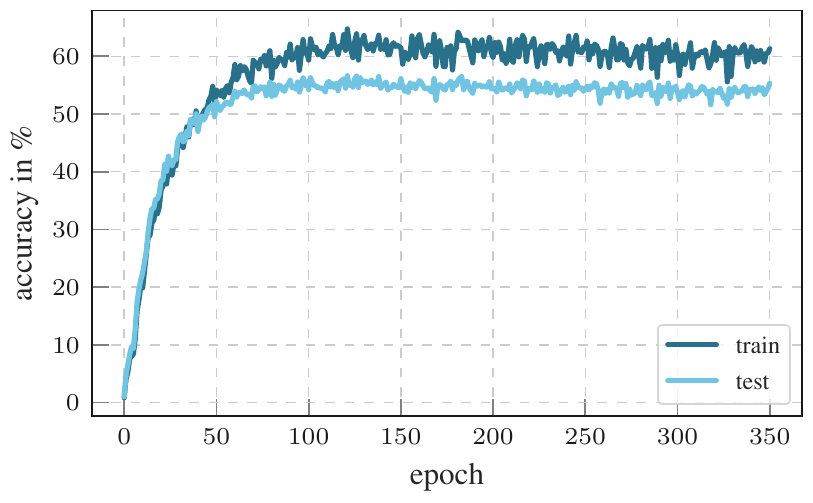}
    \end{minipage}

    \textbf{\cifarhun \allcnnc \adam}\\[1mm]
    \begin{minipage}{0.50\textwidth}
        \centering
        \includegraphics[scale=1.0]{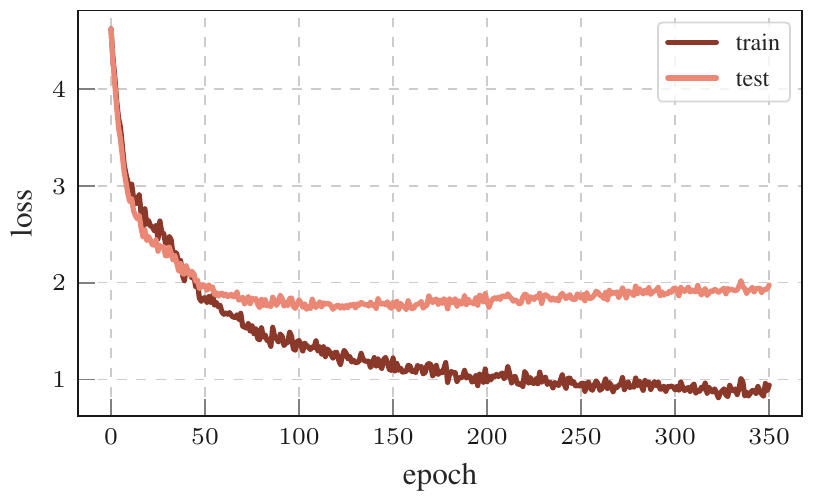}
    \end{minipage}\hfill
    \begin{minipage}{0.50\textwidth}
        \centering
        \includegraphics[scale=1.0]{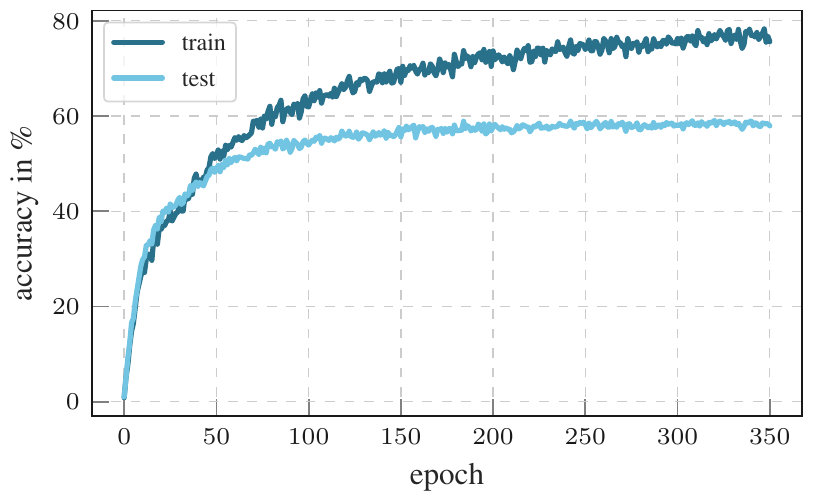}
    \end{minipage}

    \caption{\textbf{Training metrics (2):}
        Training/test loss/accuracy for all test problems.}
    \label{fig:training_metrics_2}
\end{figure}

\paragraph{Results:} 
The results for the hyperparameter grid search are reported in \Cref{tab:noise-hyperparameters}. The training metrics training/test loss/accuracy for all eight test problems are shown in \Cref{fig:training_metrics_1} and \ref{fig:training_metrics_2}. 

\begin{table}[ht]
  \centering
  \caption{
\textbf{Hyperparameter settings for training runs:}
For both \sgd and \adam, we report their learning rates $\alpha$ (taken from the baselines in \cite{dangel2020backpack} or, for \resnetthirtytwo, determined via grid search). Momentum for \sgd is fixed to $0.9$. \adam uses the default parameters $\beta_1 = 0.9$, $\beta_2 = 0.999$, $\epsilon = 10^{-8}$. We also report the batch size used for training and the number of training epochs.
}
  \label{tab:noise-hyperparameters}
  \vspace{1ex}
  \begin{normalsize}
    \begin{tabular}{lllll}
      \toprule
      Problem
      & \sgd
      & \adam
      & Batch size
      & Training epochs \\
      \midrule
      \fmnist \twoctwod
      & $\alpha \approx 2.07 \cdot 10^{-2}$ 
      & $\alpha \approx 1.27\cdot 10^{-4}$ 
      & $N = 128$
      & $100$
      \\
      \cifarten \threecthreed
      & $\alpha \approx 3.79 \cdot 10^{-3}$ 
      & $\alpha \approx 2.98 \cdot 10^{-4}$ 
      & $N = 128$
      & $100$
      \\
      \cifarten \resnetthirtytwo
      & $\alpha \approx 6.31 \cdot 10^{-2}$ 
      & $\alpha \approx 2.51 \cdot 10^{-3}$ 
      & $N = 128$
      & $180$
      \\
      \cifarhun \allcnnc
      & $\alpha \approx 4.83\cdot 10^{-1}$ 
      & $\alpha \approx 6.95\cdot 10^{-4}$ 
      & $N = 256$
      & $350$
      \\
      \bottomrule
    \end{tabular}
  \end{normalsize}
\end{table}

\subsection{\ggn vs. Hessian}
\label{sec:ggn_vs_hessian}
\paragraph{Checkpoints:} During training of the neural networks (see \Cref{sec:training_of_nns}), we store a copy of the model (i.e. the network's current parameters) at specific checkpoints.
This grid defines the temporal resolution for all subsequent computations.
Since training progresses much faster in the early training stages, we use a log-grid with $100$ grid points between $1$ and the number of training epochs and shift this grid by $-1$.

\paragraph{Overlap:} Recall from \Cref{subsec:approx_quality}:
For the set of orthonormal eigenvectors $\{ \ve_c^\mU \}_{c=1}^C$ to the $C$ largest eigenvalues of some symmetric matrix $\mU$, let
$\mP^\mU = (\ve_1^\mU, ..., \ve_C^\mU) (\ve_1^\mU, ..., \ve_C^\mU)^\top$.
As in \cite{gurari2018gradient}, the overlap between two subspaces
$\mathcal{E}^\mU = \vecspan{}(\ve_1^\mU, ..., \ve_C^\mU)$ and
$\mathcal{E}^\mV = \vecspan{}(\ve_1^\mV, ..., \ve_C^\mV)$ of the matrices $\mU$ and $\mV$ is defined by
\begin{equation*}
    \text{overlap}(\mathcal{E}^\mU, \mathcal{E}^\mV)
    = \frac{\Tr{}(\mP^\mU \mP^\mV)}
    {\sqrt{\Tr{}(\mP^\mU) \Tr{}(\mP^\mV)}}
    \in [0, 1]\, .
\end{equation*}
The overlap can be computed efficiently by using the trace's cyclic property: It holds
$\Tr{}(\mP^\mU \mP^\mV) = \Tr{}(\mW^\top \mW)$ with
$W = (\ve_1^\mU, ..., \ve_C^\mU)^\top (\ve_1^\mV, ..., \ve_C^\mV) \in \mathbb{R}^{C \times C}$. Note that this is a small $C \times C$ matrix, whereas $\mP^\mU, \mP^\mV \in \mathbb{R}^{D \times D}$.
Since
\begin{align*}
    \Tr{}(\mP^\mU)
     & = \Tr{}((\ve_1^\mU, ..., \ve_C^\mU) (\ve_1^\mU, ..., \ve_C^\mU)^\top) \\
     & = \Tr{}((\ve_1^\mU, ..., \ve_C^\mU)^\top (\ve_1^\mU, ..., \ve_C^\mU))
    \explainmath{(Cyclic property of trace)}                                 \\
     & = \Tr{}(\mI_C)
    \explainmath{(Orthonormality of the eigenvectors)}                       \\
     & = C
\end{align*}
(and analogous $\Tr{}(\mP^\mV) = C$),
the denominator simplifies to $C$.

\paragraph{Procedure:} For each checkpoint, we compute the top-$C$ eigenvalues and associated eigenvectors of the full-batch \ggn and Hessian (i.e. \ggn and Hessian are both evaluated on the entire training set) using an iterative matrix-free approach. We then compute the overlap between the top-$C$ eigenspaces as described above. The eigspaces (i.e. the top-$C$ eigenvalues and associated eigenvectors) are stored on disk such that they can be used as a reference by subsequent experiments.

\begin{figure}[p]
    \centering
    \begin{minipage}{0.50\textwidth}
        \centering
        \textbf{\fmnist \twoctwod \sgd}\\[1mm]
        \includegraphics[scale=1.0]{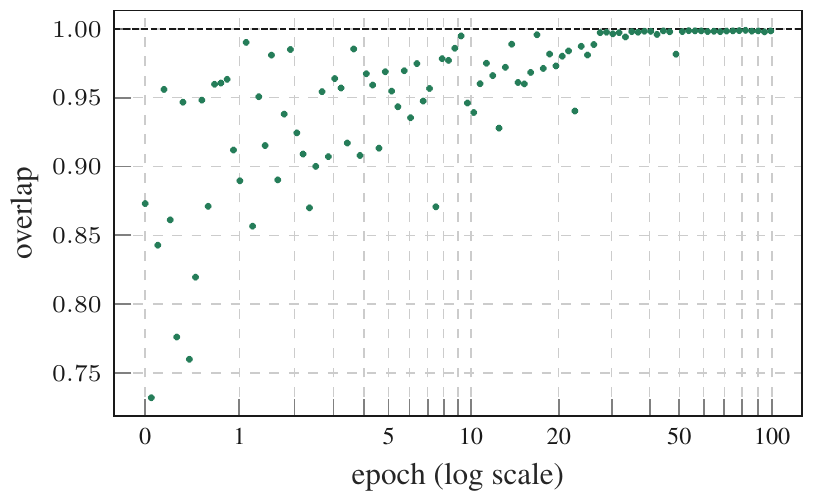}
    \end{minipage}\hfill
    \begin{minipage}{0.50\textwidth}
        \centering
        \textbf{\fmnist \twoctwod \adam}\\[1mm]
        \includegraphics[scale=1.0]{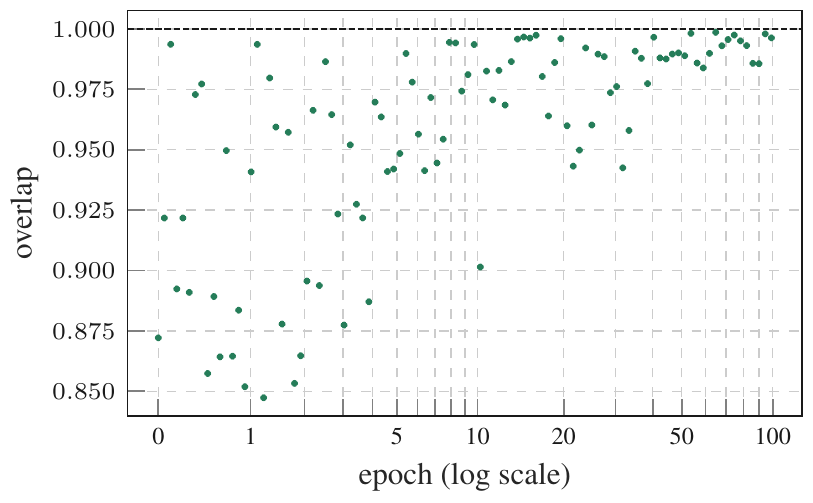}
    \end{minipage}

    \begin{minipage}{0.50\textwidth}
        \centering
        \textbf{\cifarten \threecthreed \sgd}\\[1mm]
        \includegraphics[scale=1.0]{fig/exp13_plots/eigspace_ggn_vs_hessian/cifar10_3c3d_sgd_eigenspace.pdf}
    \end{minipage}\hfill
    \begin{minipage}{0.50\textwidth}
        \centering
        \textbf{\cifarten \threecthreed \adam}\\[1mm]
        \includegraphics[scale=1.0]{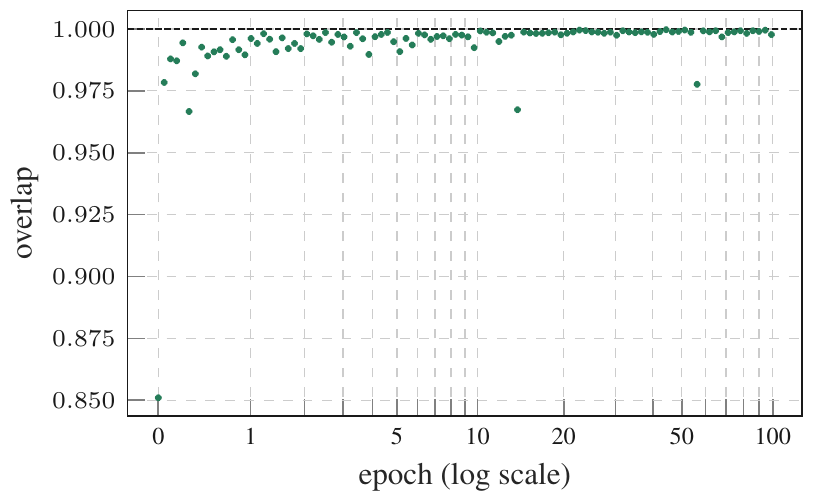}
    \end{minipage}

    \begin{minipage}{0.50\textwidth}
        \centering
        \textbf{\cifarten \resnetthirtytwo \sgd}\\[1mm]
        \includegraphics[scale=1.0]{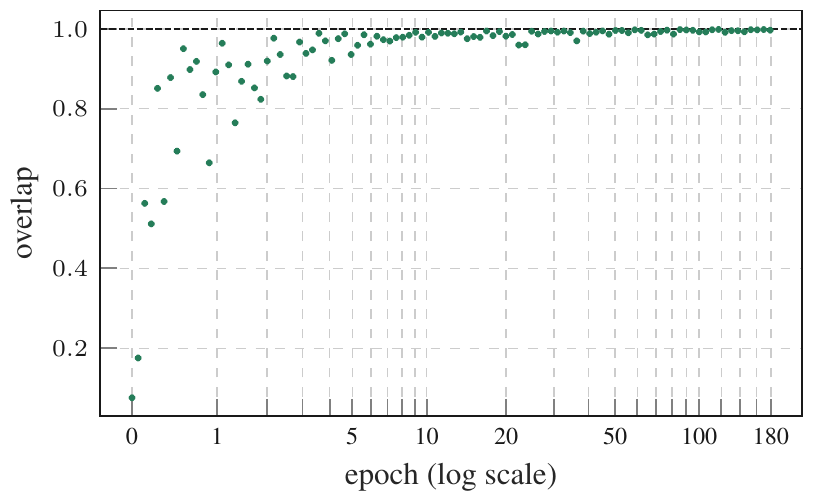}
    \end{minipage}\hfill
    \begin{minipage}{0.50\textwidth}
        \centering
        \textbf{\cifarten \resnetthirtytwo \adam}\\[1mm]
        \includegraphics[scale=1.0]{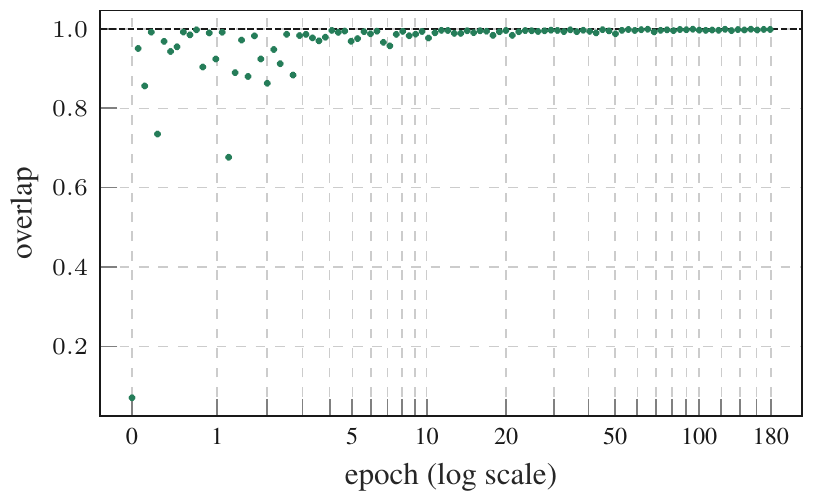}
    \end{minipage}

    \begin{minipage}{0.50\textwidth}
        \centering
        \textbf{\cifarhun \allcnnc \sgd}\\[1mm]
        \includegraphics[scale=1.0]{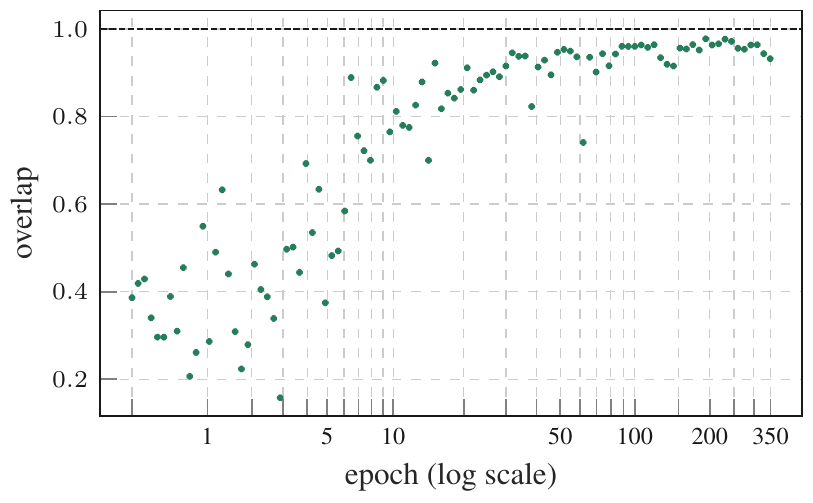}
    \end{minipage}\hfill
    \begin{minipage}{0.50\textwidth}
        \centering
        \textbf{\cifarhun \allcnnc \adam}\\[1mm]
        \includegraphics[scale=1.0]{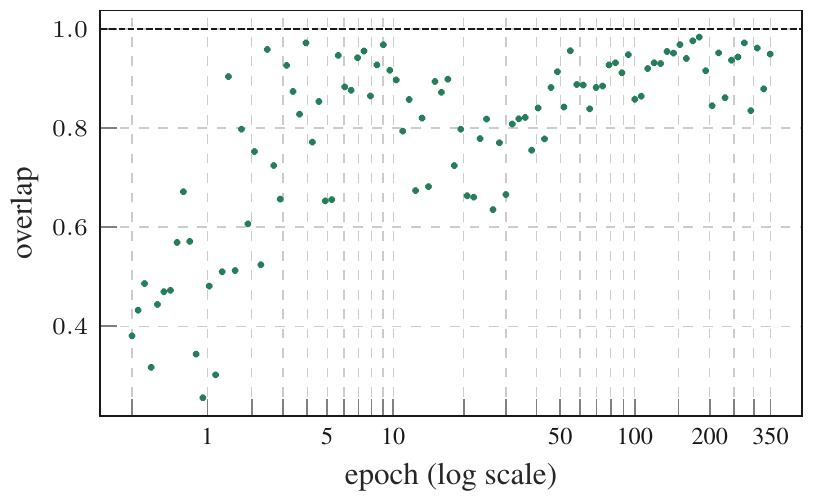}
    \end{minipage}

    \caption{
        \textbf{Full-batch \ggn vs. full-batch Hessian:}
        Overlap between the top-$C$ eigenspaces of the full-batch \ggn and full-batch Hessian during training for all test problems.
    }
    \label{fig:ggn_vs_hessian}
\end{figure}

\paragraph{Results:} The results for all test problems are presented in \Cref{fig:ggn_vs_hessian}.
Except for a short phase at the beginning of the optimization procedure (note the log scale for the epoch-axis), a strong agreement (note the different limits for the overlap-axis) between the top-$C$ eigenspaces is observed.
We make similar observations for all test problems, yet to a slightly lesser extent for \cifarhun{}.
A possible explanation for this would be that the $100$-dimensional eigenspaces differ in the eigenvectors associated with relatively small curvature. The corresponding eigenvalues already transition into the bulk of the spectrum, where the "sharpness of separation" decreases. However, since all directions are equally weighted in the overlap, overall slightly lower values are obtained.
 
\subsection{Eigenspace under noise and approximations}
\label{sec:eigenspace_noise}

\paragraph{Procedure (1):} 
We use the checkpoints and the definition of overlaps between eigenspaces from \Cref{sec:ggn_vs_hessian}. 
For the approximation of the \ggn{}, we consider the cases listed in \Cref{tab:cases_full_batch}.

\begin{table}[ht]
  \centering
  \caption{
\textbf{Considered cases for approximation of the eigenspace:}
We use a different set of cases for the approximation of the \ggn{}'s full-batch eigenspace depending on the test problem. For the test problems with $C=10$, we use $M=1$ \mc-sample, for the \cifarhun \allcnnc test problem ($C=100$), we use $M=10$ \mc-samples in order to reduce the computational costs by the same factor. 
}
  \label{tab:cases_full_batch}
  \vspace{1ex}
  \begin{normalsize}
    \begin{tabular}{ll}
      \toprule
      Problem 
      & Cases \\
      \midrule
      \makecell[tl]{
      \fmnist \twoctwod \\ 
      \cifarten \threecthreed and \\ 
      \cifarten \resnetthirtytwo}
      & \makecell[tl]{
      \textbf{mb, exact} with mini-batch sizes $N \in \{2, 8, 32, 128\}$\\
      \textbf{mb, mc} with $N=128$ and $M=1$ \mc{}-sample\\
      \textbf{sub, exact} using $16$ samples from the mini-batch\\
      \textbf{sub, mc} using $16$ samples from the mini-batch and $M=1$ \mc{}-sample
      }
      \\
      \midrule
      \cifarhun \allcnnc
      & \makecell[tl]{
      \textbf{mb, exact} with mini-batch sizes $N \in \{2, 8, 32, 128\}$\\
      \textbf{mb, mc} with $N=128$ and $M=10$ \mc{}-samples\\
      \textbf{sub, exact} using $16$ samples from the mini-batch\\
      \textbf{sub, mc} using $16$ samples from the mini-batch and $M=10$ \mc{}-samples
      }
      \\
      \bottomrule
    \end{tabular}
  \end{normalsize}
\end{table}

For every checkpoint and case, we compute the top-$C$ eigenvectors of the respective approximation to the \ggn{}. 
The eigenvectors are either computed directly using \vivit (by transforming eigenvectors of the Gram matrix into parameter space, see \Cref{sec:computing-full-ggn-eigenspectrum}) or, if not applicable (because memory requirements exceed $N_\text{crit}$, see \Cref{subsec:scalability}), using an iterative matrix-free approach. 
The overlap is computed in reference to the \ggn{}'s full-batch top-$C$ eigenspace (see \Cref{sec:ggn_vs_hessian}).
We extract $5$ mini-batches from the training data and repeat the above procedure for each mini-batch (i.e. we obtain $5$ overlap measurements for every checkpoint and case). The same $5$ mini-batches are used over all checkpoints and cases. 

\begin{figure}[p]
\centering
\textbf{\fmnist \twoctwod \sgd}\\[1mm]
\begin{minipage}{0.50\textwidth}
\centering
\includegraphics[scale=1.0]{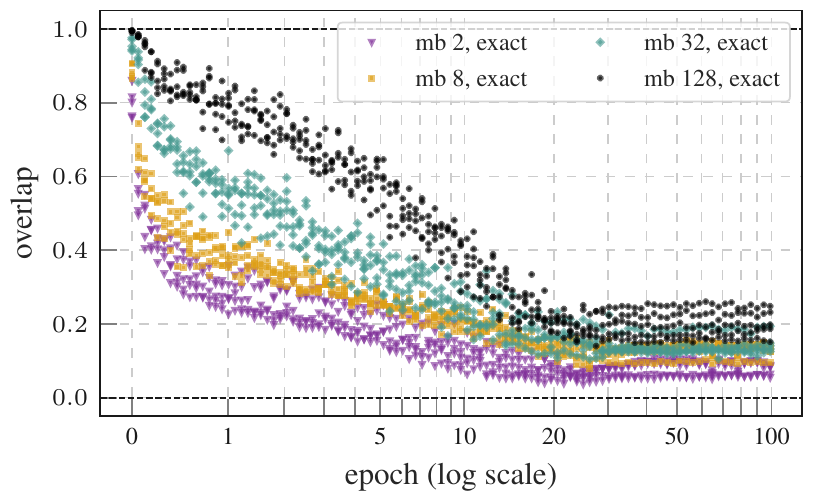} 
\end{minipage}\hfill
\begin{minipage}{0.50\textwidth}
\centering
\includegraphics[scale=1.0]{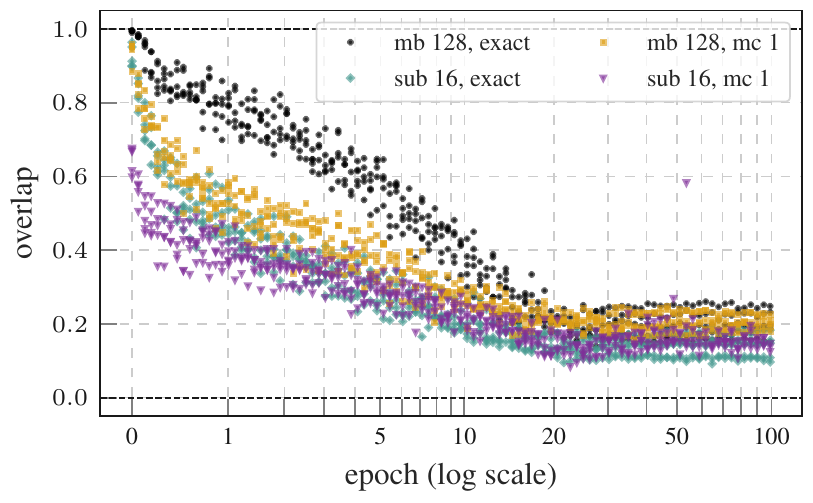} 
\end{minipage}

\textbf{\fmnist \twoctwod \adam}\\[1mm]
\begin{minipage}{0.50\textwidth}
\centering
\includegraphics[scale=1.0]{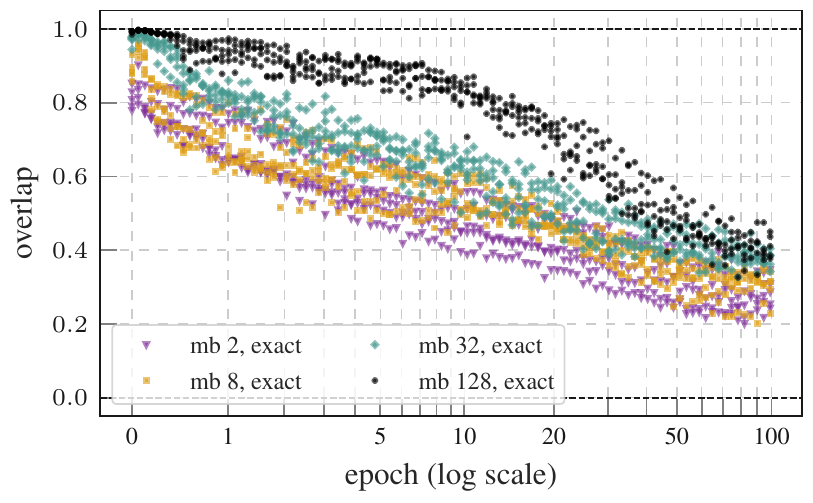} 
\end{minipage}\hfill
\begin{minipage}{0.50\textwidth}
\centering
\includegraphics[scale=1.0]{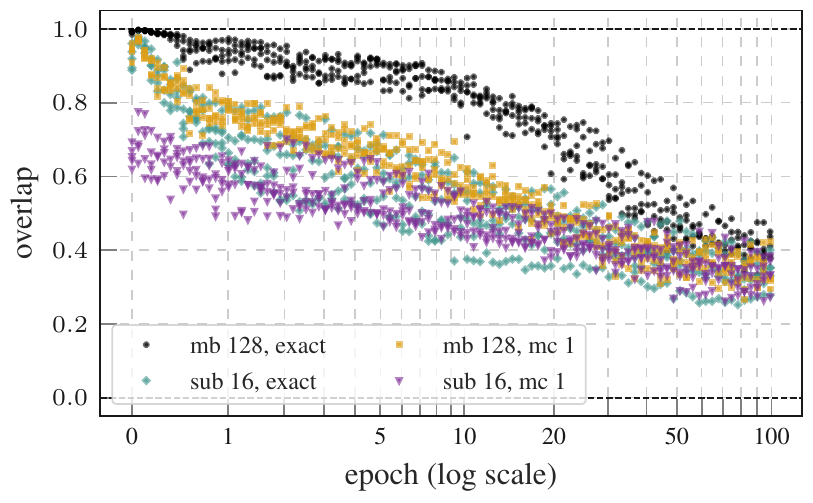} 
\end{minipage}

\textbf{\cifarten \threecthreed \sgd}\\[1mm]
\begin{minipage}{0.50\textwidth}
\centering
\includegraphics[scale=1.0]{fig/exp13_plots/eigspace_vivit_vs_fb/cifar10_3c3d_sgd_plot_bs.pdf} 
\end{minipage}\hfill
\begin{minipage}{0.50\textwidth}
\centering
\includegraphics[scale=1.0]{fig/exp13_plots/eigspace_vivit_vs_fb/cifar10_3c3d_sgd_plot_mc_sub.pdf} 
\end{minipage}

\textbf{\cifarten \threecthreed \adam}\\[1mm]
\begin{minipage}{0.50\textwidth}
\centering
\includegraphics[scale=1.0]{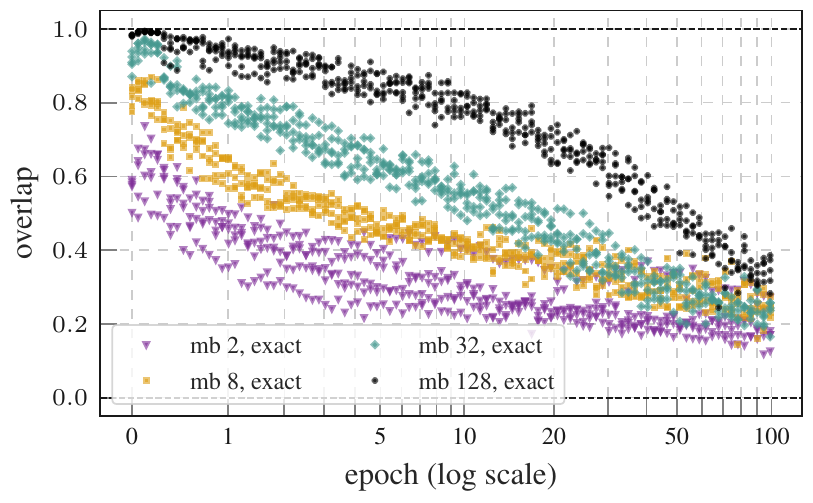} 
\end{minipage}\hfill
\begin{minipage}{0.50\textwidth}
\centering
\includegraphics[scale=1.0]{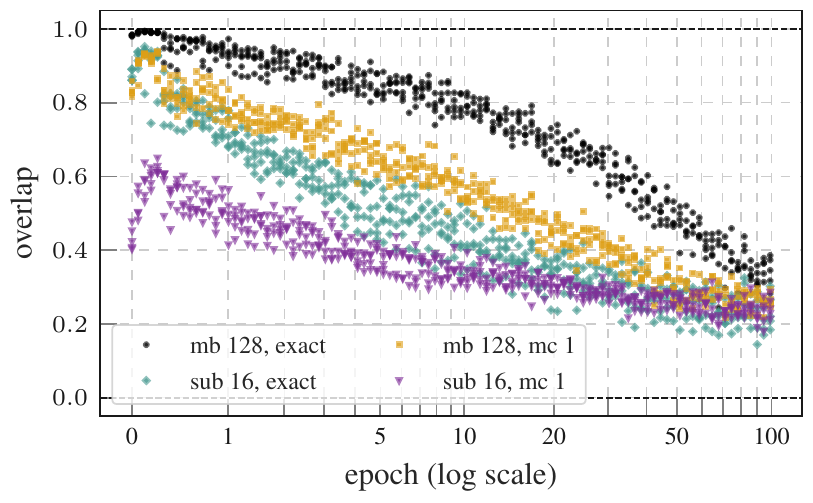} 
\end{minipage}

\caption{\textbf{\bfvivit{} vs. full-batch \ggn (1):} Overlap between the top-$C$ eigenspaces of different \ggn approximations and the full-batch \ggn during training for all test problems. 
Each approximation is evaluated on $5$ different mini-batches.}
\label{fig:vivit_vs_full_batch_ggn_1}
\end{figure}

\begin{figure}[p]
\centering
\textbf{\cifarten \resnetthirtytwo \sgd}\\[1mm]
\begin{minipage}{0.50\textwidth}
\centering
\includegraphics[scale=1.0]{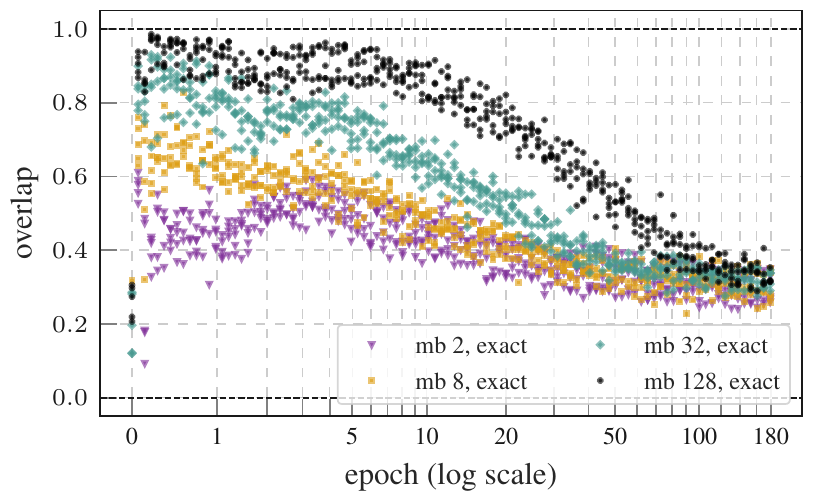} 
\end{minipage}\hfill
\begin{minipage}{0.50\textwidth}
\centering
\includegraphics[scale=1.0]{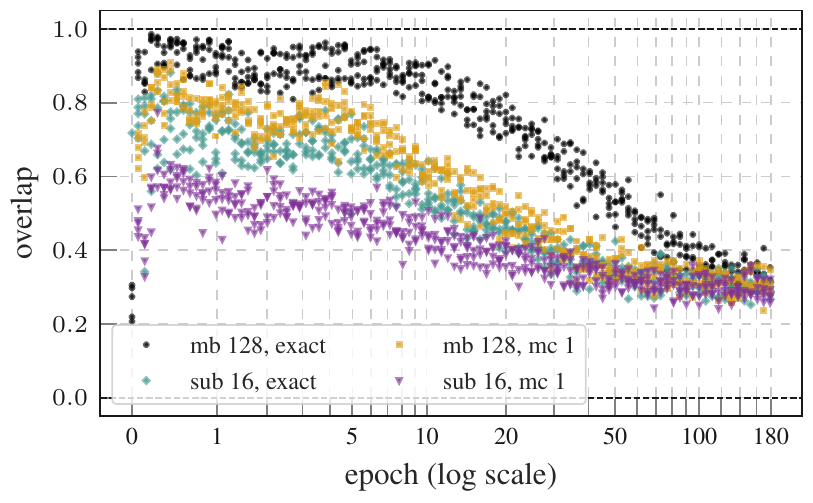} 
\end{minipage}

\textbf{\cifarten \resnetthirtytwo \adam}\\[1mm]
\begin{minipage}{0.50\textwidth}
\centering
\includegraphics[scale=1.0]{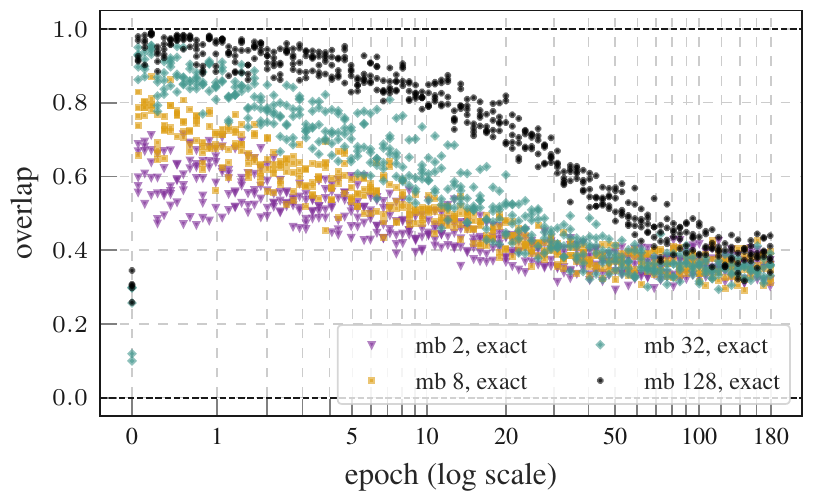} 
\end{minipage}\hfill
\begin{minipage}{0.50\textwidth}
\centering
\includegraphics[scale=1.0]{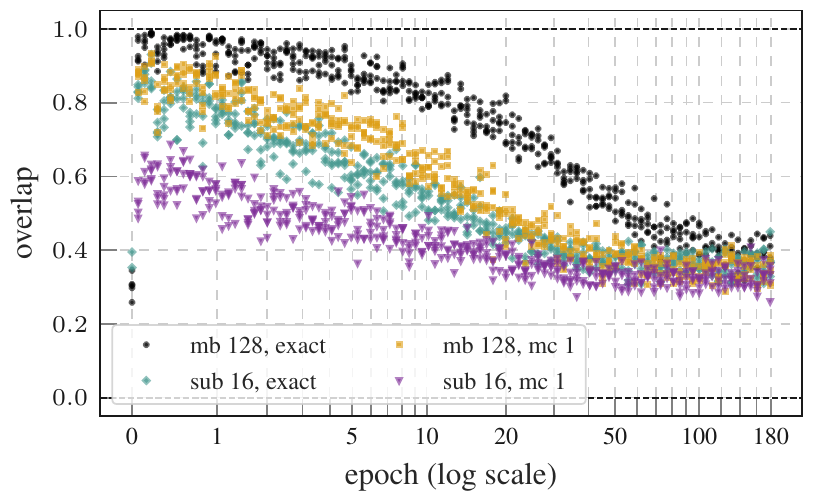} 
\end{minipage}

\textbf{\cifarhun \allcnnc \sgd}\\[1mm]
\begin{minipage}{0.50\textwidth}
\centering
\includegraphics[scale=1.0]{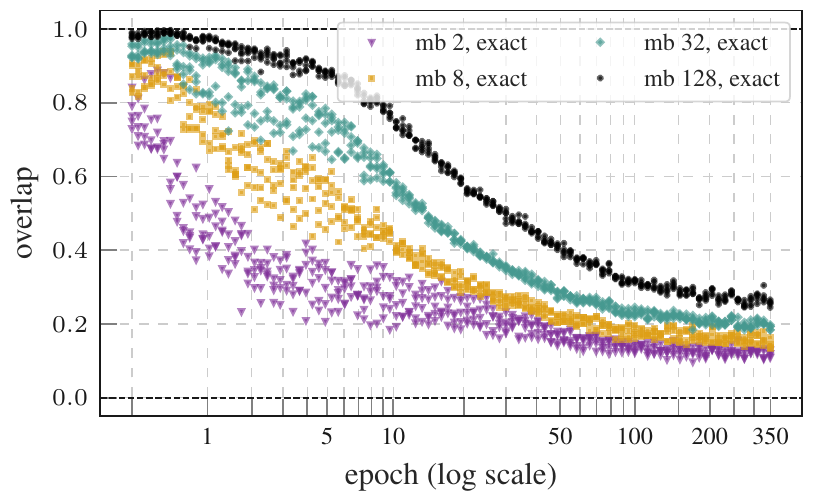} 
\end{minipage}\hfill
\begin{minipage}{0.50\textwidth}
\centering
\includegraphics[scale=1.0]{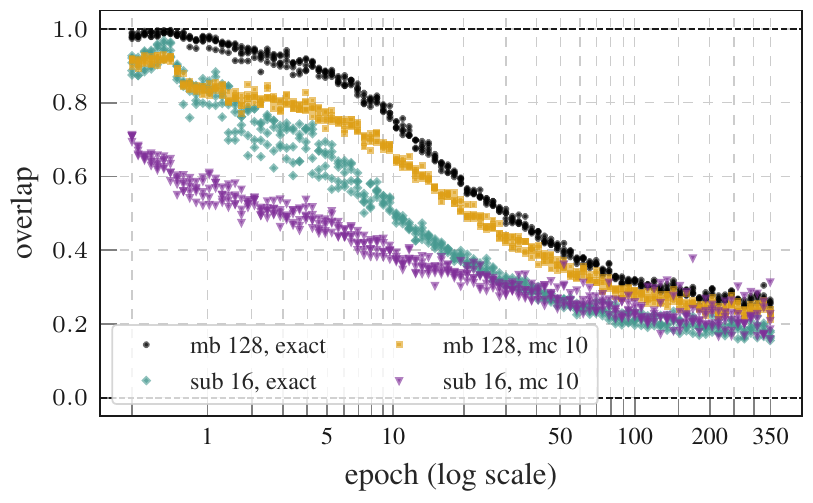} 
\end{minipage}

\textbf{\cifarhun \allcnnc \adam}\\[1mm]
\begin{minipage}{0.50\textwidth}
\centering
\includegraphics[scale=1.0]{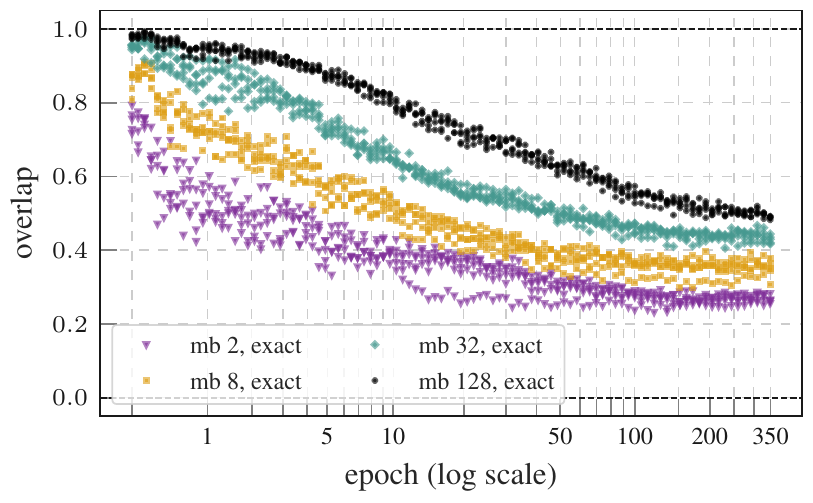} 
\end{minipage}\hfill
\begin{minipage}{0.50\textwidth}
\centering
\includegraphics[scale=1.0]{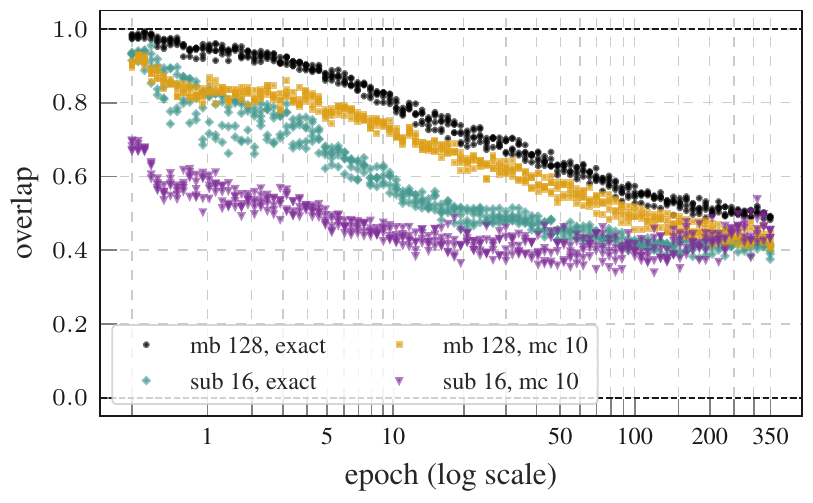} 
\end{minipage}

\caption{\textbf{\bfvivit{} vs. full-batch \ggn (2):} Overlap between the top-$C$ eigenspaces of different \ggn approximations and the full-batch \ggn during training for all test problems.
Each approximation is evaluated on $5$ different mini-batches.}
\label{fig:vivit_vs_full_batch_ggn_2}
\end{figure}

\paragraph{Results (1):} The results can be found in \Cref{fig:vivit_vs_full_batch_ggn_1} and \ref{fig:vivit_vs_full_batch_ggn_2}. 
All test problems show the same characteristics: With decreasing computational effort, the approximation carries less and less structure of its full-batch counterpart, as indicated by dropping overlaps. 
In addition, for a fixed approximation method, a decrease in approximation quality can be observed over the course of training. 

\paragraph{Procedure (2):}
Since \vivit{}'s \ggn approximations using curvature sub-sampling and/or the MC approximation (the cases \textbf{mb, mc} as well as \textbf{sub, exact} and \textbf{sub, mc} in \Cref{tab:cases_full_batch}) are based on the \textit{mini}-batch \ggn{}, we cannot expect them to perform better than this baseline. 
We thus repeat the analysis from above but use the mini-batch \ggn with batch-size $N=128$ as ground truth instead of the full-batch \ggn. 
Of course, the mini-batch reference top-$C$ eigenspace is always evaluated on the same mini-batch as the approximation.

\begin{figure}[p]
\centering
\begin{minipage}{0.50\textwidth}
\centering
\textbf{\fmnist \twoctwod \sgd}\\[1mm]
\includegraphics[scale=1.0]{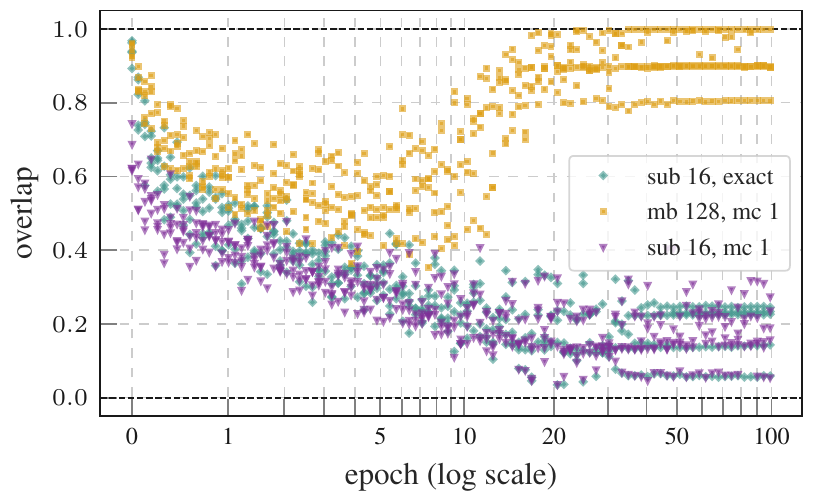} 
\end{minipage}\hfill
\begin{minipage}{0.50\textwidth}
\centering
\textbf{\fmnist \twoctwod \adam}\\[1mm]
\includegraphics[scale=1.0]{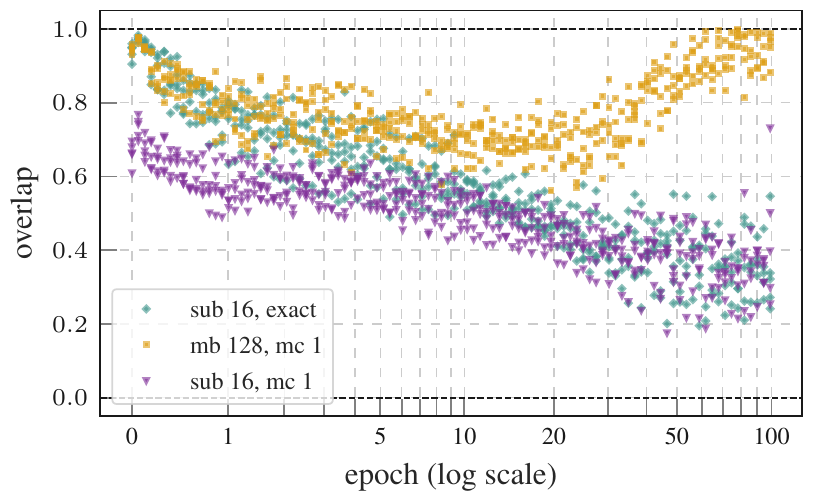} 
\end{minipage}

\begin{minipage}{0.50\textwidth}
\centering
\textbf{\cifarten \threecthreed \sgd}\\[1mm]
\includegraphics[scale=1.0]{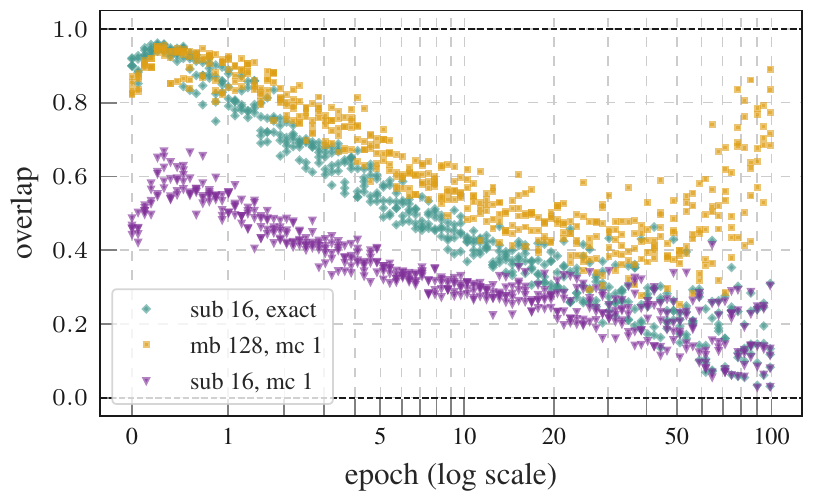} 
\end{minipage}\hfill
\begin{minipage}{0.50\textwidth}
\centering
\textbf{\cifarten \threecthreed \adam}\\[1mm]
\includegraphics[scale=1.0]{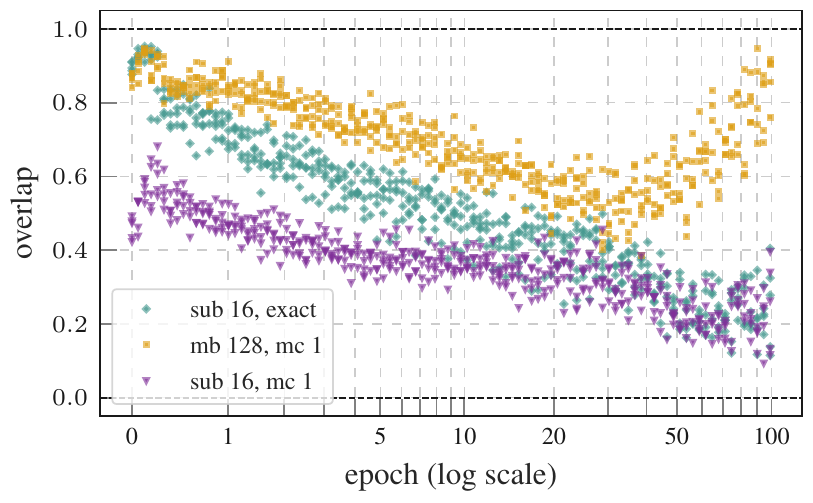} 
\end{minipage}

\begin{minipage}{0.50\textwidth}
\centering
\textbf{\cifarten \resnetthirtytwo \sgd}\\[1mm]
\includegraphics[scale=1.0]{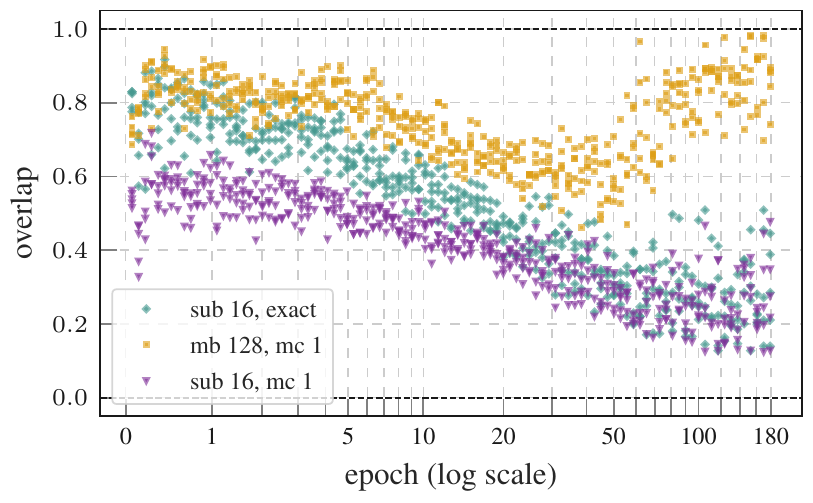} 
\end{minipage}\hfill
\begin{minipage}{0.50\textwidth}
\centering
\textbf{\cifarten \resnetthirtytwo \adam}\\[1mm]
\includegraphics[scale=1.0]{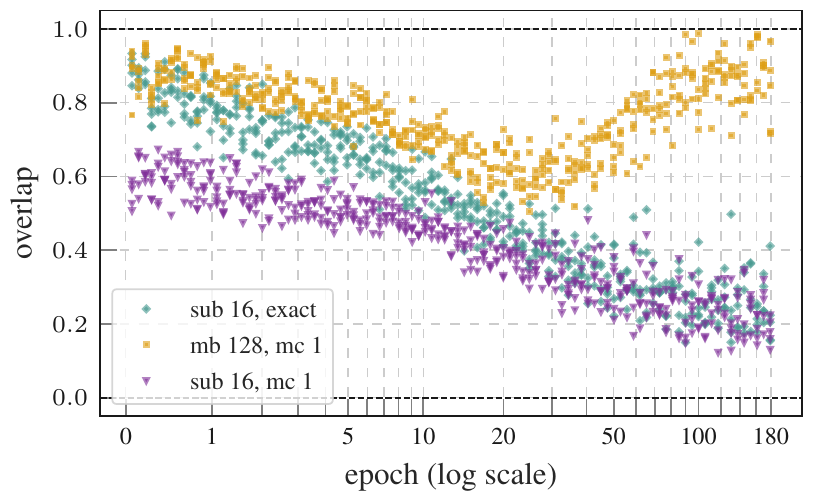} 
\end{minipage}

\begin{minipage}{0.50\textwidth}
\centering
\textbf{\cifarhun \allcnnc \sgd}\\[1mm]
\includegraphics[scale=1.0]{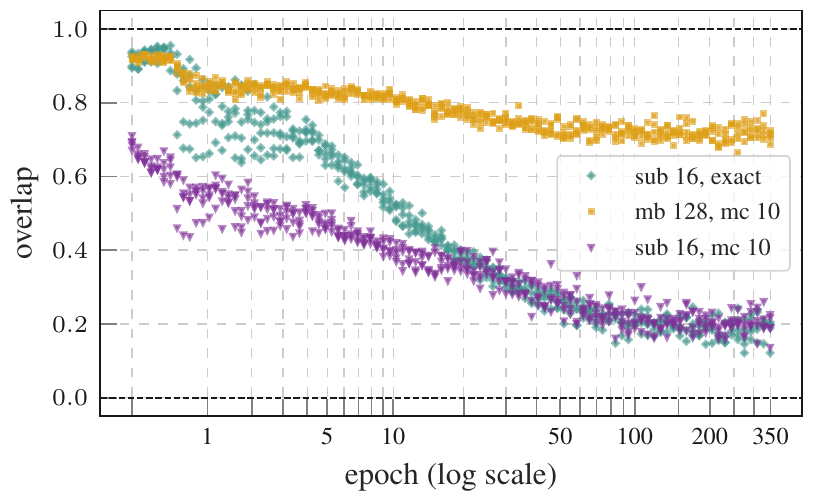} 
\end{minipage}\hfill
\begin{minipage}{0.50\textwidth}
\centering
\textbf{\cifarhun \allcnnc \adam}\\[1mm]
\includegraphics[scale=1.0]{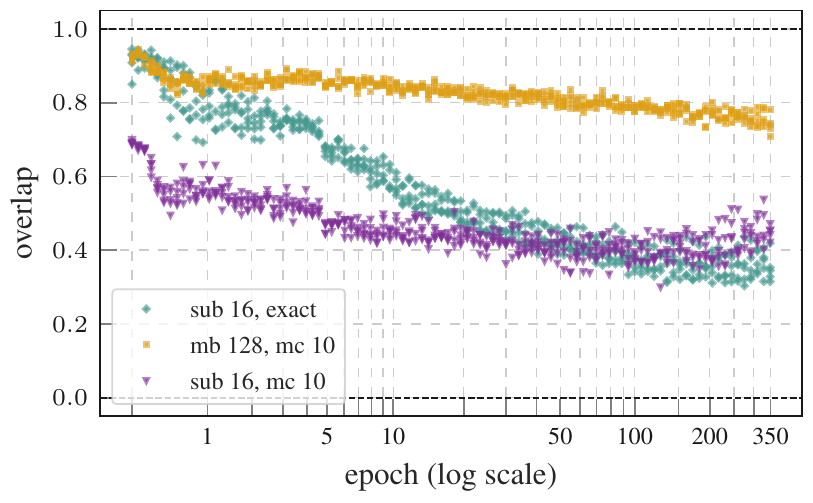} 
\end{minipage}

\caption{\textbf{\bfvivit{} vs. mini-batch \ggn{}:} 
Overlap between the top-$C$ eigenspaces of different \ggn approximations and the mini-batch \ggn during training for all test problems.
Each approximation is evaluated on $5$ different mini-batches.}
\label{fig:vivit_vs_mini_batch_ggn}
\end{figure}

\paragraph{Results (2):} The results can be found in \Cref{fig:vivit_vs_mini_batch_ggn}. 
Over large parts of the optimization (note the log scale for the epoch-axis), the \mc approximation seems to be better suited than curvature sub-sampling (which has comparable computational cost). 
For the \cifarhun \allcnnc test problem, the \mc approximation stands out particularly early from the other approximations and consistently yields higher overlaps with the mini-batch \ggn. 
  
\subsection{Curvature under noise and approximations}
\label{sec:curvature_noise}

\ggn and Hessian are predominantly used to locally approximate the loss by a quadratic model $q$ (see \Cref{eq:quadratic_model}). 
Even if the curvature's eigenspace is completely preserved in spite of the approximations, they can still alter the curvature \textit{magnitude} along the eigenvectors.

\paragraph{Procedure:} 
\Cref{tab:cases_curvature_noise} gives an overview over the cases considered in this experiment. 

\begin{table}[ht]
  \centering
  \caption{
\textbf{Considered cases for approximation of curvature:}
We use a different set of cases for the approximation of the \ggn{} depending on the test problem. For the test problems with $C=10$, we use $M=1$ \mc-sample, for the \cifarhun \allcnnc test problem ($C=100$), we use $M=10$ \mc-samples in order to reduce the computational costs by the same factor. 
}
  \label{tab:cases_curvature_noise}
  \vspace{1ex}
  \begin{normalsize}
    \begin{tabular}{llll}
      \toprule
      Problem 
      & Cases \\
      \midrule
      \makecell[tl]{
      \fmnist \twoctwod \\ 
      \cifarten \threecthreed and \\ 
      \cifarten \resnetthirtytwo}
      & \makecell[tl]{
      \textbf{mb, exact} with mini-batch size $N = 128$\\
      \textbf{mb, mc} with $N=128$ and $M=1$ \mc{}-sample\\
      \textbf{sub, exact} using $16$ samples from the mini-batch\\
      \textbf{sub, mc} using $16$ samples from the mini-batch and $M=1$ \mc{}-sample
      }
      \\
      \midrule
      \cifarhun \allcnnc
      & \makecell[tl]{
      \textbf{mb, exact} with mini-batch size $N = 128$\\
      \textbf{mb, mc} with $N=128$ and $M=10$ \mc{}-samples\\
      \textbf{sub, exact} using $16$ samples from the mini-batch\\
      \textbf{sub, mc} using $16$ samples from the mini-batch and $M=10$ \mc{}-samples
      }
      \\
      \bottomrule
    \end{tabular}
  \end{normalsize}
\end{table}

Due to the large computational effort needed for the evaluation of full-batch directional derivatives, a subset of the checkpoints from \Cref{sec:ggn_vs_hessian} is used for two test problems: 
We use every second checkpoint for \cifarten \resnetthirtytwo and every forth checkpoint for \cifarhun \allcnnc.

For each checkpoint and case, we compute the top-$C$ eigenvectors $\{\ve_k\}_{k=1}^C$ of the \ggn approximation $\mG^{(\text{ap})}$ 
either with \vivit or using an iterative matrix-free approach (as in \Cref{sec:eigenspace_noise}). 
The second-order directional derivative of the corresponding quadratic model along direction $\ve_k$ is then given by $\lambda_k^{(\text{ap})} = \ve_k^\top \mG^{(\text{ap})} \ve_k$ (see \Cref{eq:directional-derivatives}). 
As a reference, we compute the full-batch \ggn $\mG^{(\text{fb})}$ and the resulting directional derivatives along the same eigenvectors $\{\ve_k\}_{k=1}^C$, i.e. $\lambda_k^{(\text{fb})} = \ve_k^\top \mG^{(\text{fb})} \ve_k$. 
The average (over all $C$ directions) relative error is given by 
\begin{equation*}
\epsilon = \frac{1}{C} \sum_{k=1}^C \frac{
\lvert \lambda_k^{(\text{ap})} - \lambda_k^{(\text{fb})} \rvert
}{\lambda_k^{(\text{fb})}} \, .
\end{equation*}
The procedure above is repeated on $3$ mini-batches from the training data (i.e. we obtain $3$ average relative errors for every checkpoint and case) -- except for the \cifarhun \allcnnc test problem, where we perform only a single run to keep the computational effort manageable.

\begin{figure}[p]
\centering
\begin{minipage}{0.50\textwidth}
\centering
\textbf{\fmnist \twoctwod \sgd}\\[1mm]
\includegraphics[scale=1.0]{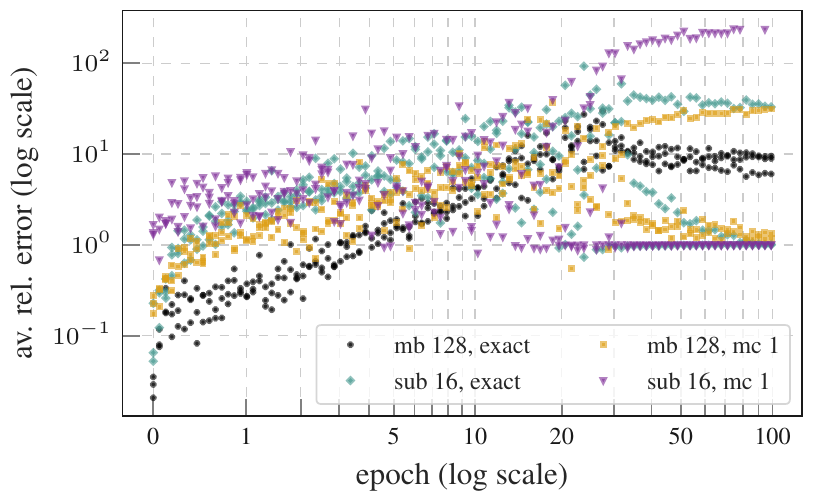} 
\end{minipage}\hfill
\begin{minipage}{0.50\textwidth}
\centering
\textbf{\fmnist \twoctwod \adam}\\[1mm]
\includegraphics[scale=1.0]{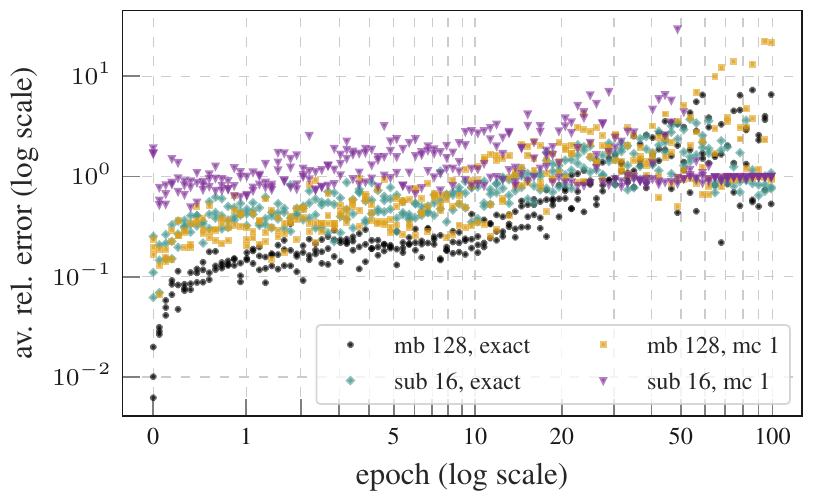} 
\end{minipage}

\begin{minipage}{0.50\textwidth}
\centering
\textbf{\cifarten \threecthreed \sgd}\\[1mm]
\includegraphics[scale=1.0]{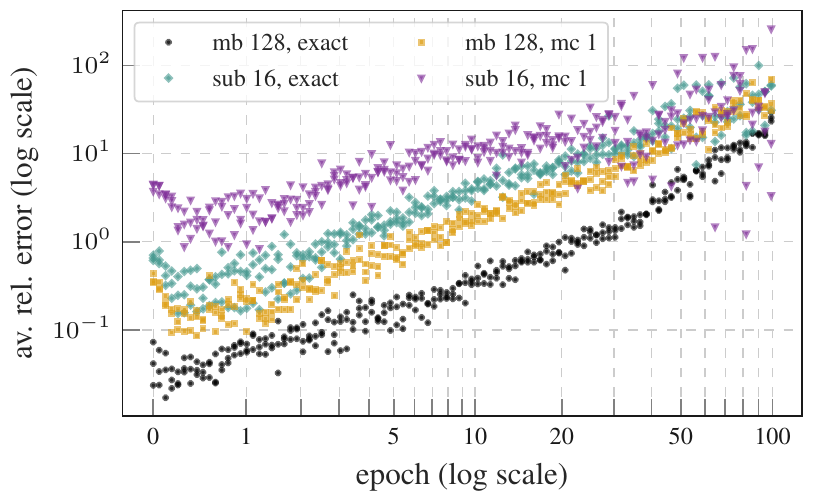} 
\end{minipage}\hfill
\begin{minipage}{0.50\textwidth}
\centering
\textbf{\cifarten \threecthreed \adam}\\[1mm]
\includegraphics[scale=1.0]{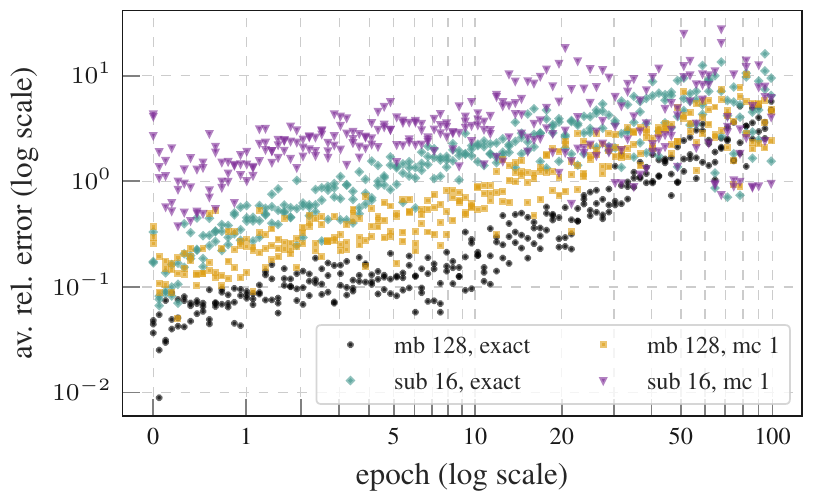} 
\end{minipage}

\begin{minipage}{0.50\textwidth}
\centering
\textbf{\cifarten \resnetthirtytwo \sgd}\\[1mm]
\includegraphics[scale=1.0]{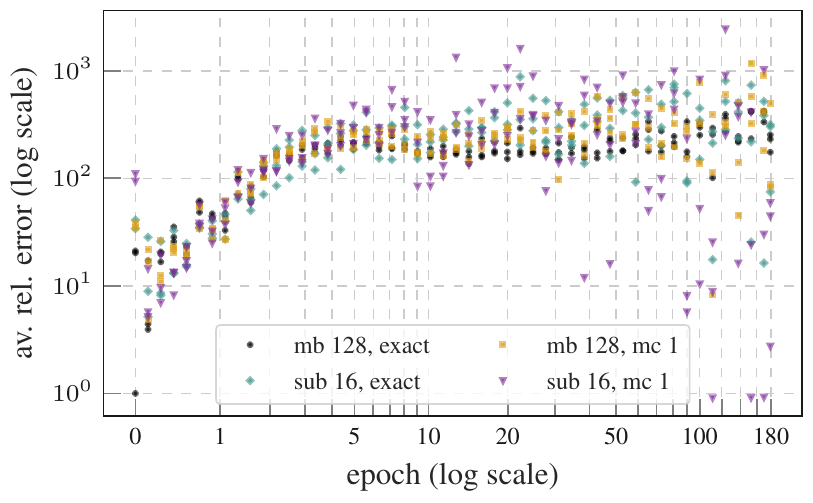} 
\end{minipage}\hfill
\begin{minipage}{0.50\textwidth}
\centering
\textbf{\cifarten \resnetthirtytwo \adam}\\[1mm]
\includegraphics[scale=1.0]{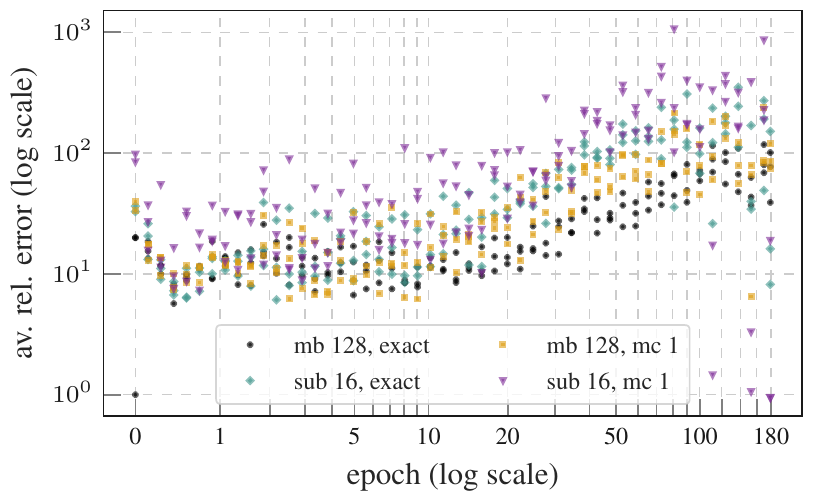} 
\end{minipage}

\begin{minipage}{0.50\textwidth}
\centering
\textbf{\cifarhun \allcnnc \sgd}\\[1mm]
\includegraphics[scale=1.0]{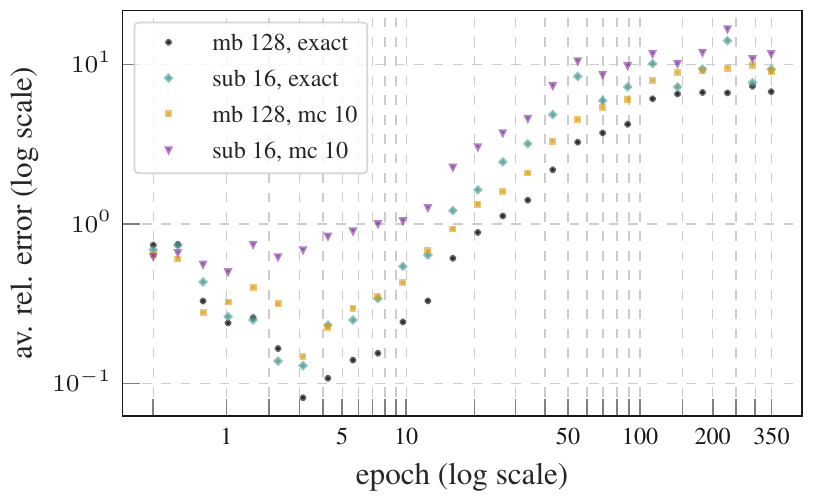} 
\end{minipage}\hfill
\begin{minipage}{0.50\textwidth}
\centering
\textbf{\cifarhun \allcnnc \adam}\\[1mm]
\includegraphics[scale=1.0]{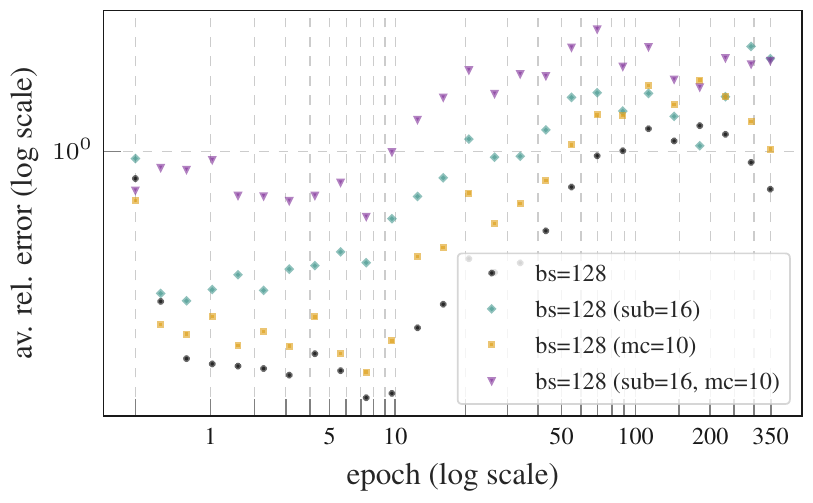} 
\end{minipage}

\caption{
\textbf{\bfvivit{}'s vs. full-batch quadratic model:}
Comparison between approximations to the quadratic model and the full-batch model in terms of the average relative error for the directional curvature during training for all test problems.
}
\label{fig:curvature_noise}
\end{figure}

\paragraph{Results:} The results can be found in \Cref{fig:curvature_noise}. We observe similar results as in \Cref{sec:eigenspace_noise}: With increasing computational effort, the approximated directional derivatives become more precise and the overall approximation quality decreases over the course of the optimization. For the \resnetthirtytwo architecture, the average errors are particularly large. 

\subsection{Directional derivatives}
\label{sec:directional_derivatives}
\paragraph{Procedure:} We use the checkpoints from \Cref{sec:ggn_vs_hessian}. For every checkpoint, we compute the top-$C$ eigenvectors of the mini-batch \ggn{} (with a mini-batch size of $N=128$) using an iterative matrix-free method. We also compute the mini-batch gradient.
The first- and second-order directional derivatives of the resulting quadratic model (see \Cref{eq:quadratic_model}) are given by \Cref{eq:gammas-lambdas}. 

We use these directional derivatives 
$\{\gamma_{nk}\}_{n=1, k=1}^{N, C}$, 
$\{\lambda_{nk}\}_{n=1, k=1}^{N, C}$ 
to compute signal-to-noise ratios (SNRs) along the top-$C$ eigenvectors. The curvature SNR along  direction $\ve_k$ is given by the squared sample mean divided by the empirical variance of the samples $\{\lambda_{nk}\}_{n=1}^{N}$, i.e.
\begin{equation*}
\text{SNR} = \frac{
\lambda_k^2
}{
\nicefrac{1}{N-1} \sum_{n=1}^N (\lambda_{nk} - \lambda_k)^2
}
\quad \text{where} \quad
\lambda_k = \frac{1}{N} \sum_{n=1}^N \lambda_{nk} \, .
\end{equation*}
(and similarly for $\{\gamma_{nk}\}_{n=1}^{N}$).

\begin{figure}[p]
\centering
\textbf{\fmnist \twoctwod \sgd}\\[1mm]
\begin{minipage}{0.50\textwidth}
\centering
\includegraphics[scale=1.0]{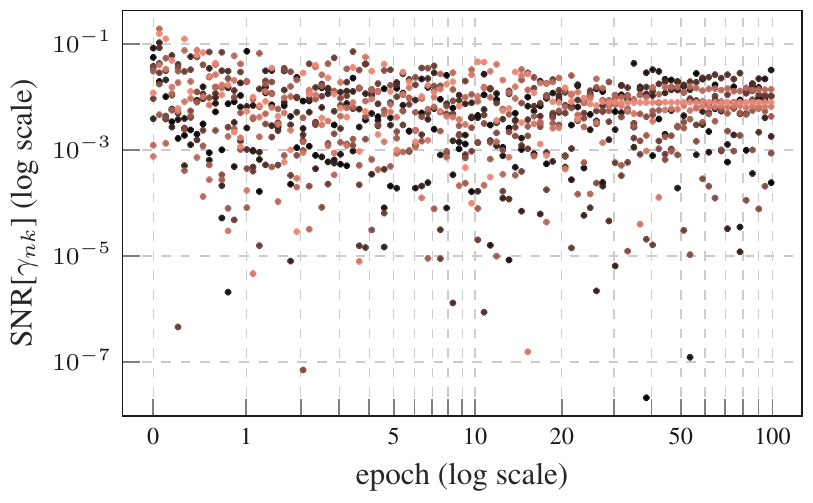} 
\end{minipage}\hfill
\begin{minipage}{0.50\textwidth}
\centering
\includegraphics[scale=1.0]{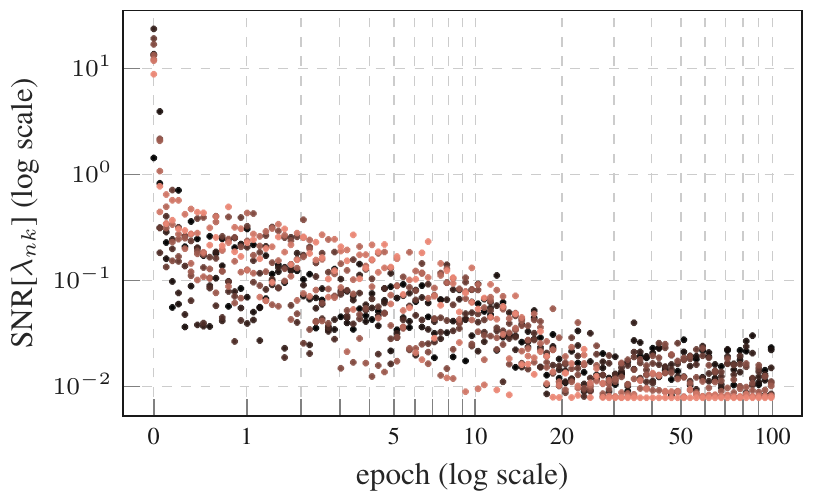} 
\end{minipage}

\textbf{\fmnist \twoctwod \adam}\\[1mm]
\begin{minipage}{0.50\textwidth}
\centering
\includegraphics[scale=1.0]{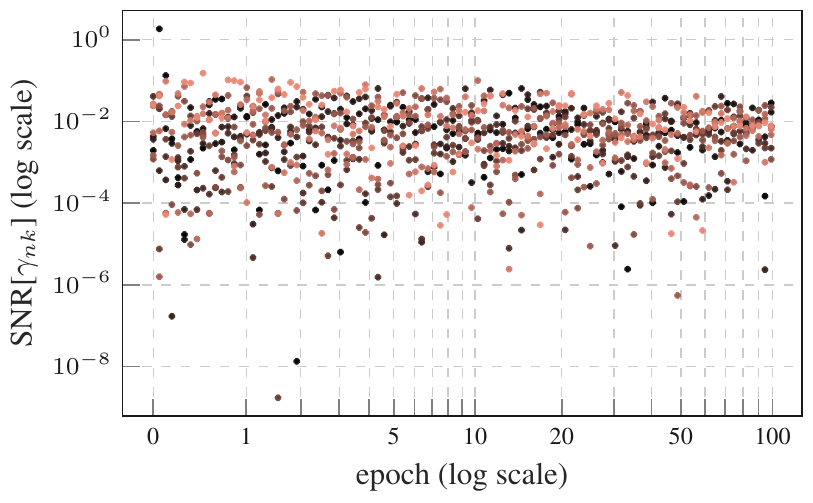} 
\end{minipage}\hfill
\begin{minipage}{0.50\textwidth}
\centering
\includegraphics[scale=1.0]{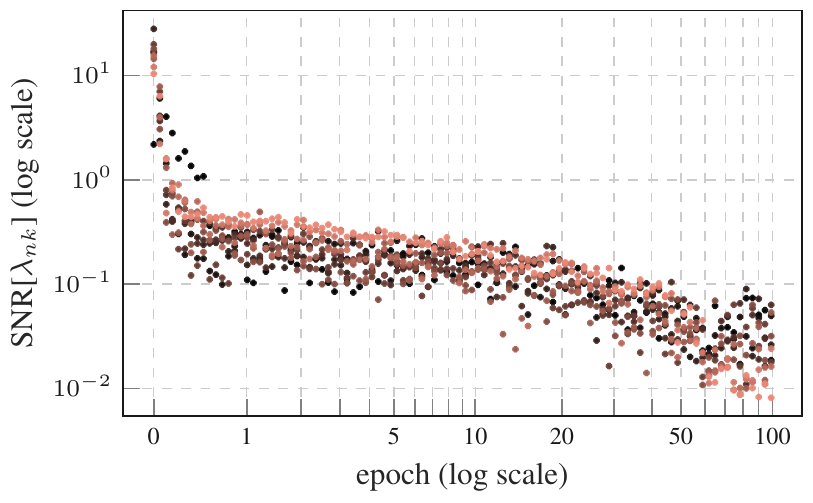} 
\end{minipage}

\textbf{\cifarten \threecthreed \sgd}\\[1mm]
\begin{minipage}{0.50\textwidth}
\centering
\includegraphics[scale=1.0]{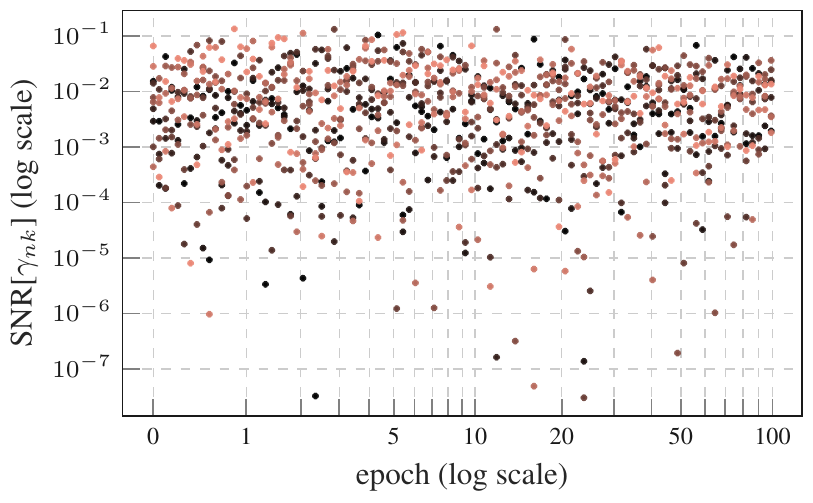} 
\end{minipage}\hfill
\begin{minipage}{0.50\textwidth}
\centering
\includegraphics[scale=1.0]{fig/exp13_plots/gammas_lambdas/cifar10_3c3d_sgd_lambdas.pdf} 
\end{minipage}

\textbf{\cifarten \threecthreed \adam}\\[1mm]
\begin{minipage}{0.50\textwidth}
\centering
\includegraphics[scale=1.0]{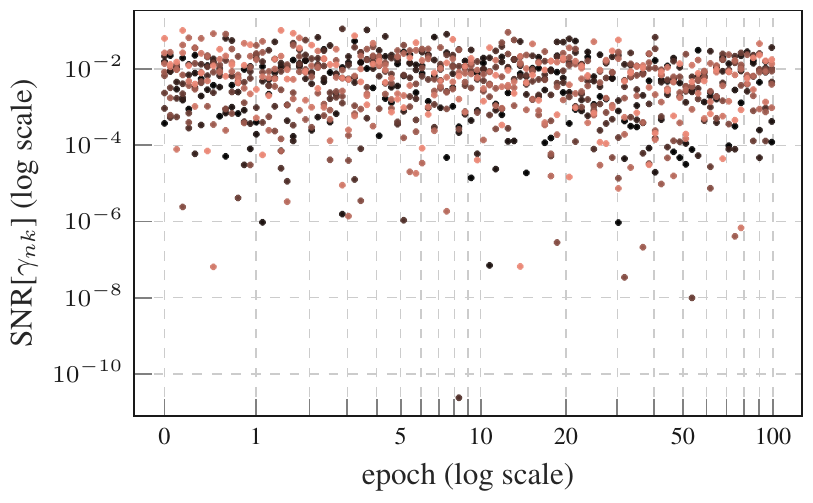} 
\end{minipage}\hfill
\begin{minipage}{0.50\textwidth}
\centering
\includegraphics[scale=1.0]{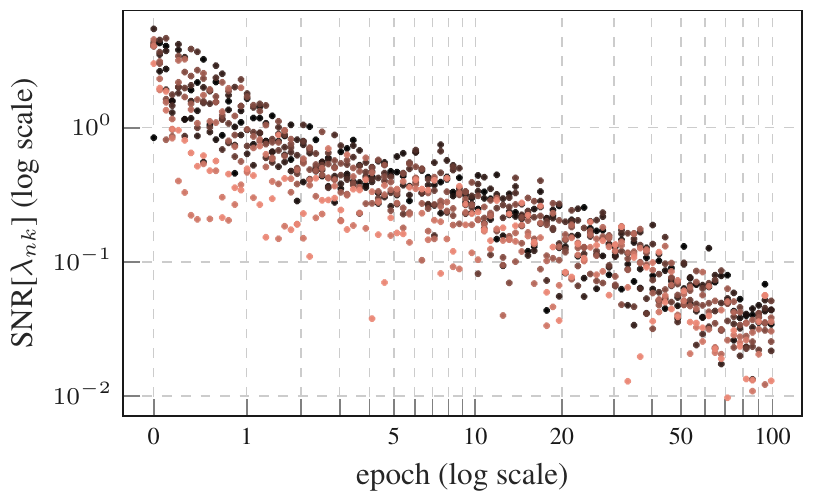} 
\end{minipage}

\caption{\textbf{Directional SNRs (1):} 
SNR along each of the mini-batch \ggn{}'s top-$C$ eigenvectors during training for all test problems. 
At fixed epoch, the SNR for the most curved direction is shown in 
\tikz\draw[white,fill={light_red},line width=0mm] (0,0) circle (.8ex); 
and for the least curved direction in 
\tikz\draw[white,fill={black}] (0,0) circle (.8ex);.
}

\label{fig:directional_derivatives_1}
\end{figure}

\begin{figure}[p]
\centering
\textbf{\cifarten \resnetthirtytwo \sgd}\\[1mm]
\begin{minipage}{0.50\textwidth}
\centering
\includegraphics[scale=1.0]{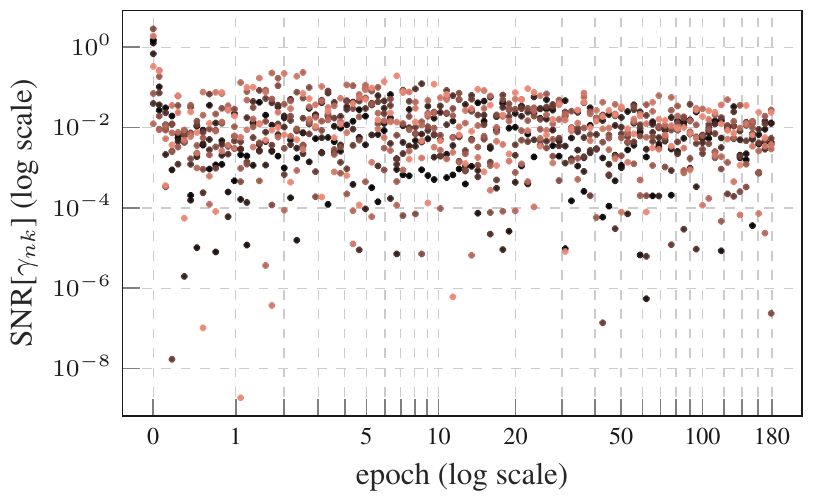} 
\end{minipage}\hfill
\begin{minipage}{0.50\textwidth}
\centering
\includegraphics[scale=1.0]{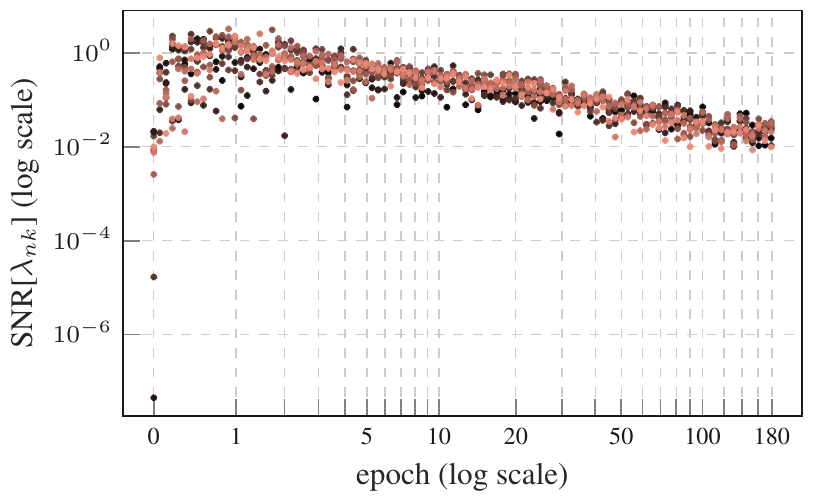} 
\end{minipage}

\textbf{\cifarten \resnetthirtytwo \adam}\\[1mm]
\begin{minipage}{0.50\textwidth}
\centering
\includegraphics[scale=1.0]{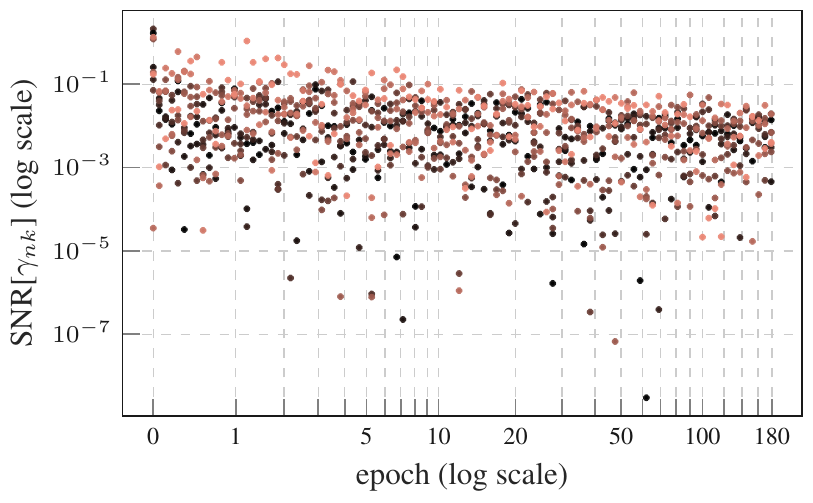} 
\end{minipage}\hfill
\begin{minipage}{0.50\textwidth}
\centering
\includegraphics[scale=1.0]{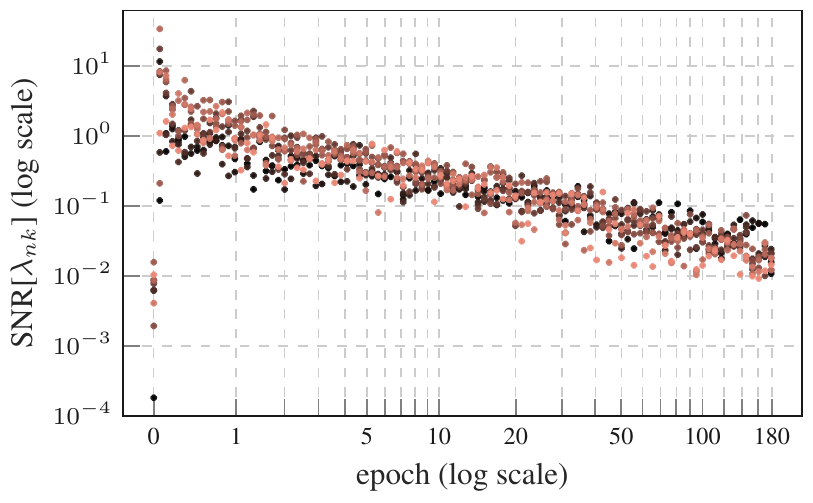} 
\end{minipage}

\textbf{\cifarhun \allcnnc \sgd}\\[1mm]
\begin{minipage}{0.50\textwidth}
\centering
\includegraphics[scale=1.0]{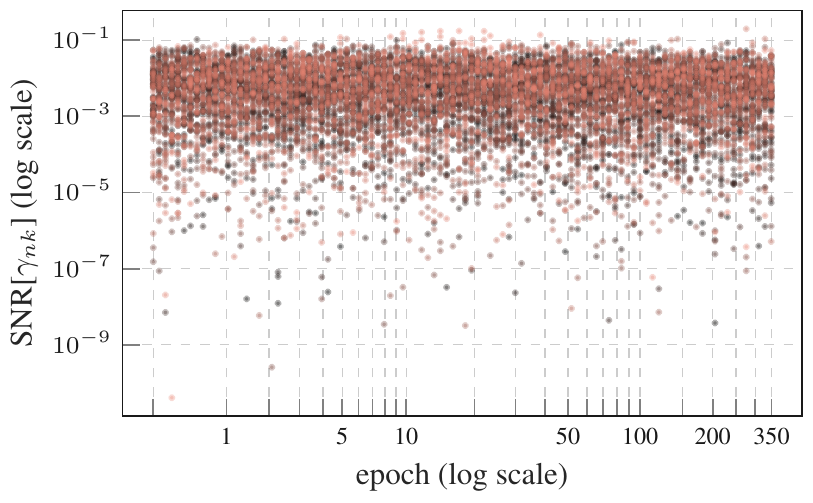} 
\end{minipage}\hfill
\begin{minipage}{0.50\textwidth}
\centering
\includegraphics[scale=1.0]{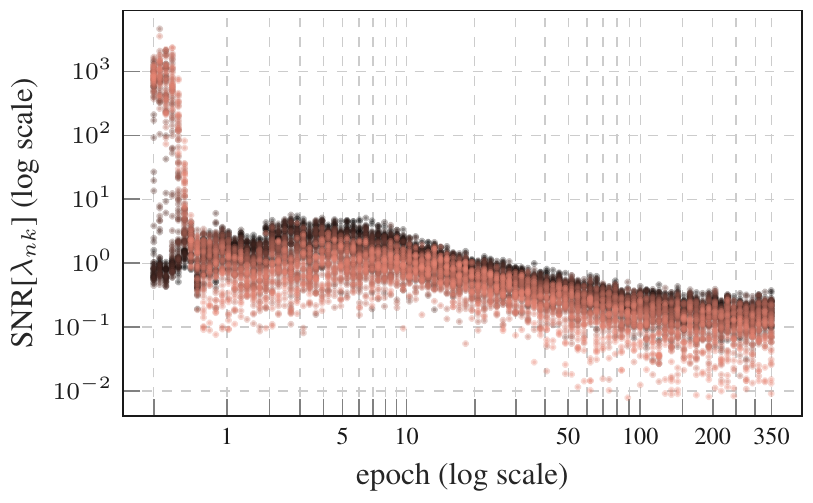} 
\end{minipage}

\textbf{\cifarhun \allcnnc \adam}\\[1mm]
\begin{minipage}{0.50\textwidth}
\centering
\includegraphics[scale=1.0]{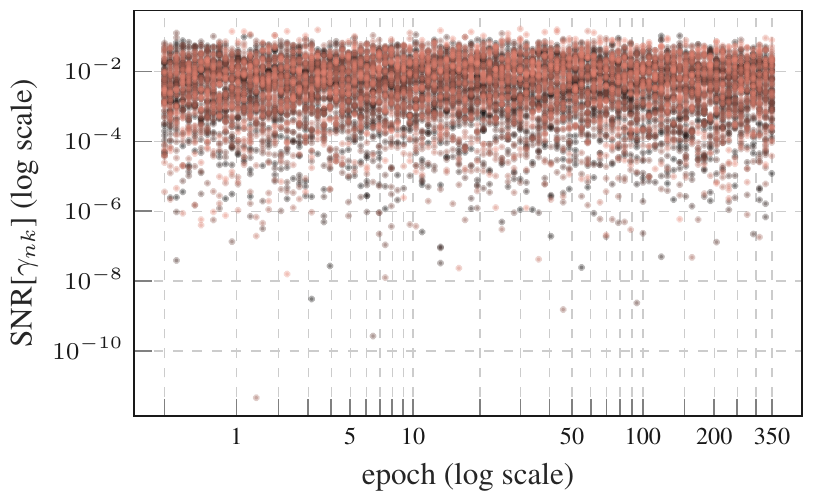} 
\end{minipage}\hfill
\begin{minipage}{0.50\textwidth}
\centering
\includegraphics[scale=1.0]{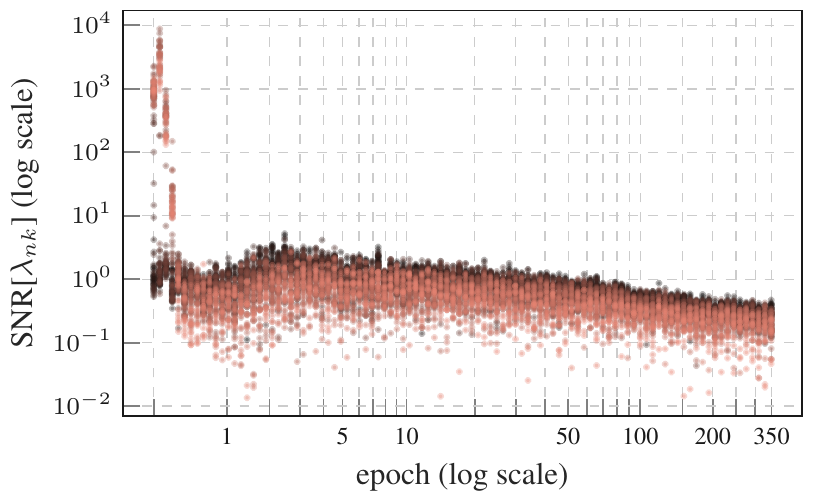} 
\end{minipage}

\caption{\textbf{Directional SNRs (2):} 
SNR along each of the mini-batch \ggn{}'s top-$C$ eigenvectors during training for all test problems. 
At fixed epoch, the SNR for the most curved direction is shown in 
\tikz\draw[white,fill={light_red},line width=0mm] (0,0) circle (.8ex); 
and for the least curved direction in 
\tikz\draw[white,fill={black}] (0,0) circle (.8ex);.
}
\label{fig:directional_derivatives_2}
\end{figure}

\paragraph{Results:} The results can be found in \Cref{fig:directional_derivatives_1} and \ref{fig:directional_derivatives_2}. 
These plots show the SNRs in $C$ distinct colors that are generated by linearly interpolating in the RGB color space from black
(\tikz\draw[white,fill={black},line width=0mm] (0,0) circle (.8ex);)
to light red
(\tikz\draw[white,fill={light_red},line width=0mm] (0,0) circle (.8ex);).
At each checkpoint, the colors are assigned based on the \textit{order} of the respective directional curvature $\lambda_k$: The SNR that belongs to the direction with the smallest curvature is shown in black and the SNR that belongs to the direction with the largest curvature is shown in light red. 
The color thus encodes only the order of the top-$C$ directional curvatures -- \textit{not} their magnitude.
We use this color encoding to reveal potential correlations between SNR and curvature. 

We find that the gradient SNR along the top-$C$ eigenvectors is consistently small (in comparison to the curvature SNR) and remains roughly on the same level during the optimization. 
The curvature signal decreases as training proceeds.
The SNRs along the top-C eigendirections do not appear to show any significant correlation with the corresponding curvatures. 
Only for the \cifarhun test problems we can suspect a correlation between strong curvature and small curvature SNR. 

\section{Implementation details}
\label{sec:implementation-details}
\paragraph{Layer view of backpropagation:}
Consider a single layer $T^{(i)}_{\vtheta^{(i)}}$ that transforms inputs
$\vz_n^{(i-1)} \in \sR^{h^{(i-1)}}$ into outputs $\vz_n^{(i)} \in \sR^{h^{(i)}}$
by means of a parameter $\vtheta^{(i)} \in \sR^{d^{(i)}}$. During
backpropagation for $\mV$, the layer receives vectors $ \vs_{nc}^{(i)} =
(\jac_{\vz_n^{(i)}}f_n)^\top \vs_{nc}$
from the previous stage (recall $\nabla_{f}^2\ell_n = \sum_{c=1}^C \vs_{nc}
\vs_{nc}^\top$). Parameter contributions $\vv_{nc}^{(i)}$ to $\mV$ are
obtained by application of its Jacobian,
\begin{align}
  \nonumber
  \vv_{nc}^{(i)}
  &=
    \left(
    \jac_{\vtheta^{(i)}} f_n
    \right)^\top
    \vs_{nc}
  \\
  \nonumber
  &=
    \left(
    \jac_{\vtheta^{(i)}} \vz_n^{(i)}
    \right)^\top
    \left(
    \jac_{\vz_n^{(i)}} f_n
    \right)^\top
    \vs_{nc}
    \explainmath{(Chain rule)}
  \\
  &=
    \left(
    \jac_{\vtheta^{(i)}} \vz_n^{(i)}
    \right)^\top
    \vs_{nc}^{(i)}\,.
    \explainmath{(Definition of $\vs_{nc}^{(i)}$)}
    \label{eq:ggn-vector-layer-i}
\end{align}
Consequently, the contribution of $\vtheta^{(i)}$ to $\mV$, denoted by
$\mV^{(i)} \in \sR^{d^{(i)} \times NC}$, is
\begin{align}
  \label{eq:layerwise-ggn-factor}
  \mV^{(i)}
  =
  \frac{1}{\sqrt{N}}
  \begin{pmatrix}
    \vv^{(i)}_{11} & \vv^{(i)}_{12} & \dots & \vv^{(i)}_{NC}
  \end{pmatrix}
                                              \quad\text{with}\quad
                                              \vv_{nc}^{(i)} =
                                              \left(
                                              \jac_{\vtheta^{(i)}} f_n\right)^\top
                                              \vs_{nc}\,.
\end{align}

\subsection{Optimized Gram matrix computation for linear layers}
\label{sec:optimized-gram-matrix}
Our goal is to efficiently extract $\vtheta^{(i)}$'s contribution to the
Gram matrix $\mGtilde$, given by
\begin{align}
  \label{eq:layerwise-gram-matrix}
  \mGtilde^{(i)} = {\mV^{(i)}}^\top \mV^{(i)} \in \sR^{NC \times NC}\,.
\end{align}

\paragraph{Gram matrix via expanding $\mV^{(i)}$:}
One way to construct $\mG^{(i)}$ is to first expand $\mV^{(i)}$
(\Cref{eq:layerwise-ggn-factor}) via the Jacobian $\jac_{\vtheta^{(i)}}
\vz_n^{(i)}$, then contract it (\Cref{eq:layerwise-gram-matrix}). This can be a
memory bottleneck for large linear layers which are common in many architectures
close to the network output. However if only the Gram matrix rather than $\mV$
is required, structure in the Jacobian can be used to construct $\mGtilde^{(i)}$
without expanding $\mV^{(i)}$ and thus reduce this overhead.

\paragraph{Optimization for linear layers:}
Now, let $T^{(i)}_{\vtheta^{(i)}}$ be a linear layer with weights
$\mW^{(i)} \in \sR^{h^{(i)}\times h^{(i-1)}}$, \ie $\vtheta^{(i)} =
\mathrm{vec}(\mW^{(i)}) \in \sR^{d^{(i)} = h^{(i)} h^{(i-1)}}$ with column
stacking convention for vectorization,
\begin{align*}
  T^{(i)}_{\vtheta^{(i)}}:\quad \vz_n^{(i)} = \mW^{(i)} \vz_n^{(i-1)}\,.
\end{align*}
The Jacobian is
\begin{align}
  \label{eq:jacobian-linear-layer}
  \jac_{\vtheta^{(i)}} \vz_n^{(i)}
  =
  {\vz_n^{(i-1)}}^\top \otimes \mI_{h^{(i)}}\,.
\end{align}
Its structure can be used to directly compute entries of the Gram
matrix without expanding $\mV^{(i)}$,
\begin{align*}
   \left[ \mGtilde^{(i)}\right]_{(nc)(n'c')}
  &=
    {\vv_{nc}^{(i)}}^\top\vv_{n'c'}^{(i)}
    \explainmath{(\Cref{eq:layerwise-gram-matrix})}
    \\
  &=
    {\vs^{(i)}_{nc}}^\top
    \left( \jac_{\vtheta^{(i)}} \vz_n^{(i)} \right)
    \left( \jac_{\vtheta^{(i)}} \vz_{n'}^{(i)} \right)^\top
    {\vs^{(i)}_{n'c'}}
  \\
  &=
    {\vs^{(i)}_{nc}}^\top
    \left( {\vz_n^{(i-1)}}^\top \otimes \mI_{h^{(i)}}\right)
    \left( {\vz_{n'}^{(i-1)}}^\top \otimes \mI_{h^{(i)}} \right)^\top
    {\vs^{(i)}_{n'c'}}
    \explainmath{(\Cref{eq:jacobian-linear-layer})}
    \\
  &=
    {\vs^{(i)}_{nc}}^\top
    \left( {\vz_n^{(i-1)}}^\top \vz_{n'}^{(i-1)} \otimes \mI_{h^{(i)}}\right)
    \vs^{(i)}_{n'c'}
    \explainmath{(\Cref{eq:ggn-vector-layer-i})}
  \\
  &=
    { \vz_n^{(i-1)}}^\top \vz_{n'}^{(i-1)} {\vs^{(i)}_{nc}}^\top
     \mI_{h^{(i)}} \vs^{(i)}_{n'c'}
    \explainmath{(${\vz_n^{(i-1)}}^\top \vz_{n'}^{(i-1)}\in \sR$)}
  \\
  &=
    \left(  { \vz_n^{(i-1)}}^\top \vz_{n'}^{(i-1)}\right)
    \left( {\vs^{(i)}_{nc}}^\top \vs^{(i)}_{n'c'} \right)\,.
\end{align*}
We see that the Gram matrix is built from two Gram matrices based on $\{
\vz_n^{(i-1)}\}_{n=1}^N$ and $\{\vs_{nc}^{(i)}\}_{n=1,c=1}^{N,C}$, that require
$\gO(N^2)$ and $\gO((NC)^2)$ memory, respectively. In comparison, the na\"ive
approach via $\mV^{(i)} \in \sR^{d^{(i)} \times N C}$ scales with the number of
weights, which is often comparable to $D$. For instance, the \threecthreed
architecture on \cifarten has $D=895,\!210$ and the largest weight matrix has
$d^{(i)}=589,\!824$, whereas $N C = 1,\!280$ during training
\citep{schneider2019deepobs}.

\subsection{Implicit multiplication with the inverse (block-diagonal) \ggn}
\label{sec:implicit-multiplication-inverse}
\paragraph{Inverse \ggn-vector products:}
A damped Newton step requires multiplication by $(\mG + \delta \mI_D)^{-1}$.\footnote{$\delta
  \mI_{D}$ can be replaced by other easy-to-invert matrices.} By means of
\Cref{eq:ggn-factorization} and the matrix inversion lemma,
\begin{align}
  \left(
  \delta \mI_D +  \mG
  \right)^{-1}
  \nonumber
  &=
    \left(
    \delta \mI_D + \mV \mV^\top
    \right)^{-1}
    \explainmath{(\Cref{eq:ggn-factorization})}
  \\
  \nonumber
  &=
    \frac{1}{\delta}
    \left(
    \mI_D + \frac{1}{\delta} \mV \mV^\top
    \right)^{-1}
  \\
  \nonumber
  &=
    \frac{1}{\delta}
    \left[
    \mI_D
    -
    \frac{1}{\delta} \mV
    \left(
    \mI_{NC}
    + \mV^\top \frac{1}{\delta} \mV
    \right)^{-1}
    \mV^\top
    \right]
    \explainmath{(Matrix inversion lemma)}
  \\
  \nonumber
  &=
    \frac{1}{\delta}
    \left[
    \mI_D
    -
    \mV
    \left(
    \delta \mI_{NC} + \mV^\top \mV
    \right)^{-1}
    \mV^\top
    \right]
    \explainmath{(Gram matrix)}
  \\
  &=
    \frac{1}{\delta}
    \left[
    \mI_D
    -
    \mV
    \left(
    \delta \mI_{NC} + \mGtilde
    \right)^{-1}
    \mV^\top
    \right]\,.
    \label{eq:inverse-ggn}
\end{align}
Inverse \ggn-vector products require inversion of the damped Gram matrix
as well as applications of $\mV, \mV^\top$ for the transformations between Gram
and parameter space.

\paragraph{Inverse block-diagonal \ggn-vector products:} Next, we replace
the full \ggn by its block diagonal approximation $\mG \approx
\mG_{\text{BDA}} = \diag(\mG^{(1)}, \mG^{(2)}, \dots)$ with
\begin{align*}
  \mG^{(i)} = \mV^{(i)} {\mV^{(i)}}^\top \in \sR^{d^{(i)} \times d^{(i)}}
\end{align*}
and $\mV^{(i)}$ as in \Cref{eq:layerwise-ggn-factor}. Then, inverse
multiplication reduces to each block,
\begin{align*}
  \mG_{\text{BDA}}^{-1} = \diag\!\left({\mG^{(1)}}^{-1}, {\mG^{(2)}}^{-1}, \dots\right)\,.
\end{align*}
If again a damped Newton step is considered, we can reuse \Cref{eq:inverse-ggn}
with the substitutions
\begin{align*}
  \left(\mG, D, \mV, \mV^\top, \mGtilde\right)
  \leftrightarrow \left( \mG^{(i)}, d^{(i)}, \mV^{(i)}, {\mV^{(i)}}^\top, \mGtilde^{(i)}\right)
\end{align*}
to apply the inverse and immediately discard the \vivit factors:
At backpropagation of layer $T^{(i)}_{\vtheta^{(i)}}$
\begin{enumerate}
\item Compute $\mV^{(i)}$ using \Cref{eq:layerwise-ggn-factor}.
\item Compute $\mGtilde^{(i)}$ using \Cref{eq:layerwise-gram-matrix}.
\item Compute $\left(\delta \mI_{NC} + \mGtilde^{(i)}\right)^{-1}$.
\item Apply the inverse in \Cref{eq:inverse-ggn} with the above substitutions to the target vector.
\item Discard $\mV^{(i)}$, ${\mV^{(i)}}^\top, \mGtilde^{(i)},$ and $\left(\delta
    \mI_{NC} + \mGtilde^{(i)}\right)^{-1}$. Proceed to layer $i-1$.
\end{enumerate}
Note that the above scheme should only be used for parameters that satisfy
$d^{(i)} > NC$, \ie $\dim(\mG^{(i)}) > \dim(\mGtilde^{(i)})$. Low-dimensional
parameters can be grouped with others to increase their joint dimension, and to
control the block structure of $\mG_{\text{BDA}}$.

\end{document}